\documentclass{egpubl}
 
\JournalPaper         
\usepackage[T1]{fontenc}
\usepackage{dfadobe}  

\biberVersion
\BibtexOrBiblatex
\usepackage[backend=bibtex,bibstyle=EG,citestyle=alphabetic,backref=true]{biblatex} 
\electronicVersion
\PrintedOrElectronic

\ifpdf \usepackage[pdftex]{graphicx} \pdfcompresslevel=9
\else \usepackage[dvips]{graphicx} \fi

\usepackage{egweblnk} 
\usepackage{subfig}

\title[EG \LaTeX\ Author Guidelines]%
      {GlassNet: Label Decoupling-based Three-stream Neural Network for Robust Image Glass Detection}

\author[C. Zheng \& D. Shi \& X. Yan \& D. Liang \& M. wei \& X. Yang \& Y. Guo \& H. Xie]
{\parbox{\textwidth}{\centering Chengyu Zheng$^{1}$\orcid{0000-0002-9790-4262}, Ding Shi$^{1}$, Xuefeng Yan$^1$, Dong Liang$^1$, Mingqiang Wei$^1$\thanks{M. Wei is the corresponding author (mqwei@nuaa.edu.cn).}, Xin Yang$^2$, Yanwen Guo$^3$
         and Haoran Xie$^4$
        }
        \\
{\parbox{\textwidth}{\centering 
         $^1$Nanjing University of Aeronautics and Astronautics, Nanjing, China\\
         $^2$Dalian University of Technology, Dalian, China\\
         $^3$Nanjing University, Nanjing, China\\
         $^4$Lingnan University, Hong Kong, China\\
       }
}
}

\begin{document}

\teaser{
      \centering
      \subfloat[Input]{\label{img_input}
      \begin{minipage}[t]{0.145\textwidth}
            \centering
            \includegraphics[width=1\linewidth]{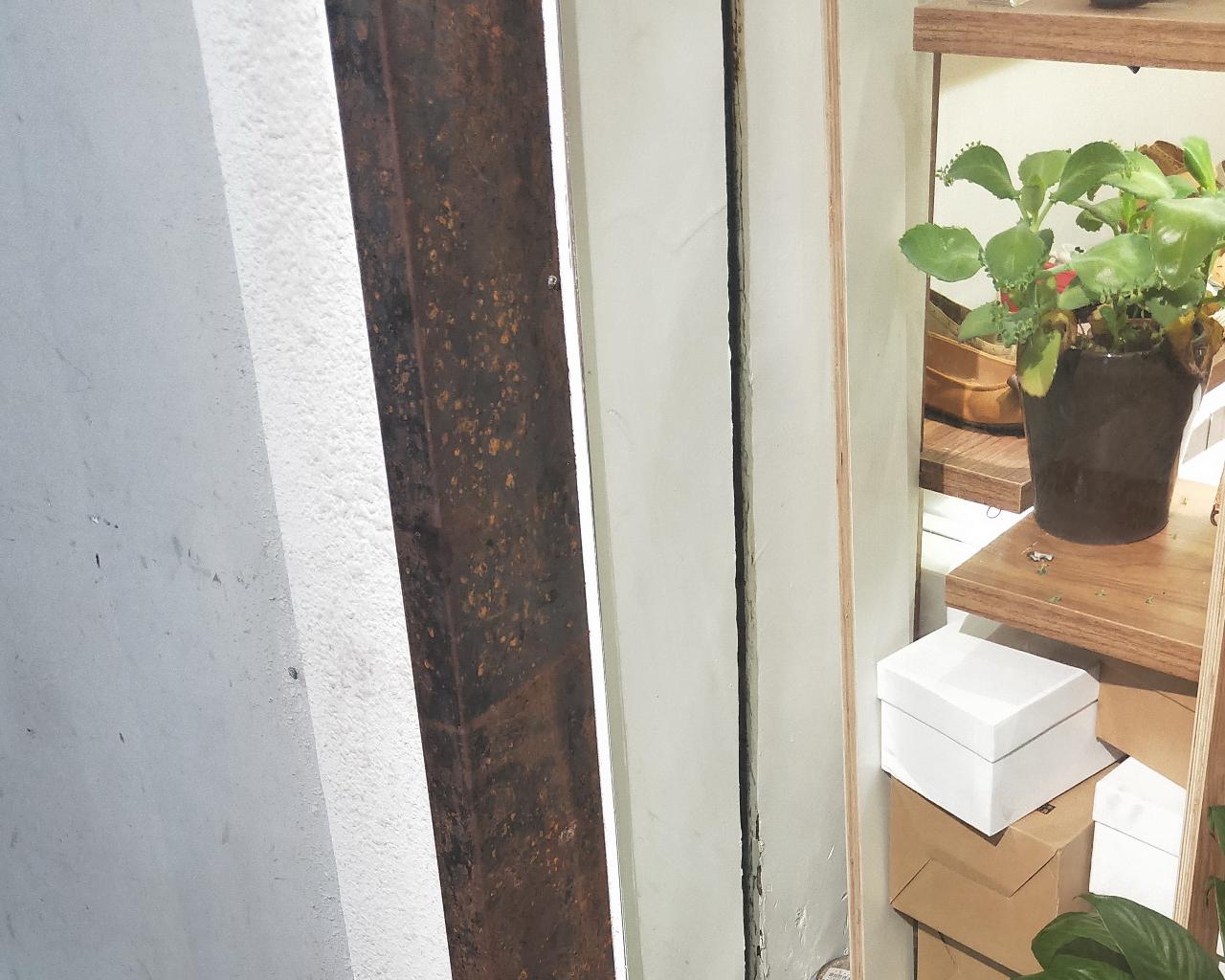}

            \includegraphics[width=1\linewidth]{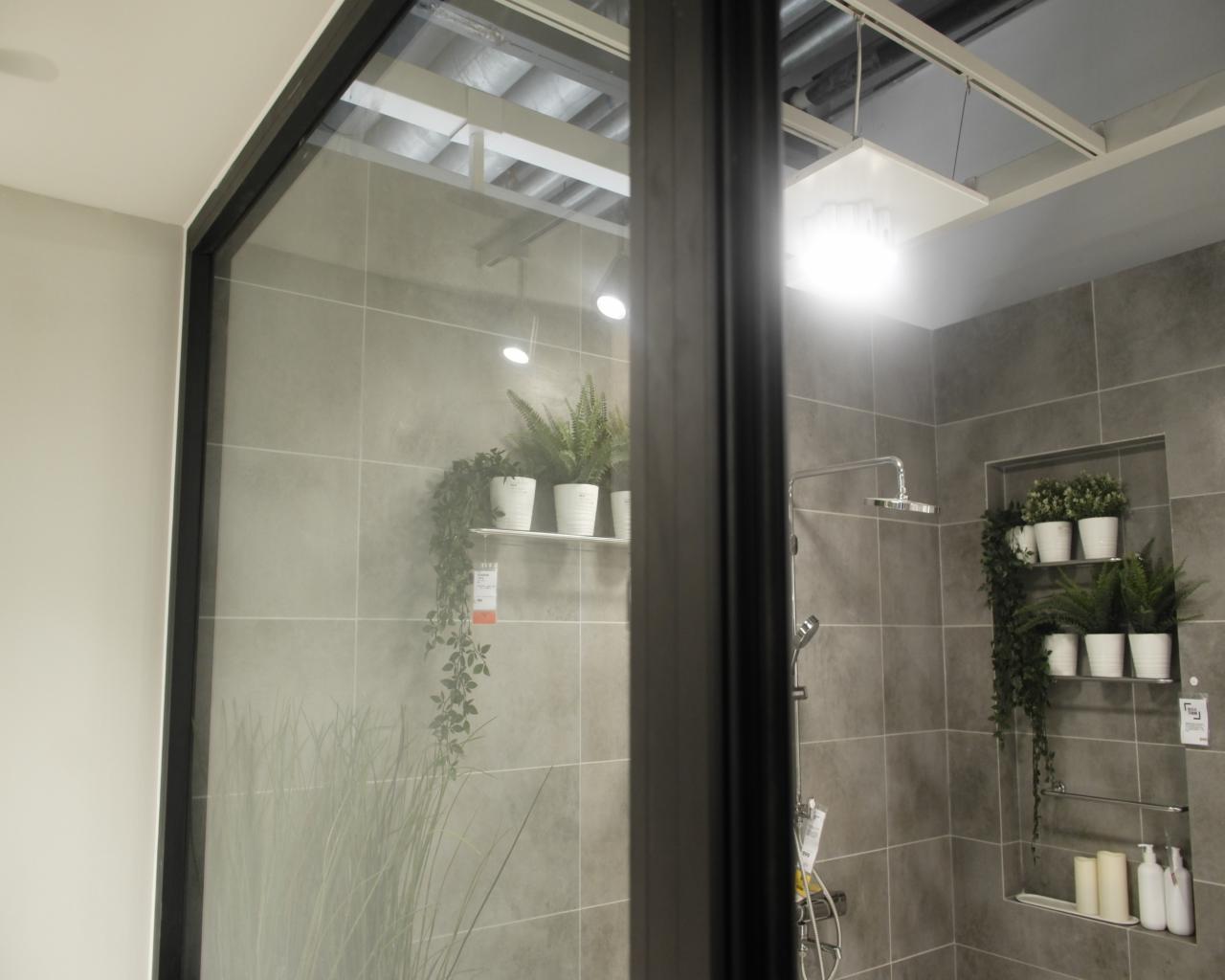}
      \end{minipage}
      }
      \subfloat[DANet \cite{DANet}]{\label{img_DA}
      \begin{minipage}[t]{0.145\textwidth}
            \centering
            \includegraphics[width=1\linewidth]{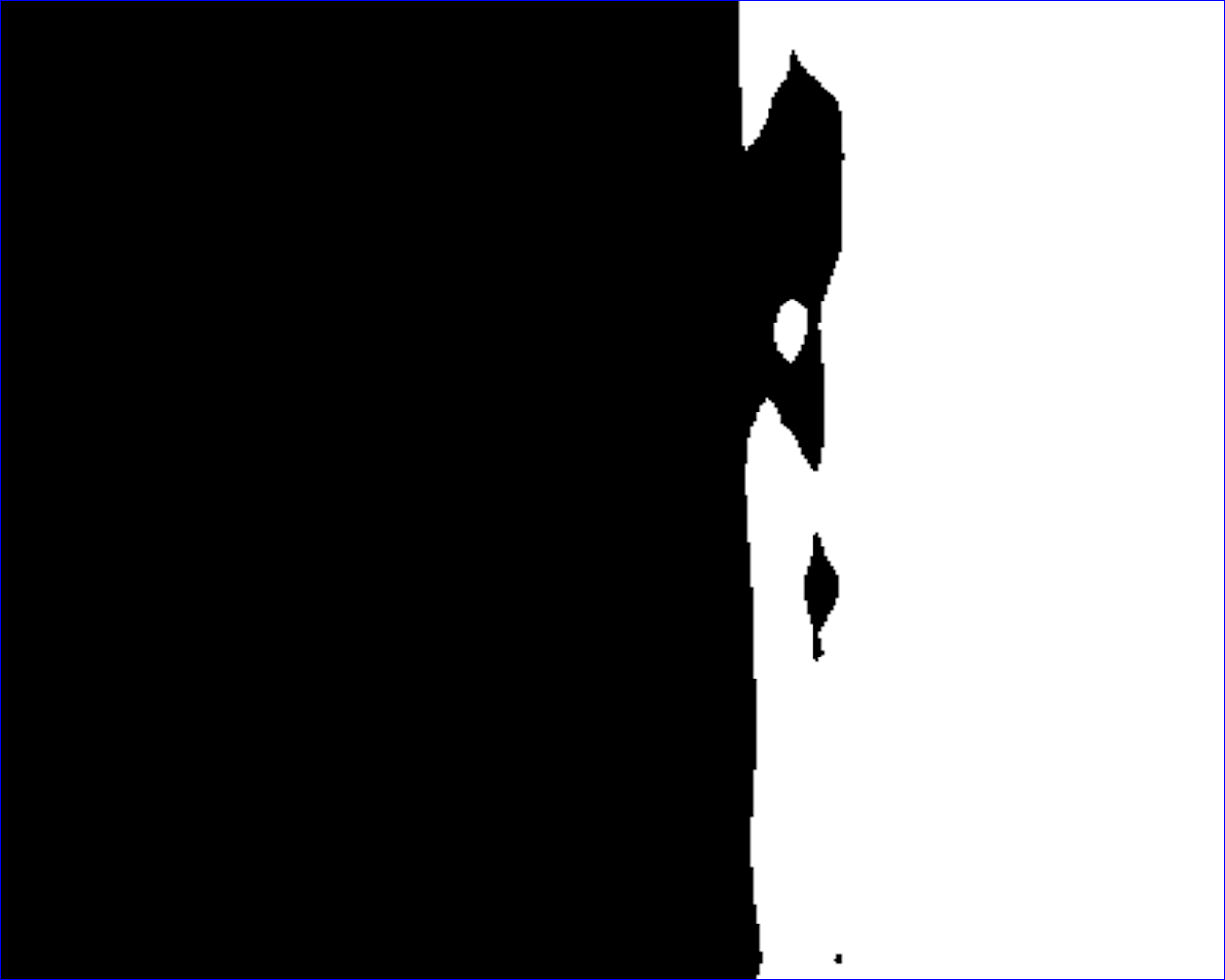}

            \includegraphics[width=1\linewidth]{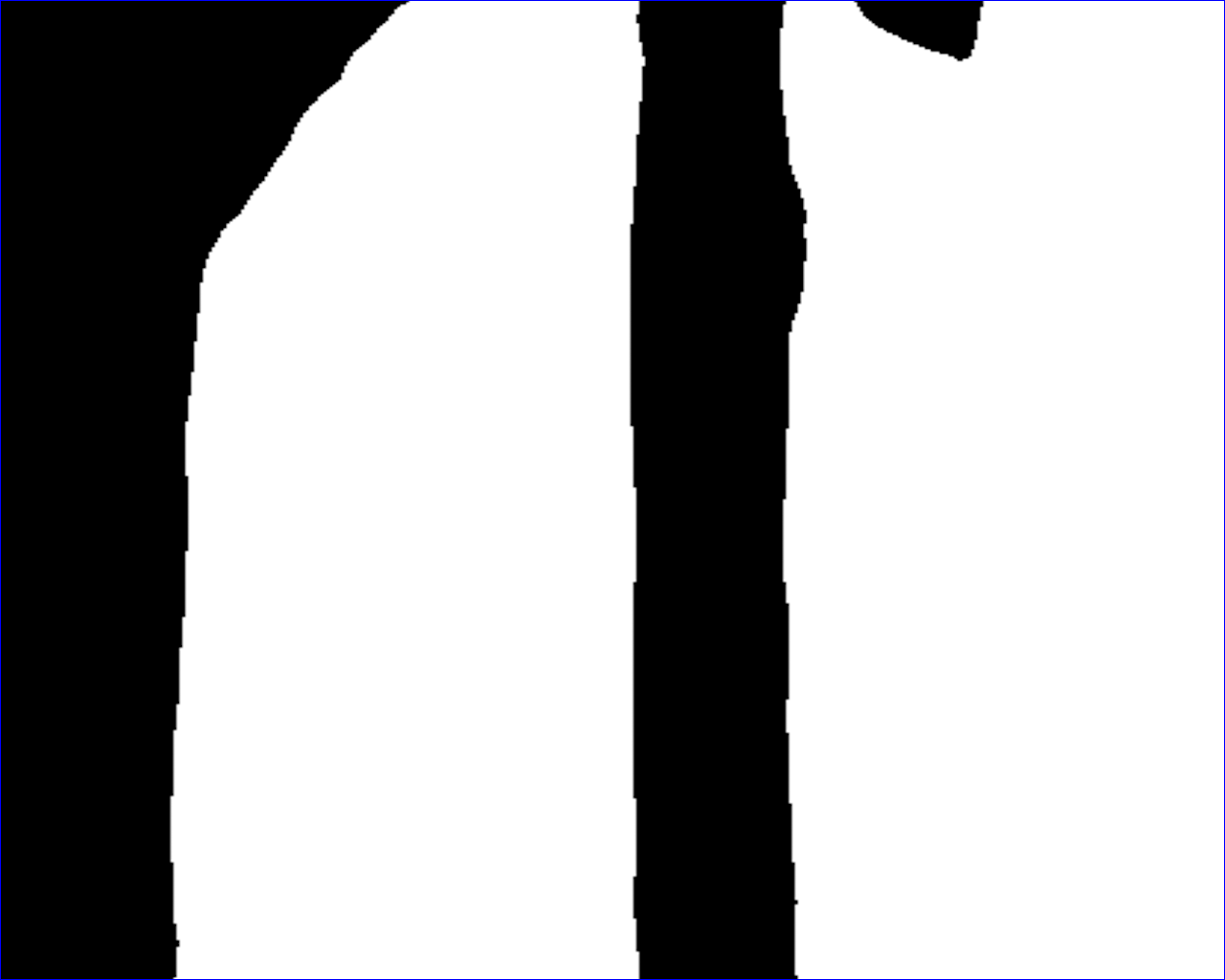}
      \end{minipage}
      }      
      \subfloat[GDNet \cite{GDNet}]{\label{img_GDNet}
      \begin{minipage}[t]{0.145\textwidth}
            \centering
            \includegraphics[width=1\linewidth]{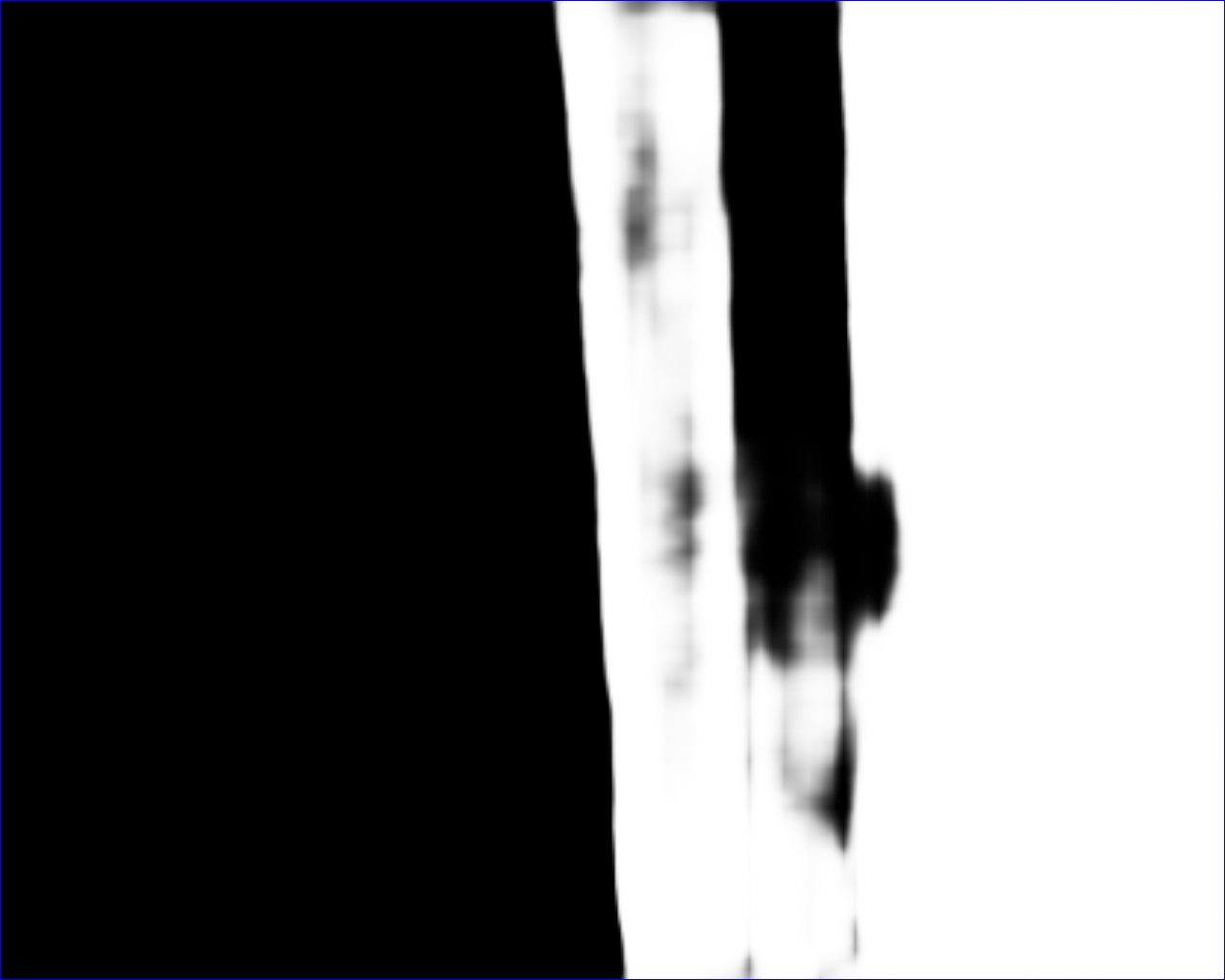}

            \includegraphics[width=1\linewidth]{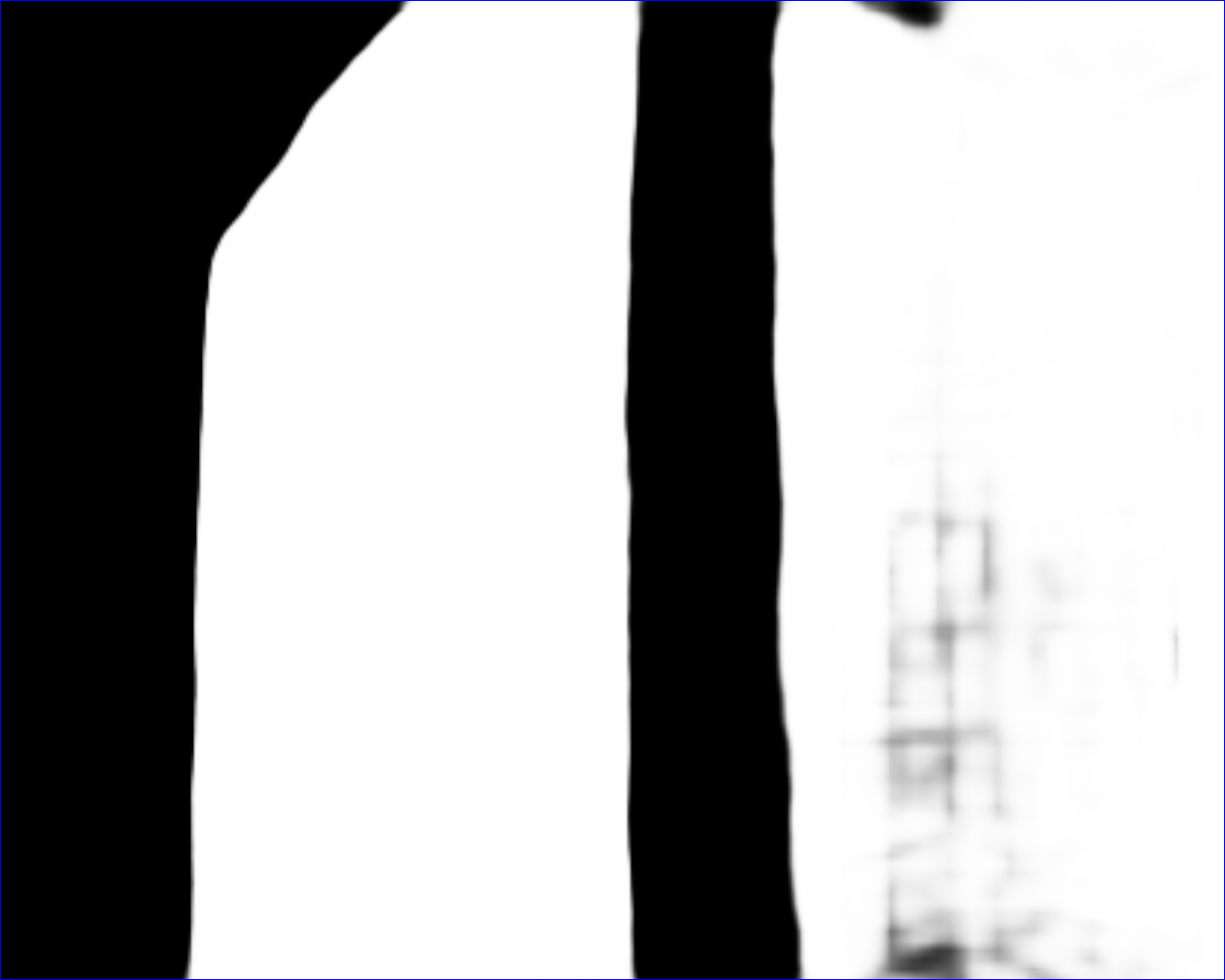}
      \end{minipage}
      }      
      \subfloat[EGNet \cite{egnet}]{\label{img_EG}
      \begin{minipage}[t]{0.145\textwidth}
            \centering
            \includegraphics[width=1\linewidth]{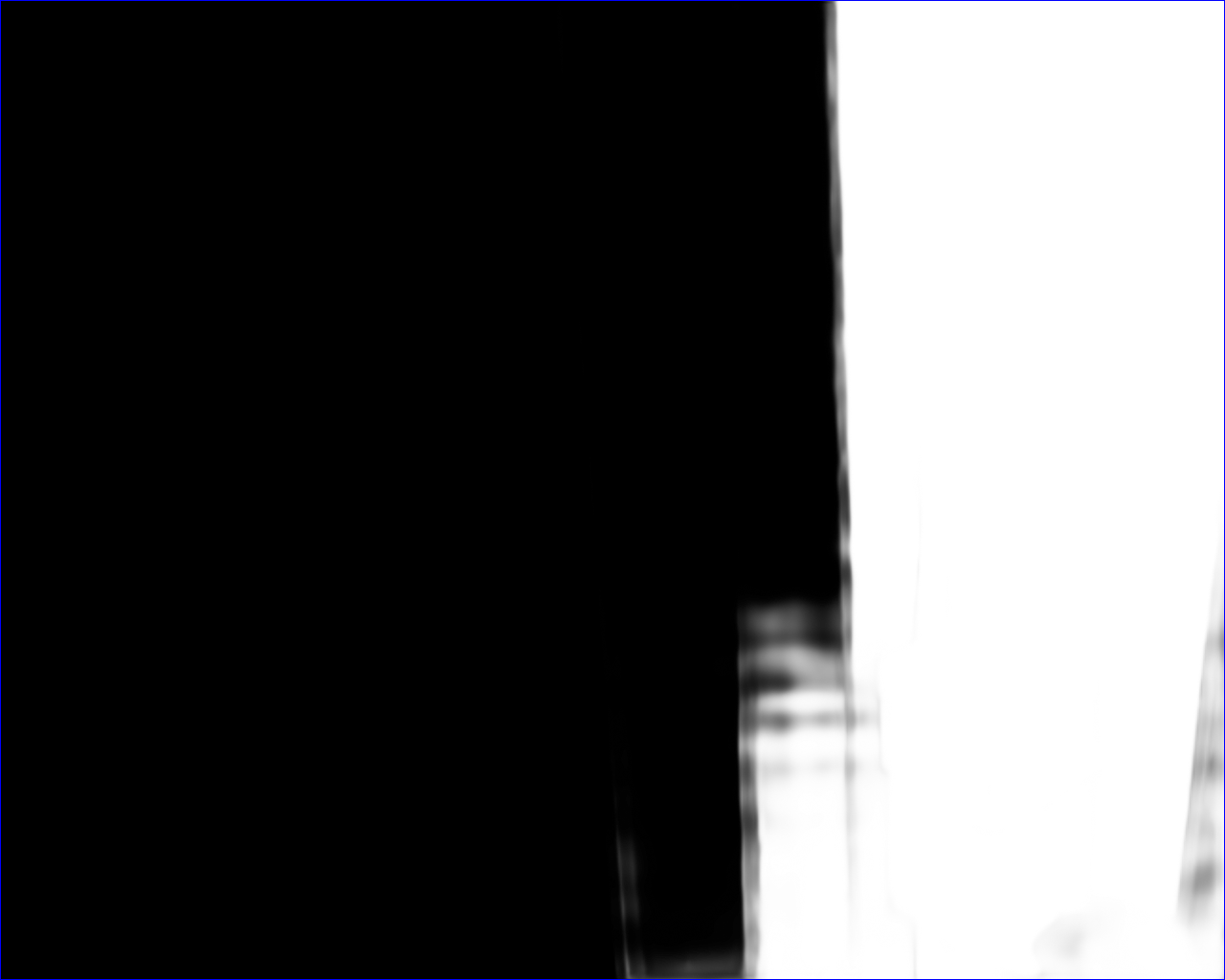}

            \includegraphics[width=1\linewidth]{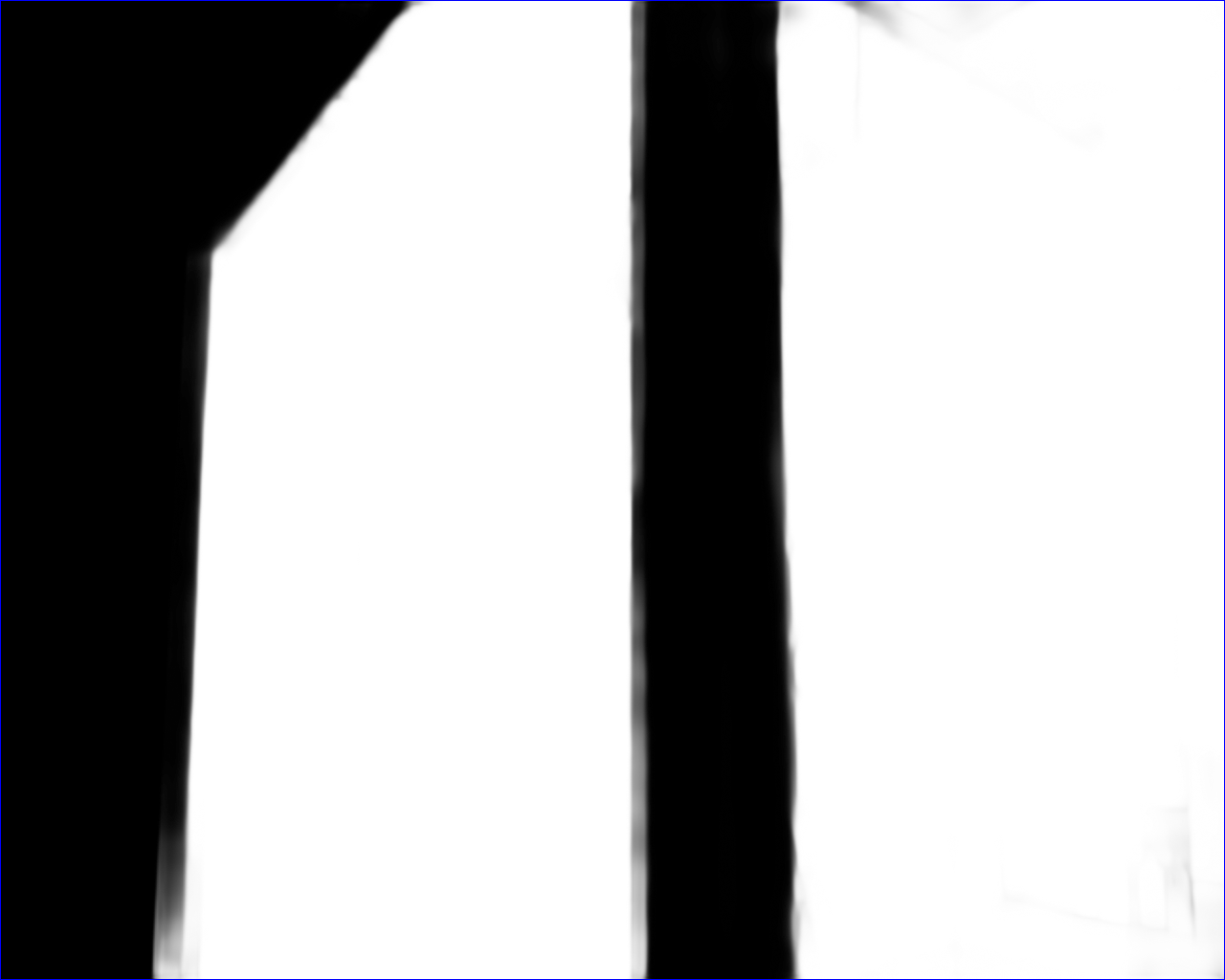}
      \end{minipage}
      }         
      \subfloat[GlassNet (ours)]{\label{img_our}
      \begin{minipage}[t]{0.145\textwidth}
            \centering
            \includegraphics[width=1\linewidth]{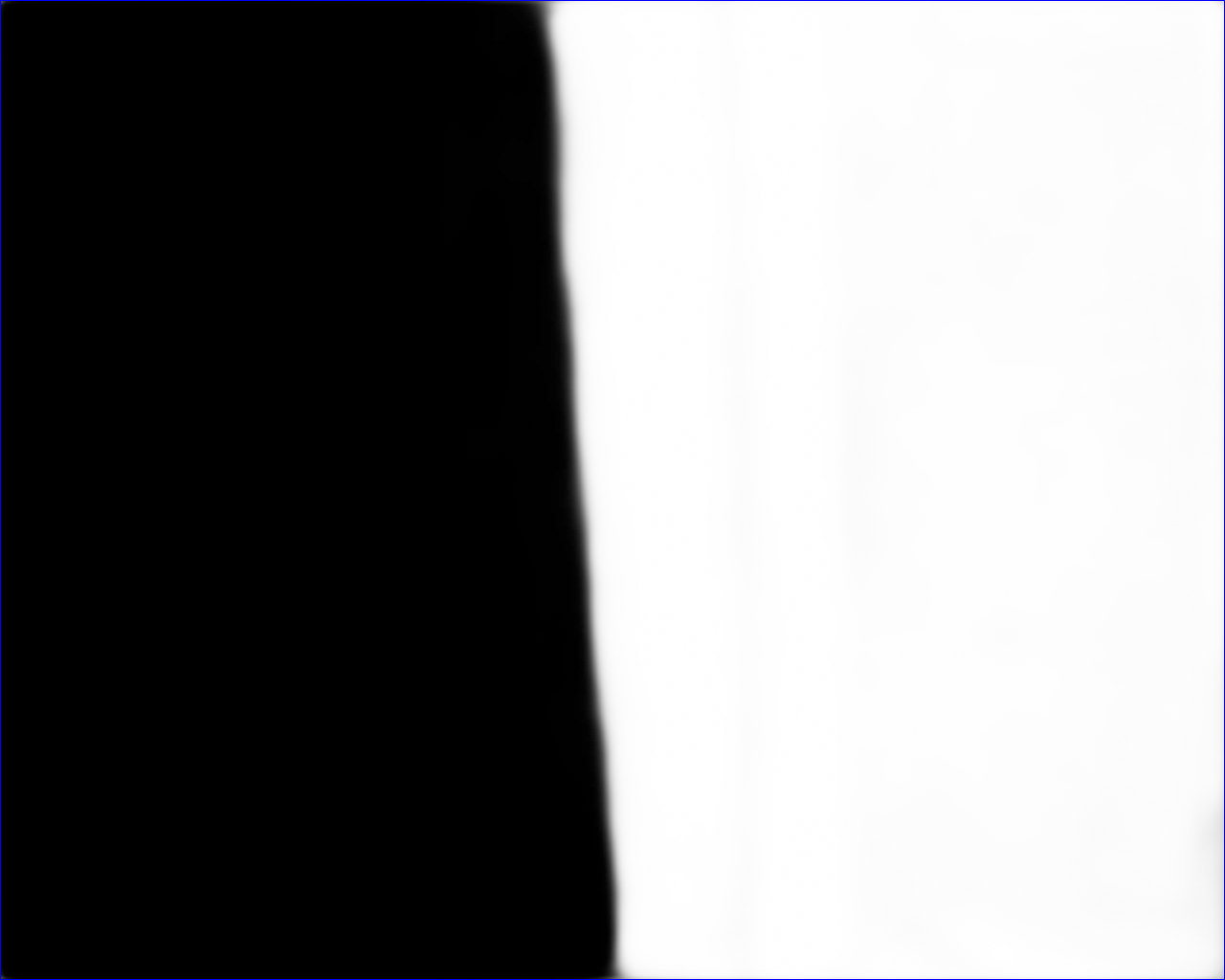}

            \includegraphics[width=1\linewidth]{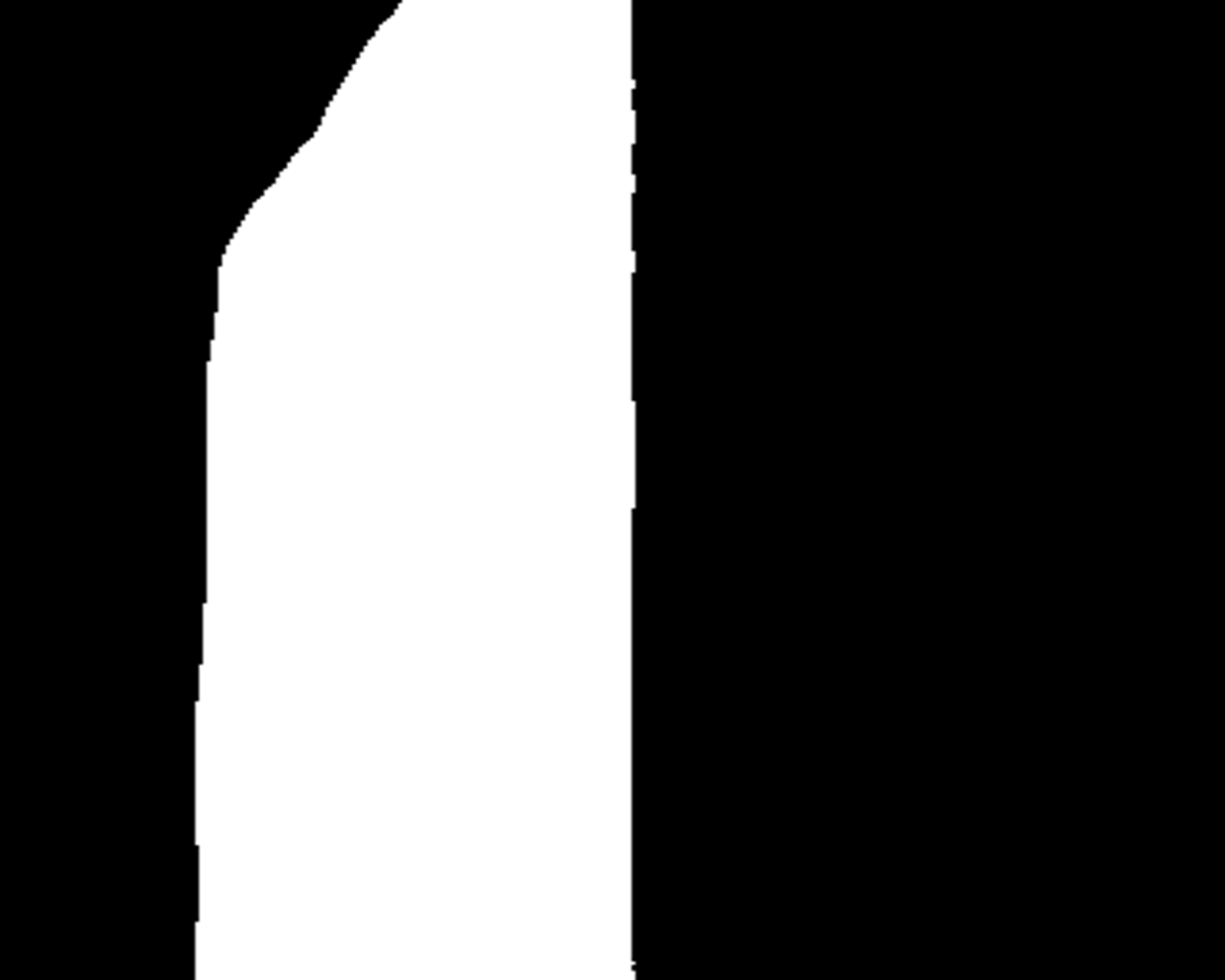}
      \end{minipage}
      }  
      \subfloat[GT]{\label{img_gt}
      \begin{minipage}[t]{0.145\textwidth}
            \centering
            \includegraphics[width=1\linewidth]{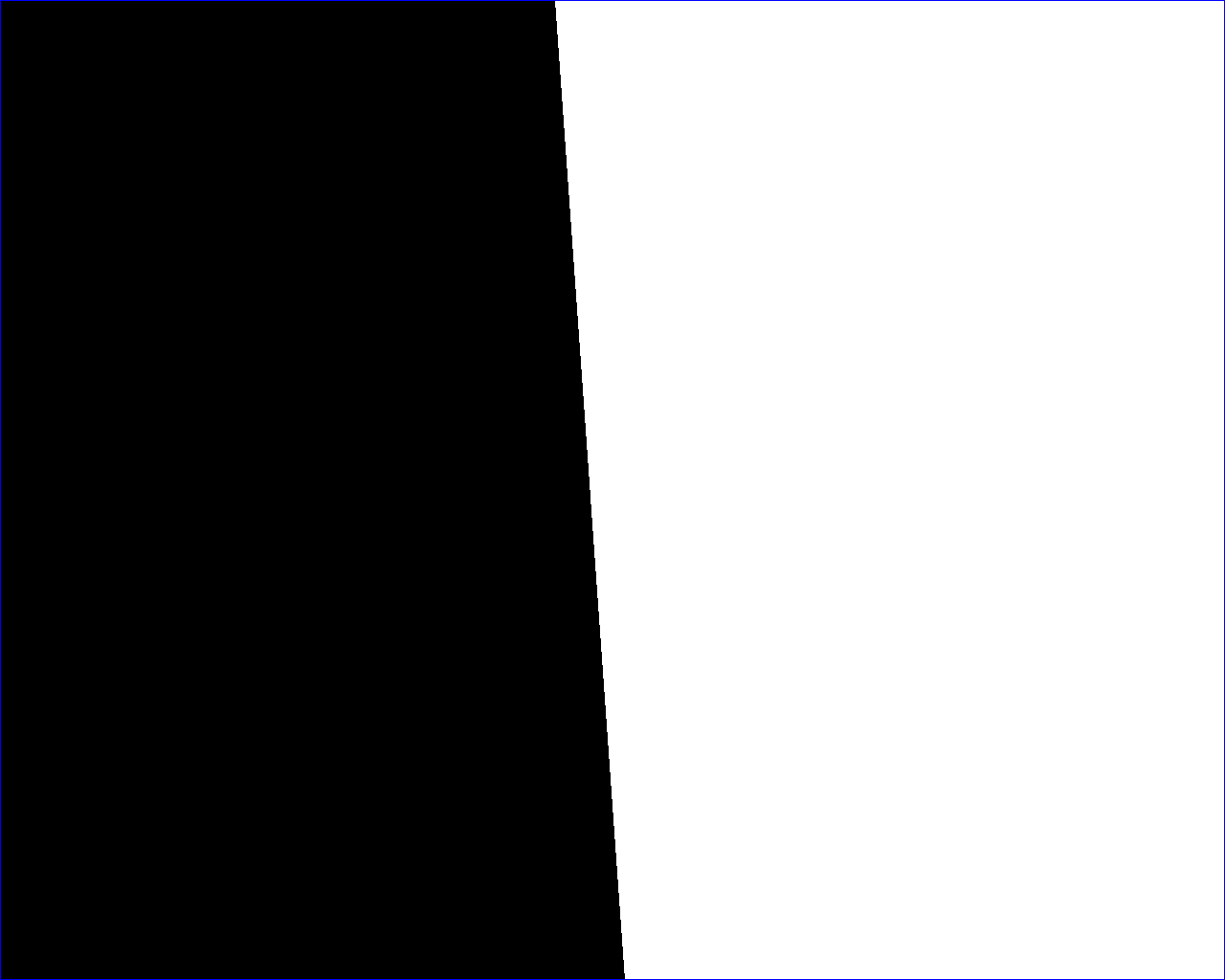}

            \includegraphics[width=1\linewidth]{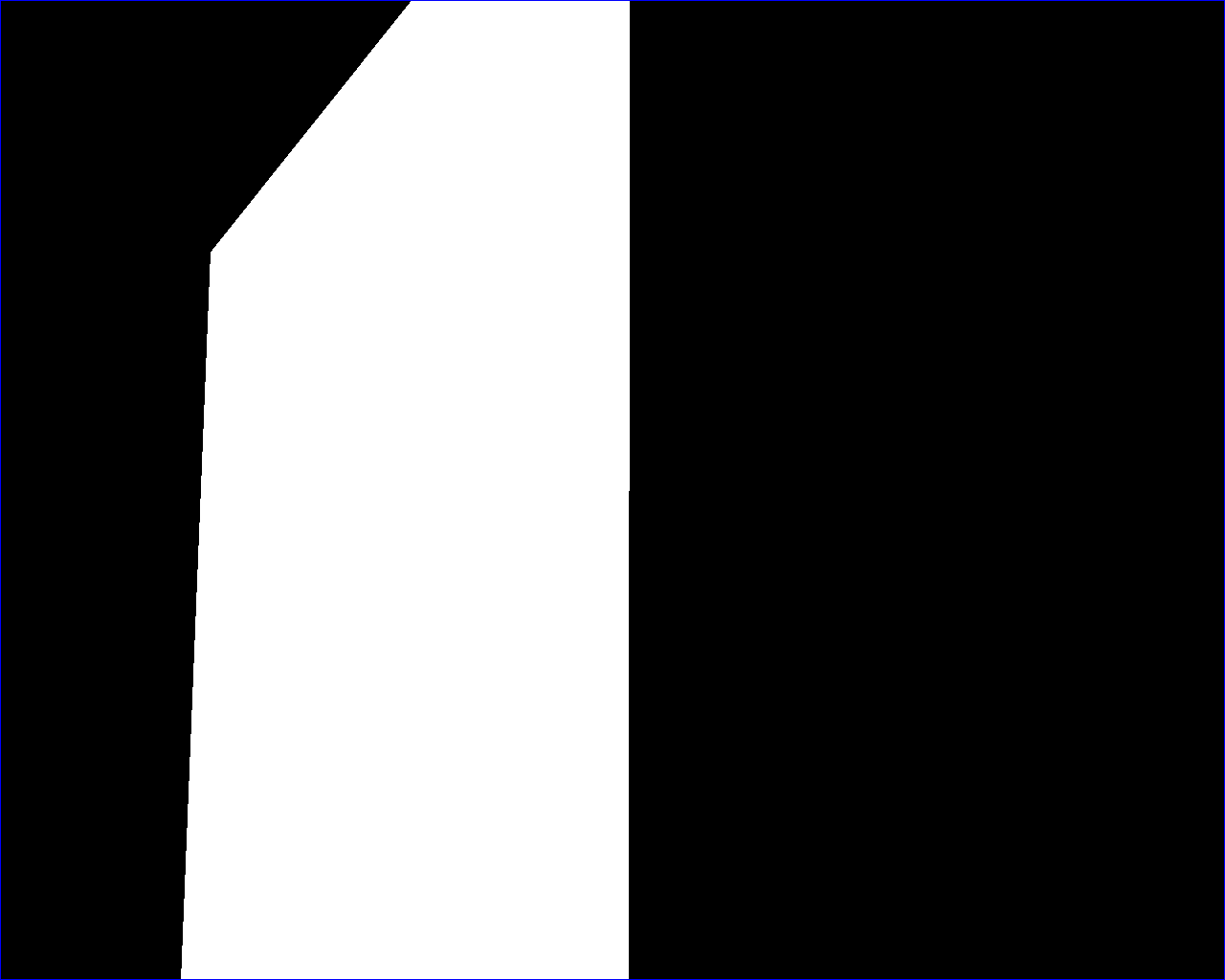}
      \end{minipage}
      }  
      \caption{\label{img1}
       GlassNet compares with its competitors on the public GDD dataset \cite{GDNet}. Current vision systems sense the presence of glass poorly, since a glass region has no fixed patterns (e.g., various objects will appear behind the glass, resulting in the same appearance of the glass and the objects behind the glass in an image). DANet commonly fails to detect the glass; GDNet leads to inaccurate glass boundaries; EGNet yields wrong detection regions; while the proposed GlassNet operates smoothly on the two challenging images, where the glass is exactly detected and its boundaries are clearer. Please note that the white region corresponds to the detected glass, and the black region means the detected background.}
}
\maketitle
\begin{abstract}
Most of the existing object detection methods generate poor glass detection results, due to the fact that the transparent glass shares the same appearance with arbitrary objects behind it in an image. Different from traditional deep learning-based wisdoms that simply use the object boundary as auxiliary supervision, we exploit label decoupling to decompose the original labeled ground-truth (GT) map into an interior-diffusion map and a boundary-diffusion map. The GT map in collaboration with the two newly generated maps breaks the imbalanced distribution of the object boundary, leading to improved glass detection quality. We have three key contributions to solve the transparent glass detection problem: (1) We propose a three-stream neural network (call GlassNet for short) to fully absorb beneficial features in the three maps. (2) We design a multi-scale interactive dilation module to explore a wider range of contextual information. (3) We develop an attention-based boundary-aware feature Mosaic module to integrate multi-modal information. Extensive experiments on the benchmark dataset exhibit clear improvements of our method over SOTAs, in terms of both the overall glass detection accuracy and boundary clearness.


\end{abstract}  

\section{Introduction}

Transparent glass is widely used in our daily life, such as glass windows/doors and many other glass products. 
However, it commonly hinders many vision-related tasks like depth prediction, instance segmentation, refection removal, and object detection, etc. For example, when an intelligent robot or an unmanned plane operates automatically, they should avoid crashing into the glass. It is, therefore, essential to accurately detect the overall glass with its boundary clearly from single images. Unfortunately, most of the existing object detection methods generate inaccurate or even wrong regions of the glass with fuzzy boundaries, due to the fact that the glass is transparent. That means, a glass region nearly has no fixed patterns; the pattern is determined by the arbitrarily appeared object behind the glass. Therefore, unlike many other objects which have relatively fixed patterns to detect more easily, the same appearance between the glass region and the objects behind it makes existing object detection methods work ineffectively. We list three representative detection approaches in Figure \ref{img1}, i.e., DANet \cite{DANet} for semantic segmentation, EGNet \cite{egnet} for edge-guided salient object detection, GDNet \cite{GDNet} for glass detection, as well as our proposed GlassNet. As shown, DANet wrongly considers the background as the glass; EGNet also yields wrong detection regions; although GDNet \cite{GDNet} pioneers to automatically detect glass from single images,
it leads to inaccurate glass boundaries; while the proposed network operates smoothly on these two challenging images: the glass is exactly detected with its clearer boundaries by GlassNet.

Intuitively, like other vision tasks, a straightforward solution to enhance the glass detection ability, is to use boundaries of the glass as auxiliary supervision.
However, in an image with glass in it, the glass-boundary pixels are much rarer than other pixels. Such a very unbalanced distribution of the glass-boundary pixels will introduce large prediction errors around the glass boundary. To this end, we arise an intriguing question that if the glass boundary diffuses itself into the glass's interior and the interior diffuses itself from its center to boundary, a deep network can better focus on regions around the glass boundary and concentrate on center areas of the glass object? To answer it, 1) we first use label decoupling (LD) \cite{LDF} to explicitly decompose the original glass map into an interior-diffusion map and a boundary-diffusion map, where the first map is concentrated in the center of glass objects and the second map focuses on regions around glass boundaries; 2) based on the three different types of label information, we propose a three-stream neural network for robust glass detection (GlassNet). For the interior-diffusion stream, we only use the highest two-level image features with rich semantic information to locate the glass region; for the boundary-diffusion stream, all levels of information are aggregated to make the detection result more accurate; for the original glass stream, we utilize the lowest two-level image features with more detailed information and highest-level image features to predict the final glass maps.

Meanwhile, we design a multi-scale interactive dilation module with a large receiving field to integrate the features from adjacent levels. And we propose an attention-based boundary fusion module to merge the boundary and glass features.
We have tested all the approaches on the benchmark dataset GDD \cite{GDNet} and our GlassNet achieves a very competitive performance. In summary, our contributions are mainly four-fold:
\begin{itemize}
	\item We observe that in an image with glass in it, the glass-boundary pixels are much rarer than other pixels. Such a very imbalanced distribution of the glass-boundary pixels introduces large prediction errors around the glass boundary when performing object detection. To break such an imbalanced distribution between glass-boundary pixels and non-glass-boundary pixels, we utilize the label decoupling procedure to decompose a glass label into an interior-diffusion map and a boundary-diffusion map to supervise the network training.
	\item We propose a three-stream network, called GlassNet, which is enhanced by label decoupling features to produce more precise glass maps.
	\item We design a multi-scale interactive dilation module to explore a wider range of contextual information and an attention-based boundary-aware feature Mosaic module to integrate multi-modal information.
	\item Extensive experiments on the benchmark dataset exhibit clear improvements of our method over SOTAs, in terms of both the overall glass detection accuracy and boundary clearness.
\end{itemize}

\section{Related Work}

In the past two years, glass detection had begun to attract mucch attention, but little work has been done on this topic. 
In this section, we briefly introduce the methods used in glass detection and the methods that can assist in solving this problem from relevant fields, including semantic segmentation, salient object detection, and mirror detection. 

\textbf{Semantic segmentation.} 
Semantic segmentation is a key problem in the computer vision community, which aims at assigning semantic class labels to each pixel in the given image. 
With the development of deep neural networks, an end-to-end training architecture method called fully convolutional networks (FCNs) \cite{FCN} has been proposed to solve this problem, which uses multi-scale context fusion to achieve high segmentation performance. 
However, the fixed geometric structures of convolution operations in those deep neural networks make the pixels capture local information and short-range contextual information inherently. 
Thus, Chen et al. \cite{DeepLab} introduce an atrous spatial pyramid pooling module (ASPP) with multi-scale dilation convolutions for contextual information aggregation. 
Zhao et al. \cite{PSP} further propose PSPNet to capture a wider range of contextual information by using a pooling operation and the pyramid structure. 
In addition, the encoder-decoder structures, like U-Net \cite{UNet}, are widely used to fuse middle-level and high-level semantic features. 

However, the dilated convolution-based methods \cite{CCF, RAC} fail to capture global contextual information and cause sparse local information due to their structures. 
The pool-based methods \cite{CE, SAC} aggregate context information in a non-adaptive way to make image pixels use the homogeneous contextual information. 
Therefore, Wang et al. \cite{non-local} introduce non-local networks utilizing a self-attention mechanism \cite{attention, LSM}, which calculate the relationship between each pixel and all other pixels in an image, thus harvesting global contextual information. 
To solve the problem that self-attention based methods have high computation complexity and occupy a huge amount of GPU memory, Huang et al. \cite{CCNet} and Fu et al. \cite{DANet} respectively propose CCNet and DANet to reduce the parameters.
After that, Carion et al. \cite{Transformer} adopt Transformer that is widely used in the NLP field, which replaces the convolution layer with the self-attention layer, for semantic segmentation.

\textbf{Salient object detection (SOD).} 
SOD aims at identifying the most visually distinctive objects or regions in an image, which is widely applied as a pre-processing procedure for downstream tasks \cite{rdcfp, ids}. Early salient object detection methods are mainly based on hand-crafted features (e.g., color, texture, and contrast) to segment salient objects in the scene \cite{gmr, rbd, om, mb}. Recent convolutional neural networks (CNNs) \cite{imgnet, vdcn, ResNet} are extensively used and achieve very remarkable performance. 

Ronneberger et al. \cite{UNet} propose U-Net, a representative network widely used in a variety of graphics processing tasks, which effectively generates more accurate detection results by using a skip connection operation and an encoder-decoder structure. 
Based on U-Net, many other methods adopt different decoders, combined with multi-level CNN features, and have achieved remarkable performance. 
Zhang et al. \cite{Amulet} introduce an AmuletNet for salient object detection that aggregates an another-level convolutional feature at each different level. 
Zhang et al. \cite{PAGR} add an attention module to the decoder, which can guide the network to selectively integrate multi-level features. 
Zhao et al. \cite{pyramid} propose a pyramid feature attention network (PFAN) to enhance the high-level context features and the low-level spatial structural features. 
Pang et al. \cite{MINet} propose aggregate interaction modules to integrate the features from adjacent levels by using a more complex decoder structure. 
Besides, more efforts utilize boundary information to improve the accuracy of saliency maps. 
Zhao et al. \cite{egnet} focus on the complementarity between salient edge information and salient object information and present an edge guidance network (EGNet) for salient object detection. 
Zhou et al. \cite{ITSD} analyze the correlation between saliency and boundary and introduce an interactive two-stream decoder to explore multiple cues, including saliency, boundary, and their correlation. 
Furthermore, Wei et al. \cite{LDF} propose a label decoupling framework (LDF) that exploits more boundary information to enhance salient object detection performance.  

\textbf{Mirror detection.} 
Similar to other image detection tasks, mirror detection aims at segmenting mirror regions in single images. 
Yang et al. \cite{MirrorNet} make the first attempt to automatically detect mirrors and propose MirrorNet by utilizing inconsistencies between the inside and outside of the mirror region, called contextual contrasted features, to segment mirrors from the real scene.
The reason is the performance difference between the mirror region and other non-mirror regions.
The mirror region reflects the scene in front of the mirror, which makes the semantic and low-level discontinuities often occur at the boundary of the mirror.
But not all mirrors have a great distinction between inside and outside. 
Some of them have little contextual contrasted information.
Thus, Lin et al. \cite{PMD} introduce a model to progressively learn the content similarity between the inside and outside of the mirror while explicitly detecting the mirror boundaries. 
The scenes reflected by a mirror often exhibit similarities to scenes outside the mirror, which can aid to detect mirror regions by enlarging the receptive fields of the convolution operation. 
Glass detection is very similar to mirror detection that also has the problem of similar foreground and background.

\textbf{Transparent object detection (TOD).} 
Similar to glass detection, transparent object detection (TOD) aims to segment transparent object regions in single images. 
Xie et al. \cite{TransNet} propose a large-scale dataset for TOD named Trans10K and a novel boundary-aware segmentation method termed TransLab to address the transparent object detection problem. 
However, there exists a difference between TOD and GD: TOD is a multi-label segmentation problem, while GD is a binary segmentation problem. 
This fact indicates that TOD does not operate smoothly on the GD task and vice versa. This is why we do not compare our method with those TOD methods.

\textbf{Glass detection.} 
Glass regions in an image do not have a fixed pattern since they depend on what appears behind the glass, and the content of the glass region is the content of the background region. 
This situation makes it difficult to distinguish between the glass and the background region, even using state-of-the-art segmentation methods. 
Meanwhile, other object detection methods are also not suitable for glass detection tasks on account of the difference between glass and other objects. 
Mei et al. \cite{GDNet} pioneer to propose a novel glass detection network (GDNet) by exploring abundant contextual features from a large receptive field. 
They utilize multiple well-designed large-field contextual feature integration modules for the precise positioning of the glass region, but this method has poor performance in some cases where the glass boundary region or scene is very complex or the background inside and outside the glass is insufficient. 
Lin et al. \cite{richNet} observe that humans often rely on identifying reflections to sense the existence of glass and also rely on locating the boundary in order to determine the extent of the glass. 
They propose a rich context aggregation module (RCAM) to extract multi-scale boundary features and a reflection-based refinement module (RRM) to detect reflection. 
Then, they utilize two modules for glass surface detection to solve the problem of insufficient contexts in part of the scene.

\section{Methodology}

\textbf{Motivation.} 
Due to the unbalanced distribution between boundary pixels and background pixels, only using boundary pixels for glass detection will lead to larger prediction errors of pixels close to the boundary than those far away from the glass.
Therefore, the glass boundary should diffuse itself into the glass's interior to amplify its influence. Conversely, the glass's interior should diffuse itself from the center to the boundary to loosen its influence. 
Based on this observation, we propose to decouple the glass label into the interior-diffusion component and the boundary-diffusion component, both of which are auxiliary supervisions to enhance the overall glass detection quality and boundary clearness. 
To make full use of the decoupled supervisions, we further present a three-stream network, which consists of the proposed multi-scale interactive dilation modules to effectively integrate large-field contextual features for detecting glass of different sizes. 
Also, our proposed network learns to fuse multi-modal information to further enhance the performance.

\subsection{Label Decoupling}

\begin{figure}[htb]
      \centering
      \subfloat[Boundary]{\label{edge}
      \begin{minipage}[t]{0.25\textwidth}
            \centering
            \includegraphics[width=1\linewidth]{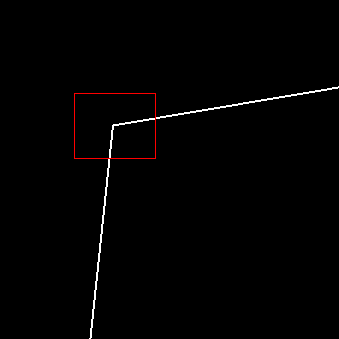}
      \end{minipage}
      }  
      \subfloat[Boundary detail]{\label{edge_detail}
      \begin{minipage}[t]{0.25\textwidth}
            \centering
            \includegraphics[width=1\linewidth]{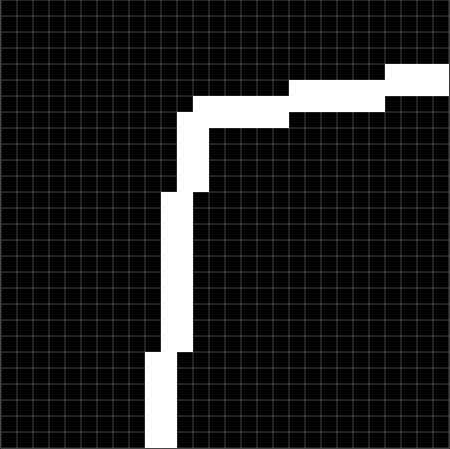}
      \end{minipage}
      }

      \subfloat[Hard example]{\label{hard_example}
      \begin{minipage}[t]{0.5\textwidth}
            \centering
            \includegraphics[width=1\linewidth]{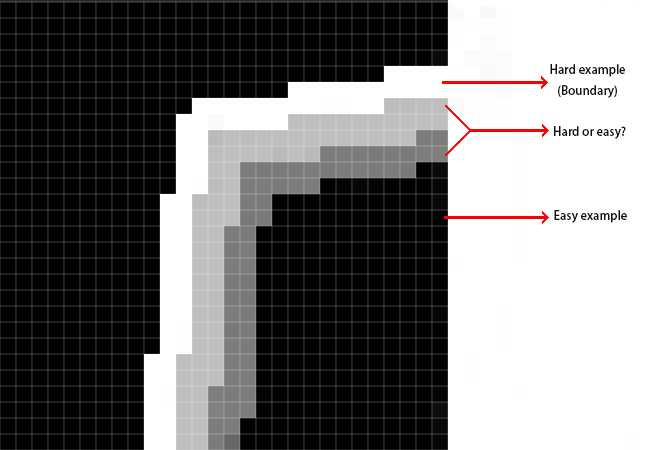}
      \end{minipage}
      }
      \caption{\label{img2}
      Illustration of hard examples: (a) the boundary obtained by the boundary extraction algorithm in EGNet \cite{egnet}, (b) the boundary detail which only has two pixels wide, and (c) it is challenging to determine the pixels near the boundary to be hard examples or not.}
\end{figure}

\begin{figure}[htb]
      \centering
      \subfloat[\centering Input]{\label{image}
      \begin{minipage}[t]{0.11\textwidth}
            \centering
            \includegraphics[width=1\linewidth]{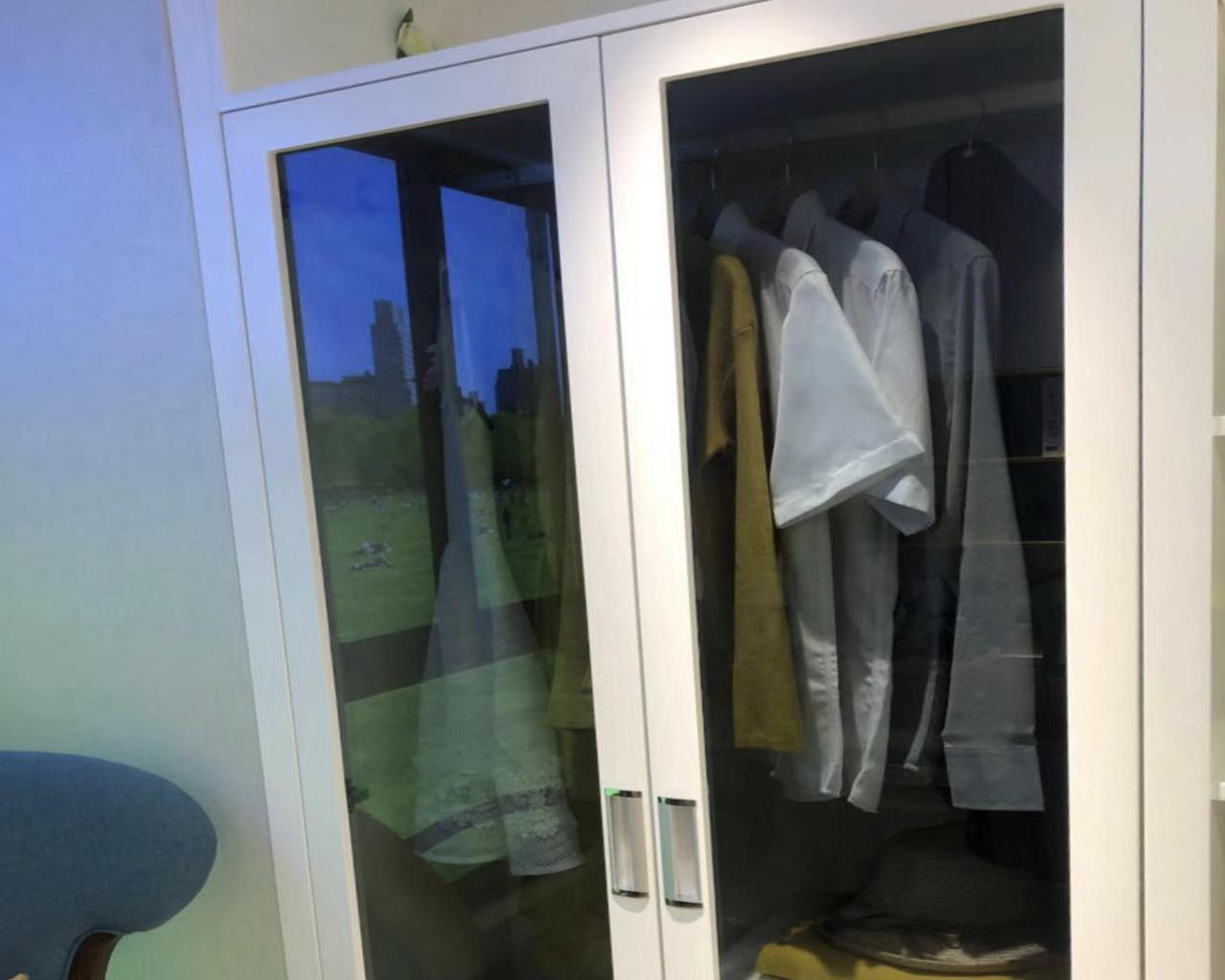}

            \includegraphics[width=1\linewidth]{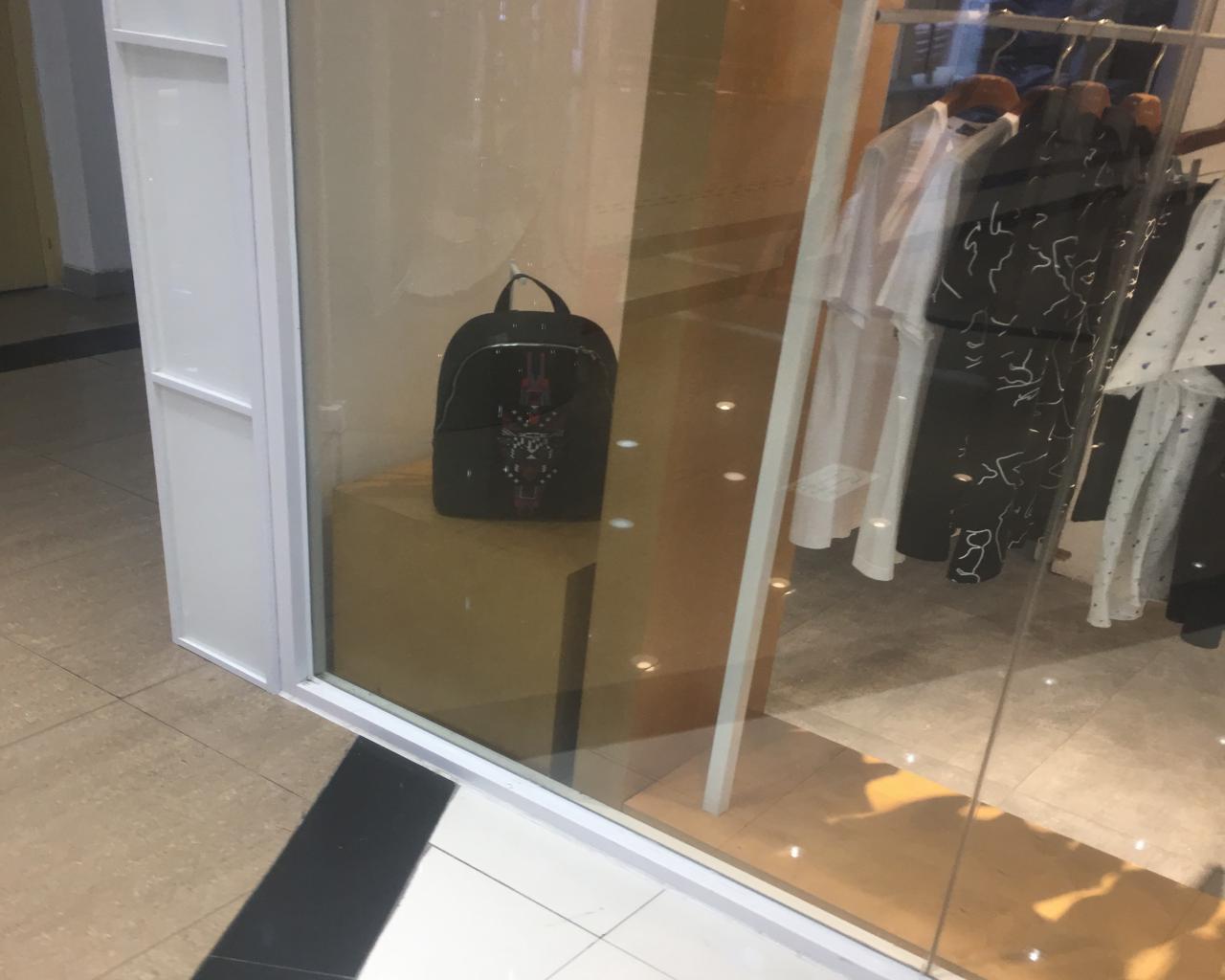}

            \includegraphics[width=1\linewidth]{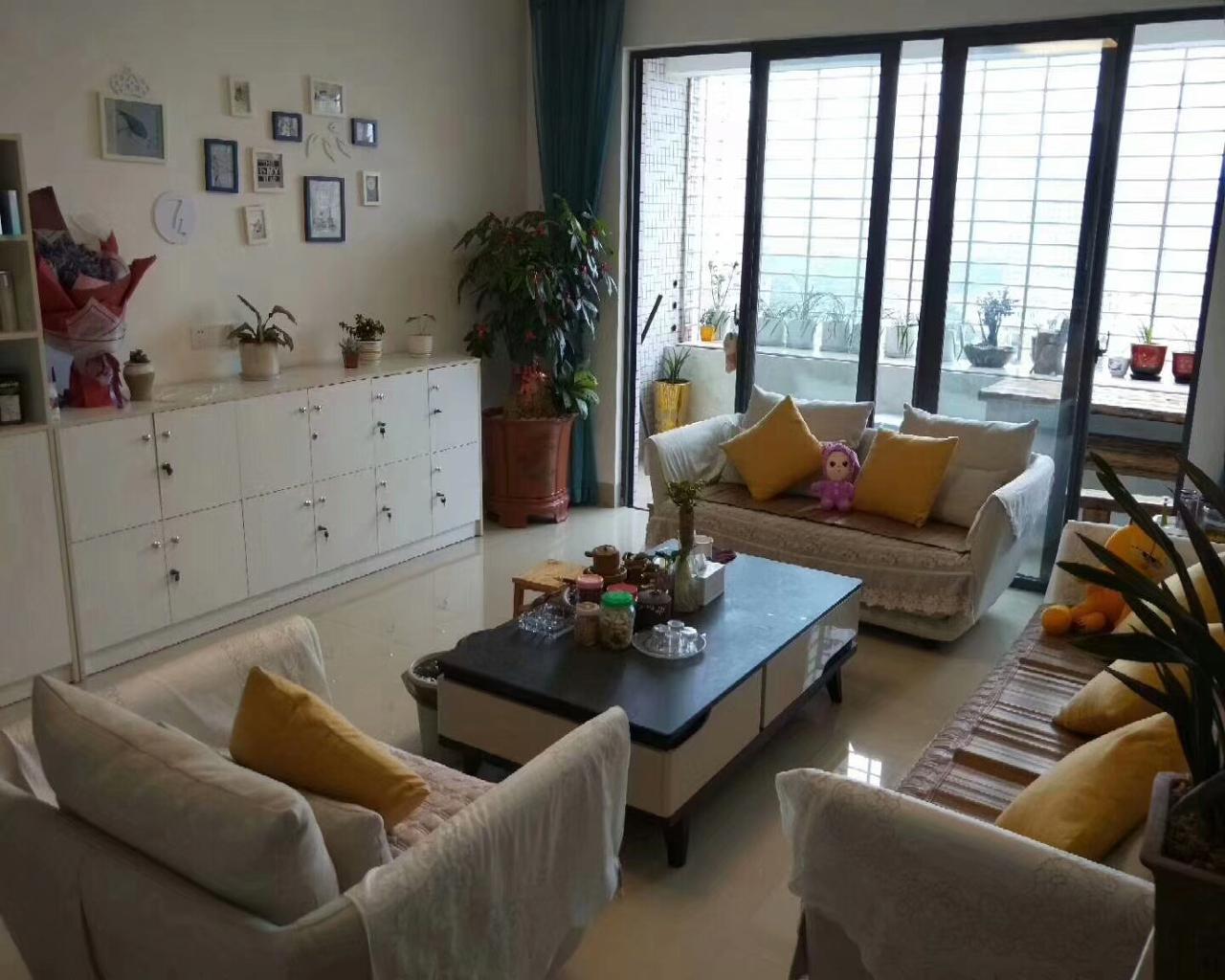}
      \end{minipage}
      }
      \subfloat[GT]{\label{GT}
      \begin{minipage}[t]{0.11\textwidth}
            \centering
            \includegraphics[width=1\linewidth]{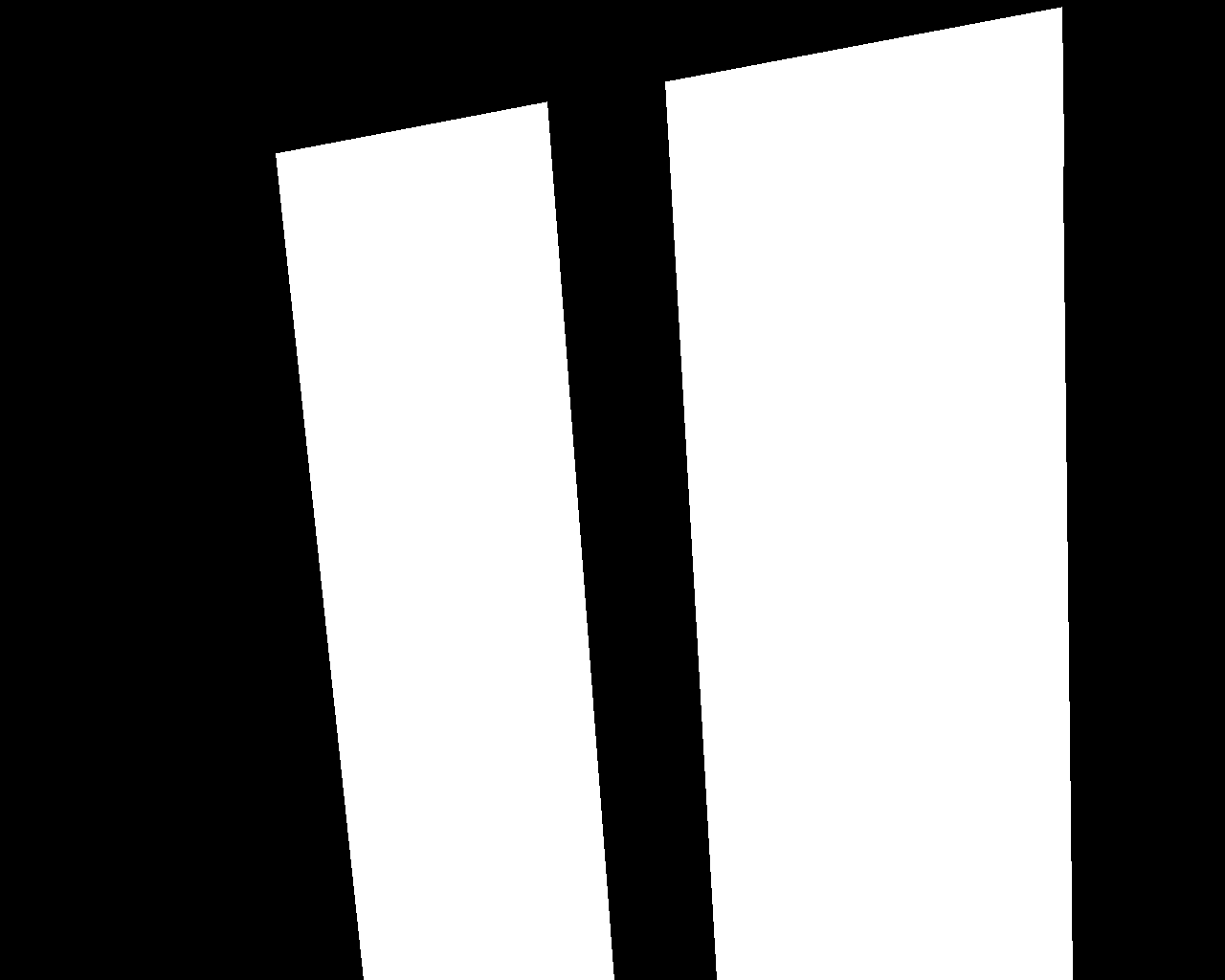}

            \includegraphics[width=1\linewidth]{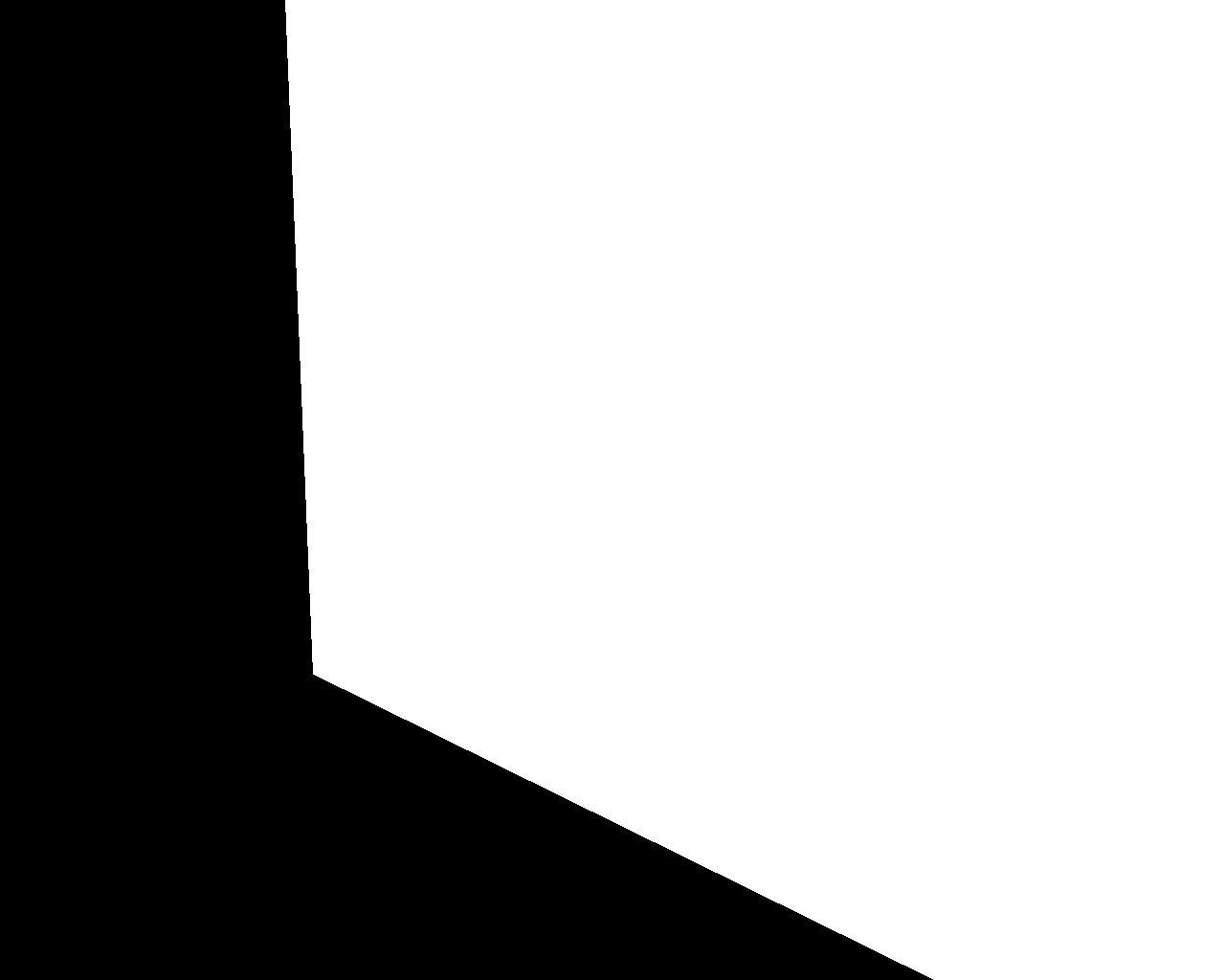}

            \includegraphics[width=1\linewidth]{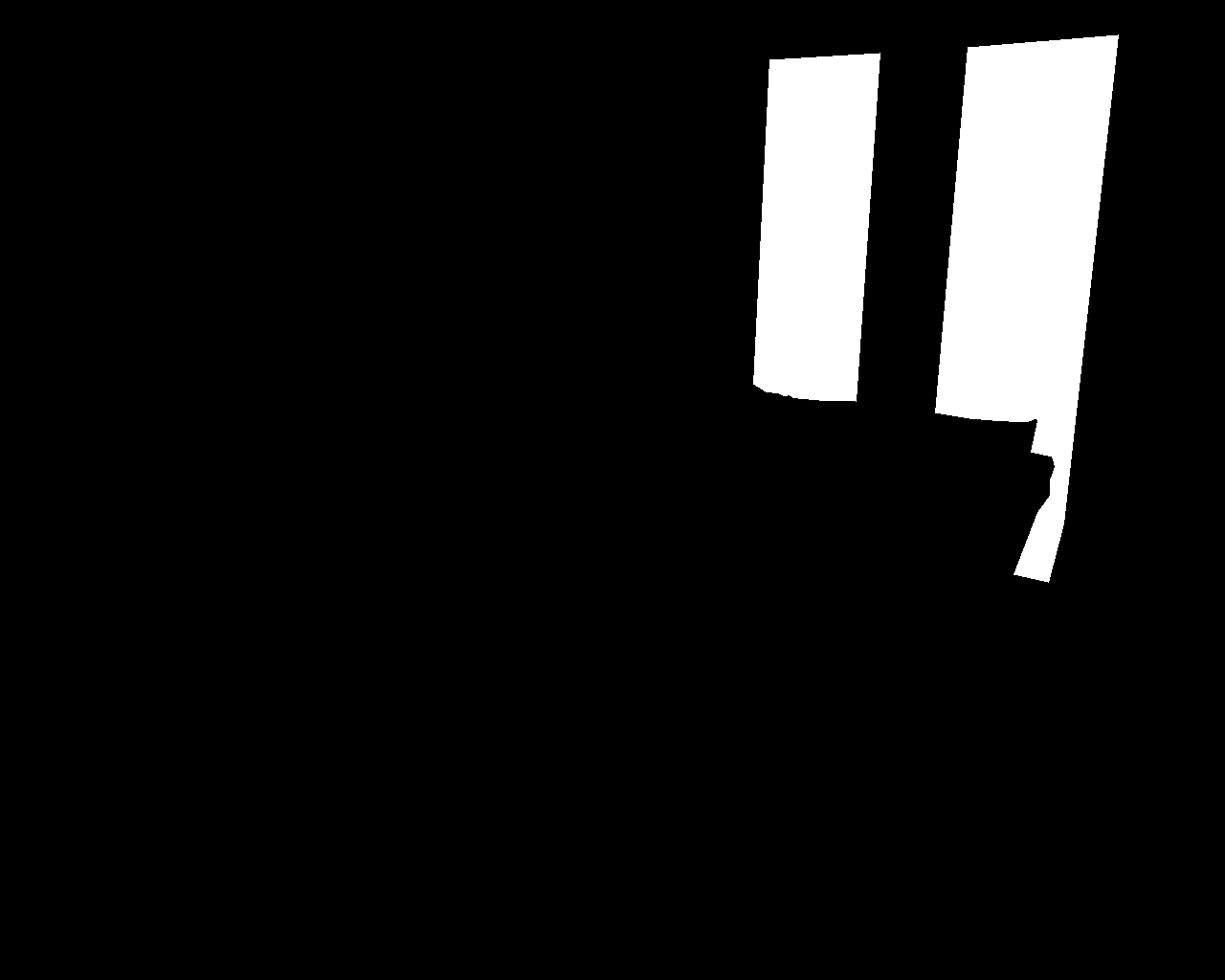}
      \end{minipage}
      }      
      \subfloat[\centering Interior-diffusion label]{\label{inner}
      \begin{minipage}[t]{0.11\textwidth}
            \centering
            \includegraphics[width=1\linewidth]{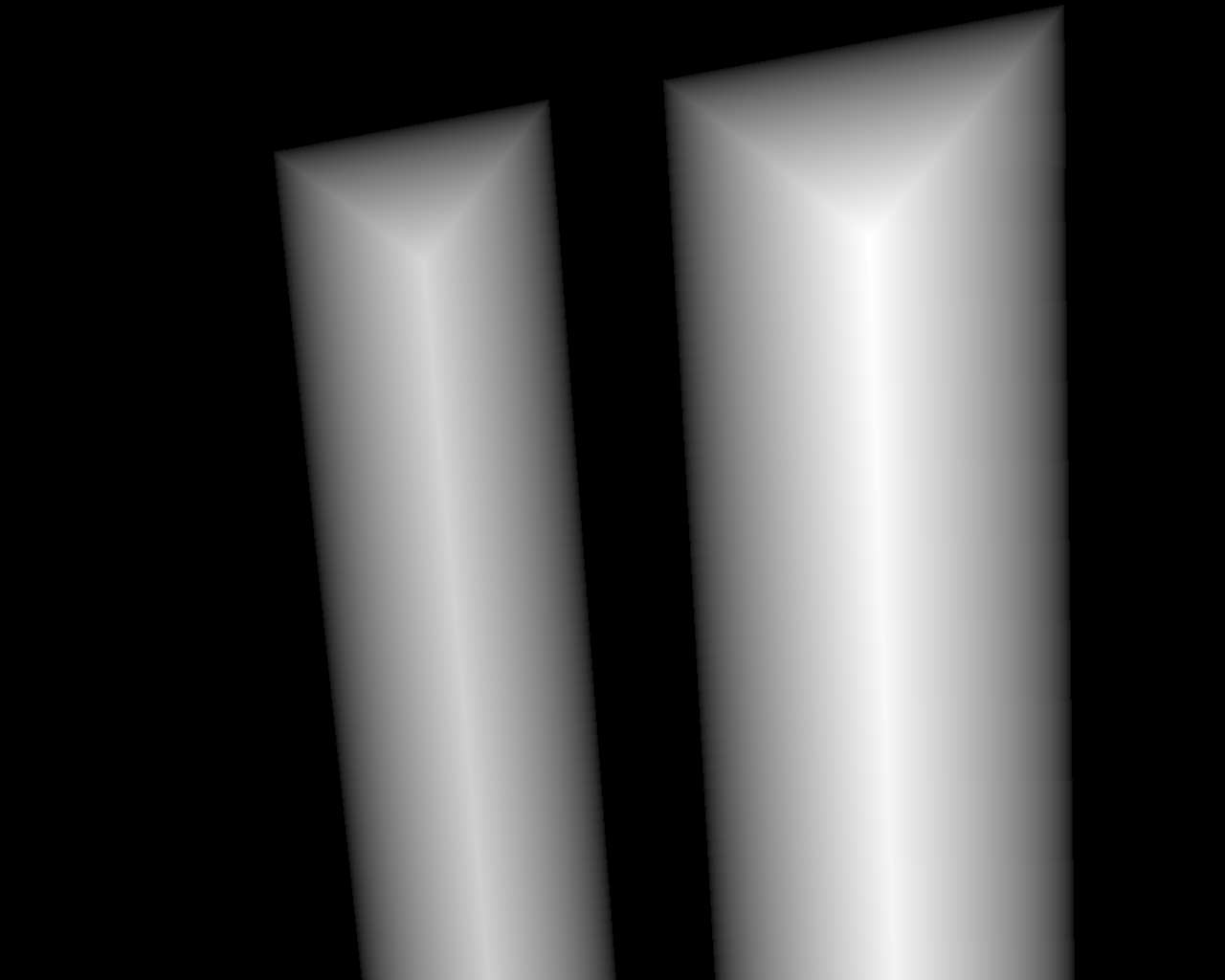}

            \includegraphics[width=1\linewidth]{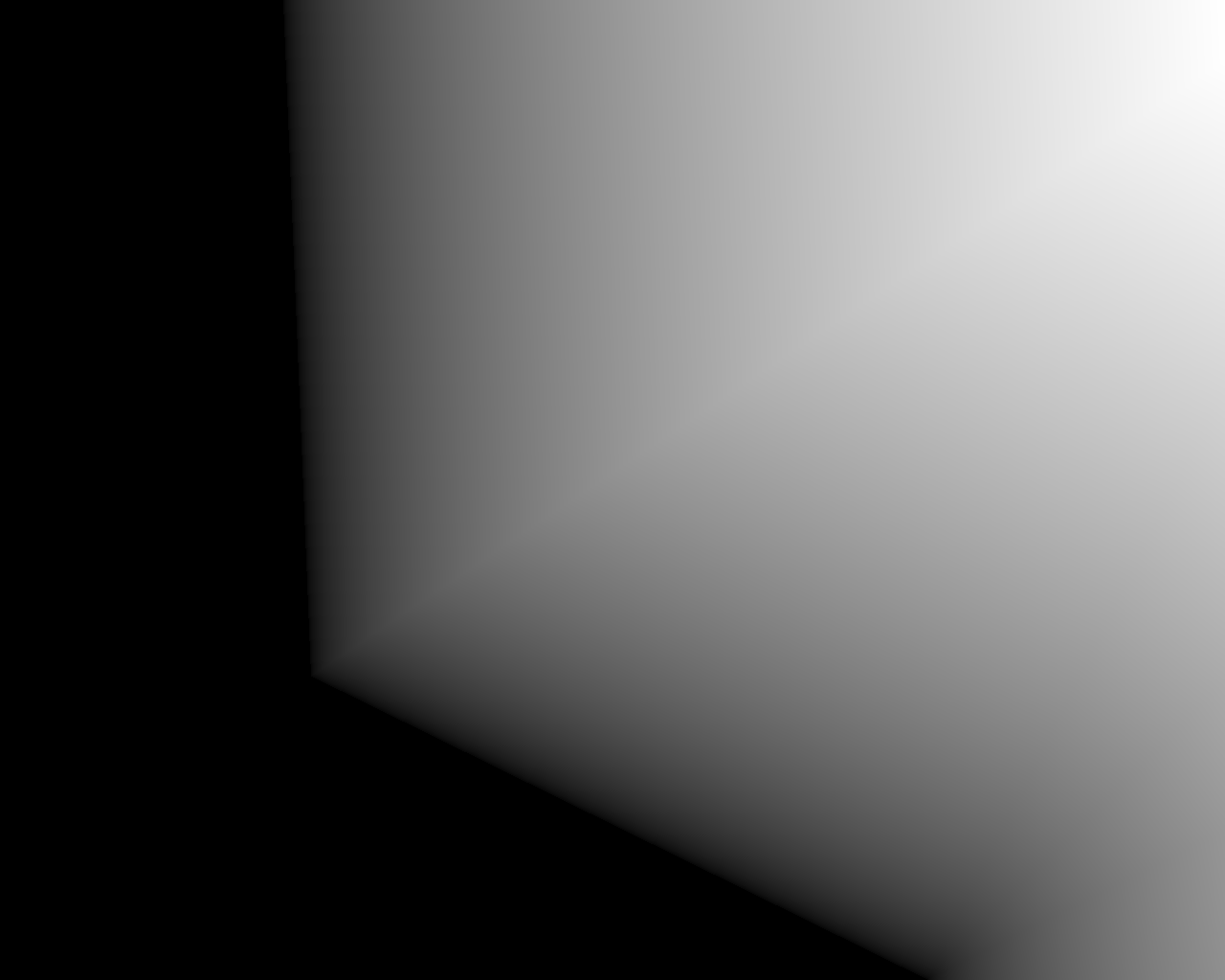}

            \includegraphics[width=1\linewidth]{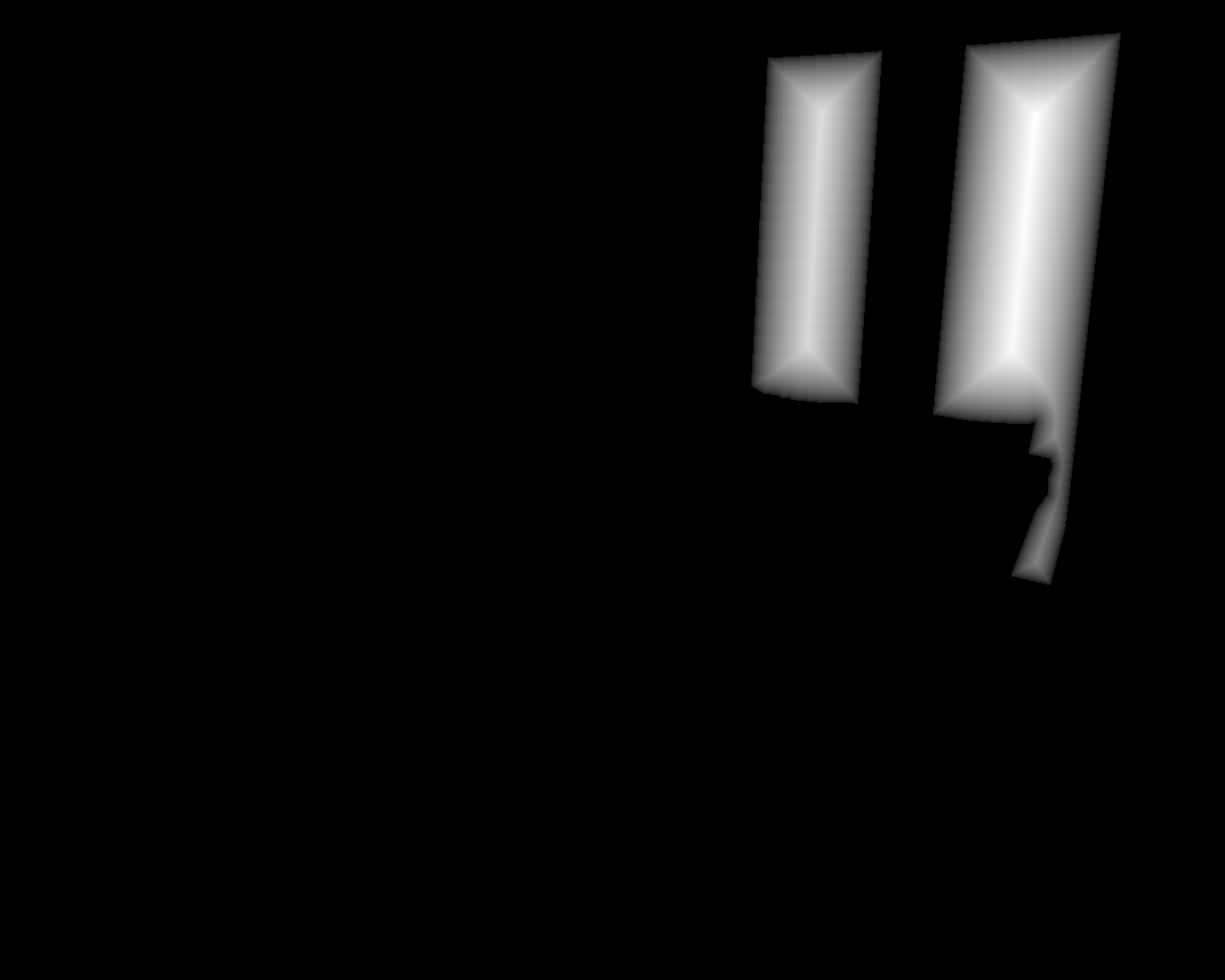}
      \end{minipage}
      }      
      \subfloat[\centering Boundary-diffusion label]{\label{boundary}
      \begin{minipage}[t]{0.11\textwidth}
            \centering
            \includegraphics[width=1\linewidth]{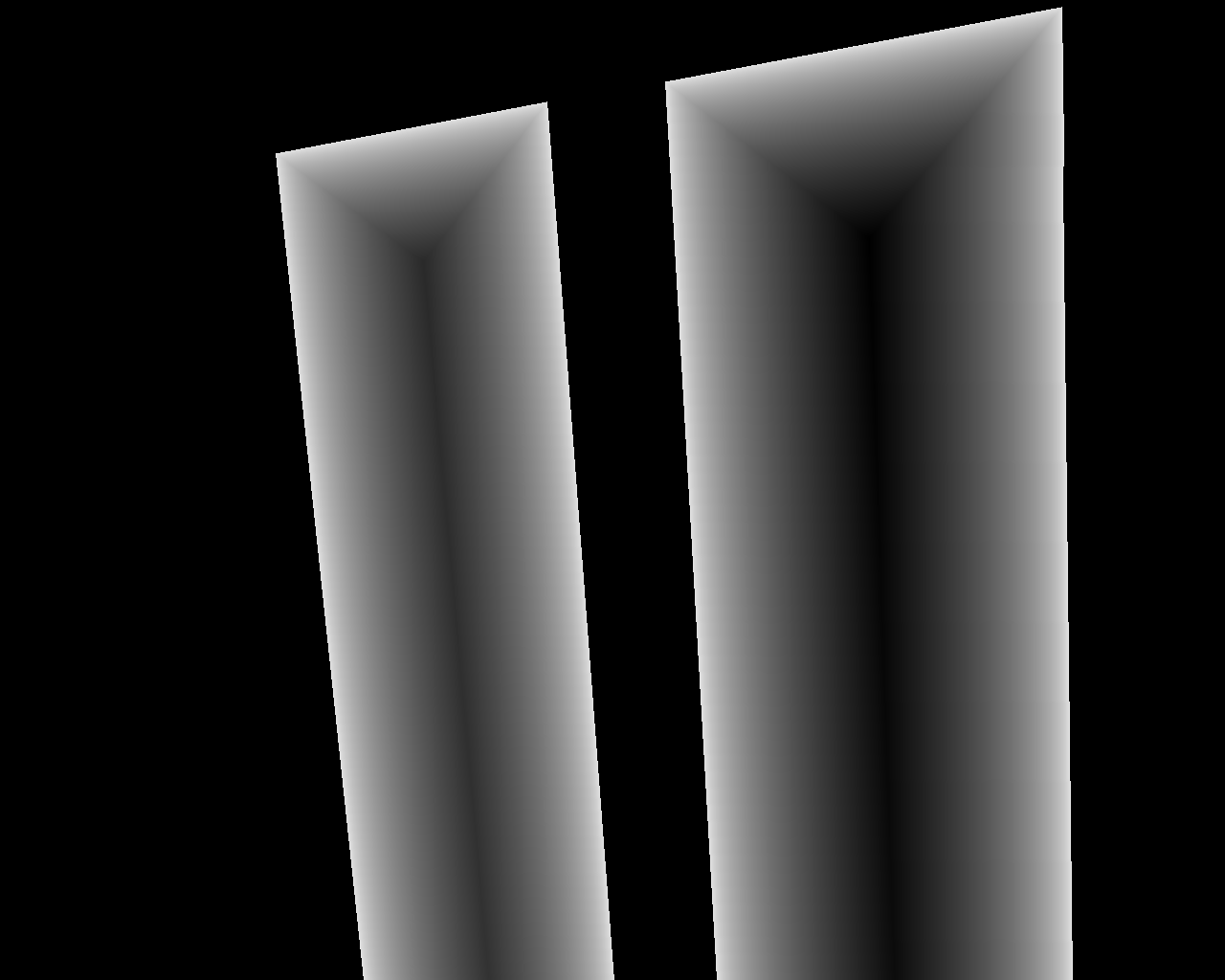}

            \includegraphics[width=1\linewidth]{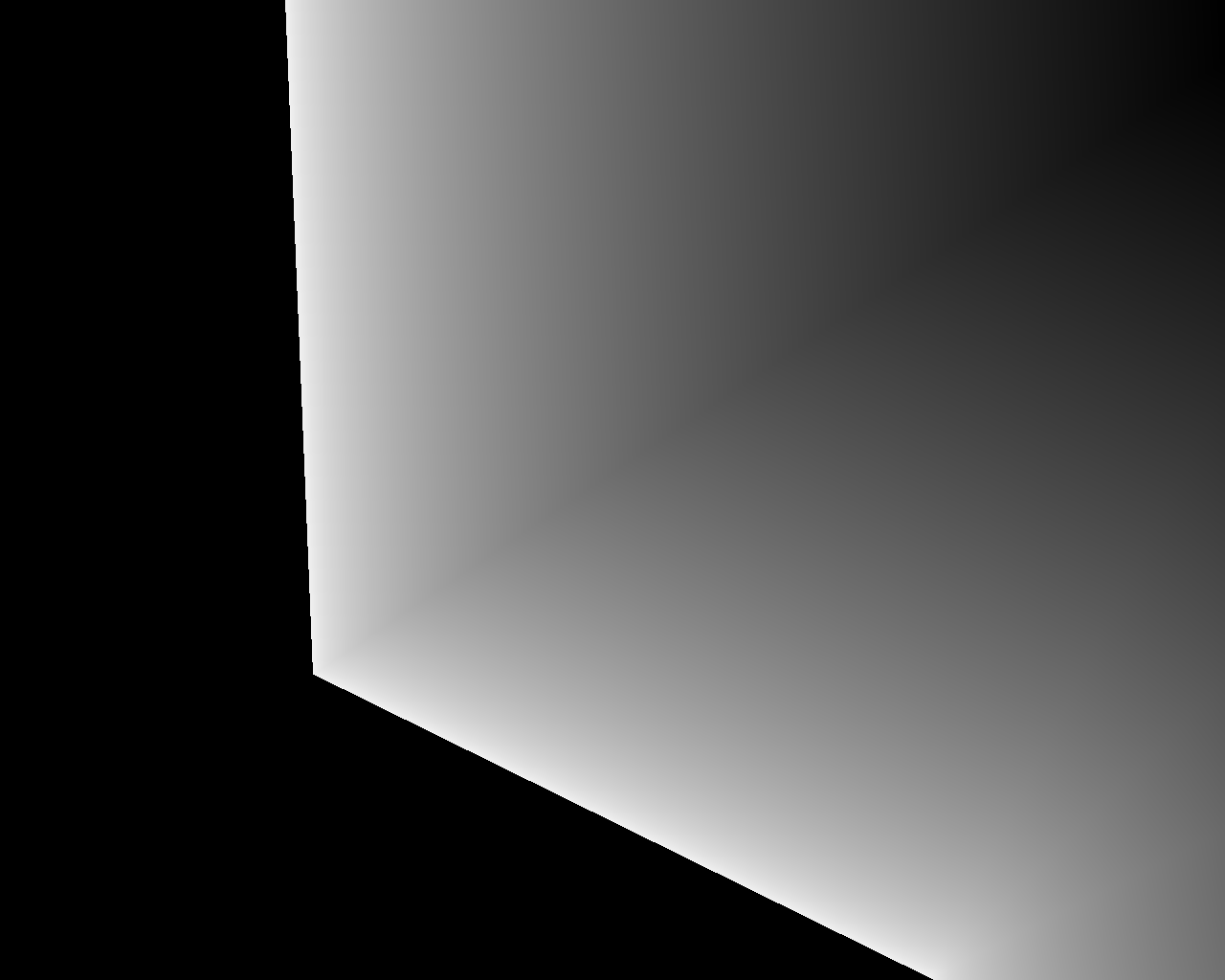}

            \includegraphics[width=1\linewidth]{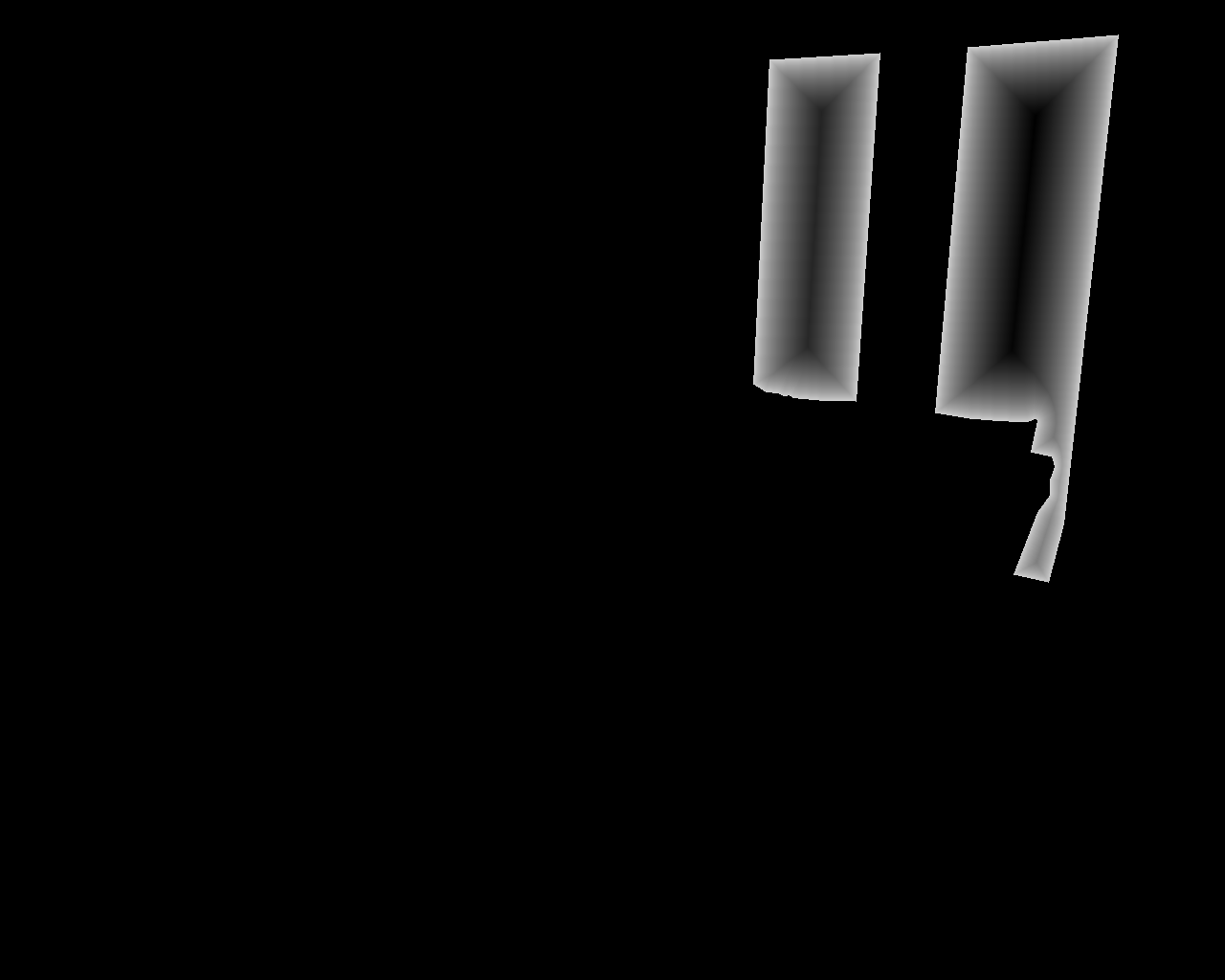}
      \end{minipage}
      }         
      \caption{\label{img3}
      Examples of label decoupling. In the interior-diffusion label (c) of GT (b), pixels close to the center of the glass have larger values. In the boundary-diffusion label (d) of GT (b), pixels near the boundary of the glass have larger values. The sum of (c) and (d) is equal to (b).}
\end{figure}

Many object detection methods pay attention to boundary information for the enhancement of detection accuracy, but the prediction difficulty of boundary pixels is closely related to their locations. 
Therefore, it is difficult to classify the pixels near the boundary correctly, which is called ``hard examples'' \cite{hard}. 
In contrast, the consistency of interior regions makes the central region easier to detect. 
Therefore, a strategy of dealing with boundary pixels and interior pixels differently will make the detection results more reasonable. 
However, it is difficult to claim which pixels are hard examples or not, as illustrated in Figure \ref{img2}. 
We adopt the label decoupling (LD) strategy proposed by Wang et al. \cite{egnet} to decouple the original glass map into an interior-diffusion map and a boundary-diffusion map, as shown in Figure \ref{img3}. 
In more detail, LD uses the simple Distance Transformation (DT) to convert the ground-truth glass map into a new image, where the value of the foreground pixel is the minimum distance from the background obtained by the distance function. 
Please note that the foreground herein refers to the glass region, and the background means the remaining non-glass region.

Distance Transformation (DT) calculates the distance from the nearest zero points to itself for each non-zero point in an image. Its input is a binary graph such as the ground truth of the image detection task, which can be divided into two groups (i.e., the foreground $ I_{fg} $ and the background $ I_{bg} $). 
The original metric function is defined as $ f(p,q)=\sqrt{(p_x-q_x)^2+(p_y-q_y)^2} $ to calculate the distance between two pixels, and here we modify $ f(p,q)$ to fit our approach. 
The new distance function is formulated as:
\begin{equation}
      I'(p)=\left\{ 
            \begin{array}{rcl}
            \min\limits_{q\in I_{bg}}{f(p,q)}, &     & p \in I_{fg}\\
                                            0, &     & p \in I_{bg} 
            \end{array} \right.
\end{equation}   
For the foreground pixel $p$, DT calculates the original distance function $f(p,q)$ by looking for its nearest pixel $q$ in the background pixel, and directly sets the value to 0 for the background pixel. 
We use a linear normalization function $I'=\frac{I'-\min{(I')}}{\max{(I')}-\min{(I')}} $, which normalizes the image generated from the new distance function. 
Compared with the original image, the new image obtained by defining the distance function depends not only on its foreground or the background but also on its relative position. 
Therefore, the new image corresponds to the inner part of the original image, and the closer to the center is, the larger the pixel value will be. 
The boundary images obtained by subtracting the new image from the original image can help deal with the hard examples. 
To remove the background interference, we process the new image and the original ground truth to generate the interior-diffusion label and boundary-diffusion label as:
\begin{equation}
Label\Rightarrow \left\{ 
                  \begin{array}{lcl}
                  BL=I\ast I'    & \\
                  DL=I\ast (1-I')&
                  \end{array}  
                  \right. 
\end{equation}
where $BL$ means the interior-diffusion label and $DL$ represents the boundary-diffusion label. Thus, we decouple the original label into two different kinds of labels, to work in learning both interior and boundary features with different characteristics. 

\subsection{Network Overview}

\begin{figure*}[htb]
      \centering
      \includegraphics[width=1\linewidth]{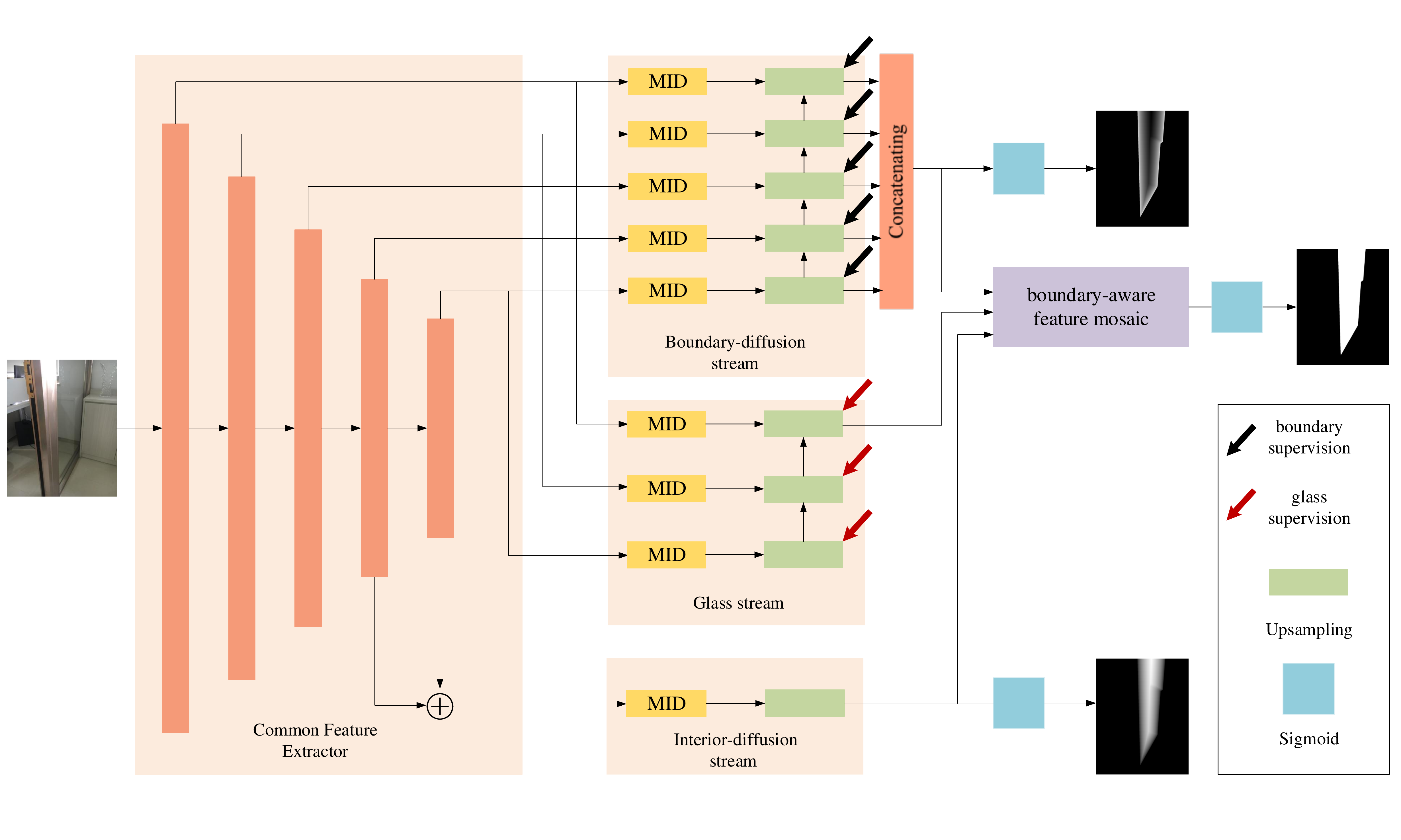}
      \caption{\label{img4}
      Overview of the proposed GlassNet. The pre-trained ResNet-50 \cite{ResNet} is employed as the backbone network to extract multi-level image features. 
      The extracted image features are fed into the three streams. 
      In each of the three streams, we use Multi-scale Interactive Dilation Module (MID) to extract large-field contextual features, to obtain glass features, interior-diffusion features, and boundary-diffusion features respectively through supervision. 
      The three different features are fused through an attention-based Boundary-aware Feature Mosaic Module (BFM) and fed themselves into the predict block to generate the final glass map.}
\end{figure*}

The overview of the proposed model is illustrated in Figure \ref{img4}, which consists of three parallel streams: an interior-diffusion stream, a boundary-diffusion stream, and a glass stream. 
We first feed an image to the backbone network to extract multi-scale backbone features. 
Then, the features of each level are fed into the three streams supervised by the decoupled labels to generate different features. 
In each stream, we use the Multi-scale Interactive Dilation Module (MID) to extract large-field contextual features, then obtain the glass features, interior-diffusion features, and boundary-diffusion features for each stream. 
Finally, we use the attention-based Boundary-aware Feature Mosaic Module (BFM) to integrate the boundary features and interior-diffusion features into the glass prediction maps to generate the final glass map of the whole network. Details of the proposed approach are described as follows.

\textbf{Feature encoder.} 
We use ResNet-50 \cite{ResNet} as the backbone network to extract common multi-level image features for the three streams as suggested by \cite{stagerefine, dgrl, Picanet}. 
In particular, as a backbone network, we remove the last global pooling and fully connected layers and only use the five residual blocks.
For the sake of simplification, we represent this five blocks as $ f_i{(w_i)}$, $i \in \{1,...,5\}$, where $w_i$ is the weight parameters pre-trained on ImageNet \cite{Imagenet} of the $f_i(\cdot)$ operation, and the output of the i-th layer $f_i(\cdot)$ is the input of $f_{i+1}(\cdot)$, $\forall i\in\{1,...,4\}$. 
We feed an input image with the shape $H\times W$ into it to generate different-scale features denoted as $ EF=\{EF_i|i=1,2,3,4,5\} $ by utilizing $ f_i{(w_i)}$, $i \in \{1,...,5\}$, i.e., $EF_{i+1}=f_i(EF_i)$. 
Then, we input the different levels of features into the three streams for processing.
The features $EF_5$ and $EF_4$ are fed into the interior-diffusion stream decoder to roughly locate the glass region.
In order to obtain a finer glass boundary, the features $\{EF_i|i=1,2,3,4,5\}$ are fed into the boundary-diffusion stream decoder. 
In addition, we utilize the features $\{EF_i|i=1,2,5\}$ for the glass map generation in the glass stream.

\textbf{Three-stream decoder.} 
As shown in Figure \ref{img4}, we built a three-stream network to use the label decoupling information. 
We utilize a label decoupling procedure to decompose a glass label into an interior-diffusion map and a boundary-diffusion map to supervise the model separately. 
Through the supervision of these three different labels, better detection results can be obtained.

In each stream, we use a Multi-scale Interactive Dilation Module (MID) to extract large-field contextual features. Figure \ref{img4} illustrates the detailed structure of the decoder. 
For the glass stream and the boundary-diffusion stream, we employ the short connections \cite{dssc} to merge feature maps $EF_i$ at different CNN layers, resulting in new feature maps (denoted as $DF_i$). 
Specifically, the merged feature map $DF_i$ at the $k$-th CNN layer $(i = 1, ..., 5)$ is computed by
\begin{equation}
DF_i=Conv(Concat(MF_i,...,MF_5))
\end{equation}
Then, we use MID to generate large-field contextual features $MF_i$, which can be formulated as: 
\begin{equation}
MF_i = MID(DF_i)
\end{equation}
where $MID(\cdot )$ is the Multi-scale Interactive Dilation Module. 
Then, we integrate the different levels of MF through  a decoder structure. 
In particular, we concatenate the output of each level in the boundary-diffusion stream to get the final boundary map. For the interior-diffusion stream, we add $EF_4$ and $EF_5$ by using the element-wise addition operation and entering them into MID to generate MF. 
In this way, different-level feature maps are jointly fused, which is beneficial for semantic segmentation.  
\begin{figure}[htb]
      \centering
      \includegraphics[width=1\linewidth]{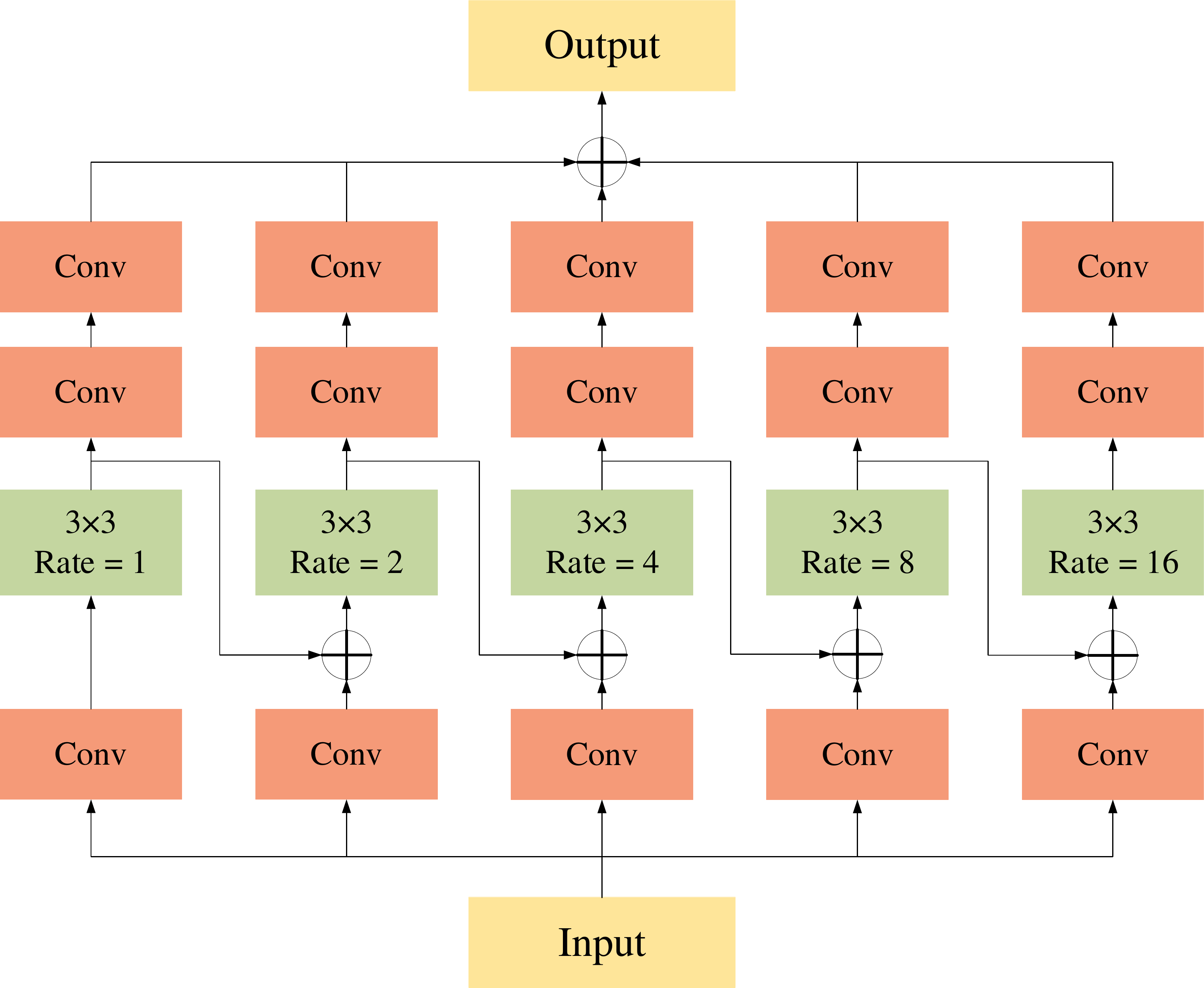}
      \caption{\label{img5}
      The structure of the Multi-scale Interactive Dilation (MID) module. $Conv$ represents a convolutional layer with a kernel size of $3\times 3$, a normalized layer, and a ReLU layer. $3\times 3$ means the convolutional kernel size, and rate represents the dilation rate in a dilated convolution.}
\end{figure}

\subsection{Multi-scale Interactive Dilation Module}
For glass detection, the key to accurately locate glass regions is to aggregate a wide range of contextual features at different scales. 
GDNet \cite{GDNet} exploits the Large-field Contextual Feature Integration (LCFI) module to efficiently extract abundant contextual information from a large field. 
It utilizes convolutions with large kernels and dilated convolutions to enlarge the receiving field, and spatially separable convolution to reduce the parameters of convolution with large kernels.
Inspired by them, we propose a multi-scale interactive dilation module (named as MID, as shown in Figure \ref{img5}) to efficiently aggregate different-scale contextual information for enhancing the glass detection performance.

Specifically, we propose to utilize dilated convolution with different dilation rates to expand the receiving field of each pixel so that it can obtain a wider range of contextual information. 
After passing the feature map into this module, the maps first pass through a convolution layer for feature extraction at each of the branches. 
Then, we use dilated convolution to extract a wide range of context features. 
Different from LCFI  \cite{GDNet}, we use the dilation convolution with larger dilation rates, which are set to 2, 4, 8, and 16, respectively.
And an additional branch is added which uses a $3 \times 3$ convolution to obtain more dense local information.
Besides, in order to reduce the parameters of the module in the whole network, we remove the large kernel convolution of LCEI.
Meanwhile, we adopt Short Connections \cite{dssc} to transfer the output of the smaller receiving field branch into the other larger branch for getting dense contextual information. 
Finally, we integrate the context feature maps of each branch through convolution layers and obtain the feature map $MF$ by utilizing an element-wise addition operation.

\begin{figure}[htb]
      \centering
      \includegraphics[width=1\linewidth]{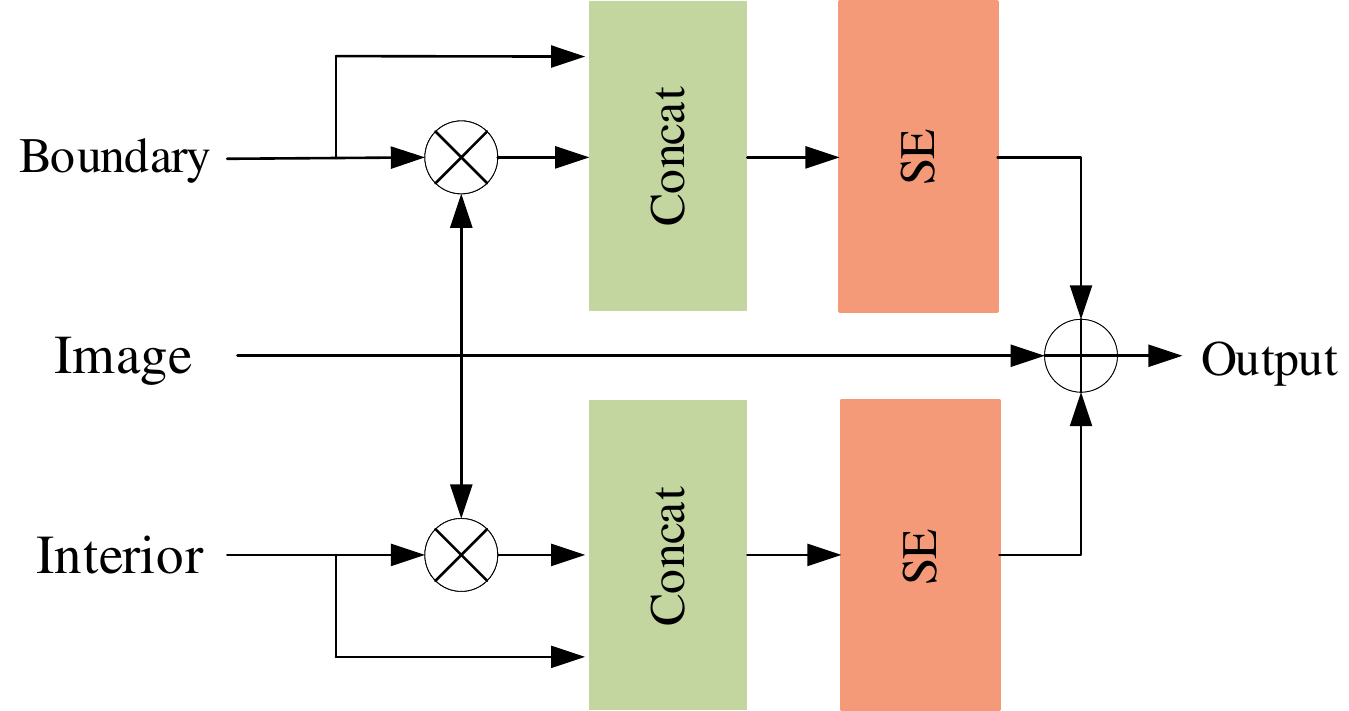}
      \caption{\label{img6}
      Attention-based boundary-aware feature Mosaic module.}
\end{figure}

\subsection{Attention-based Boundary-aware Feature Mosaic Module}
After obtaining a high-quality boundary prediction map using the boundary-diffusion stream, we utilize a Boundary-aware Feature Mosaic Module (BFM) to integrate the boundary maps into the predicted glass maps generated by the glass stream, as shown in Figure \ref{img6}. 
BFM first takes the predicted boundary map, interior-diffusion map, and glass feature map as inputs. 
Through using the predicted boundary map and the interior-diffusion map as attention maps, we integrate them into the feature maps of the glass stream by using an element-wise product operation $\otimes $.
We concatenate the interior-diffusion map and the boundary-diffusion map and input them into the SE module, respectively to enhance the corresponding boundary- and interior-diffusion features. 
Here, the SE module is an architectural unit proposed by Hu et al. \cite{hu2018squeeze}, which is termed as the ``Squeeze-and-Excitation'' (SE) block, that adaptively re-calibrates channel-wise feature responses by explicitly modelling inter-dependencies between channels.
Finally, we add the enhanced features to the input glass feature map to generate the final output.

\begin{table*}[htb]
      \centering
      \begin{tabular}{l|c|c|c|c|c|c}
            \hline
            Method                  &   CRF    &  acc$\uparrow$  &  IoU$\uparrow$  &   $F_\beta\uparrow$  &  MAE$\downarrow$  &  BER$\downarrow$  \\
            \hline
            \hline
            PSPNet\cite{PSP}           &  $\times$  &  0.916          &  0.841          &   0.906              &   0.084           &   8.79            \\
            \hline
            DenseASPP\cite{DenseASPP}  &  $\times$  &  0.919          &  0.837          &   \textcolor{blue}{0.911}              &   0.081           &   8.66            \\
            \hline
            PSANet\cite{psanet}        &  $\times$  &  0.918          &  0.835          &   0.909              &   0.082           &   9.09             \\
            \hline
            CCNet\cite{CCNet}          &  $\times$  &  0.915          &  0.843          &   0.904              &   0.085           &   8.63           \\
            \hline
            DANet\cite{DANet}          &  $\times$  &  0.911          &  0.842          &   0.901              &   0.089           &   8.96            \\
            \hline
            \hline
            R$^3$Net\cite{r3net}       &  $\surd$   &  0.869          &  0.767          &   0.869              &   0.132           &   13.85            \\
            \hline
            CPD\cite{cpd}              &  $\times$  &  0.907          &  0.825          &   0.903              &   0.095           &   8.87            \\
            \hline
            BASNet\cite{basnet}        &  $\times$  &  0.907          &  0.829          &   0.896              &   0.094           &   8.70            \\
            \hline
            EGNet\cite{egnet}          &  $\times$  &  0.885          &  0.788          &   0.858              &   0.115          &   10.87            \\
            \hline
            LDF\cite{LDF}              &  $\times$  &  \textcolor{blue}{0.921}          &  0.843          &   0.908              &  0.079           &  \textcolor{blue}{ 7.52}              \\
            \hline
            \hline
            BDRAR\cite{BDRAR}          &  $\surd$   &  0.902          &  0.800          &  0.908               &   0.098            &   9.87            \\
            \hline
            DSC\cite{DSC}              &  $\times$  &  0.914          &  0.836          &  \textcolor{blue}{0.911}               &   0.090            &   7.97            \\
            \hline
            \hline
            MirrorNet\cite{MirrorNet}  &  $\surd$   &  0.918          &  \textcolor{blue}{0.851}          &  0.903               &   0.083            &   7.67            \\
            \hline
            PMD\cite{PMD}              &  $\surd$   &  \textcolor{blue}{0.921}          &  0.836          &  0.894               &   \textcolor{blue}{0.078}            &   8.34             \\
            \hline
            \hline
            GDNet\cite{GDNet}          &  $\times$  &  \textcolor{red}{0.939}          &   \textcolor{red}{0.876}          &  \textcolor{red}{0.920}               &   \textcolor{red}{0.061}            &    \textcolor{red}{5.62}            \\
            \hline
            \hline
           GlassNet (ours)                       &  $\times$  &   \textbf{0.946}          &  \textbf{0.887}         &   \textbf{0.937}               &     \textbf{0.054}            &   \textbf{5.42}            \\
            \hline

      \end{tabular}
      \caption{\label{table1}
      Quantitative results on the GDD dataset. CRF indicates whether CRF \cite{crf} is used as a post-processing step. The first, second, and third best results are marked in bold, red, and blue, respectively.}
\end{table*}
\begin{figure*}[htb]
      \centering
      \subfloat[Input]{\label{img}
      \begin{minipage}[t]{0.07\textwidth}
            \centering
            \includegraphics[width=1\linewidth]{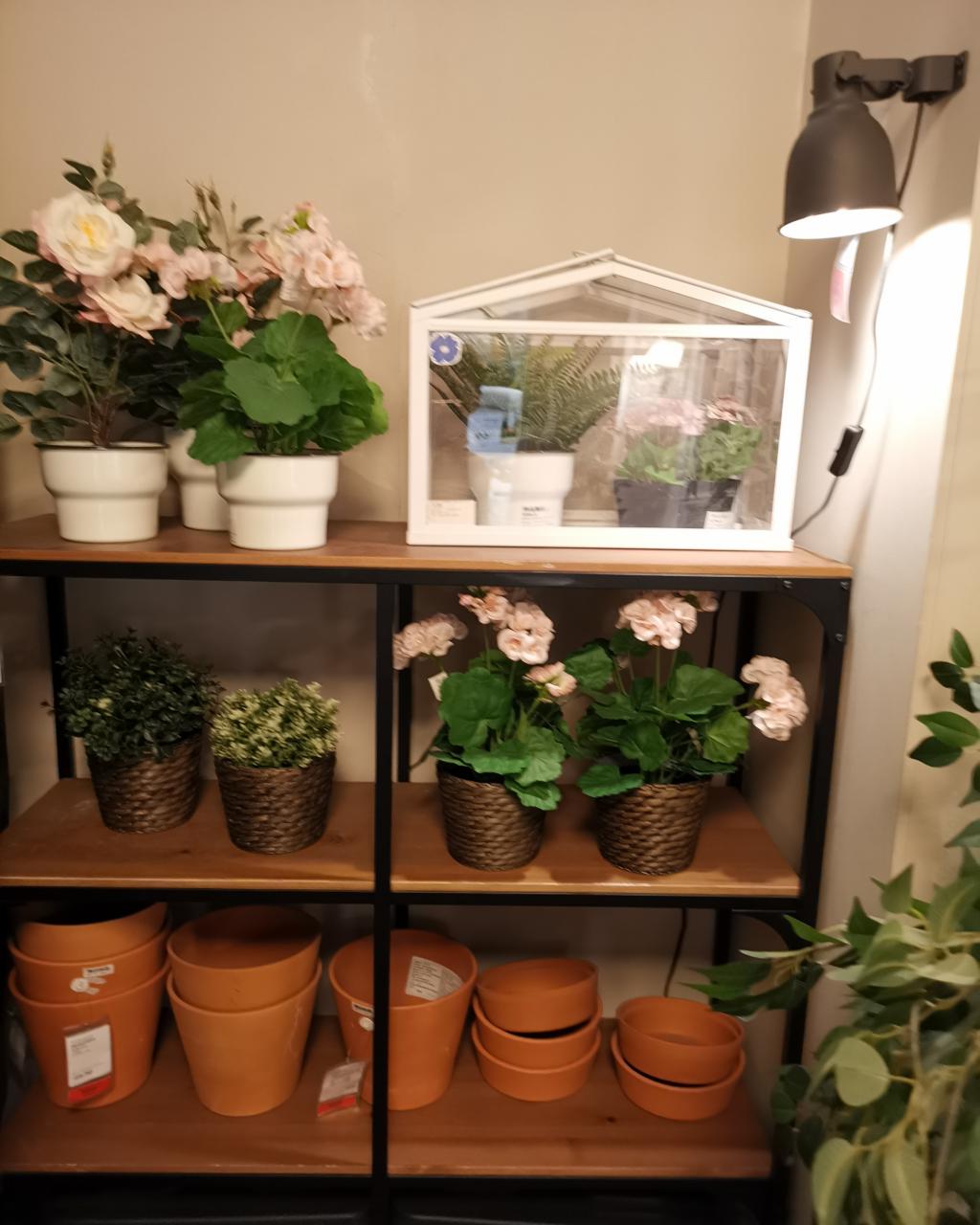}

            \includegraphics[width=1\linewidth]{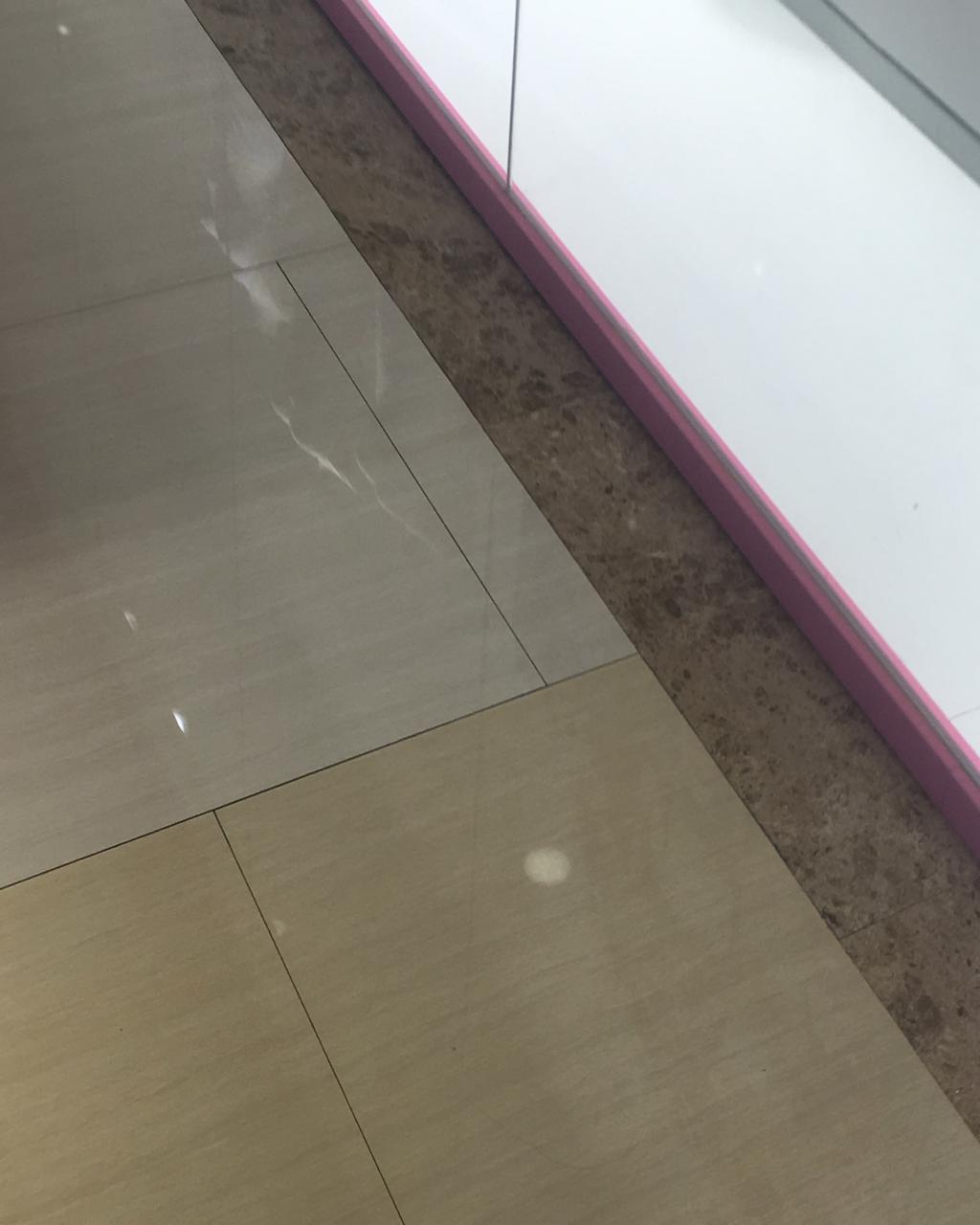}

            \includegraphics[width=1\linewidth]{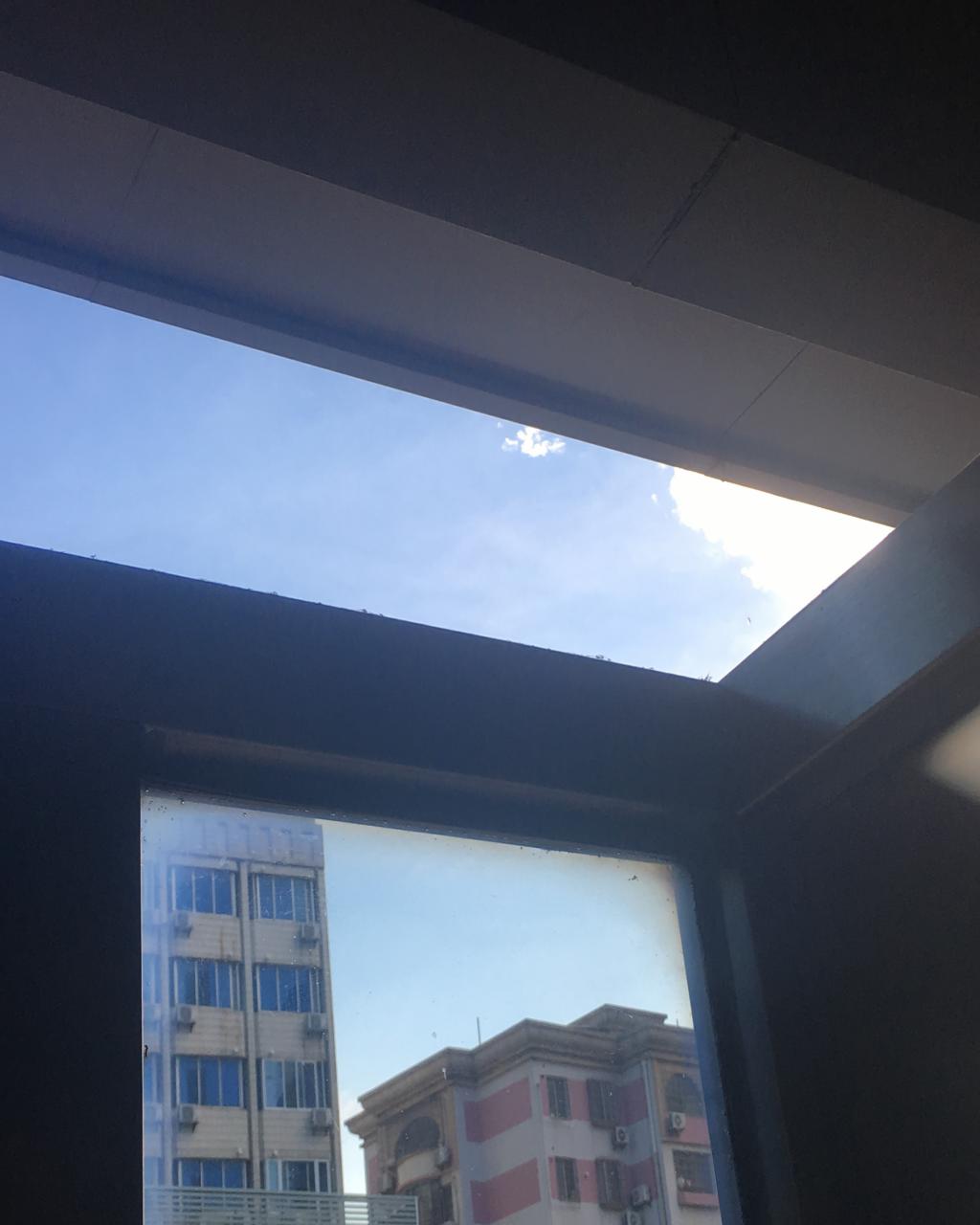}

            \includegraphics[width=1\linewidth]{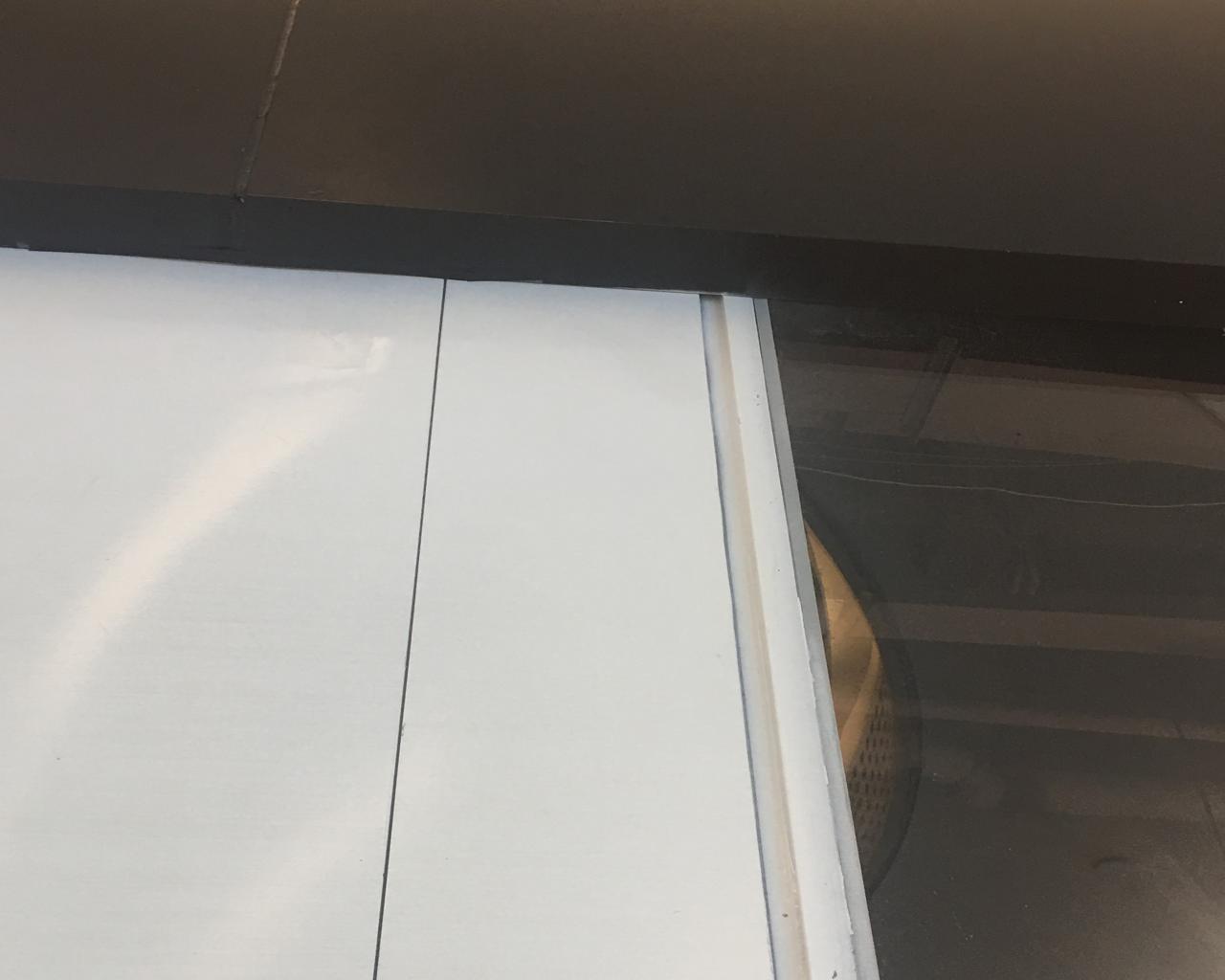}

            \includegraphics[width=1\linewidth]{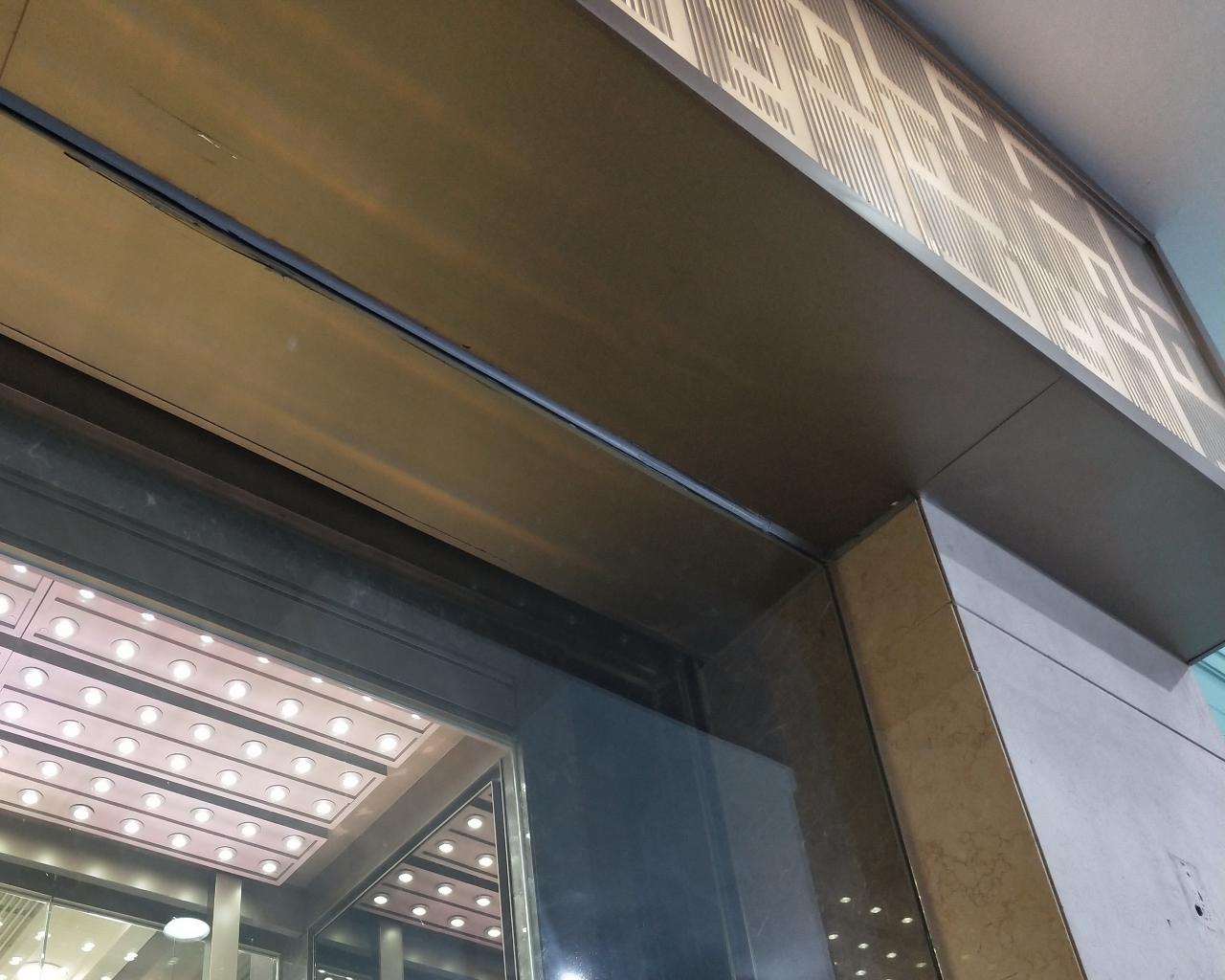}

            \includegraphics[width=1\linewidth]{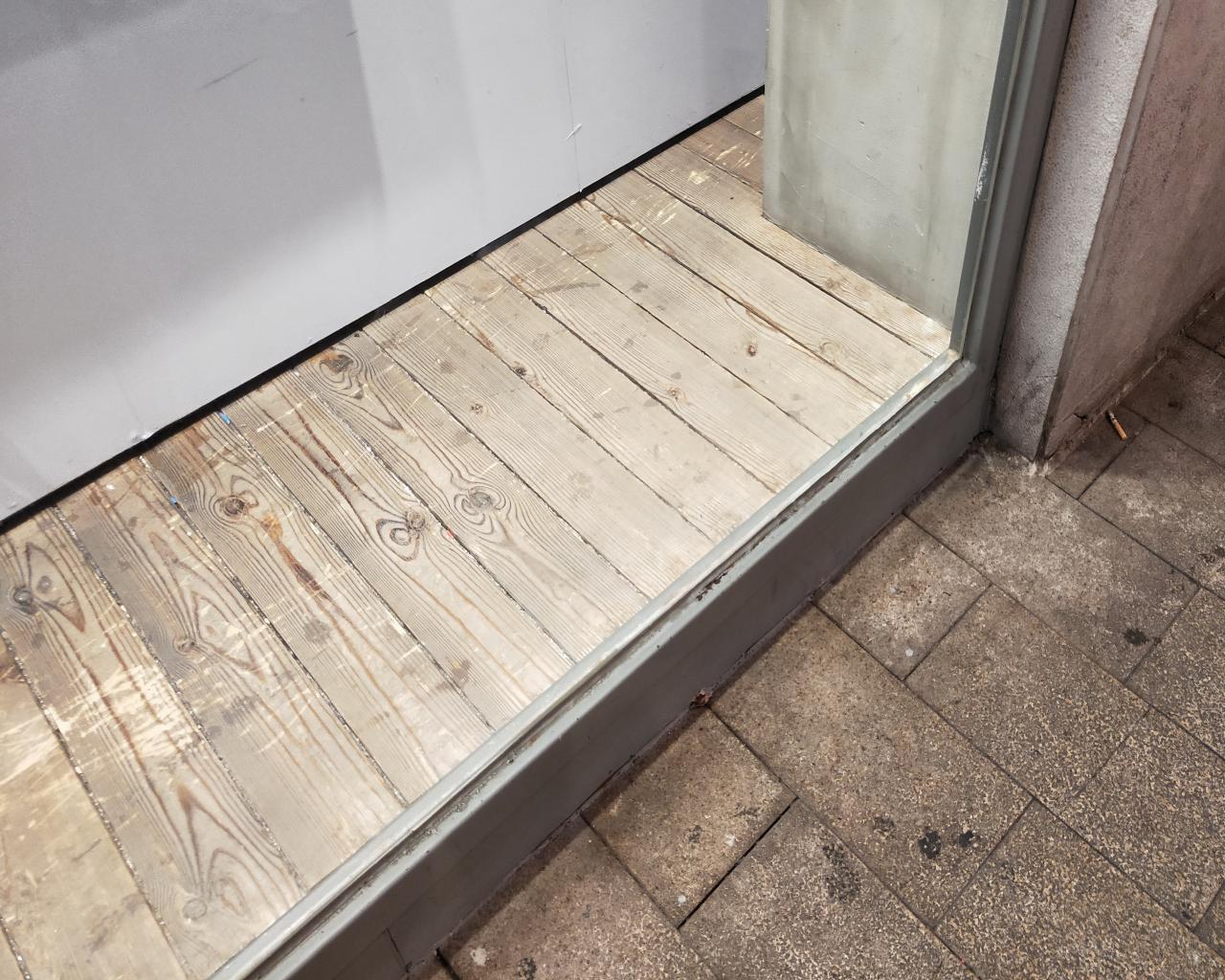}

            \includegraphics[width=1\linewidth]{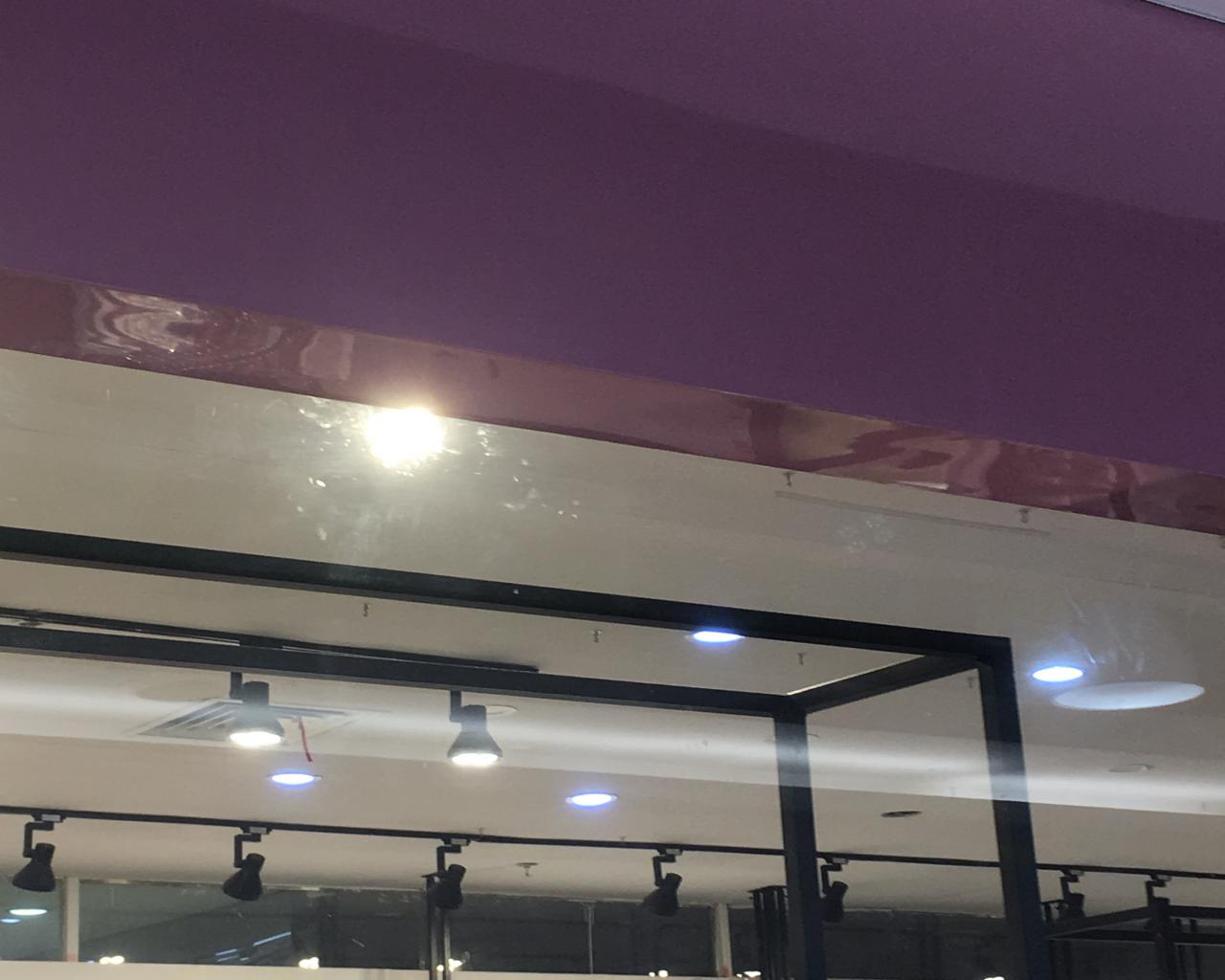}
            
            \includegraphics[width=1\linewidth]{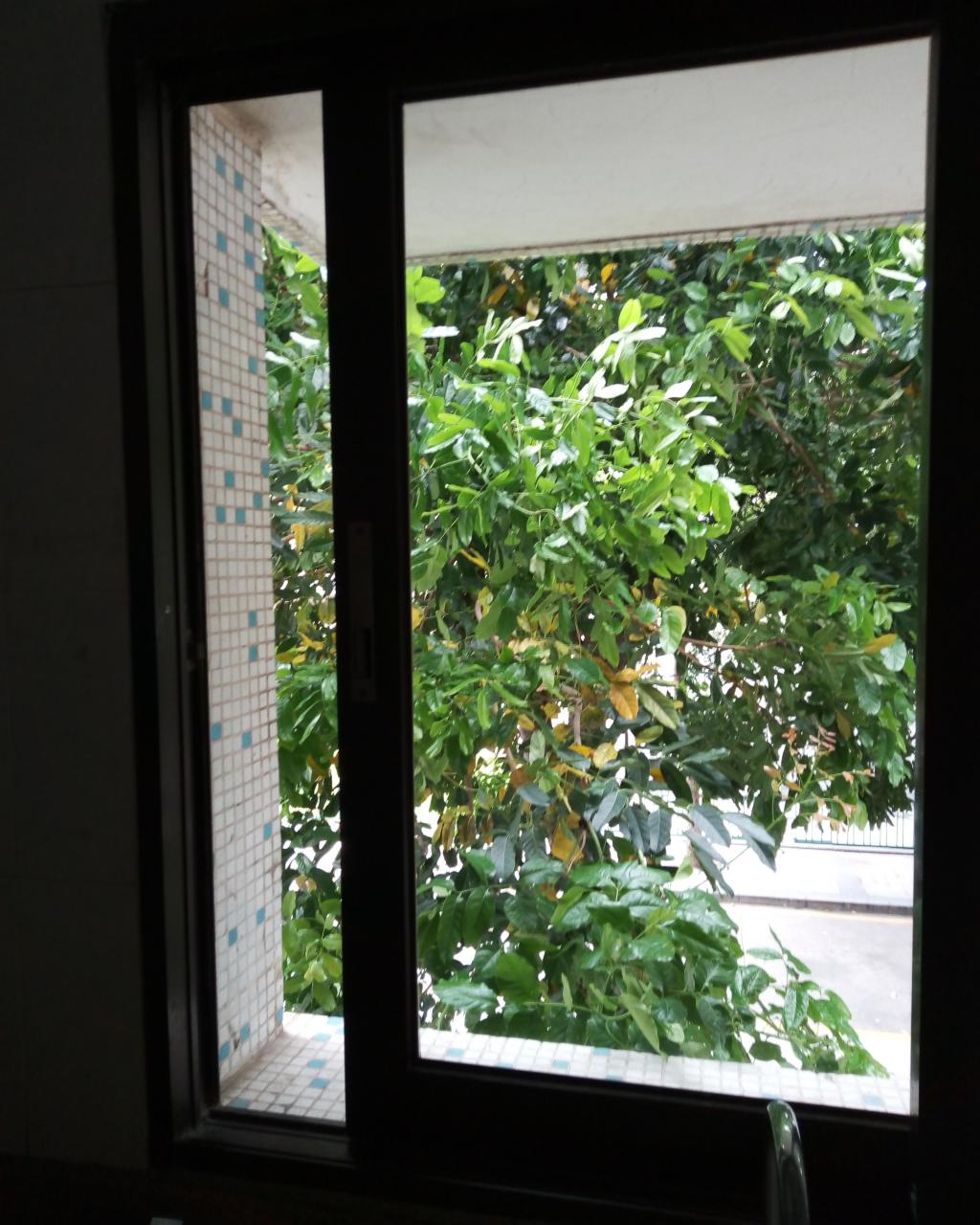}

            \includegraphics[width=1\linewidth]{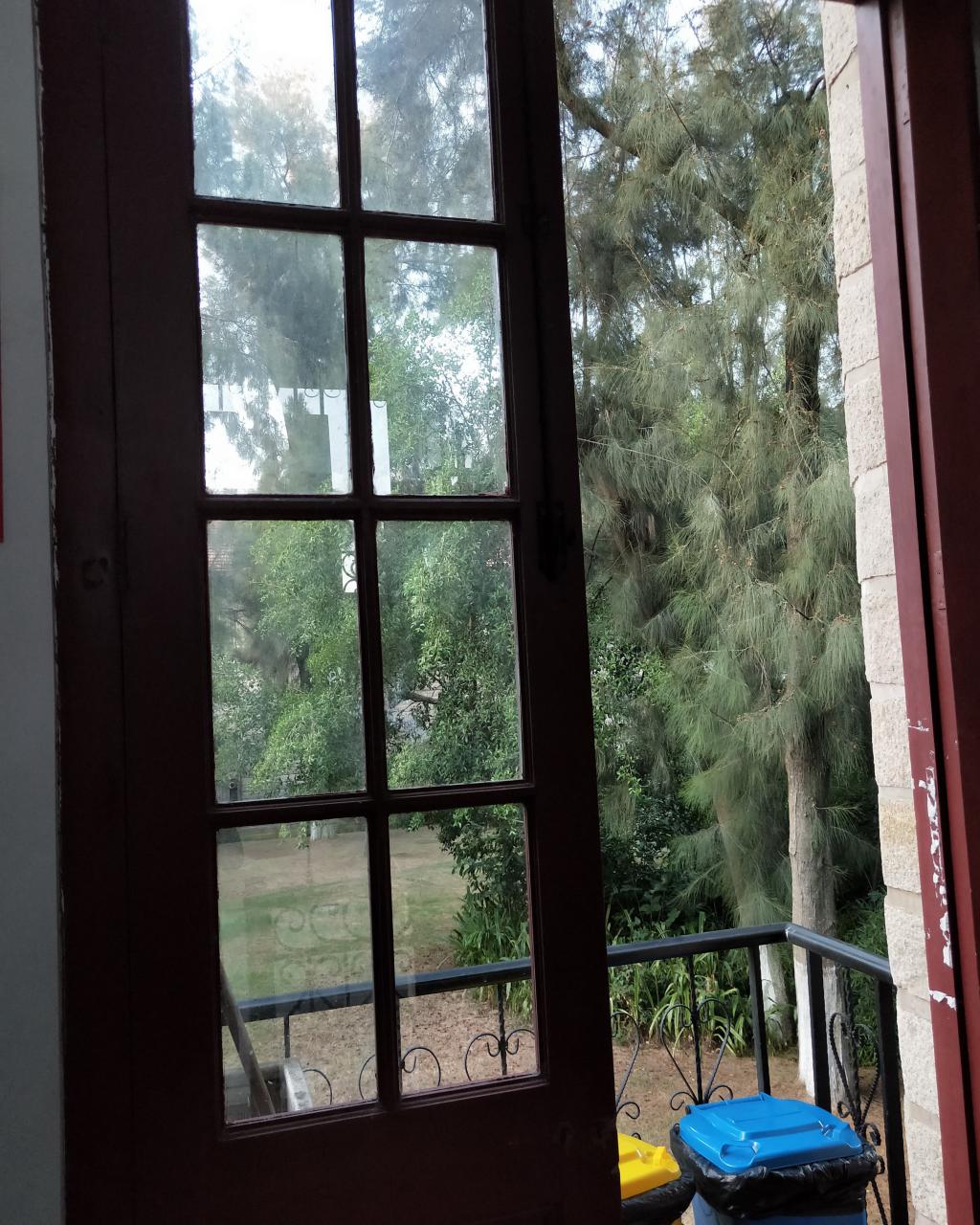}
      \end{minipage}
      }
      \subfloat[DANet]{\label{DA}
      \begin{minipage}[t]{0.07\textwidth}
            \centering
            \includegraphics[width=1\linewidth]{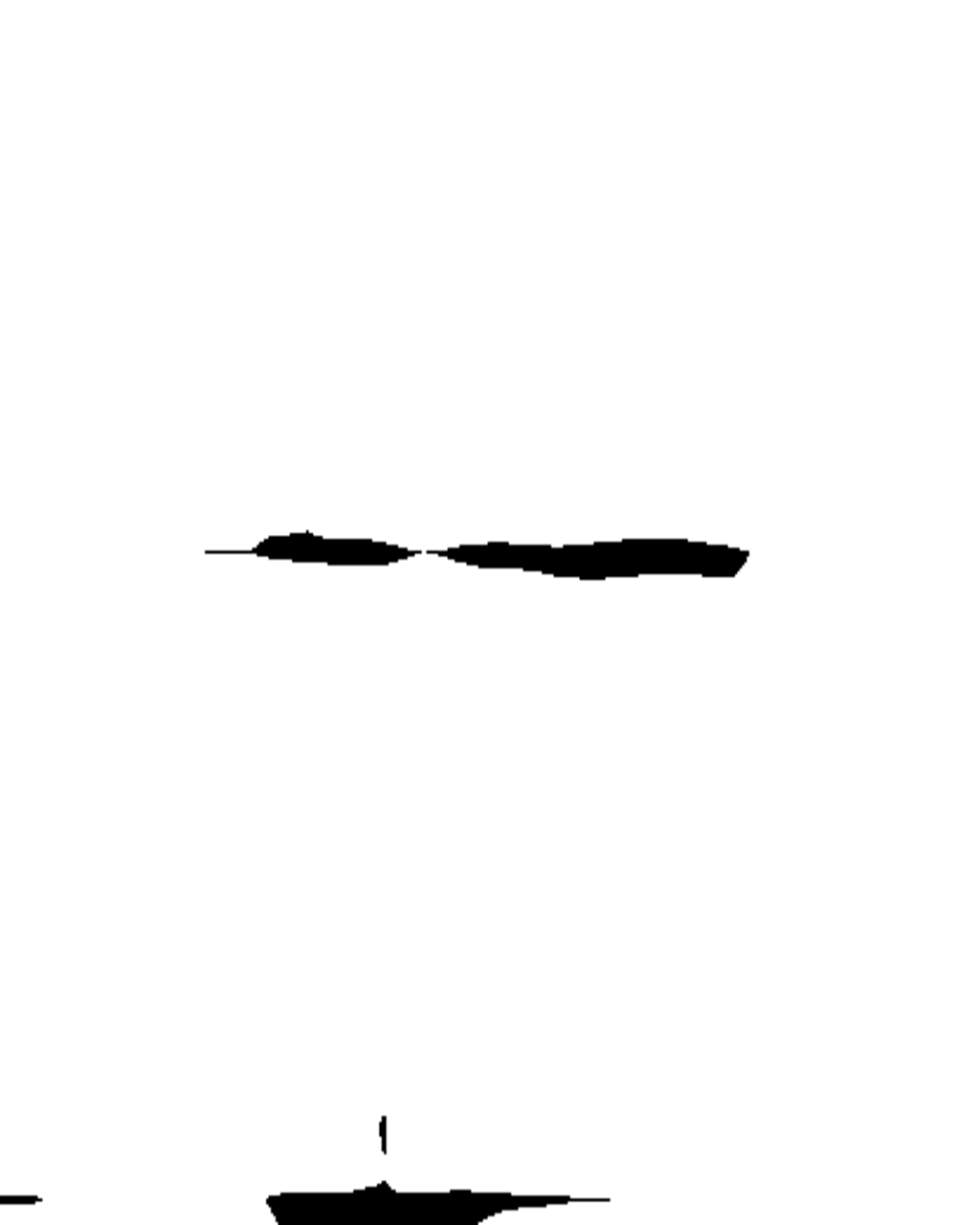}
            
            \includegraphics[width=1\linewidth]{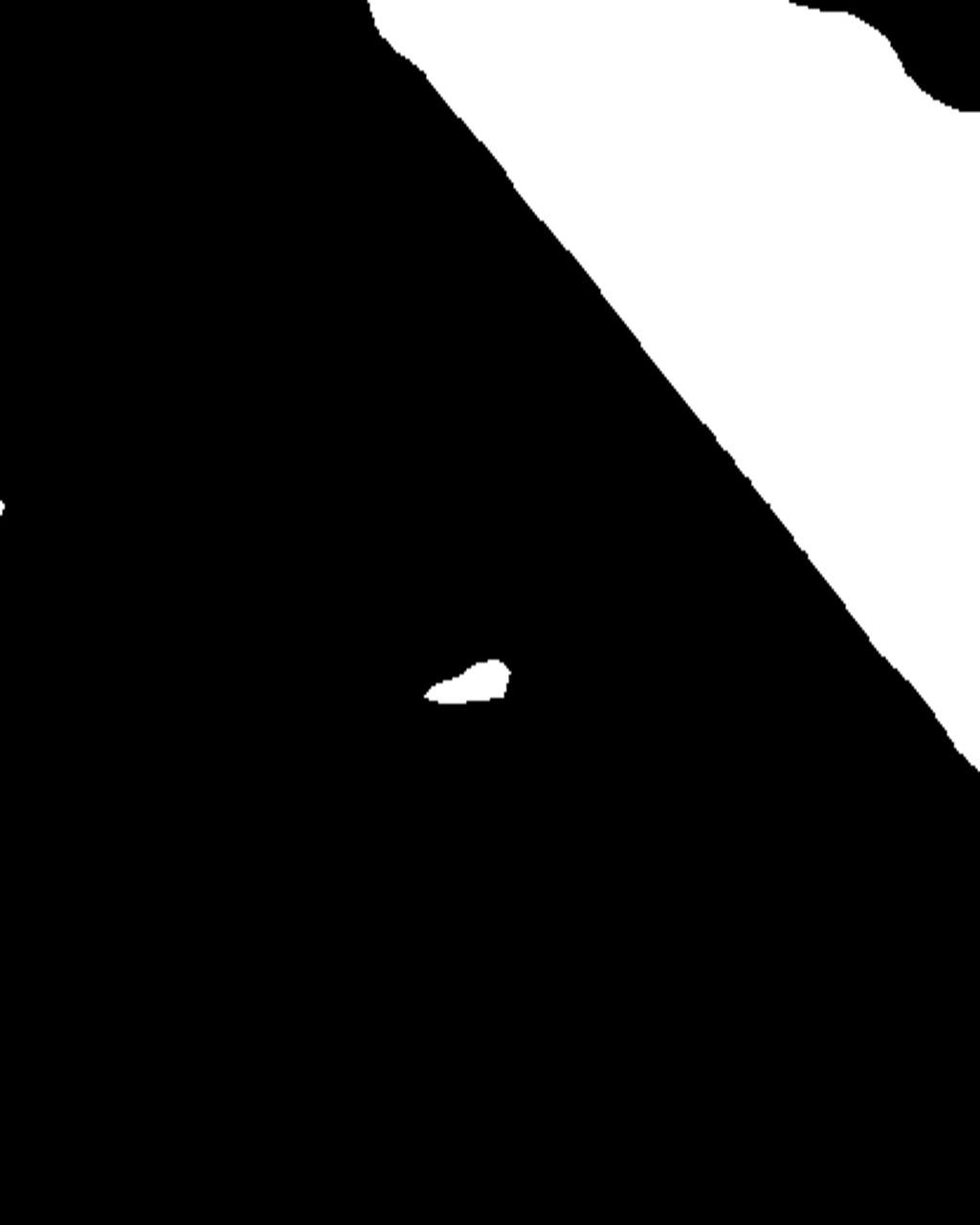}

            \includegraphics[width=1\linewidth]{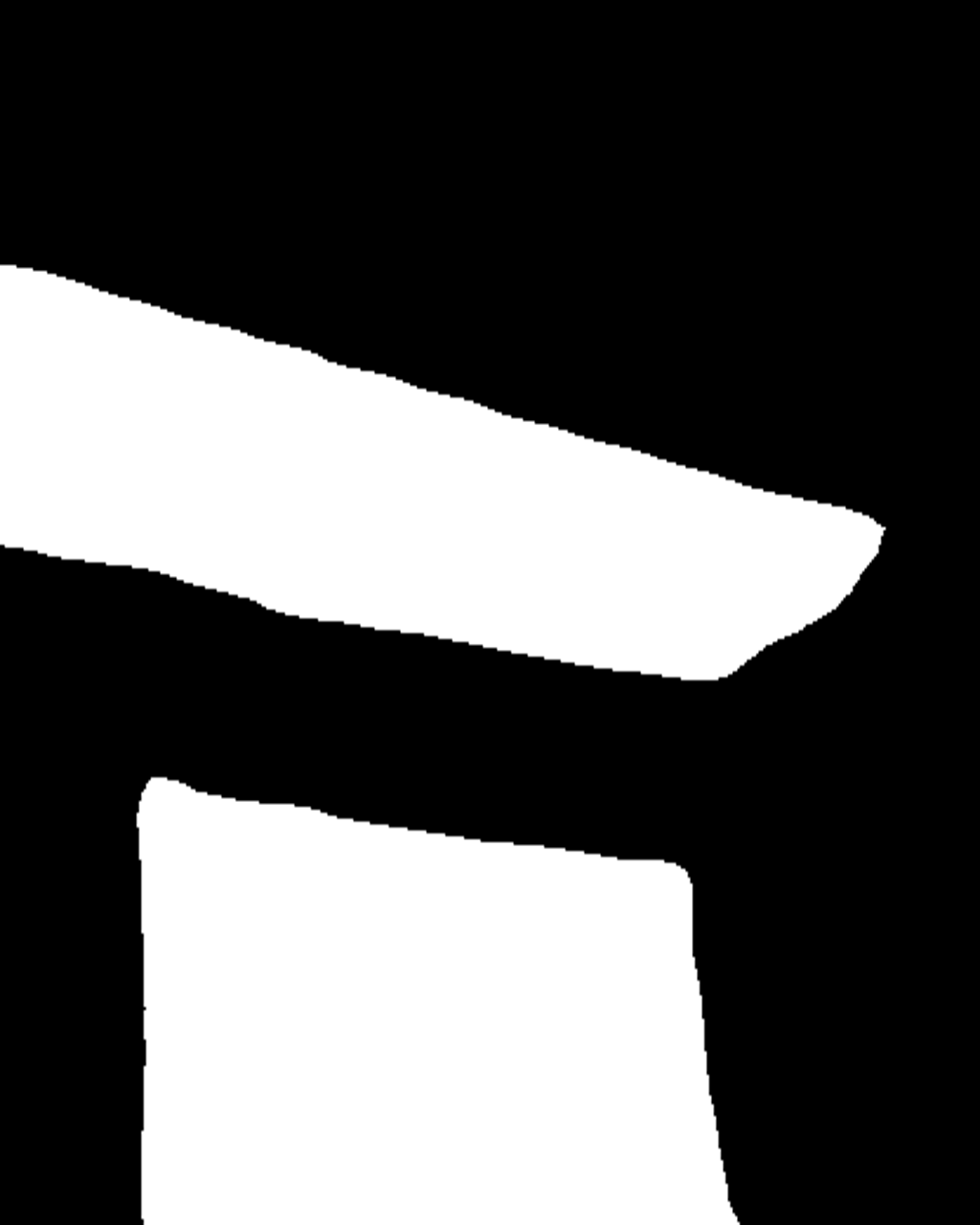}

            \includegraphics[width=1\linewidth]{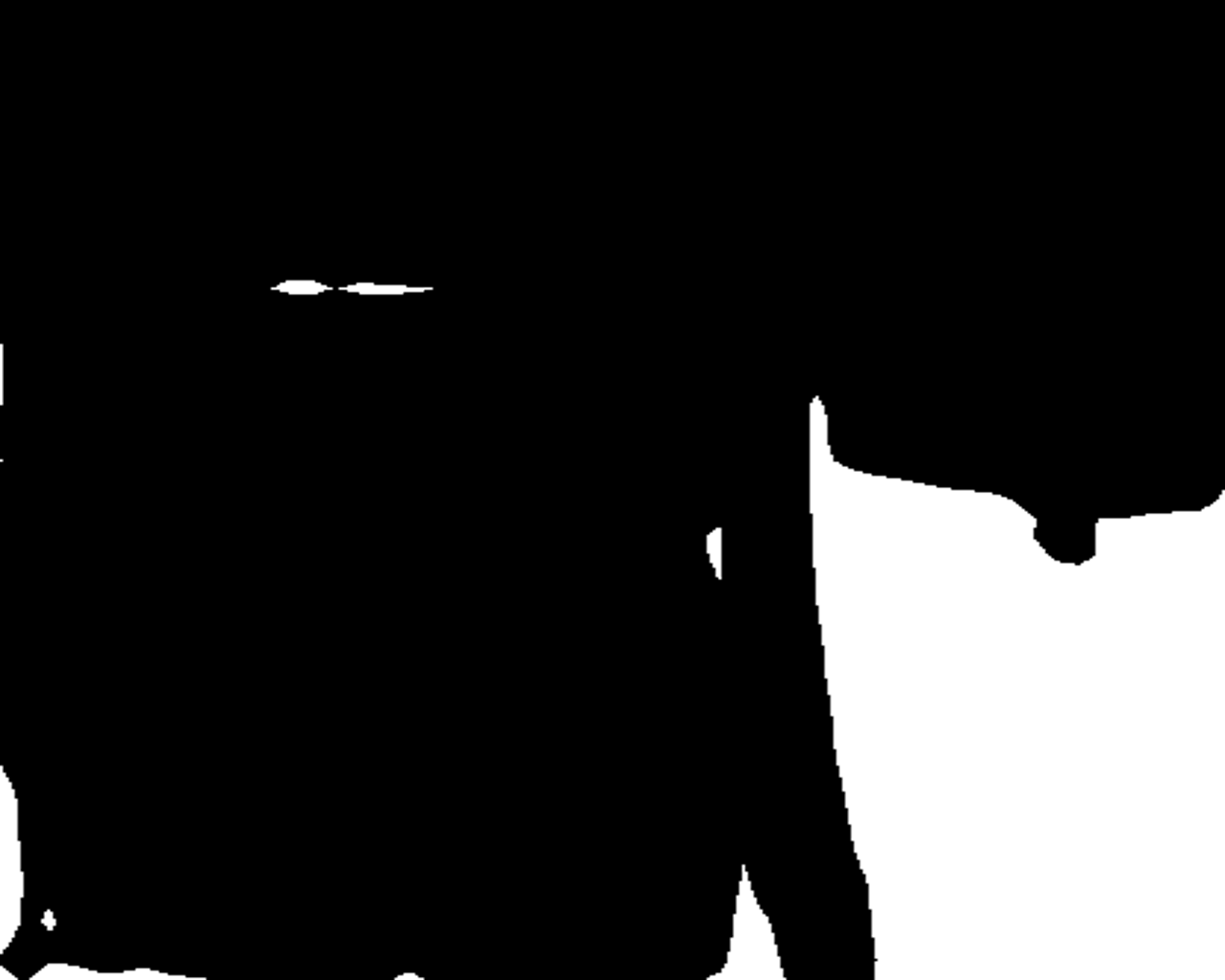}

            \includegraphics[width=1\linewidth]{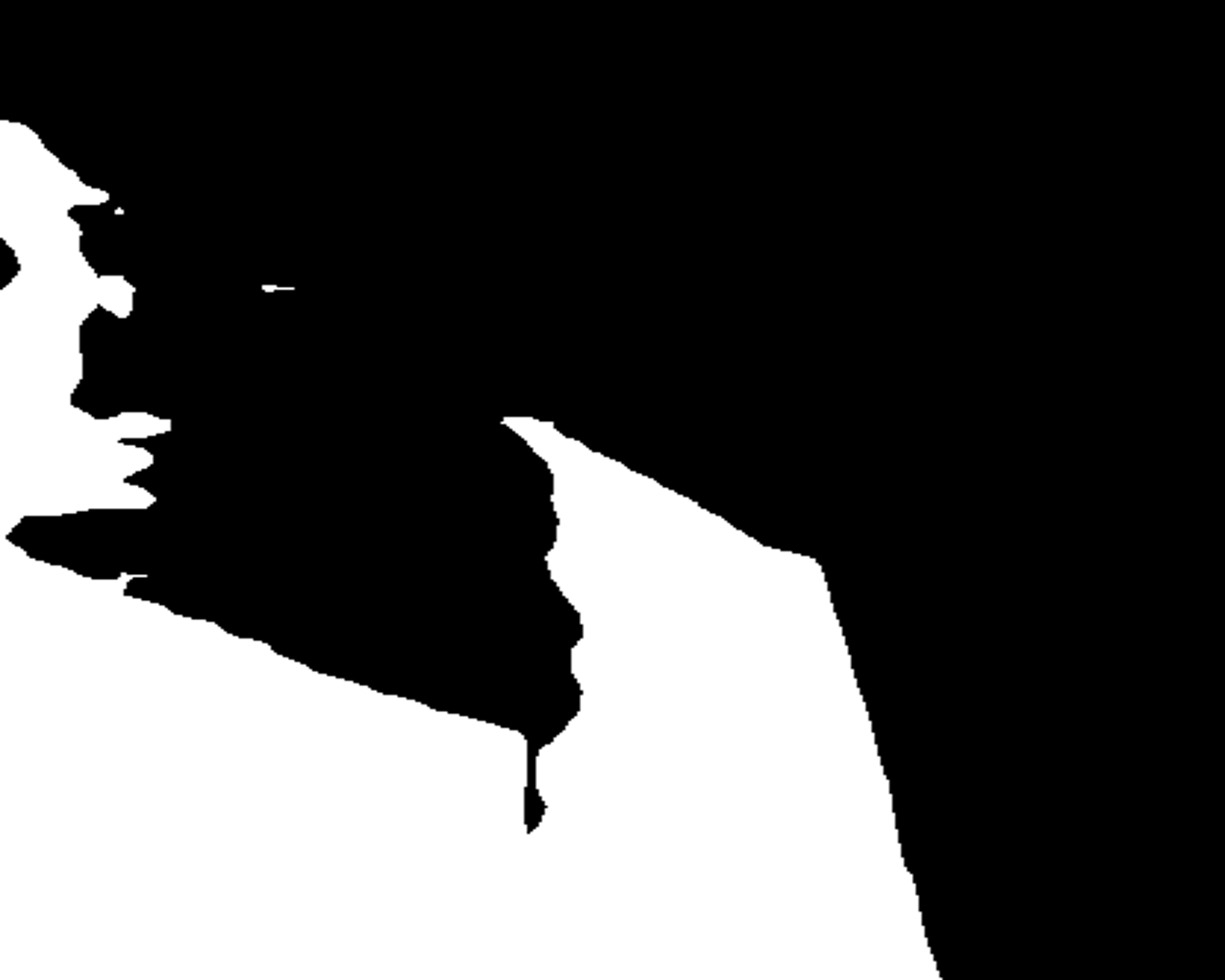}

            \includegraphics[width=1\linewidth]{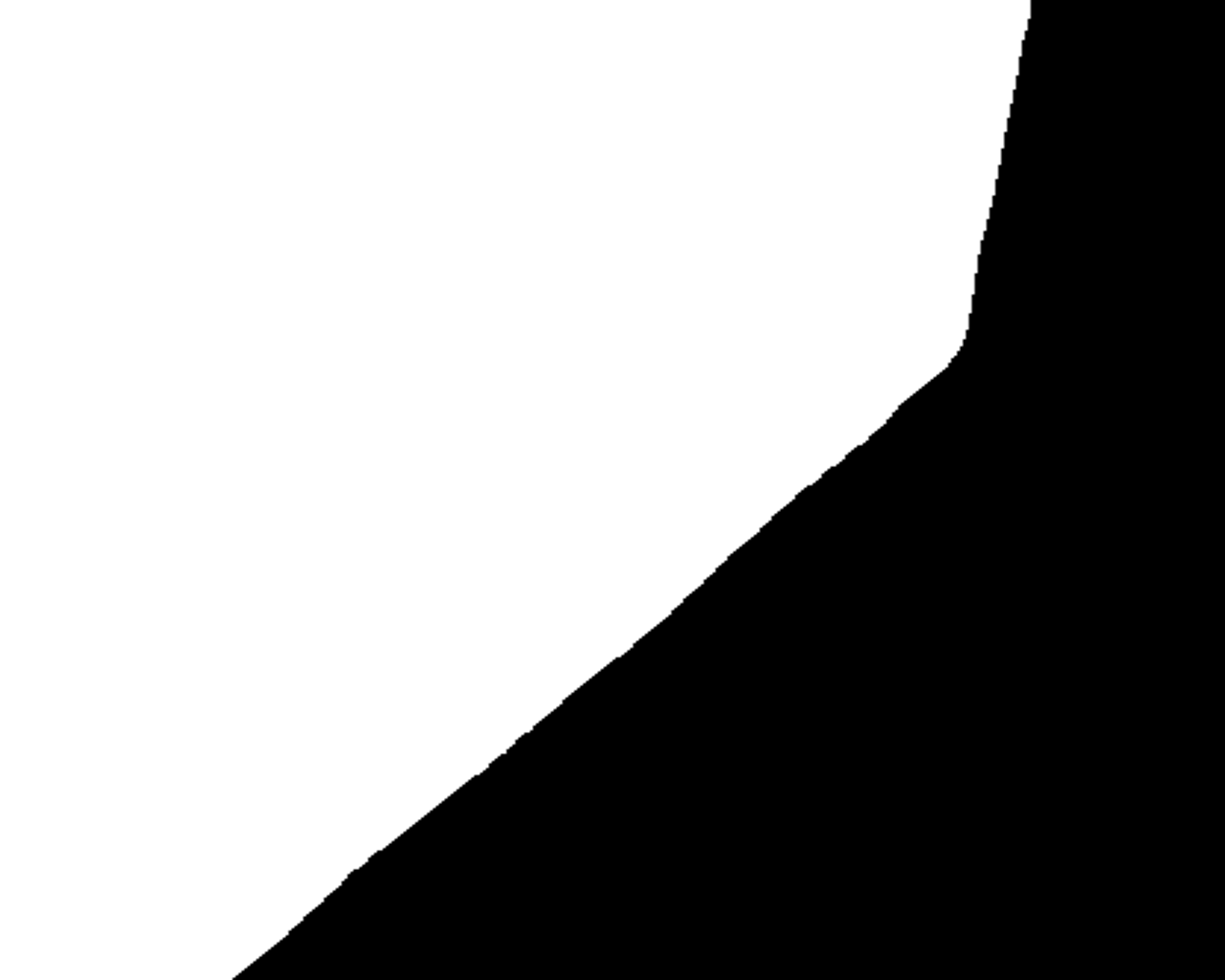}

            \includegraphics[width=1\linewidth]{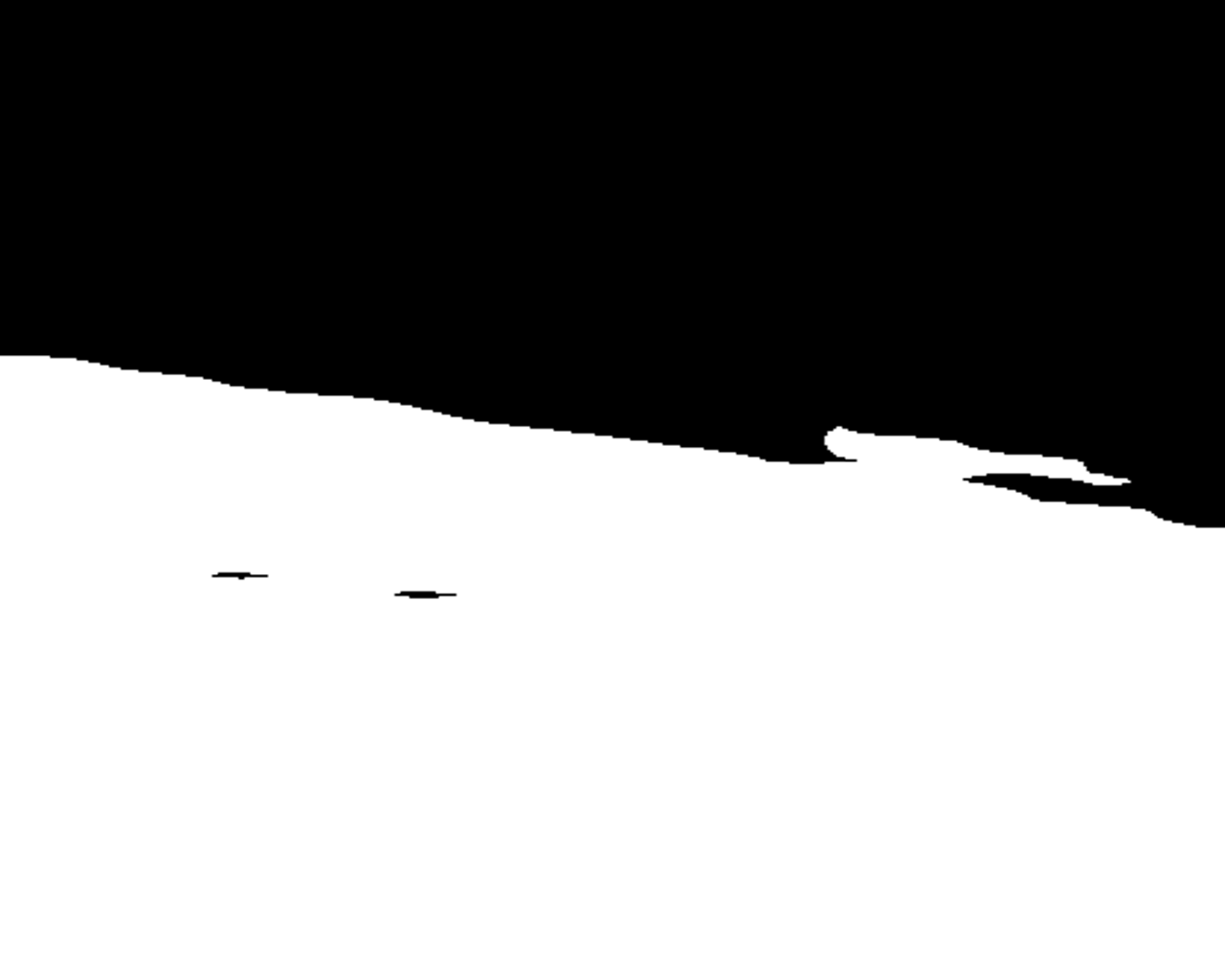}
            
            \includegraphics[width=1\linewidth]{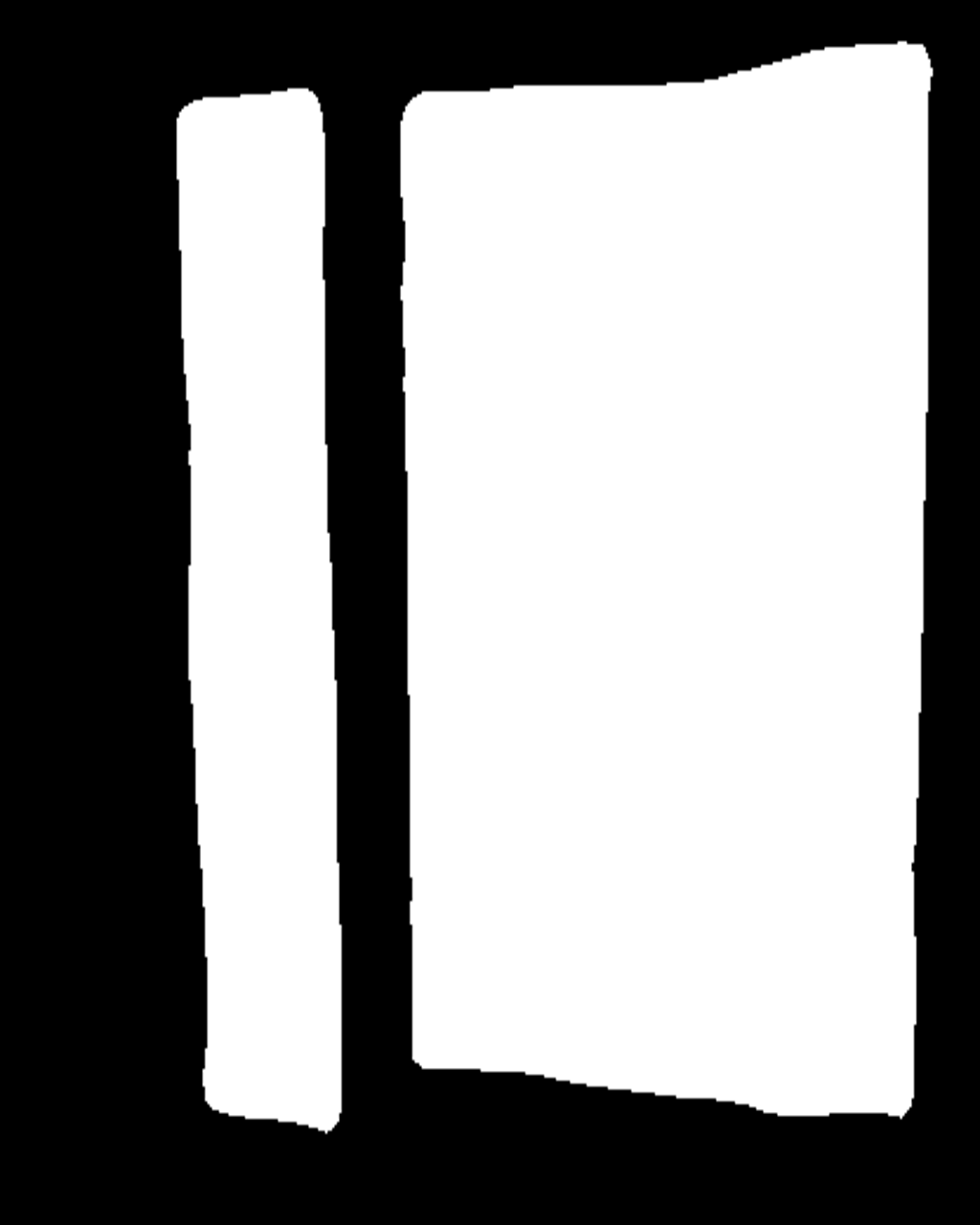}

            \includegraphics[width=1\linewidth]{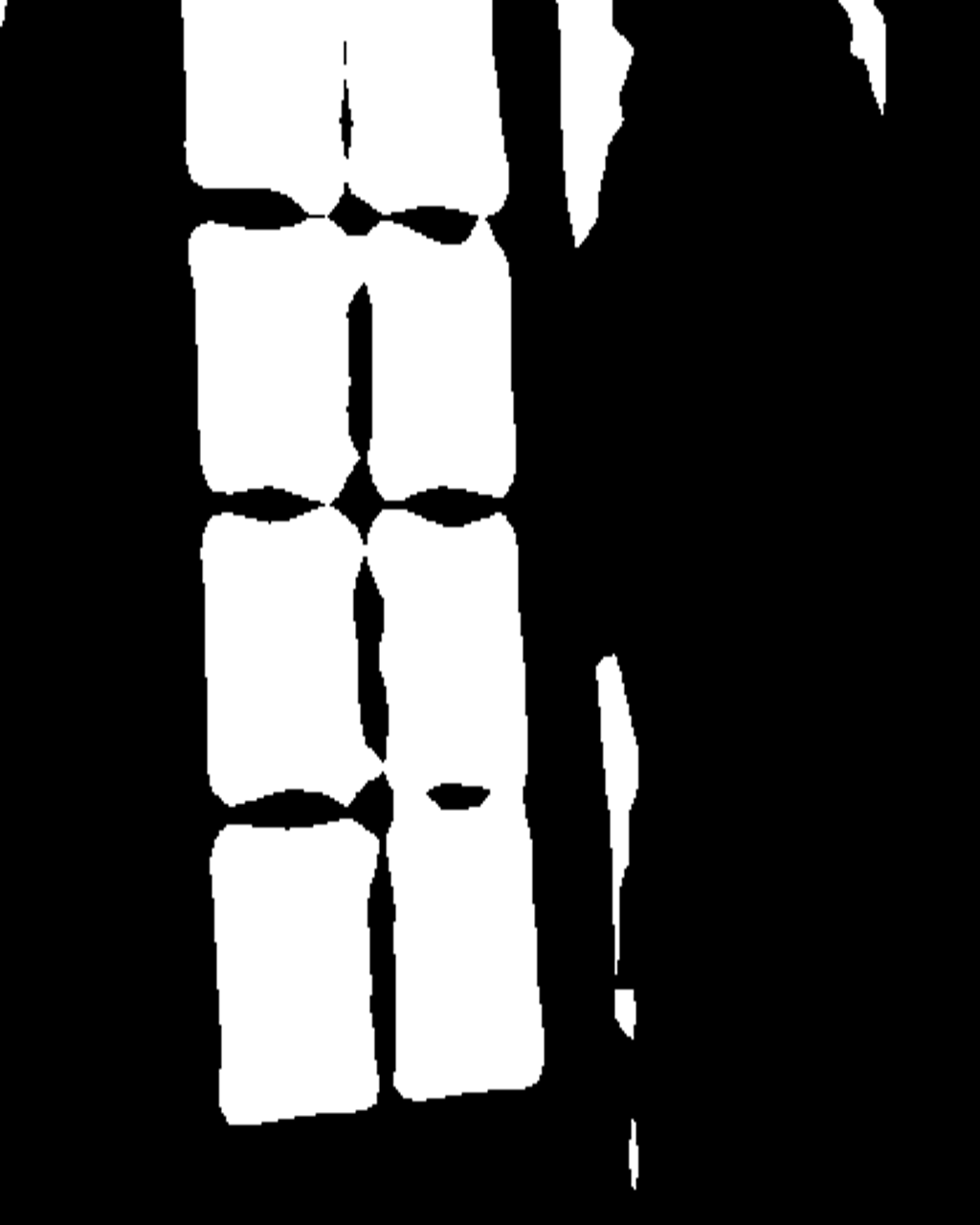}
      \end{minipage}
      }      
      \subfloat[CCNet]{\label{CC}
      \begin{minipage}[t]{0.07\textwidth}
            \centering
            \includegraphics[width=1\linewidth]{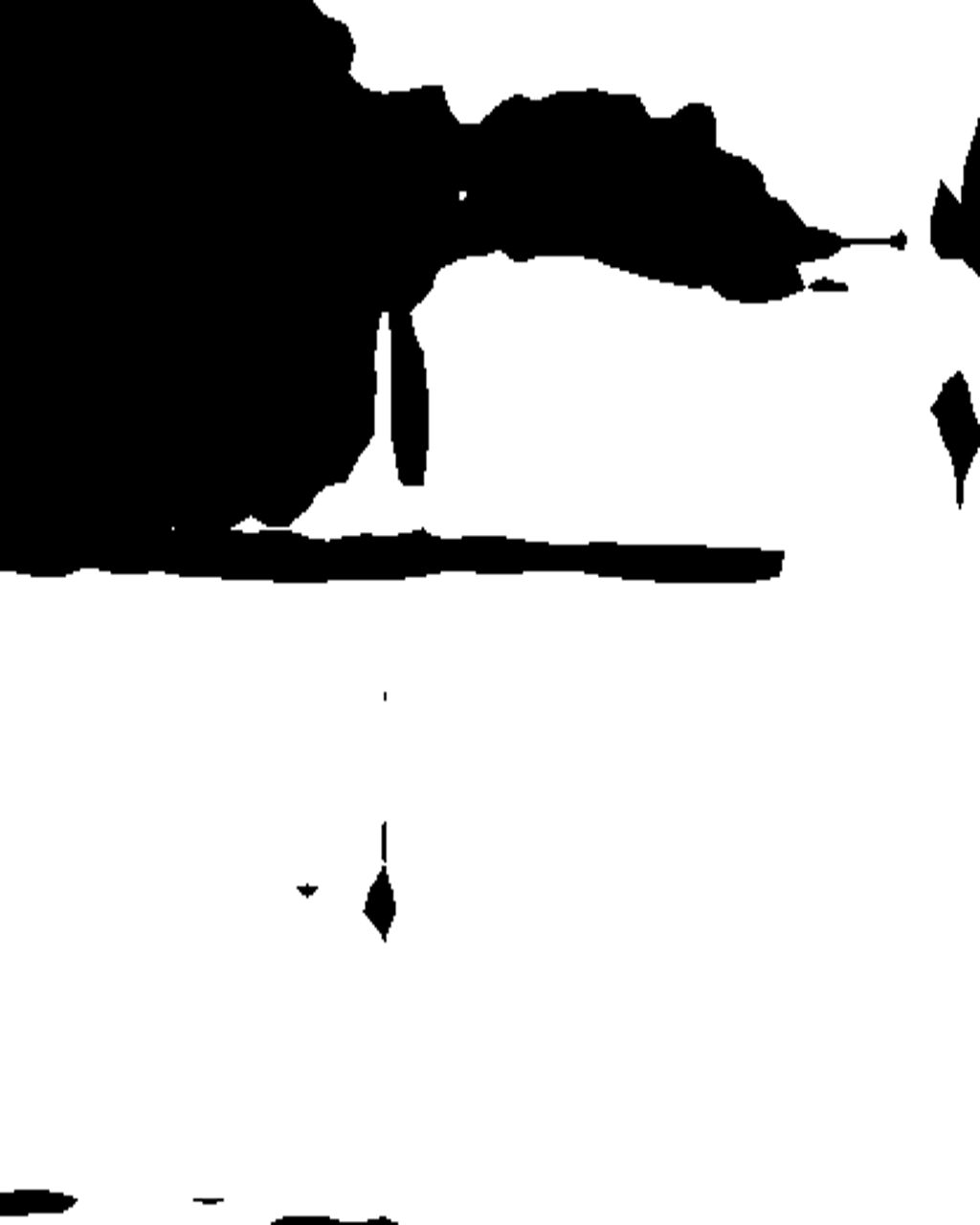}

            \includegraphics[width=1\linewidth]{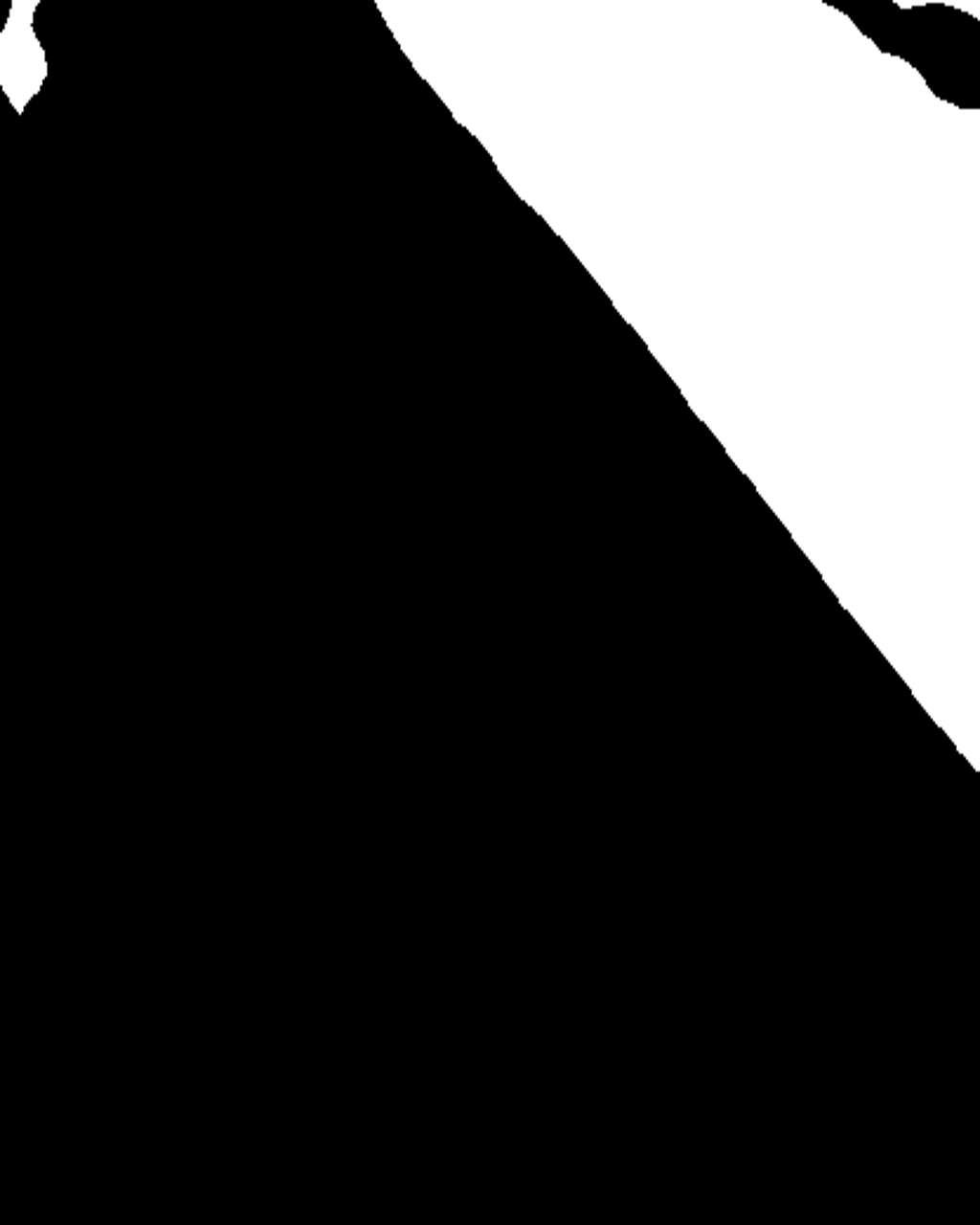}

            \includegraphics[width=1\linewidth]{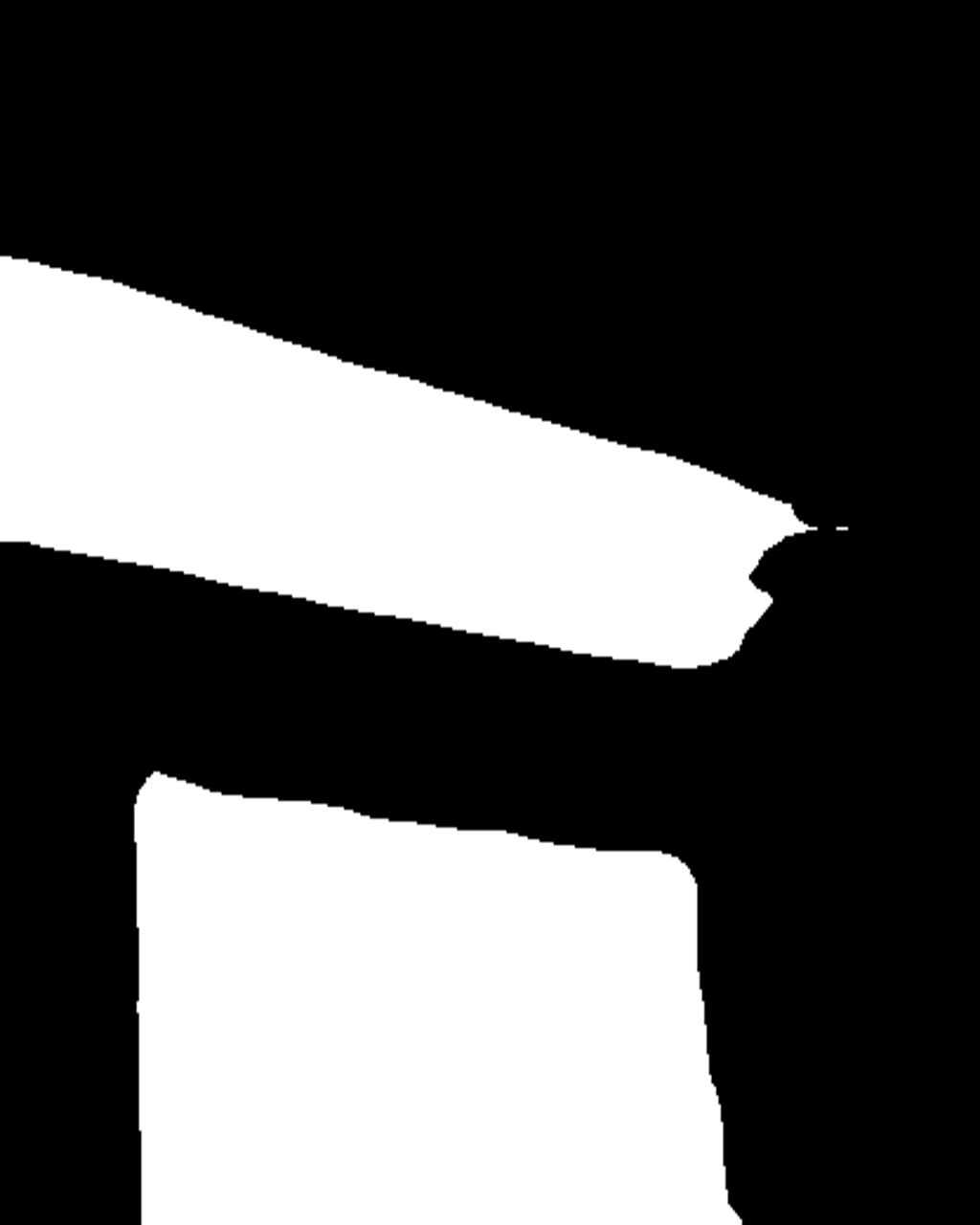}

            \includegraphics[width=1\linewidth]{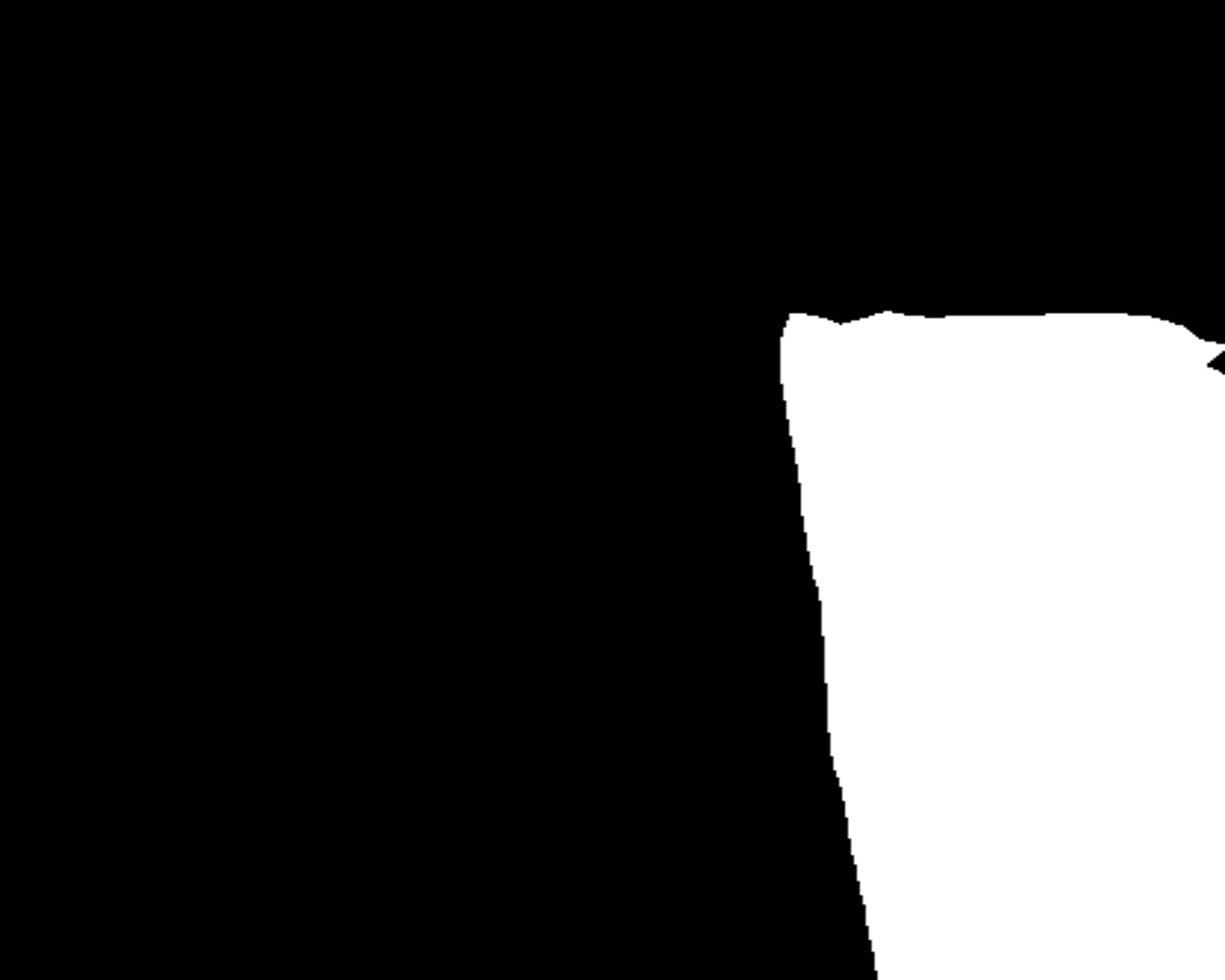}

            \includegraphics[width=1\linewidth]{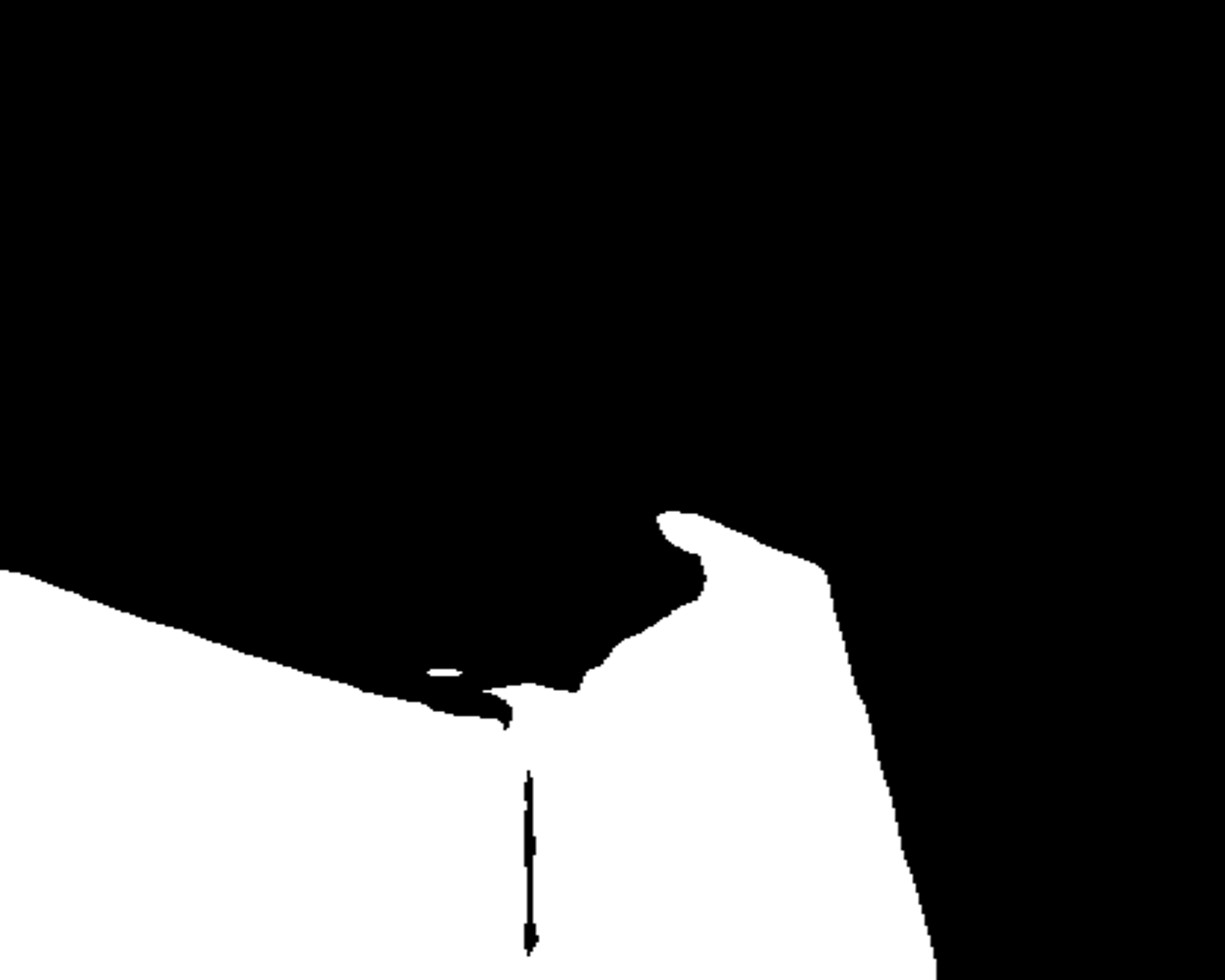}

            \includegraphics[width=1\linewidth]{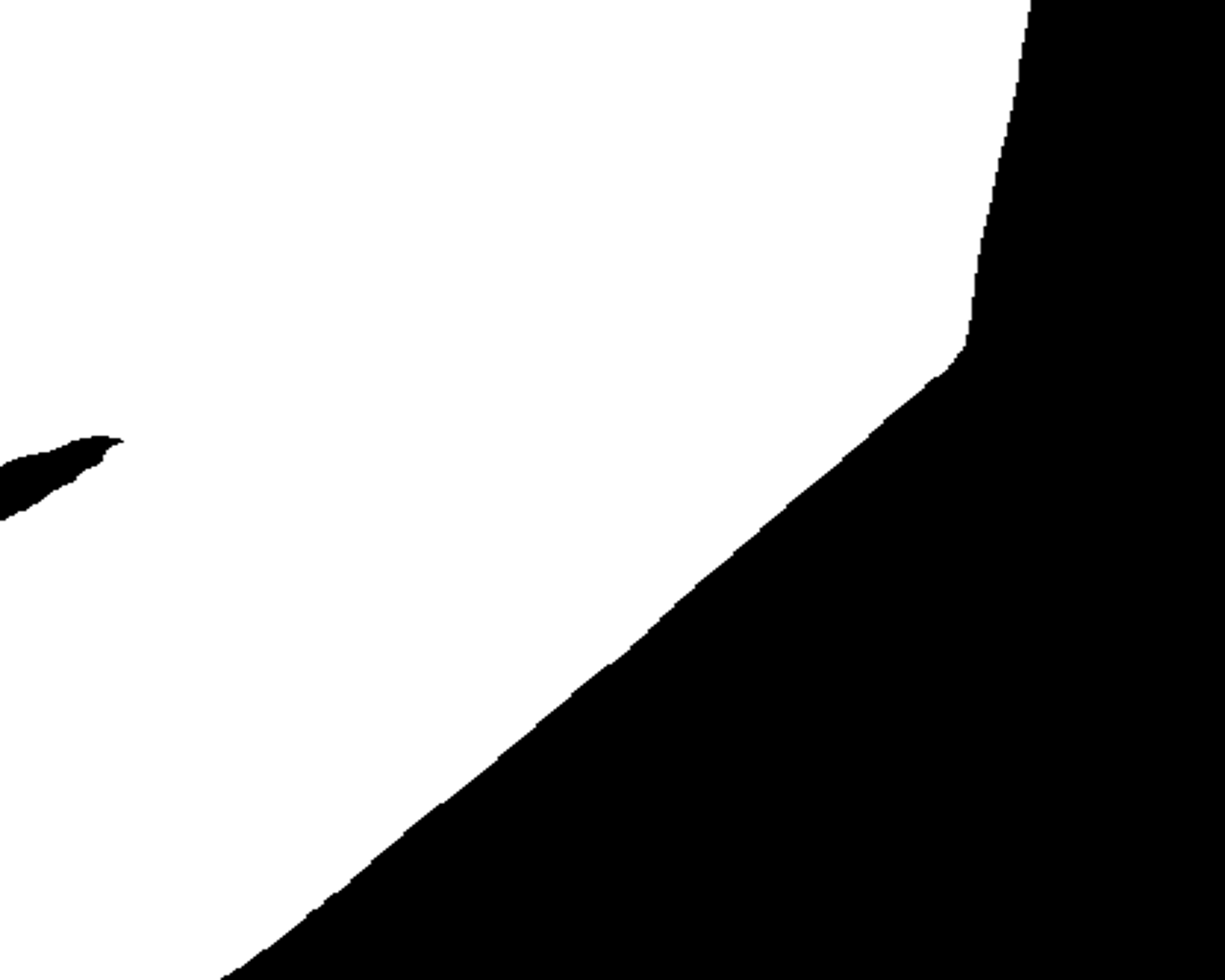}

            \includegraphics[width=1\linewidth]{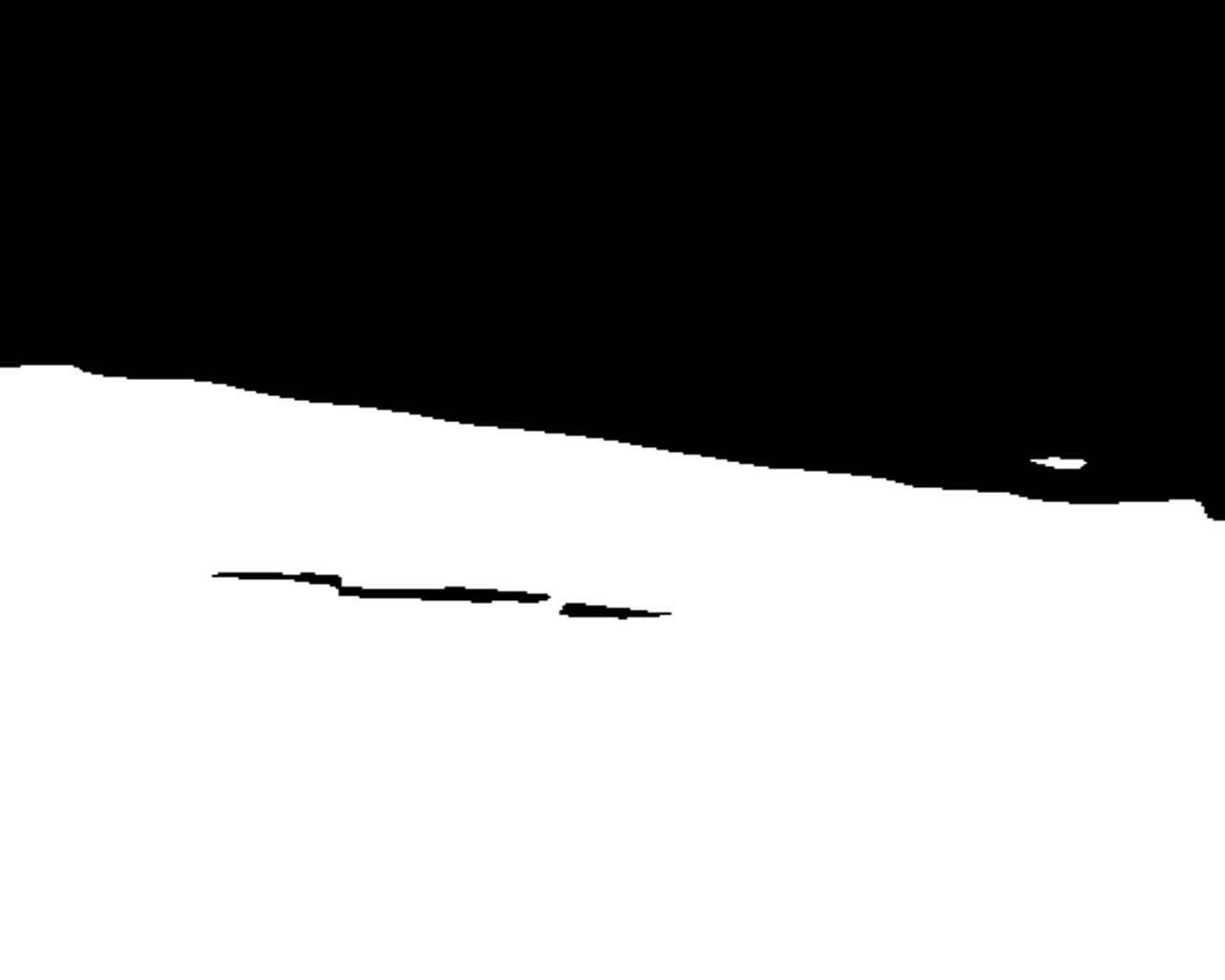}
            
            \includegraphics[width=1\linewidth]{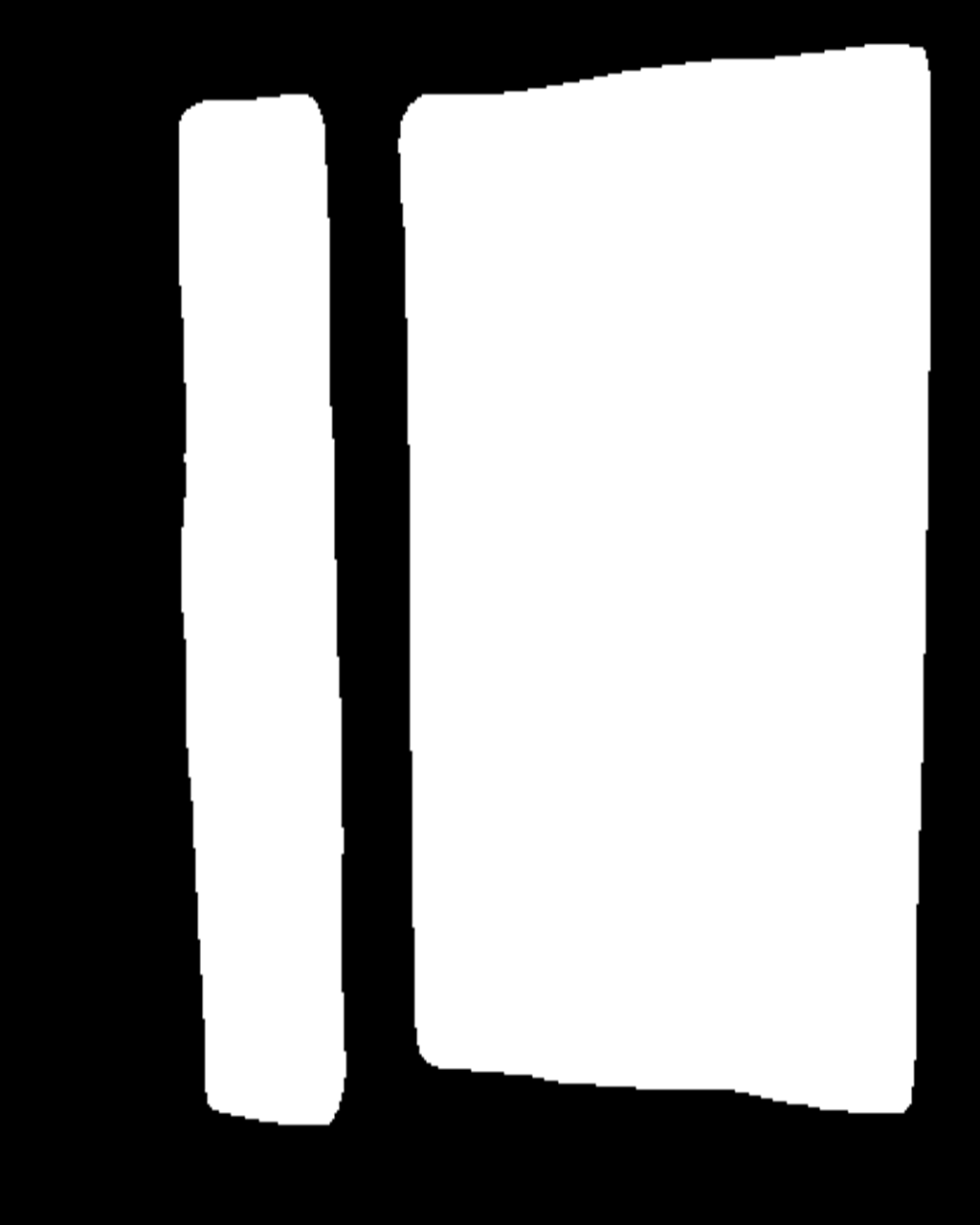}

            \includegraphics[width=1\linewidth]{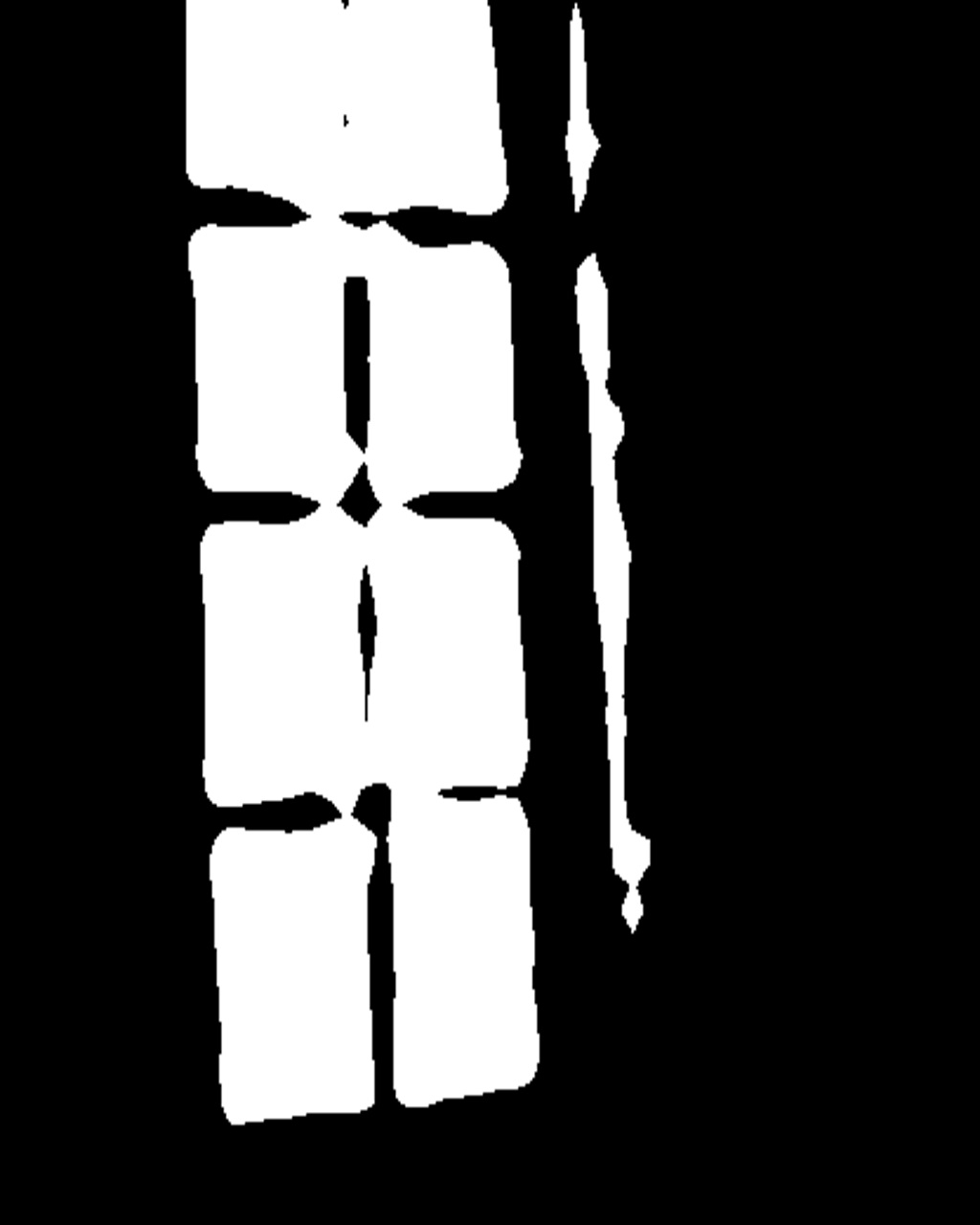}
      \end{minipage}
      }      
      \subfloat[EGNet]{\label{EG}
      \begin{minipage}[t]{0.07\textwidth}
            \centering
            \includegraphics[width=1\linewidth]{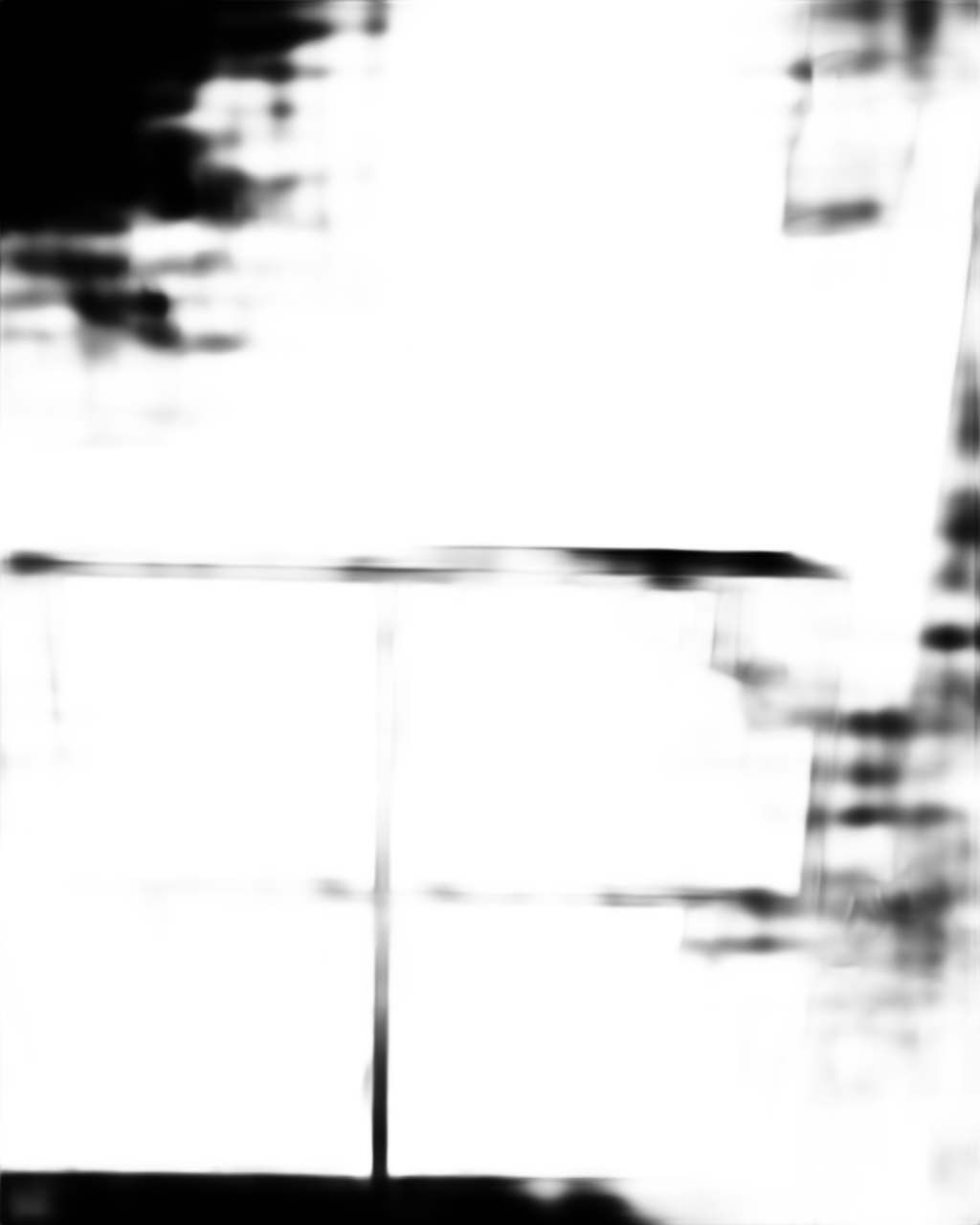}

            \includegraphics[width=1\linewidth]{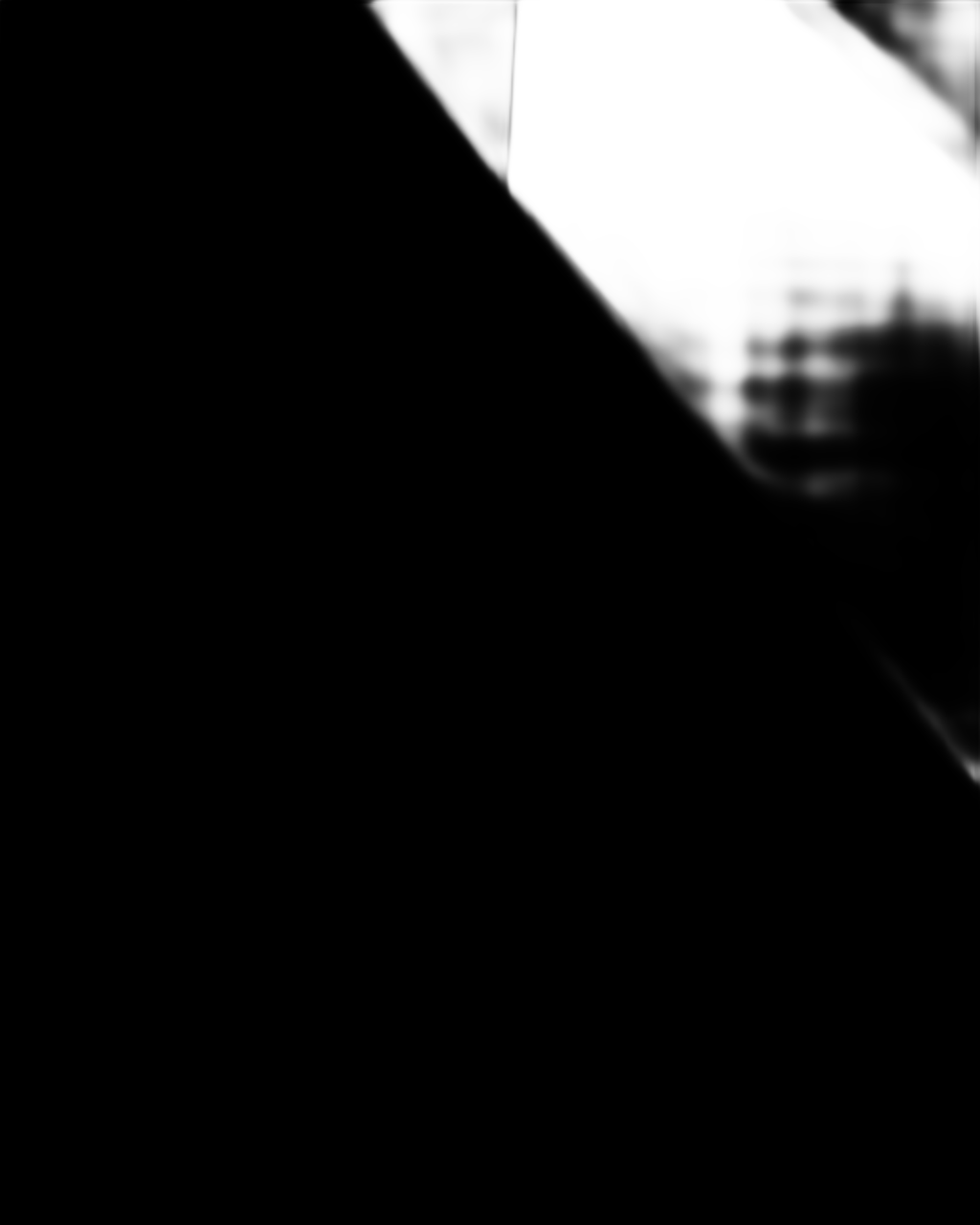}

            \includegraphics[width=1\linewidth]{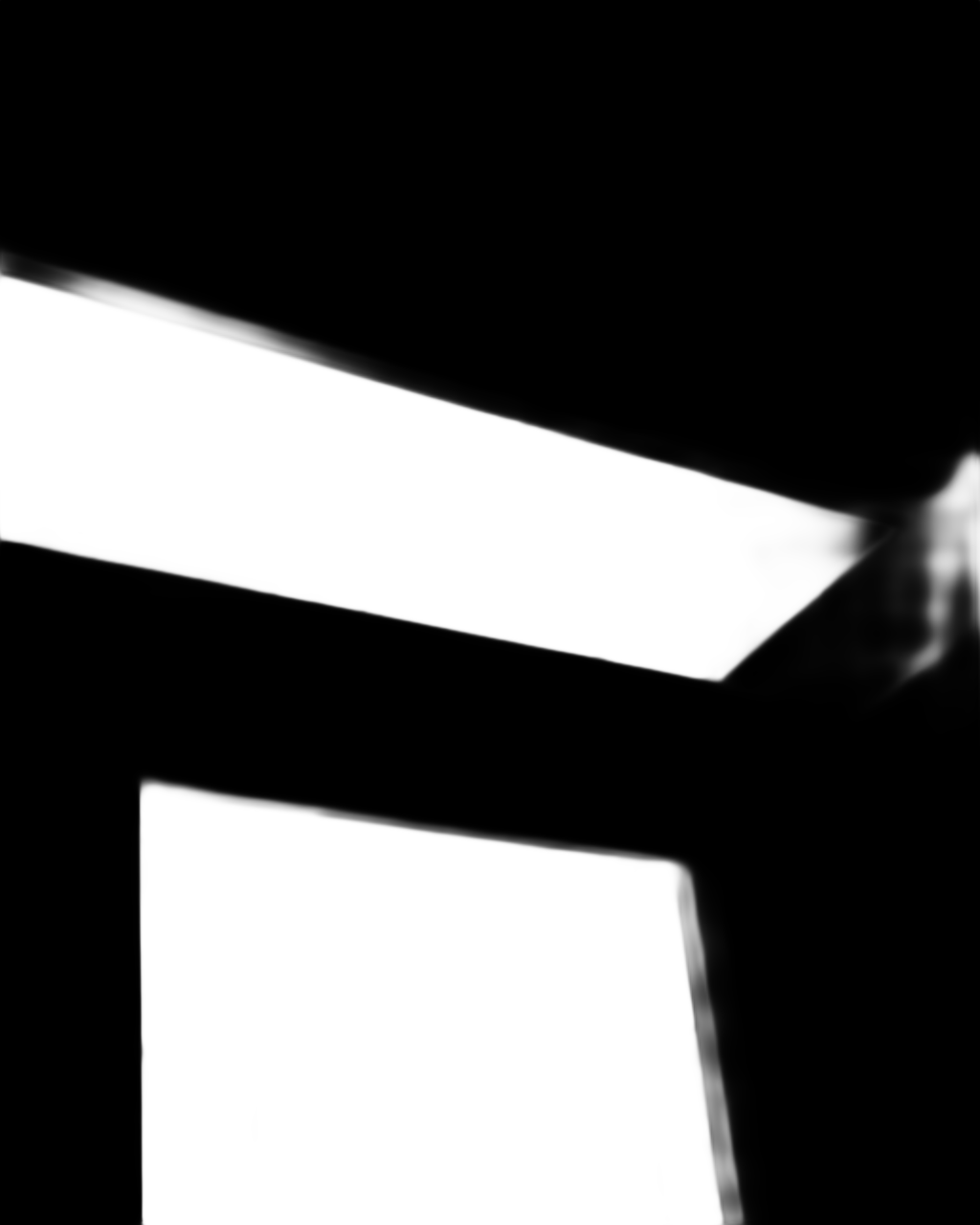}

            \includegraphics[width=1\linewidth]{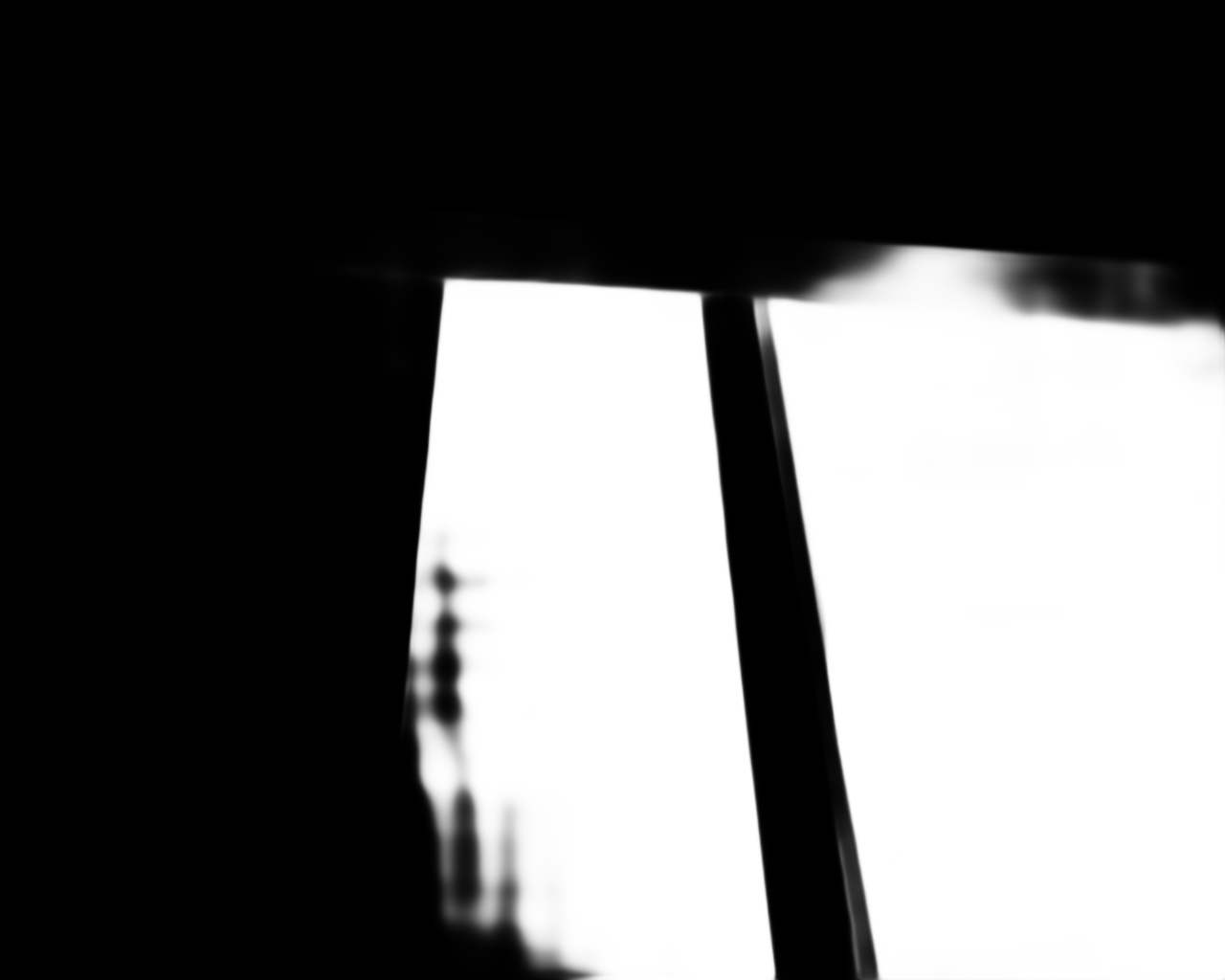}

            \includegraphics[width=1\linewidth]{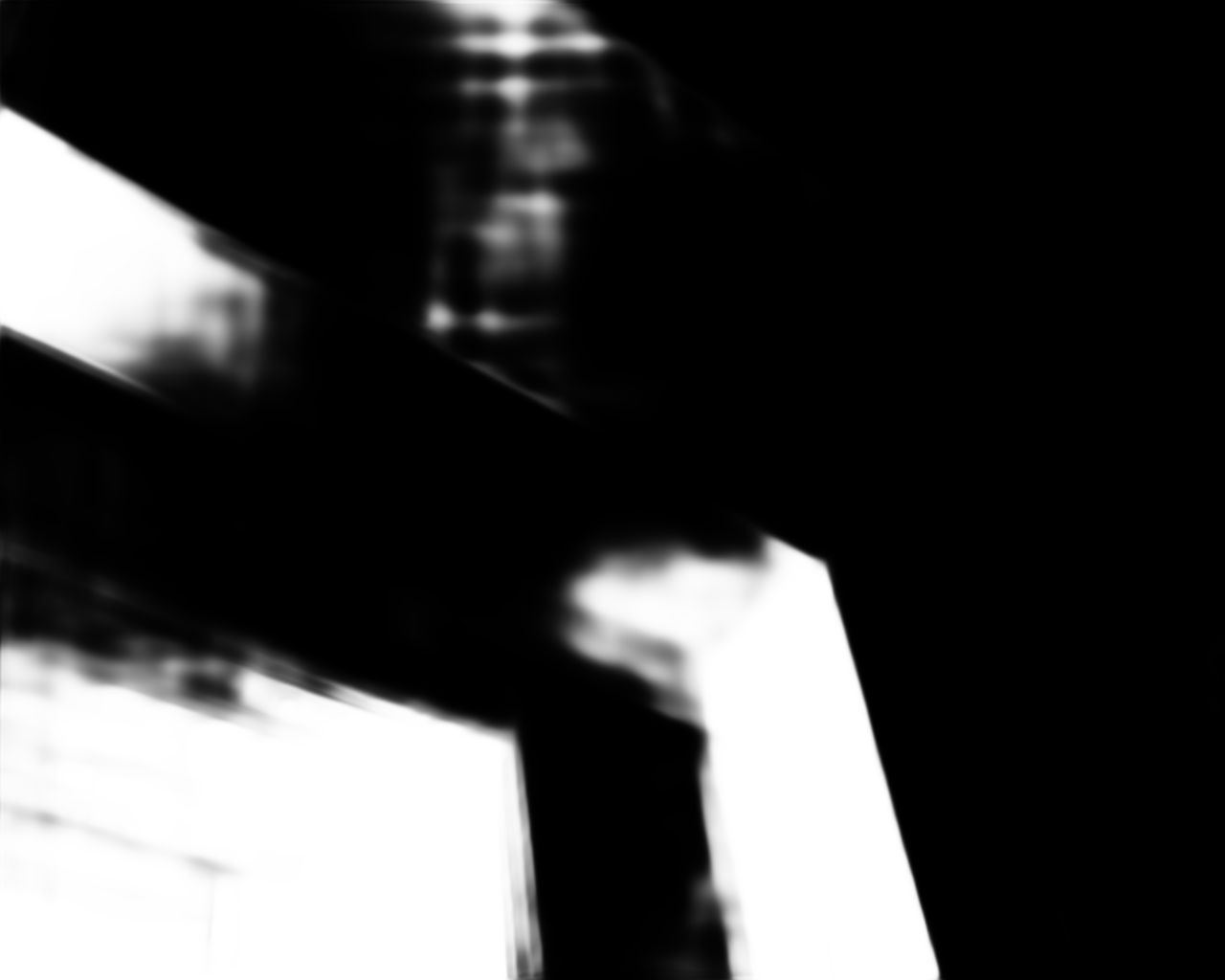}

            \includegraphics[width=1\linewidth]{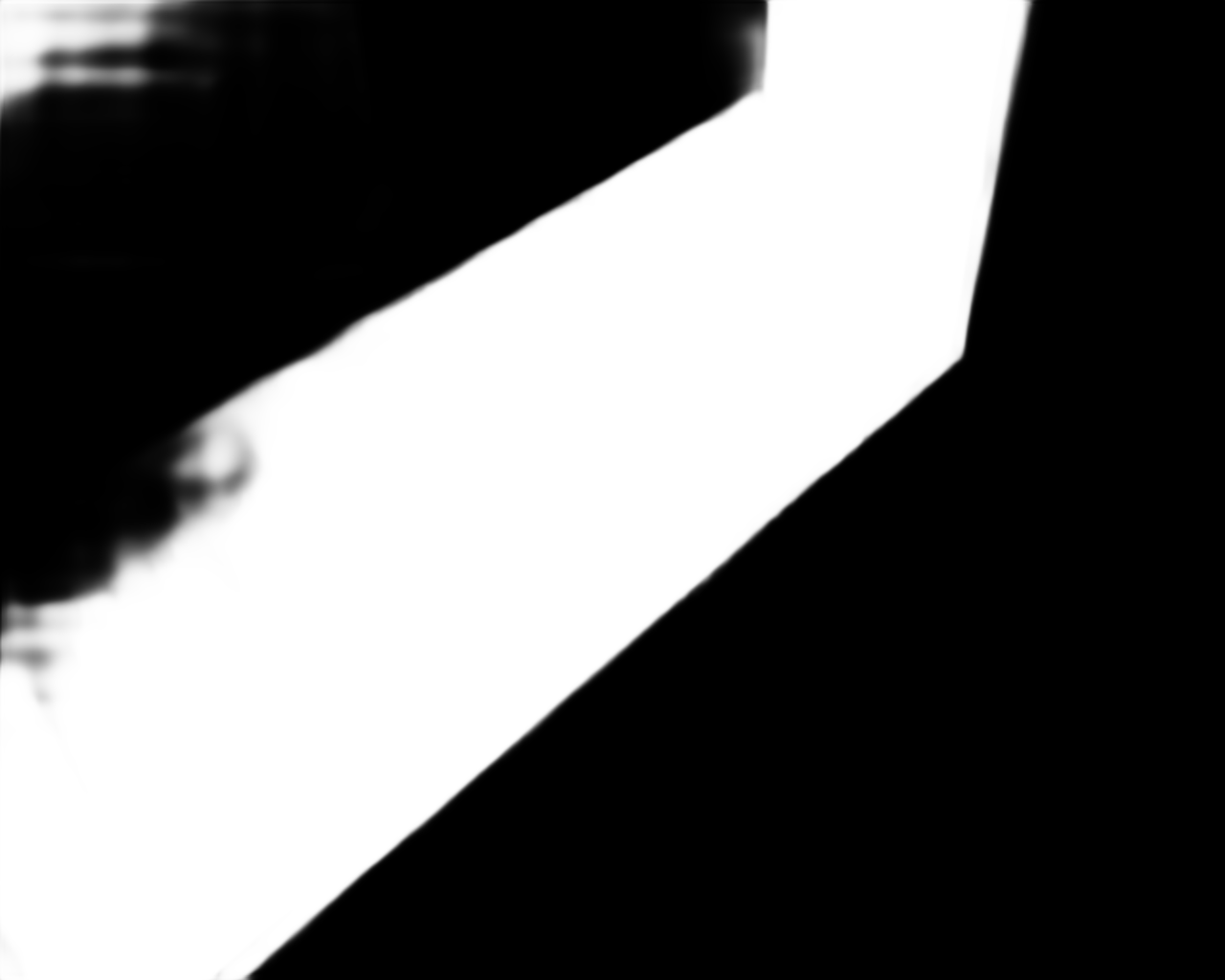}

            \includegraphics[width=1\linewidth]{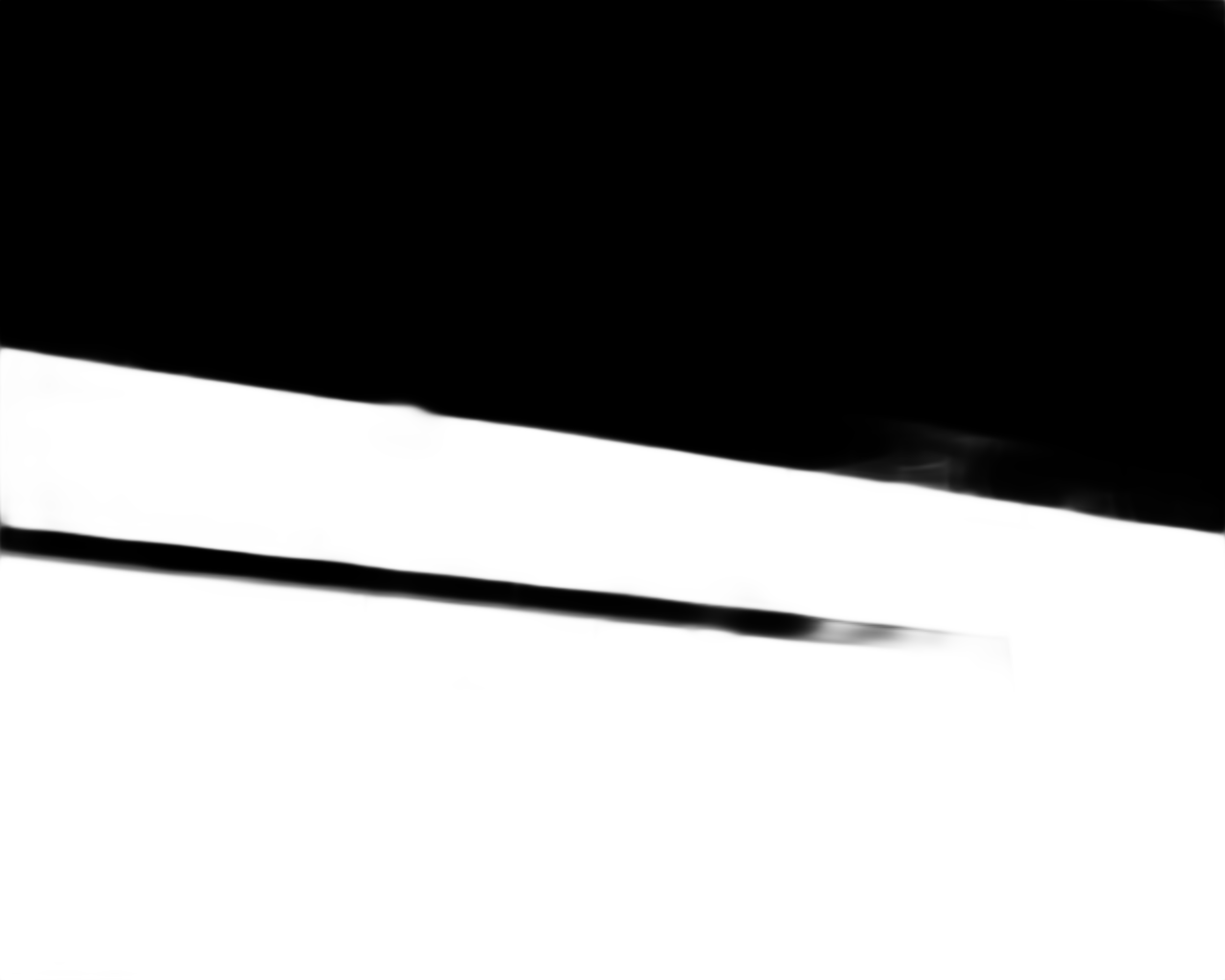}
            
            \includegraphics[width=1\linewidth]{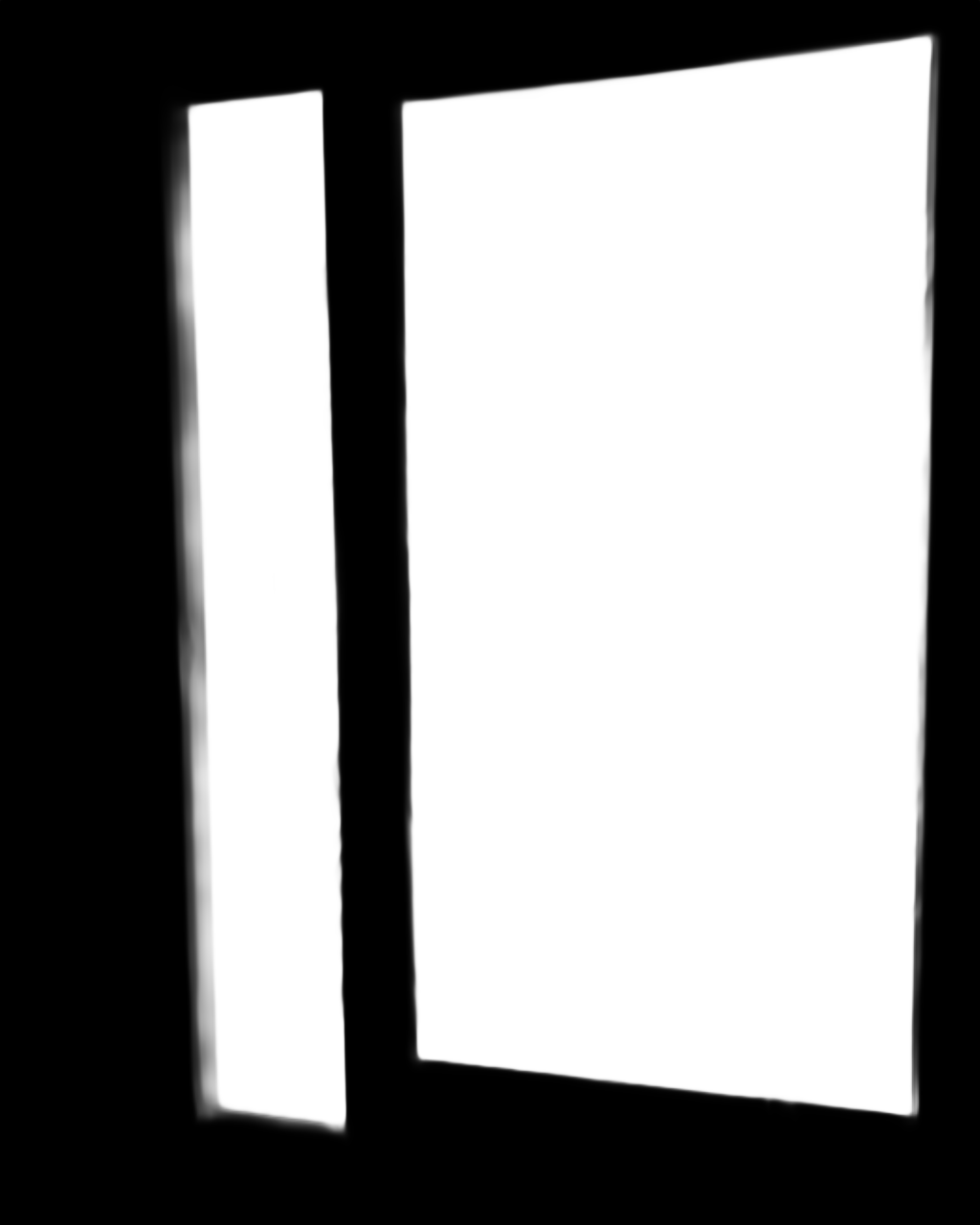}

            \includegraphics[width=1\linewidth]{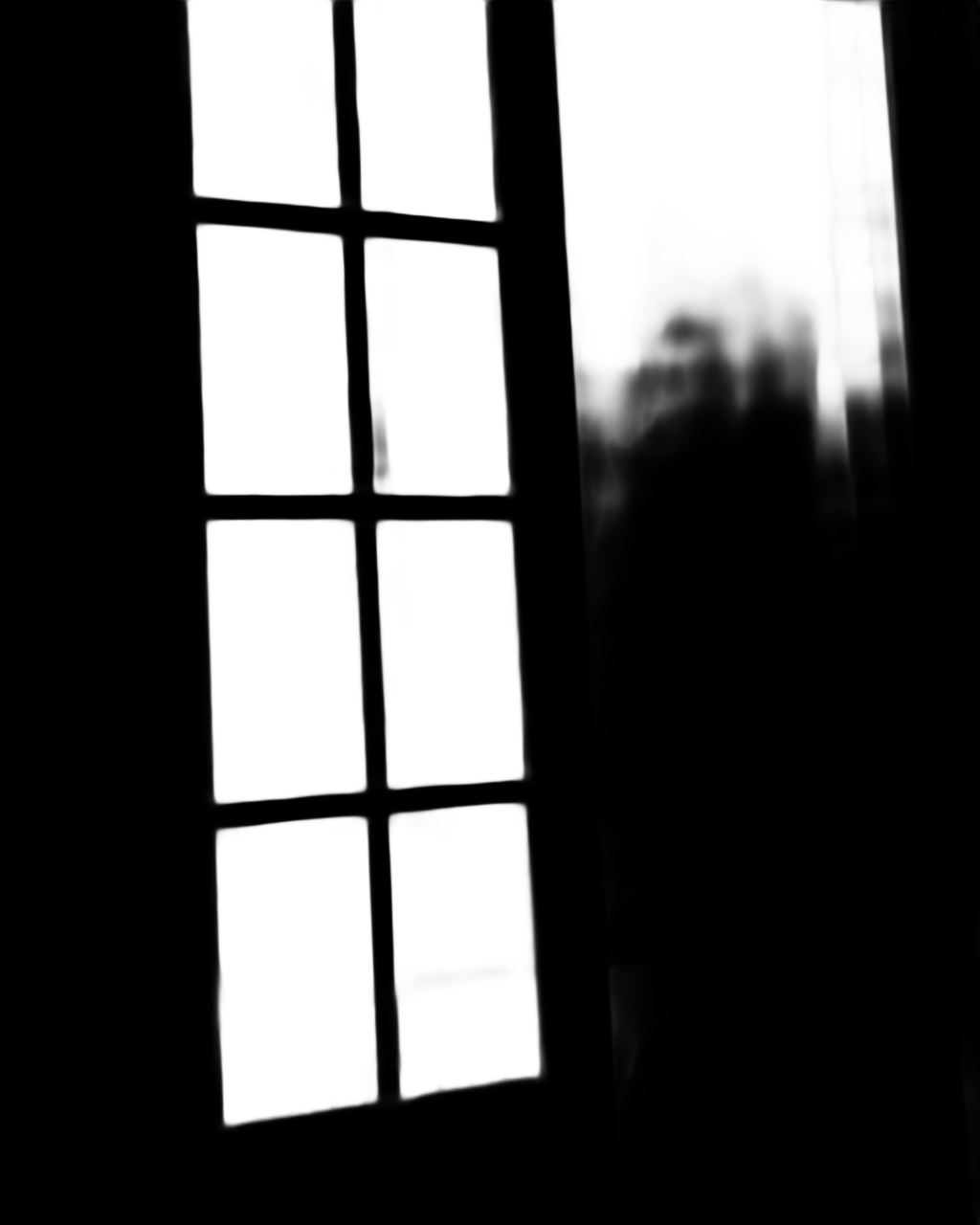}
      \end{minipage}
      }         
      \subfloat[LDF]{\label{LDF}
      \begin{minipage}[t]{0.07\textwidth}
            \centering
            \includegraphics[width=1\linewidth]{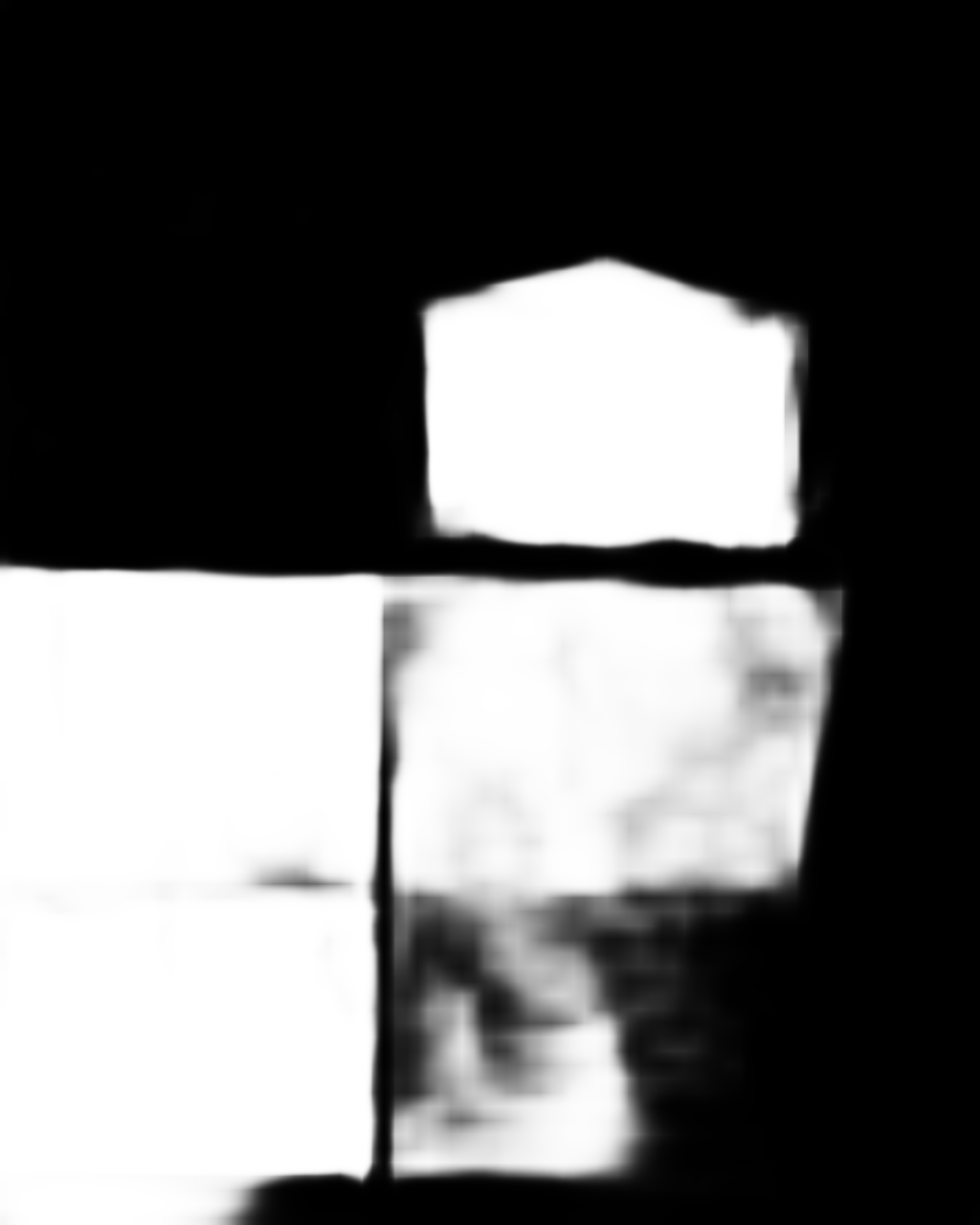}

            \includegraphics[width=1\linewidth]{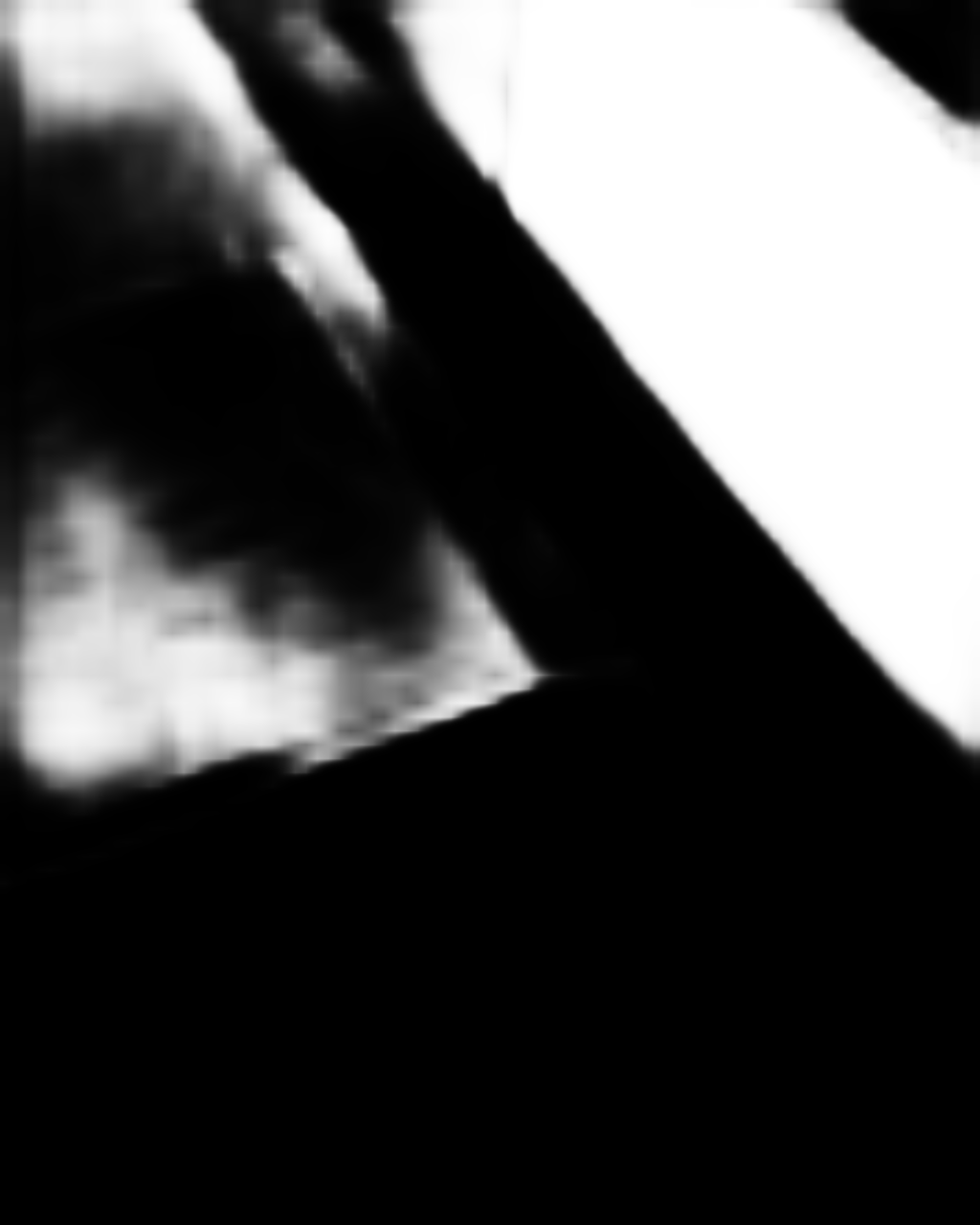}

            \includegraphics[width=1\linewidth]{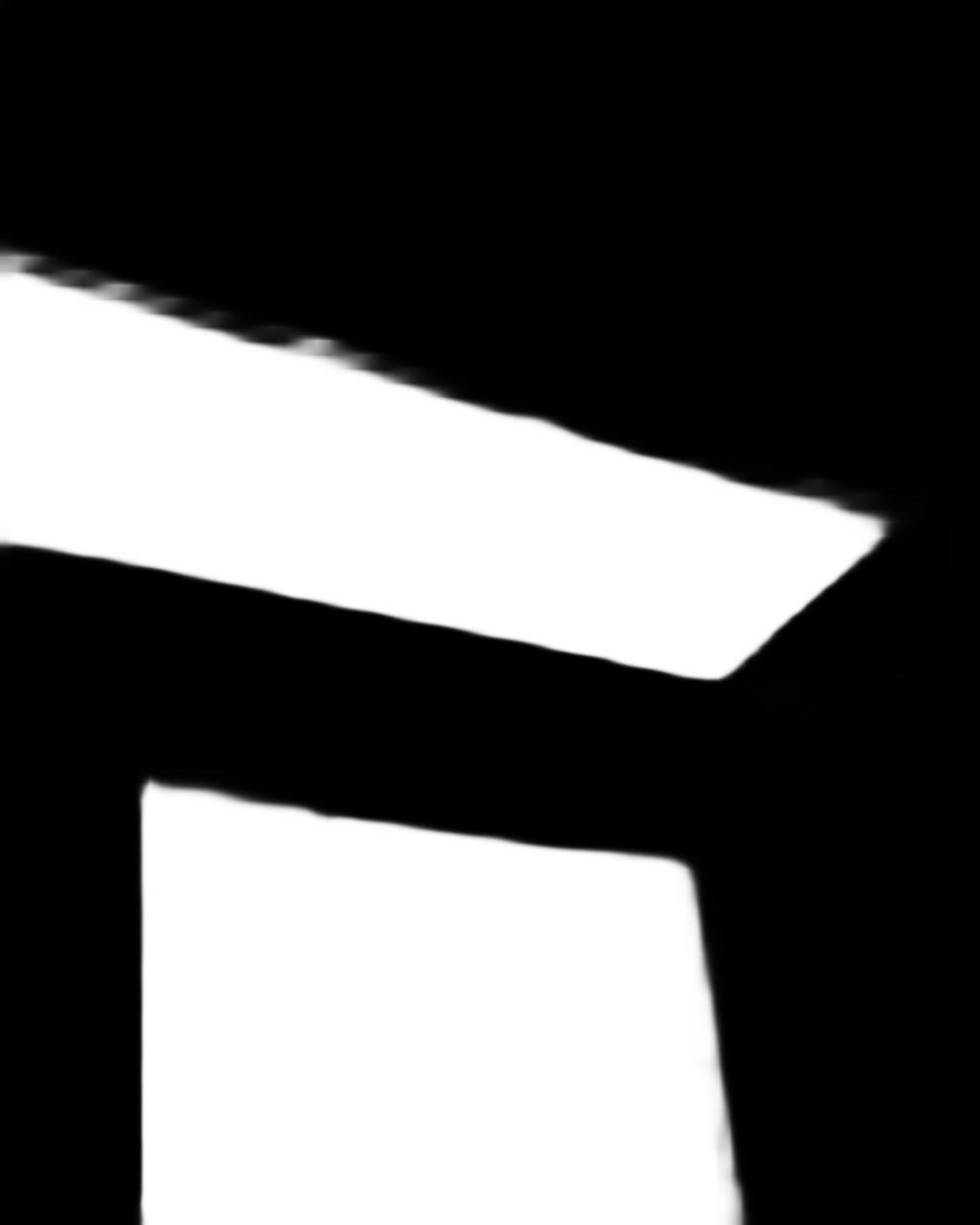}

            \includegraphics[width=1\linewidth]{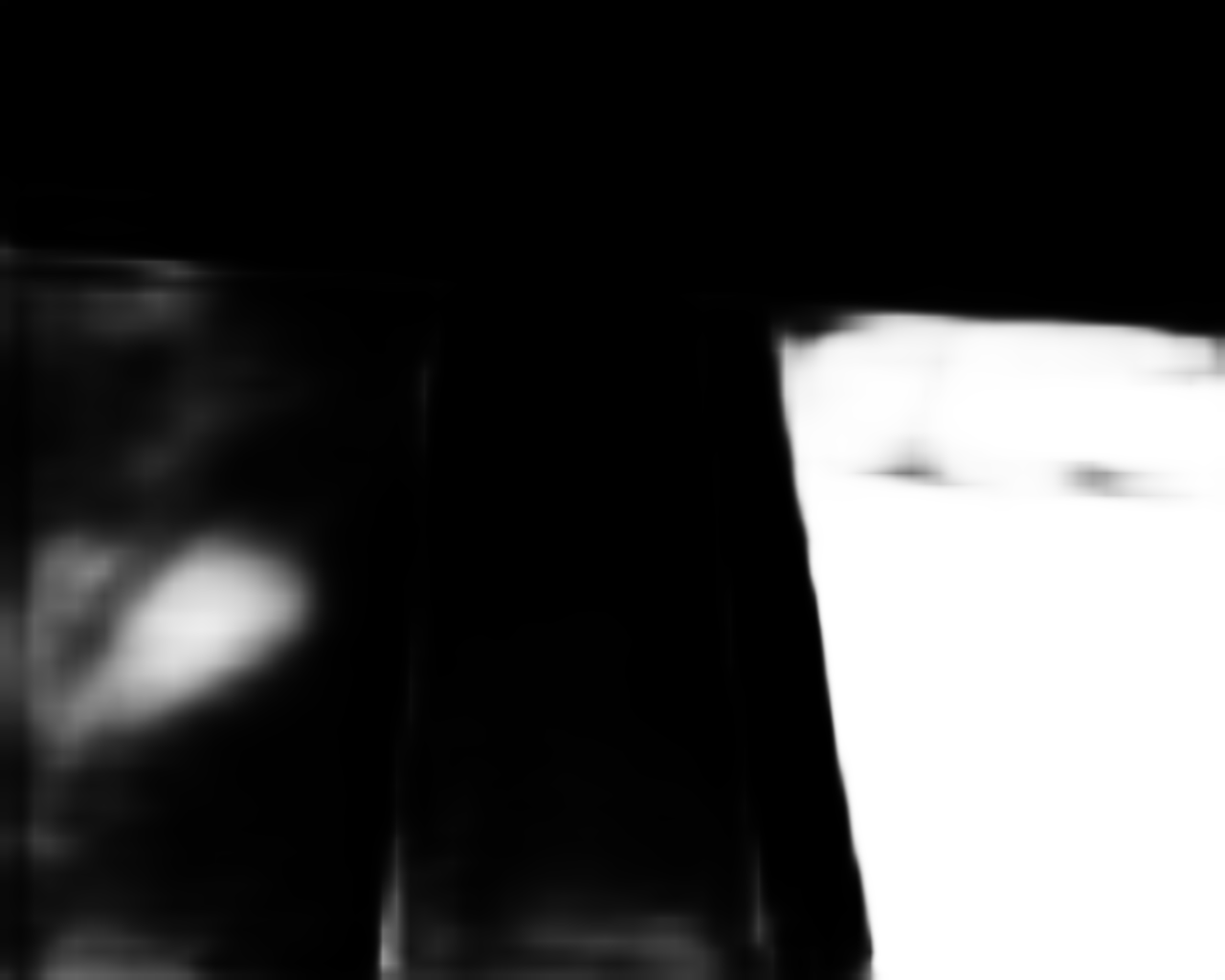}

            \includegraphics[width=1\linewidth]{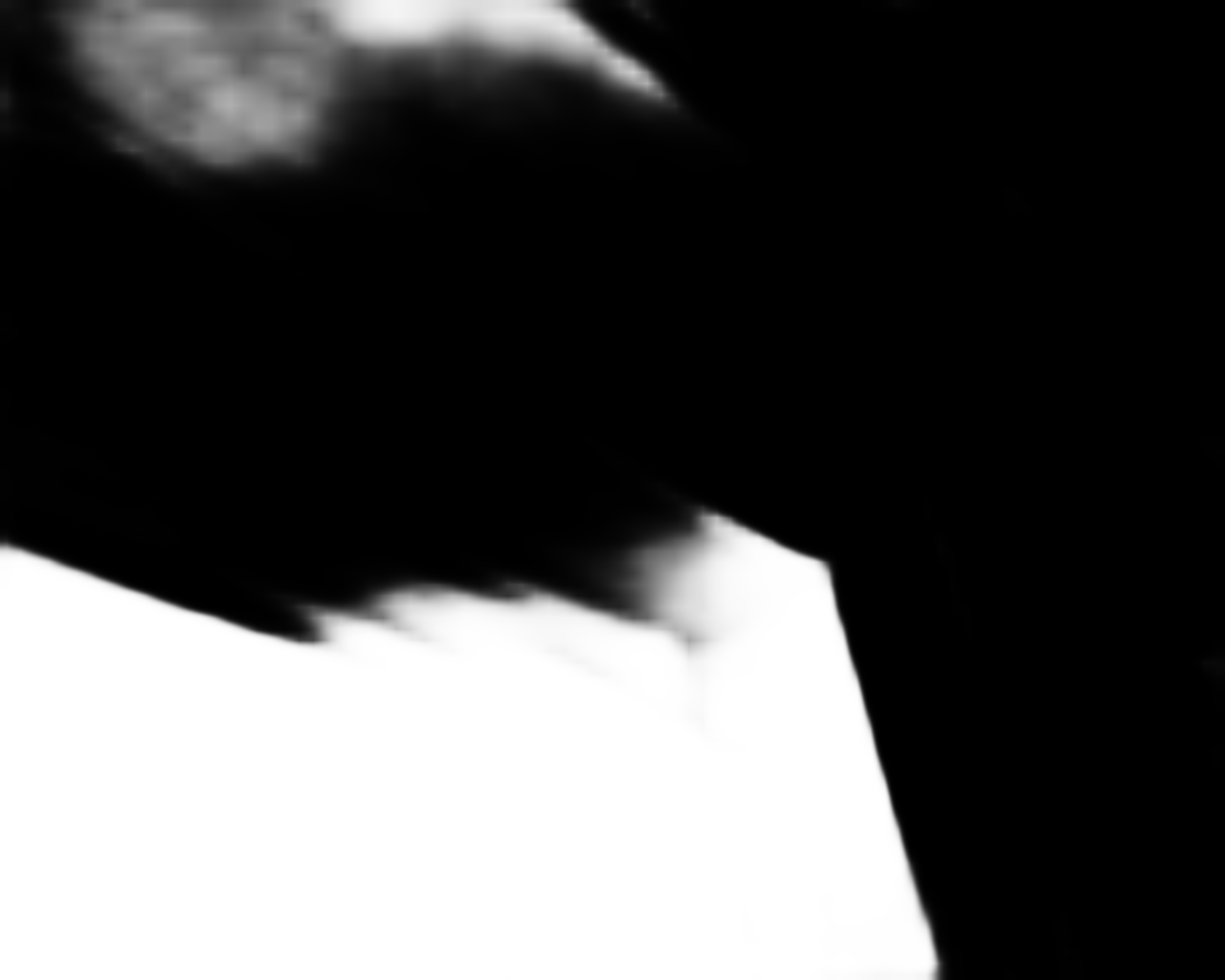}

            \includegraphics[width=1\linewidth]{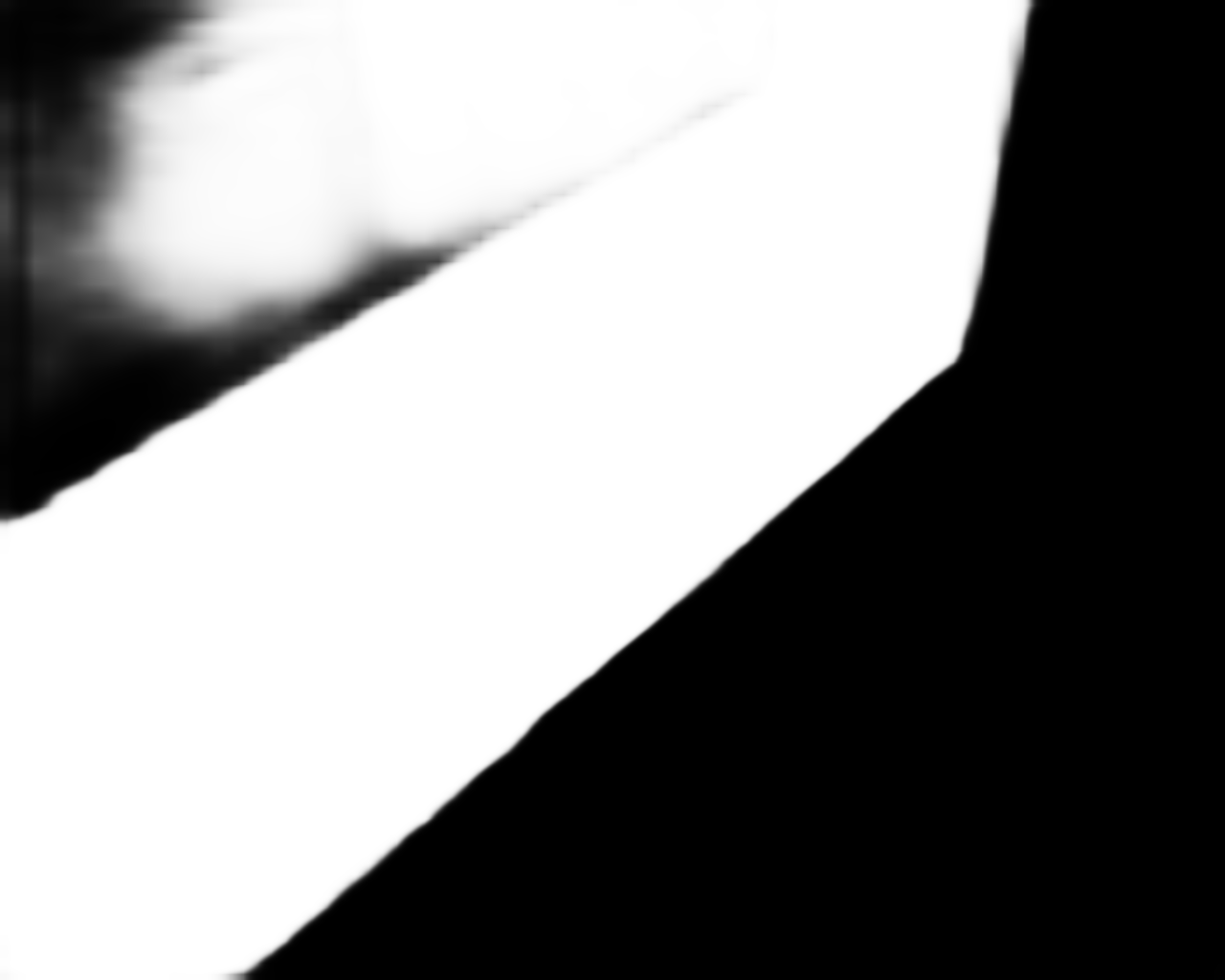}

            \includegraphics[width=1\linewidth]{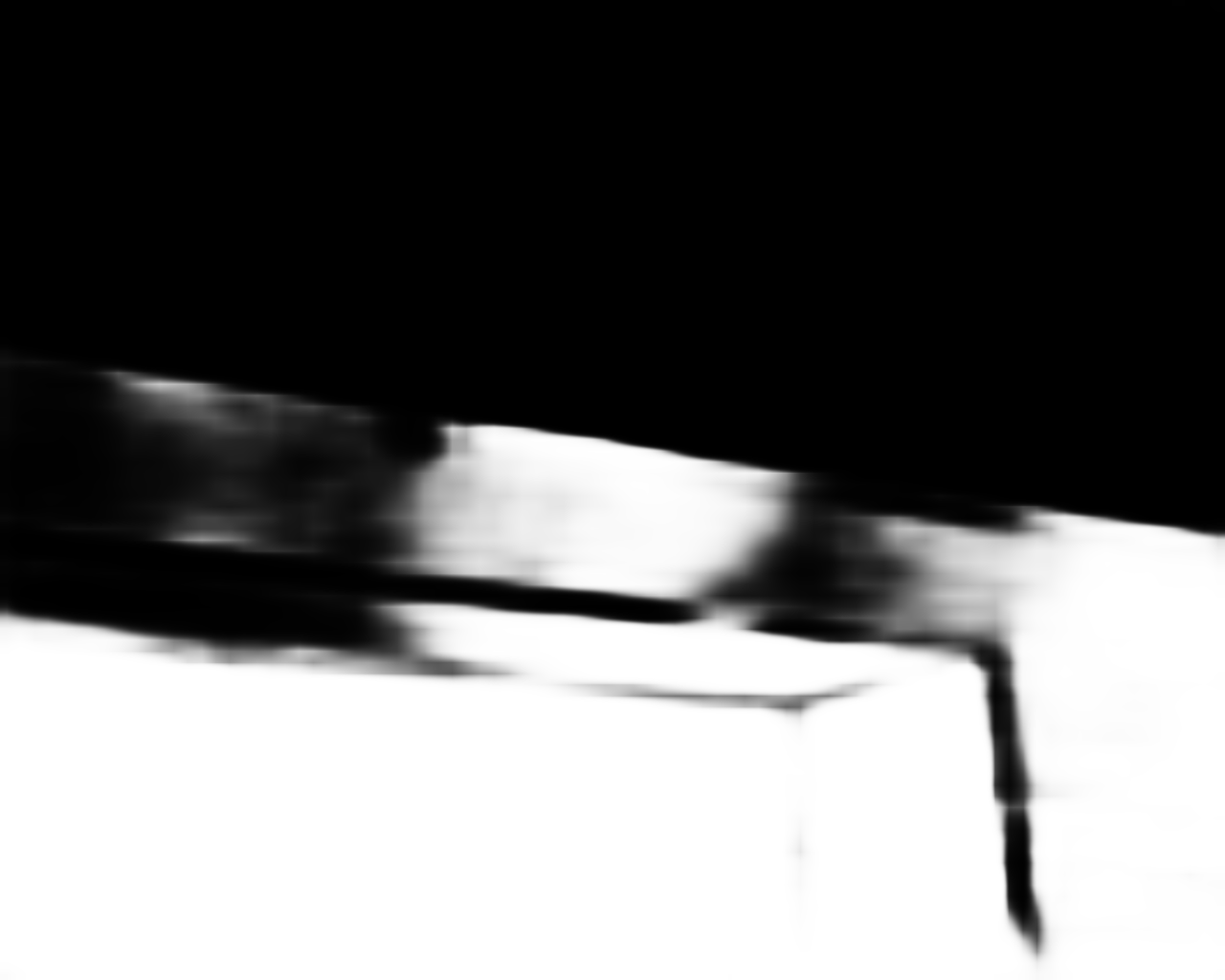}
            
            \includegraphics[width=1\linewidth]{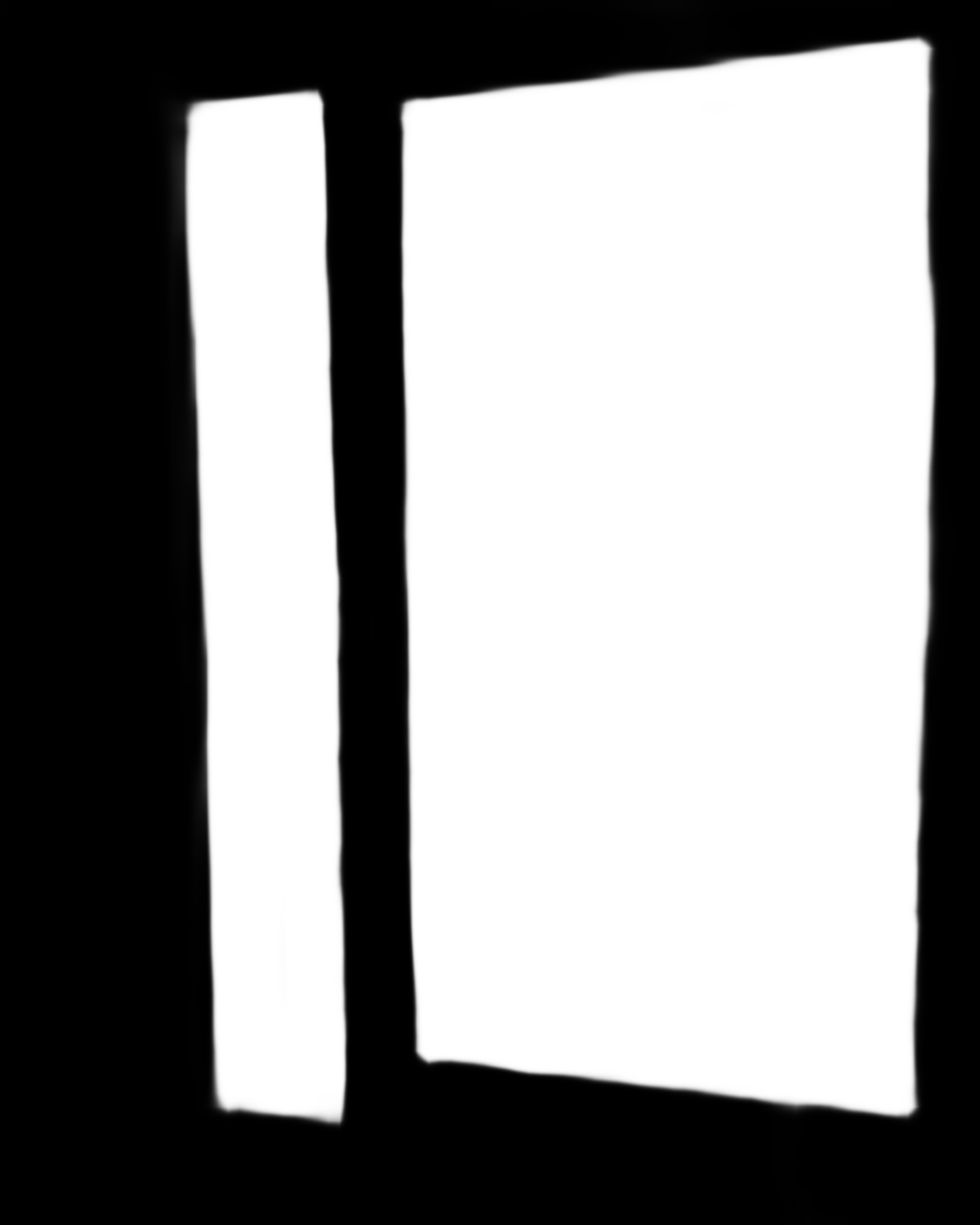}

            \includegraphics[width=1\linewidth]{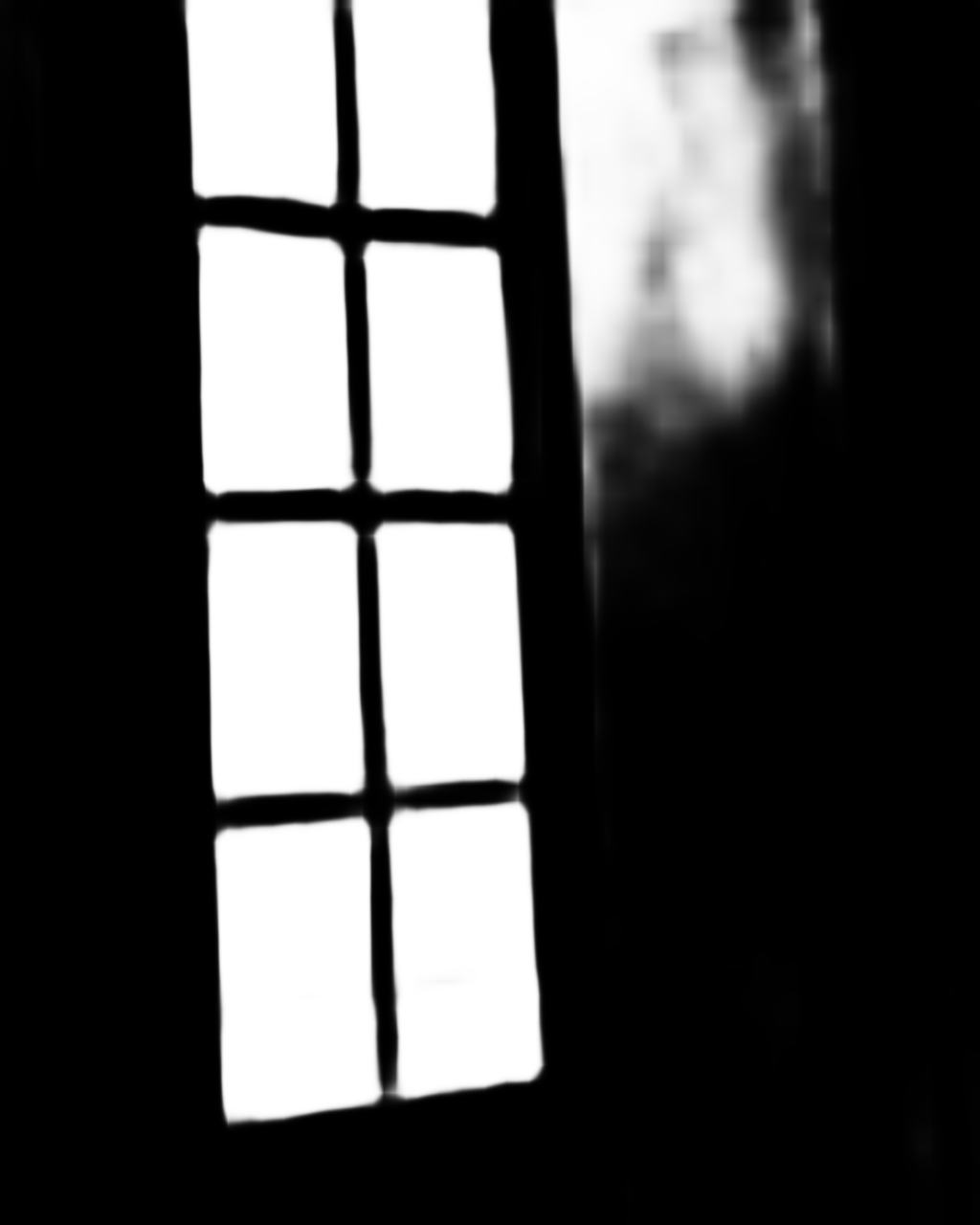}
      \end{minipage}
      }  
      \subfloat[BDRAR]{\label{BDRAR}
      \begin{minipage}[t]{0.07\textwidth}
            \centering
            \includegraphics[width=1\linewidth]{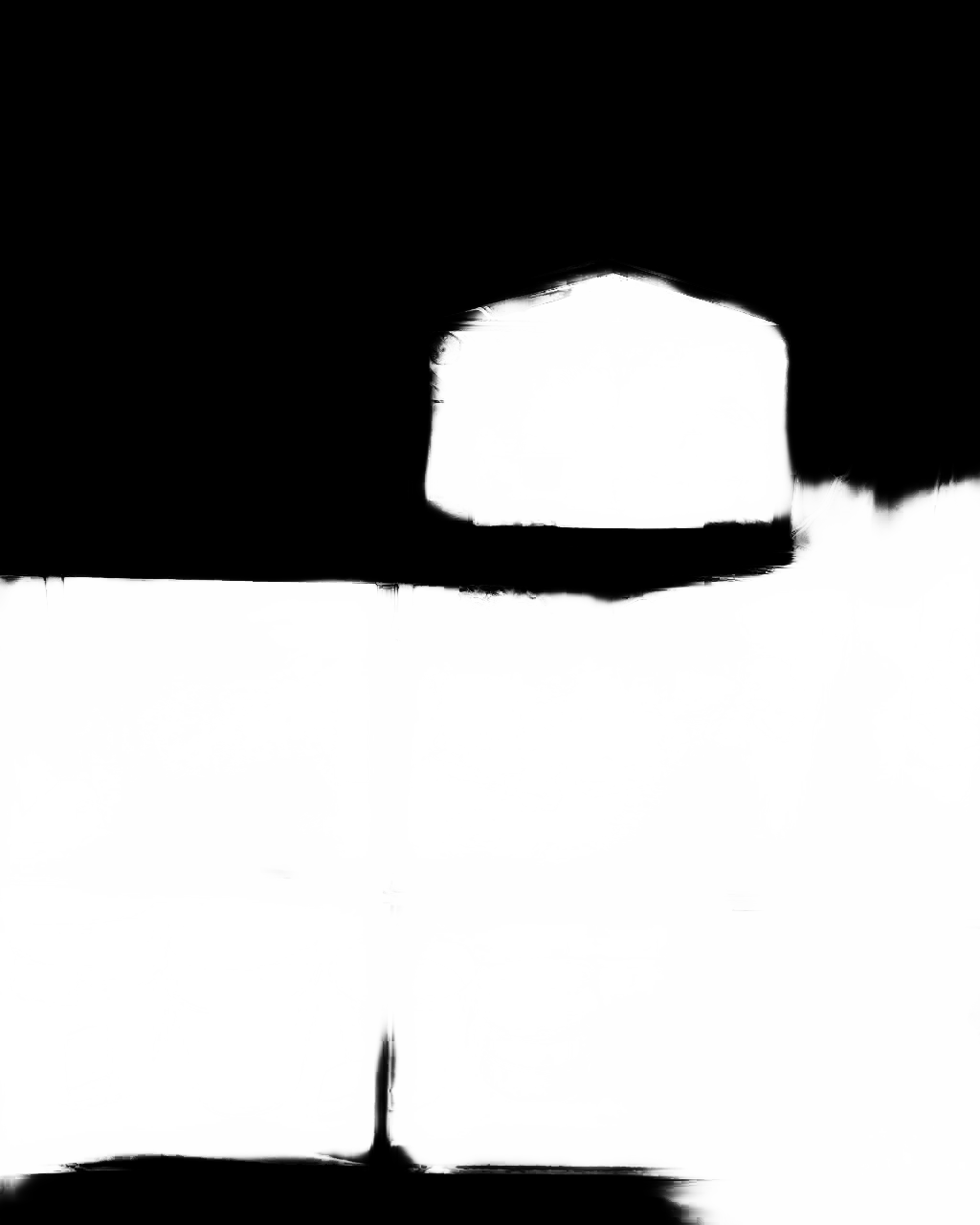}

            \includegraphics[width=1\linewidth]{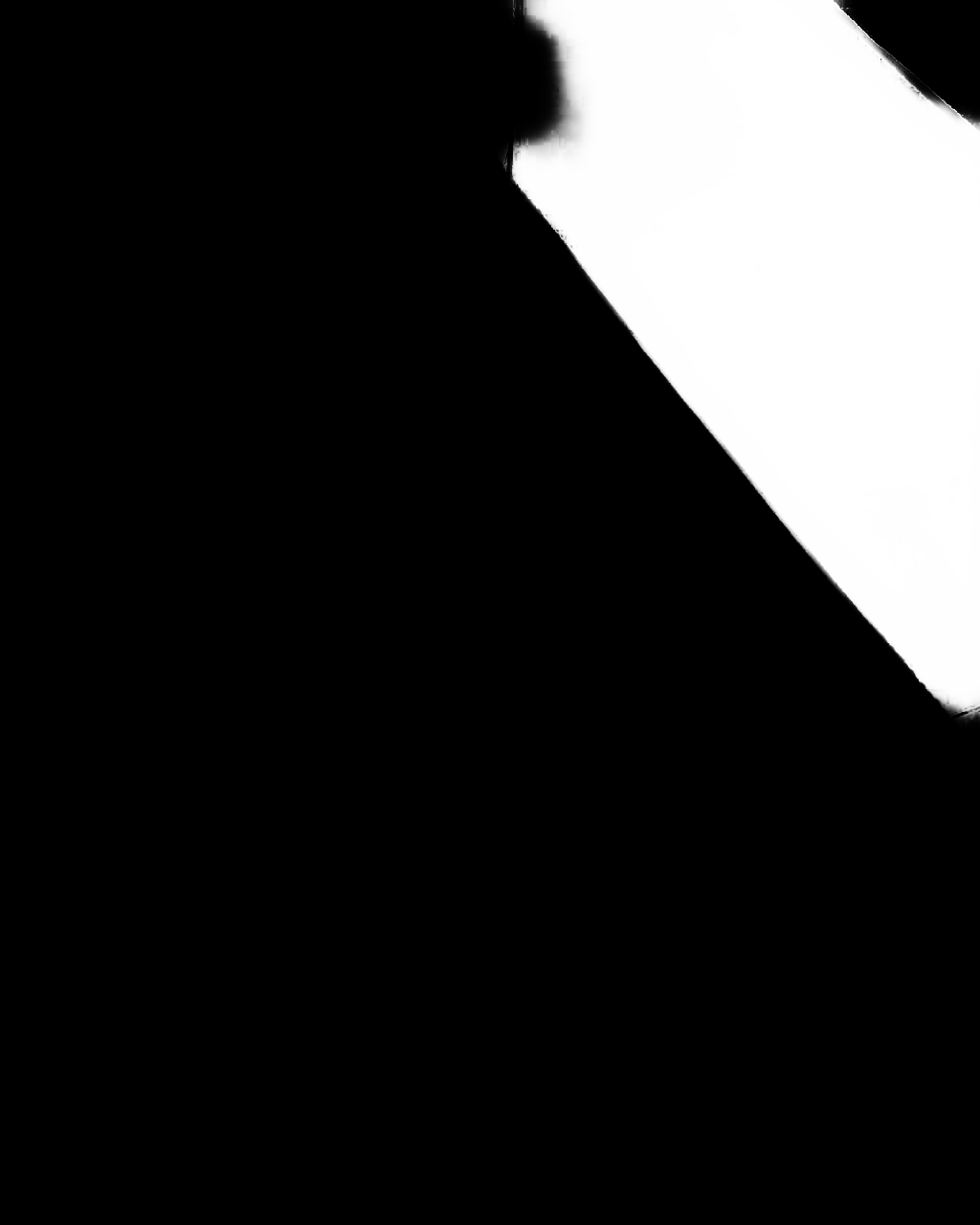}

            \includegraphics[width=1\linewidth]{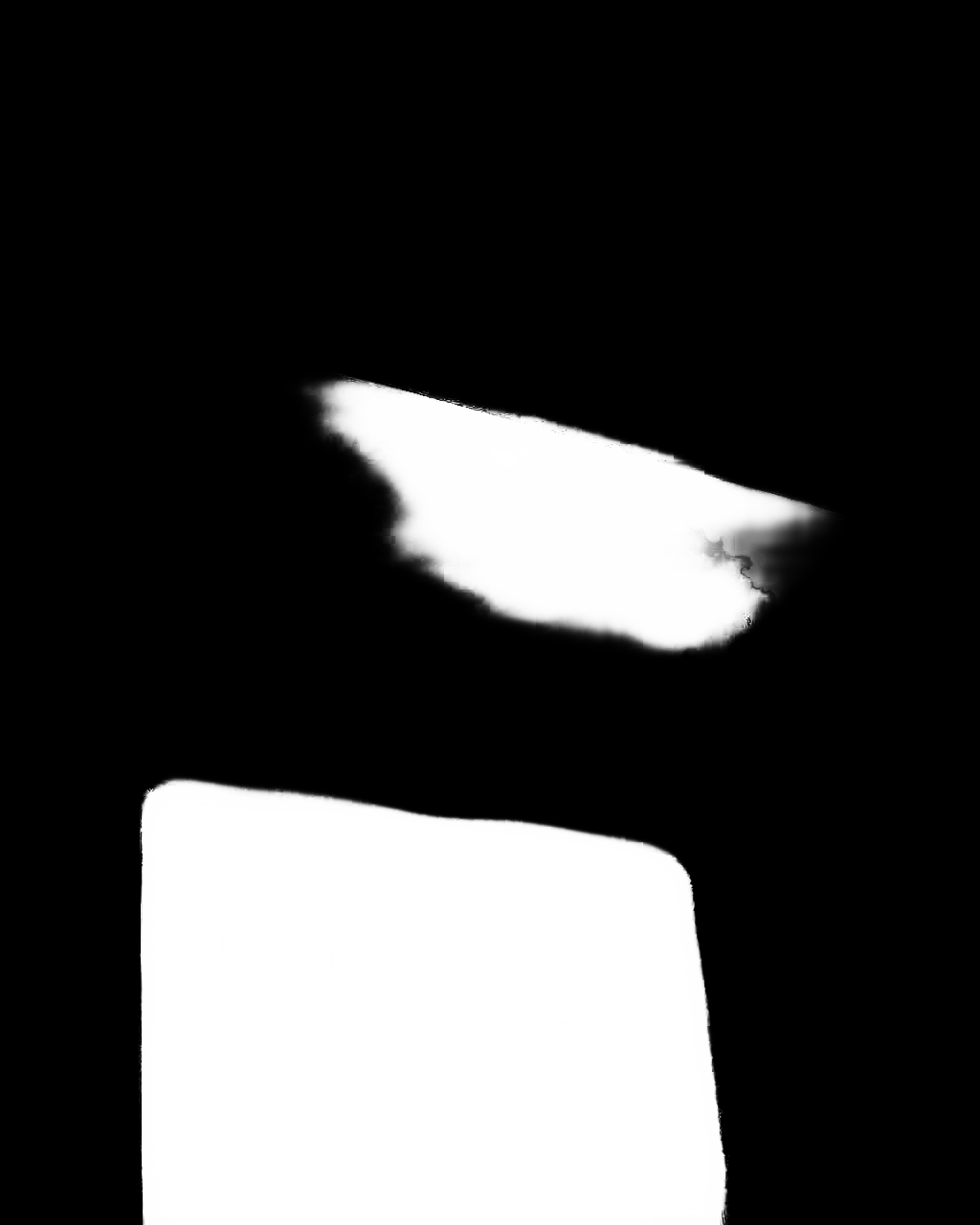}

            \includegraphics[width=1\linewidth]{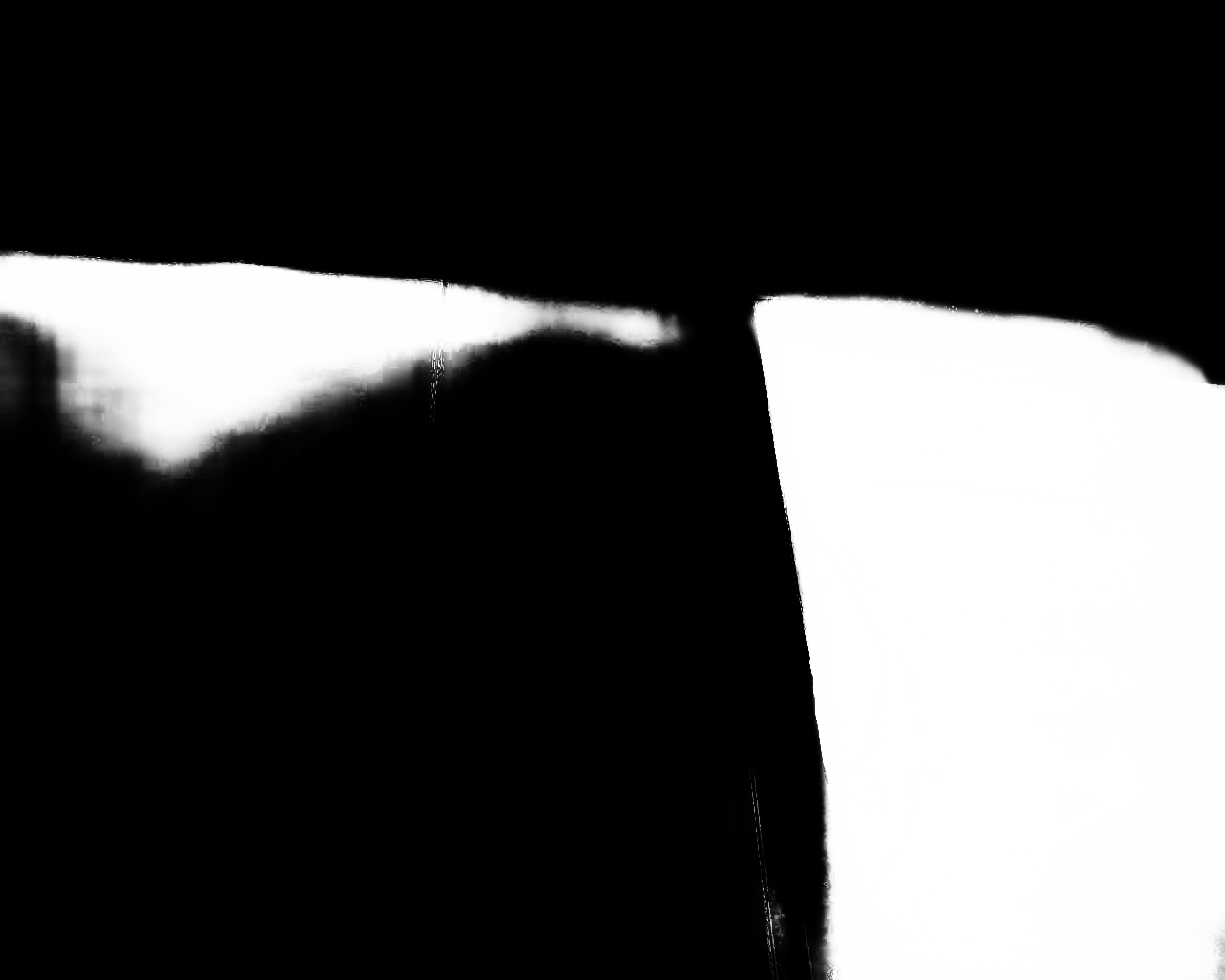}

            \includegraphics[width=1\linewidth]{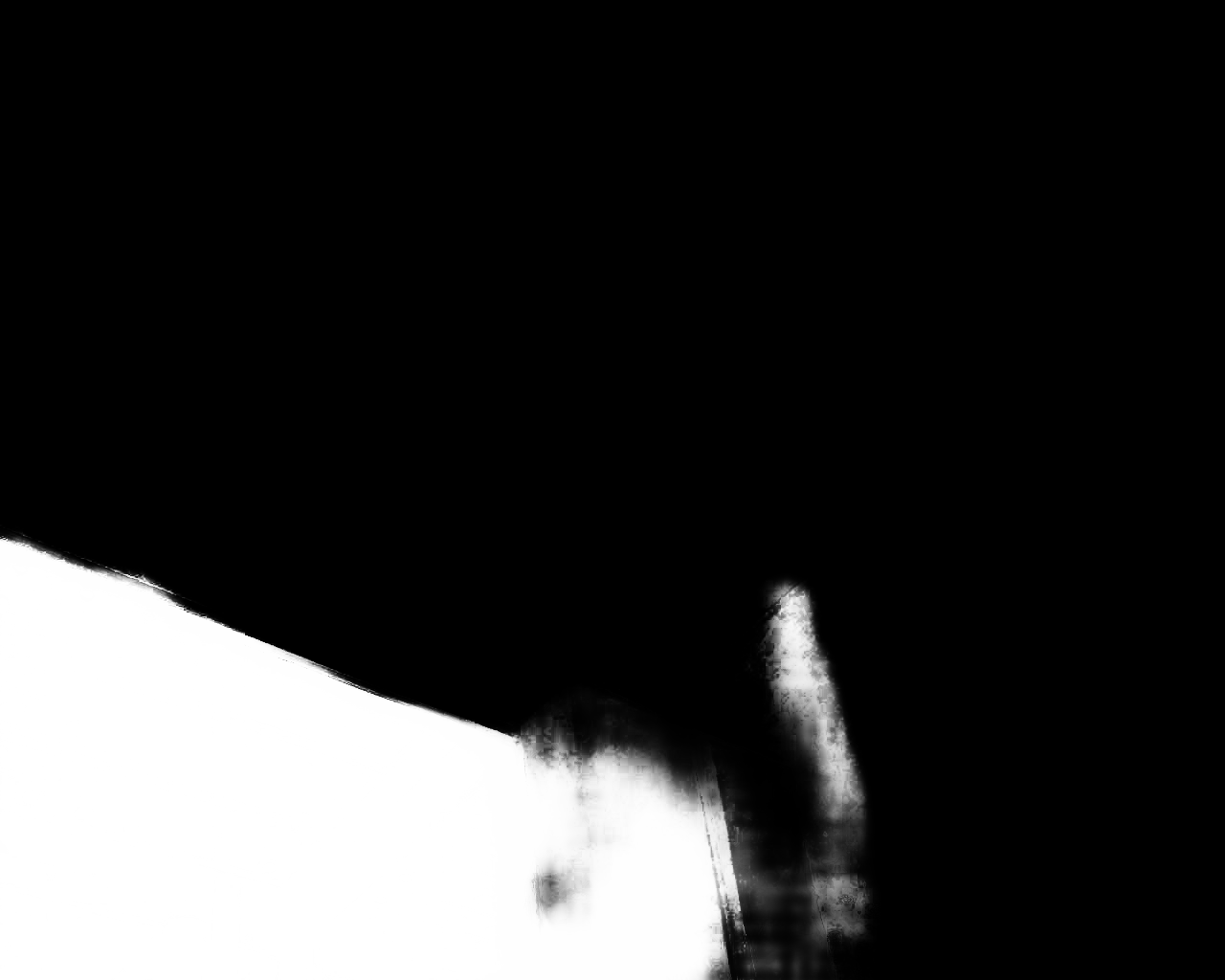}

            \includegraphics[width=1\linewidth]{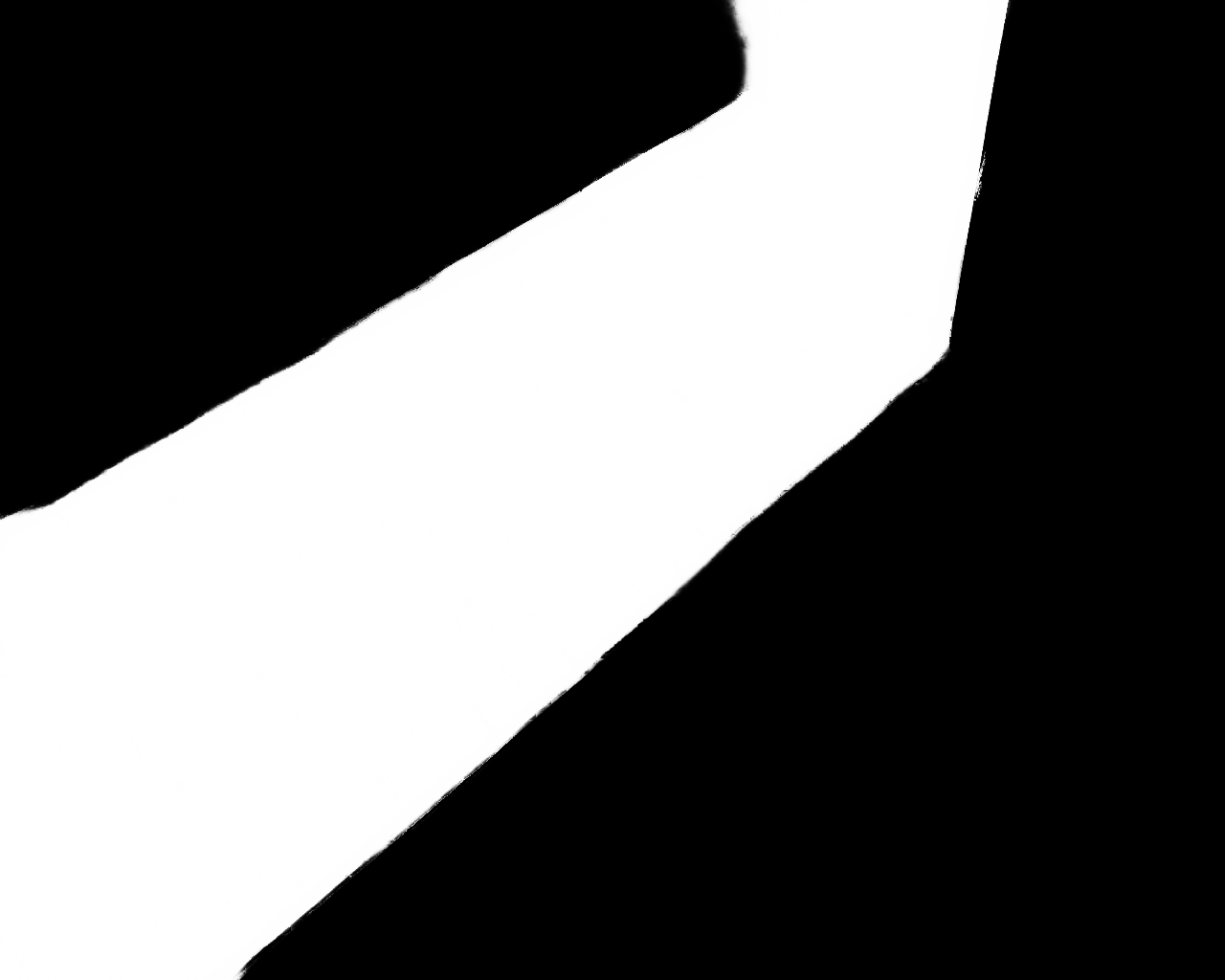}

            \includegraphics[width=1\linewidth]{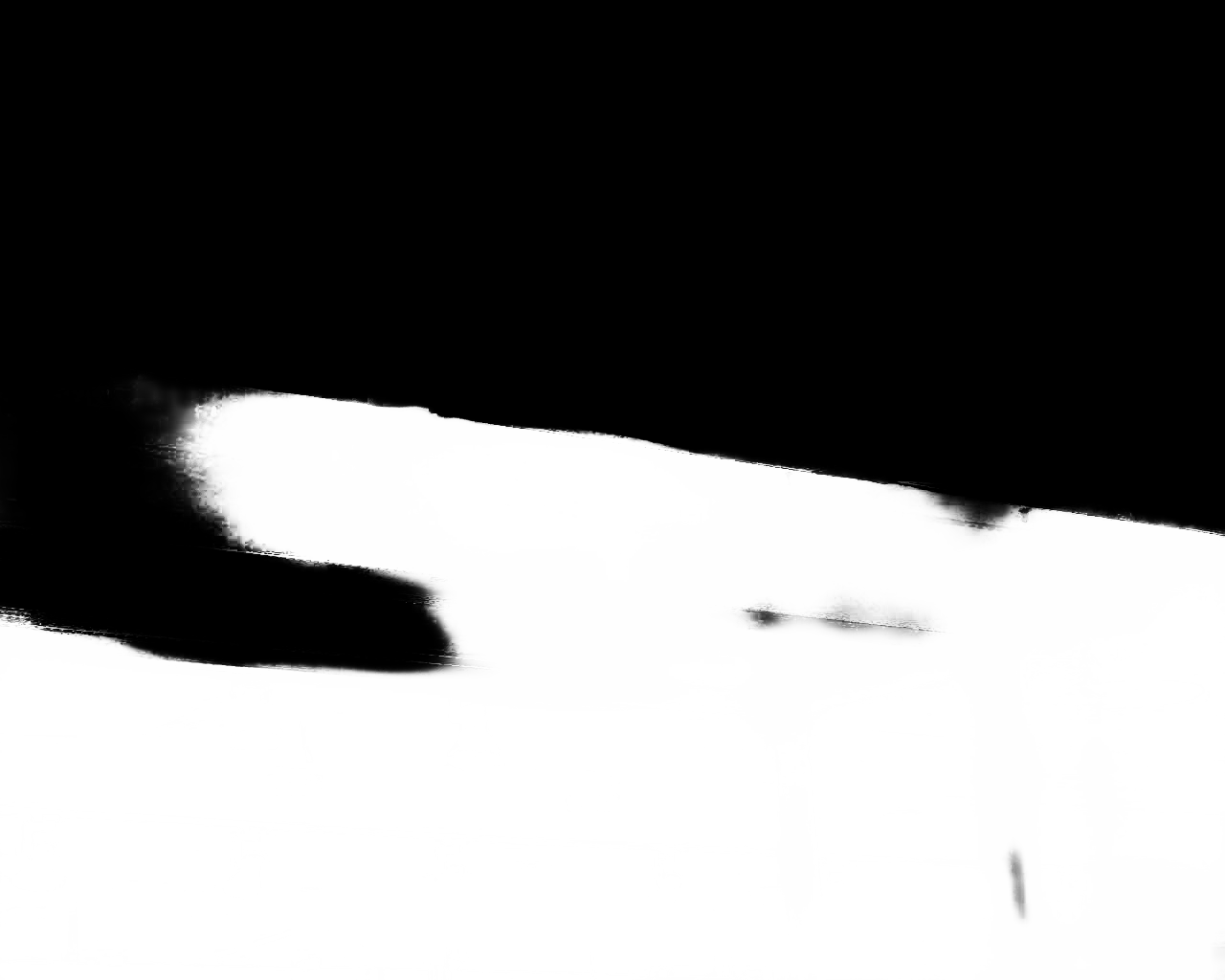}
            
            \includegraphics[width=1\linewidth]{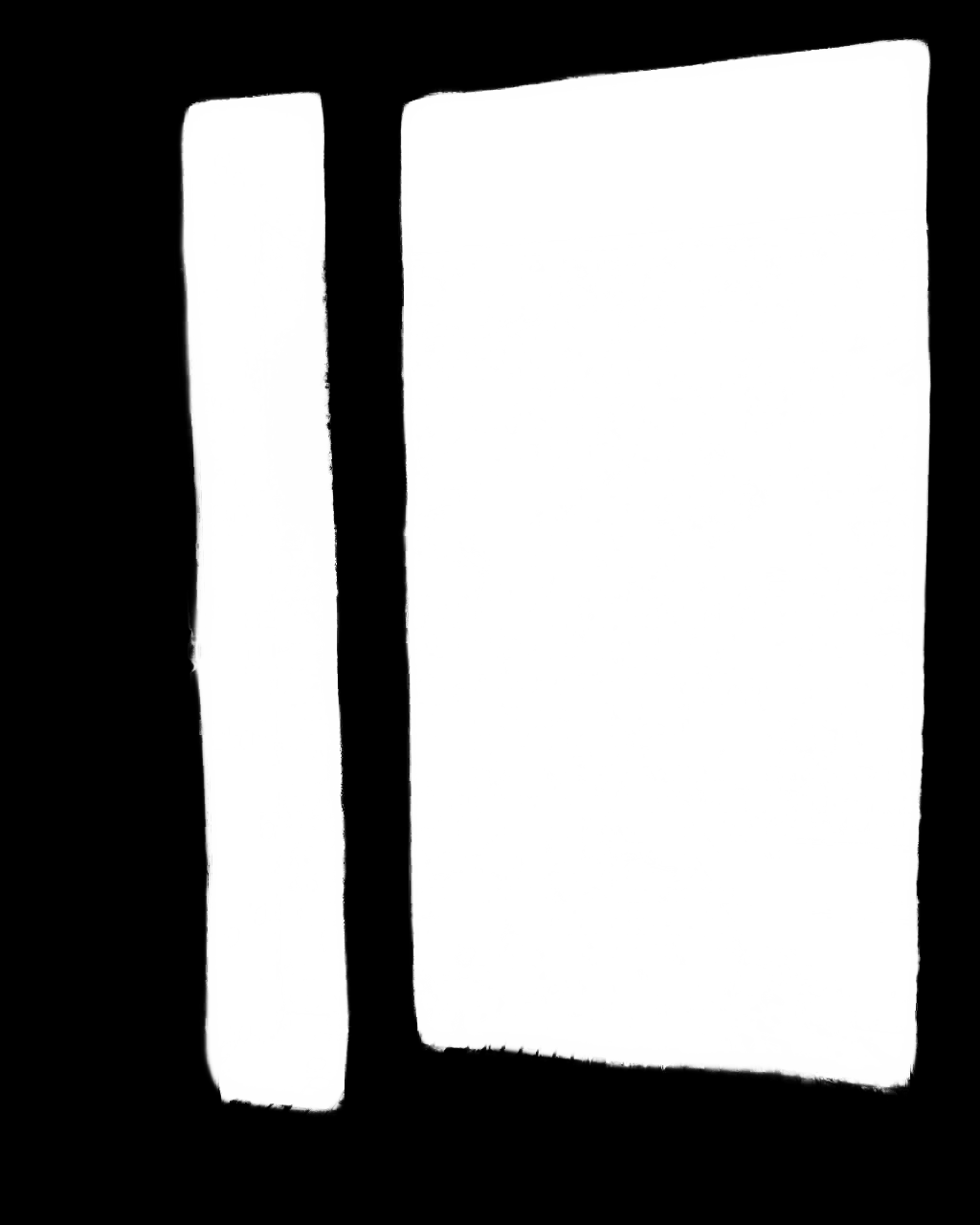}

            \includegraphics[width=1\linewidth]{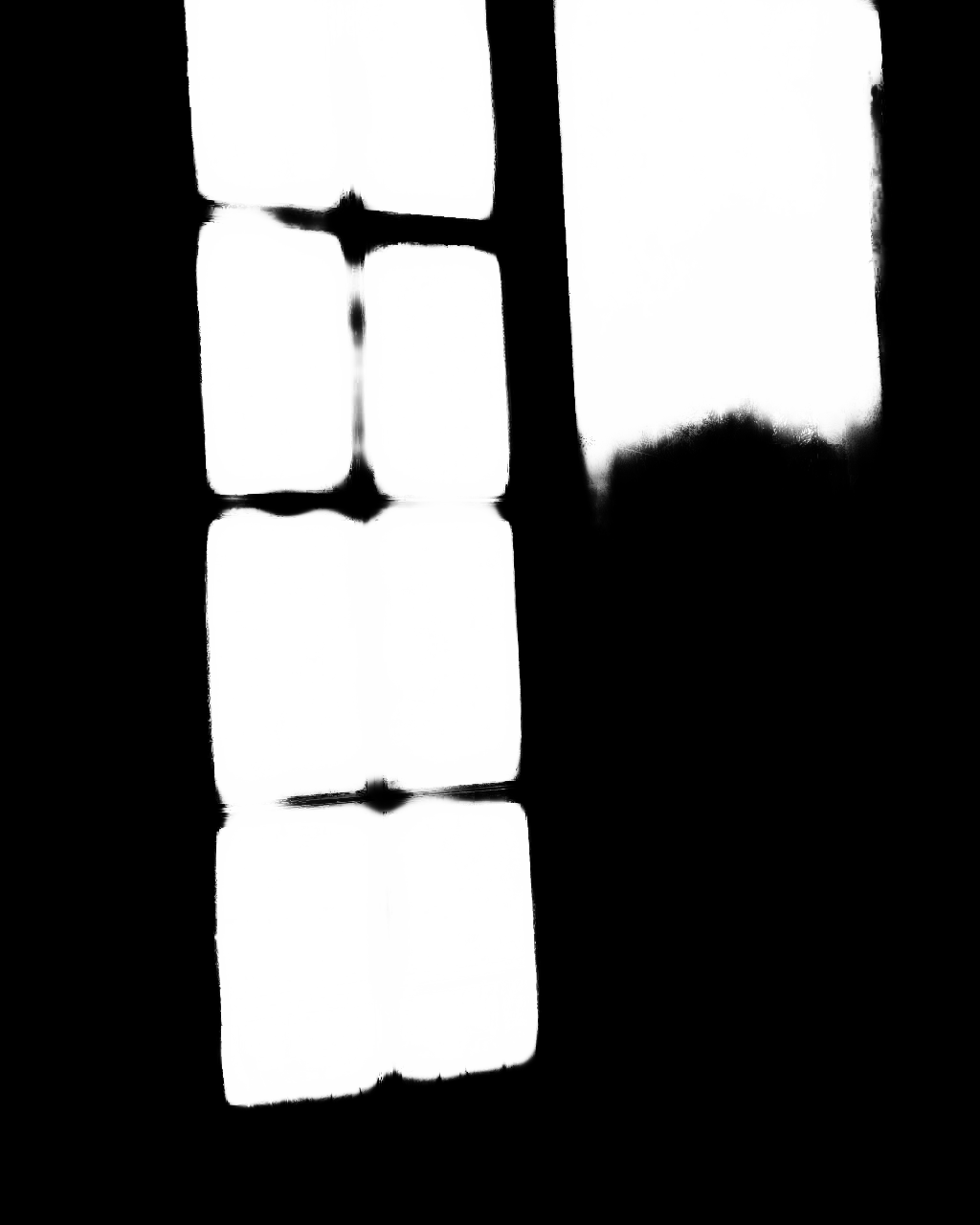}
      \end{minipage}
      }  
      \subfloat[DSC]{\label{DSC}
      \begin{minipage}[t]{0.07\textwidth}
            \centering
            \includegraphics[width=1\linewidth]{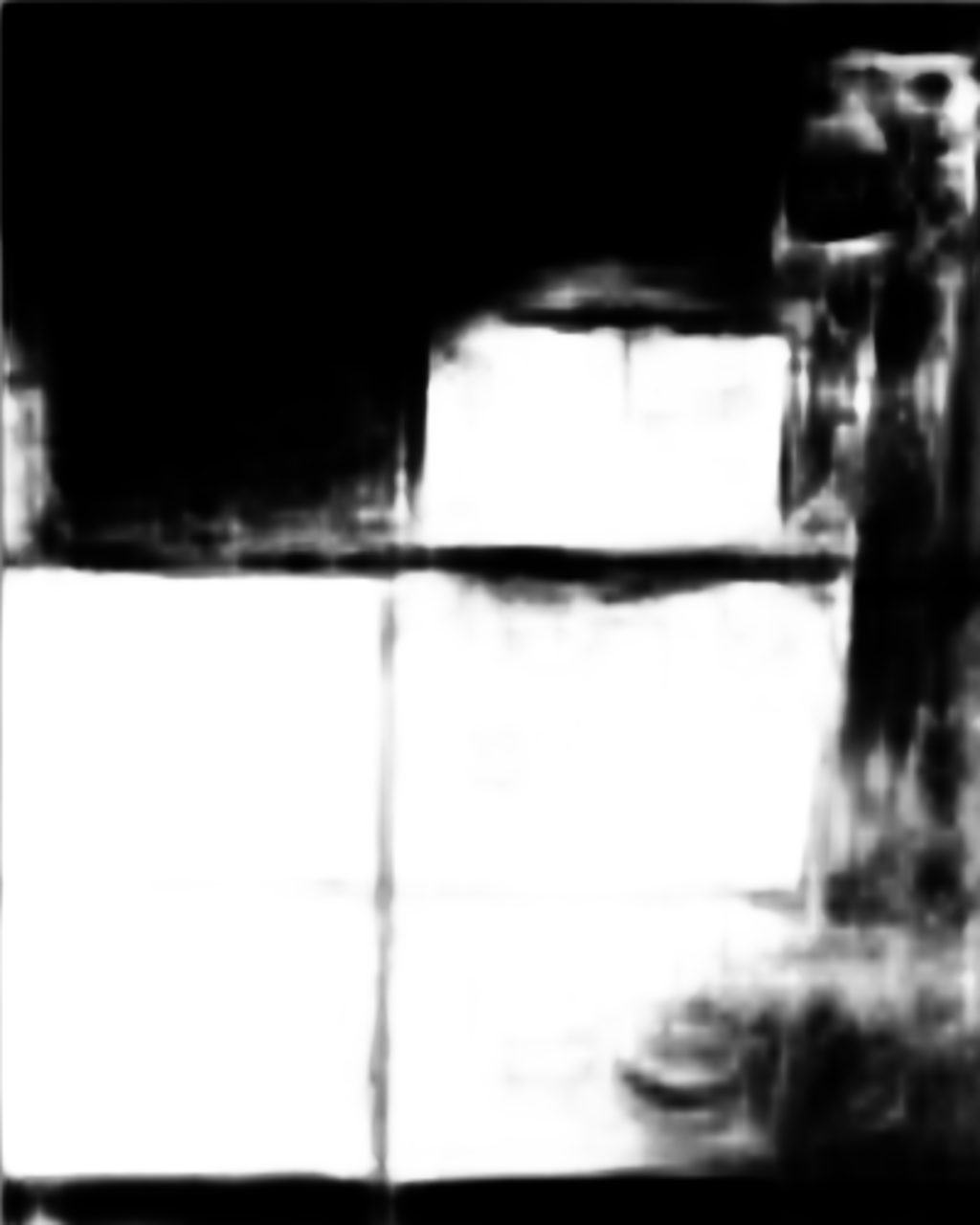}

            \includegraphics[width=1\linewidth]{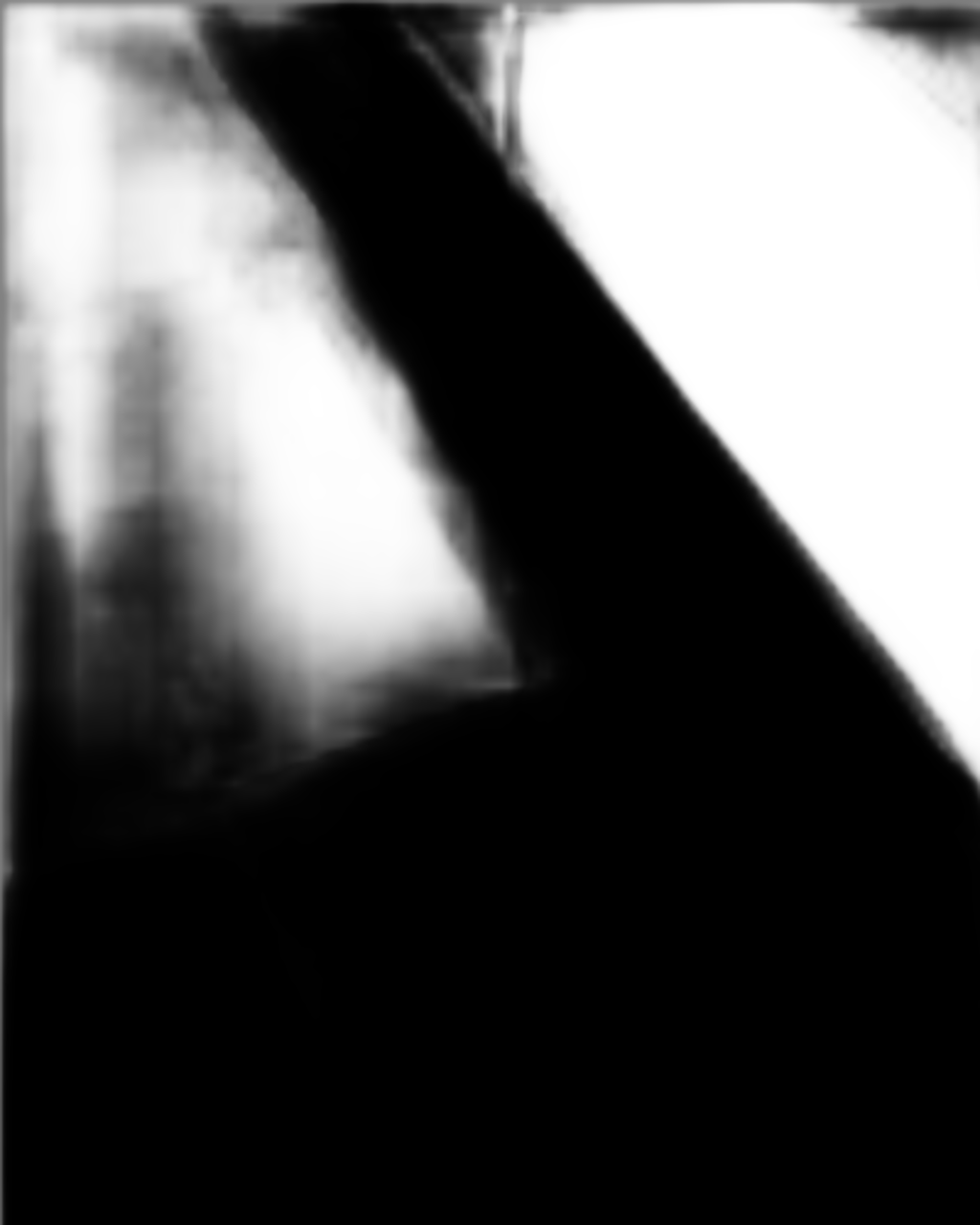}

            \includegraphics[width=1\linewidth]{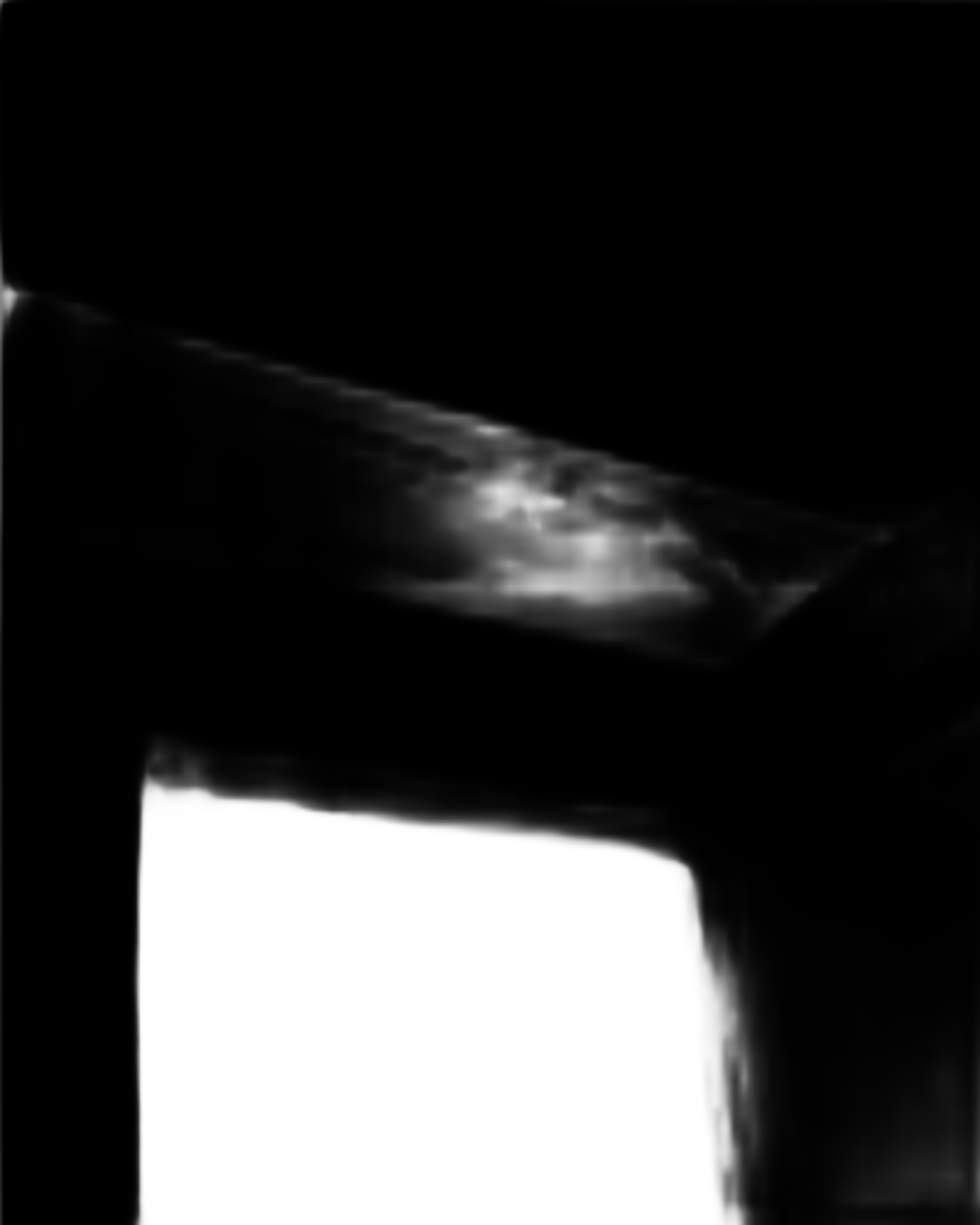}

            \includegraphics[width=1\linewidth]{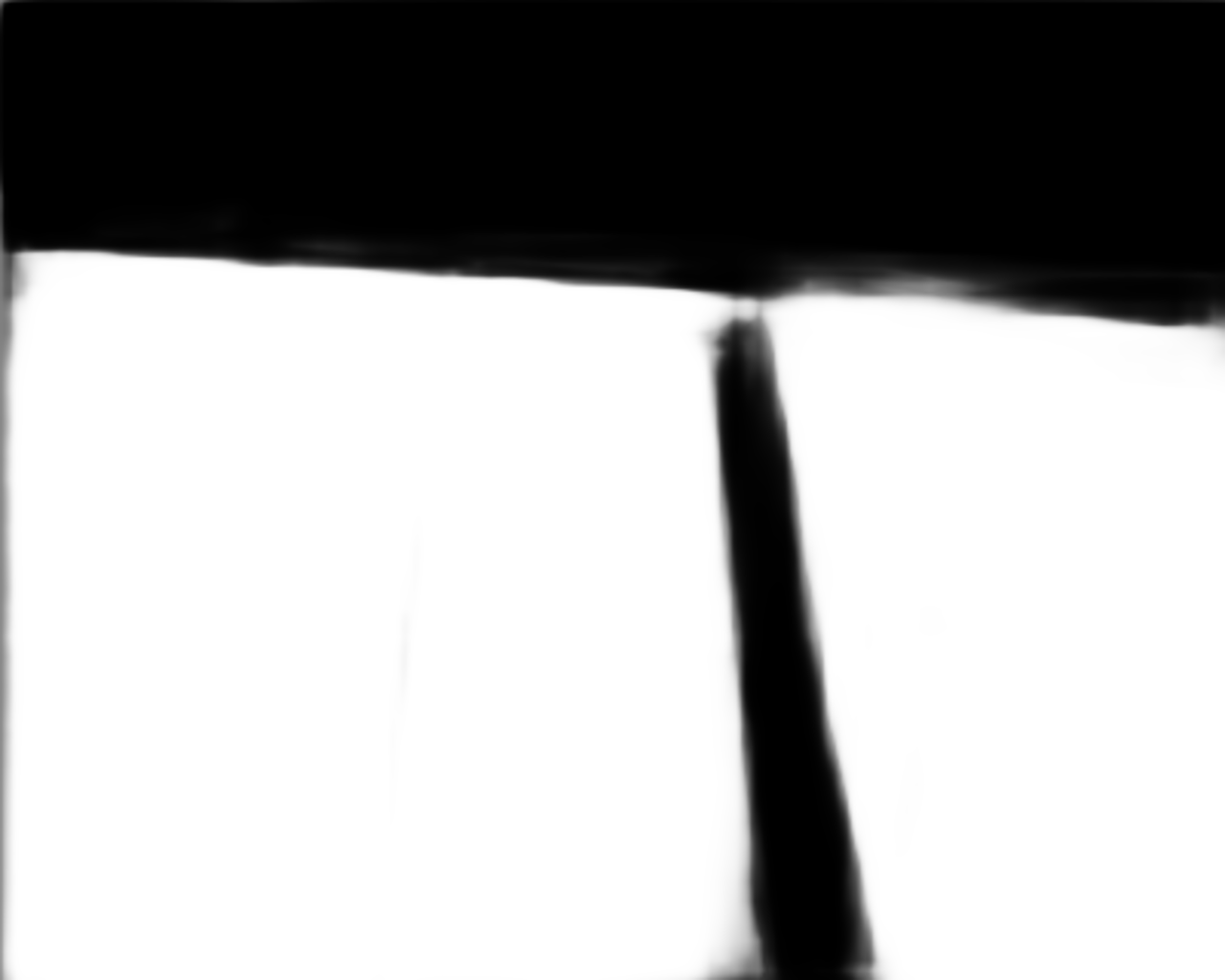}

            \includegraphics[width=1\linewidth]{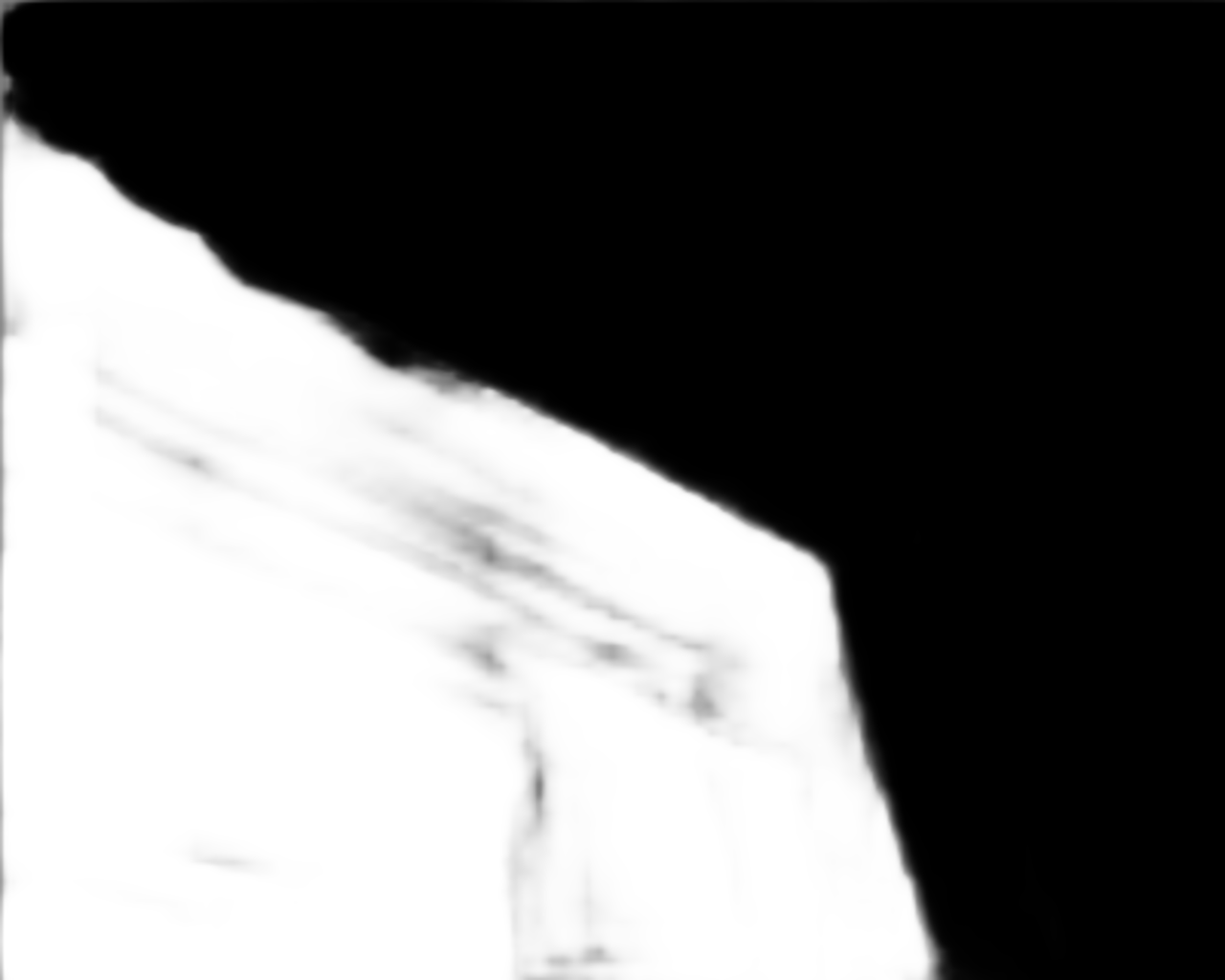}

            \includegraphics[width=1\linewidth]{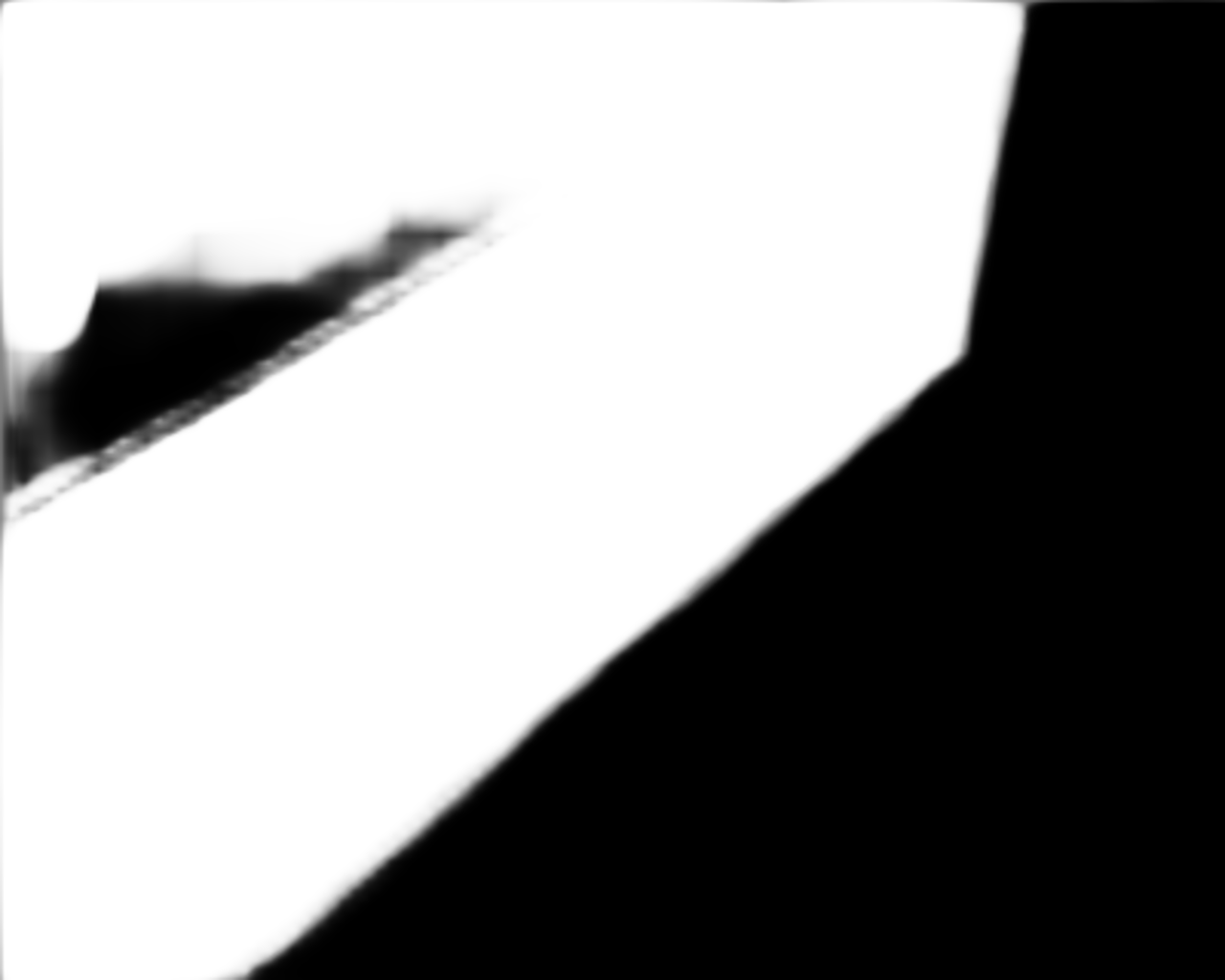}

            \includegraphics[width=1\linewidth]{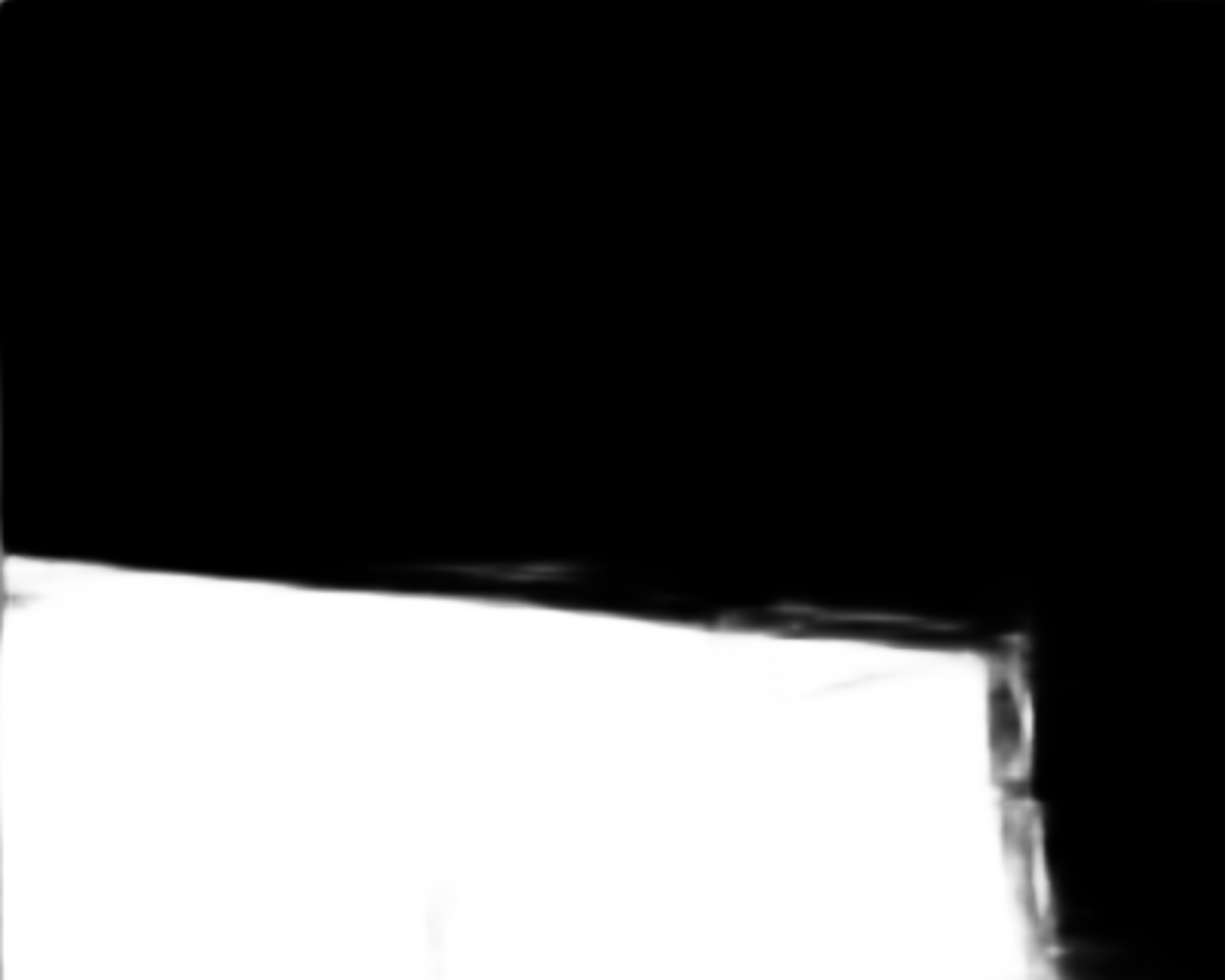}
            
            \includegraphics[width=1\linewidth]{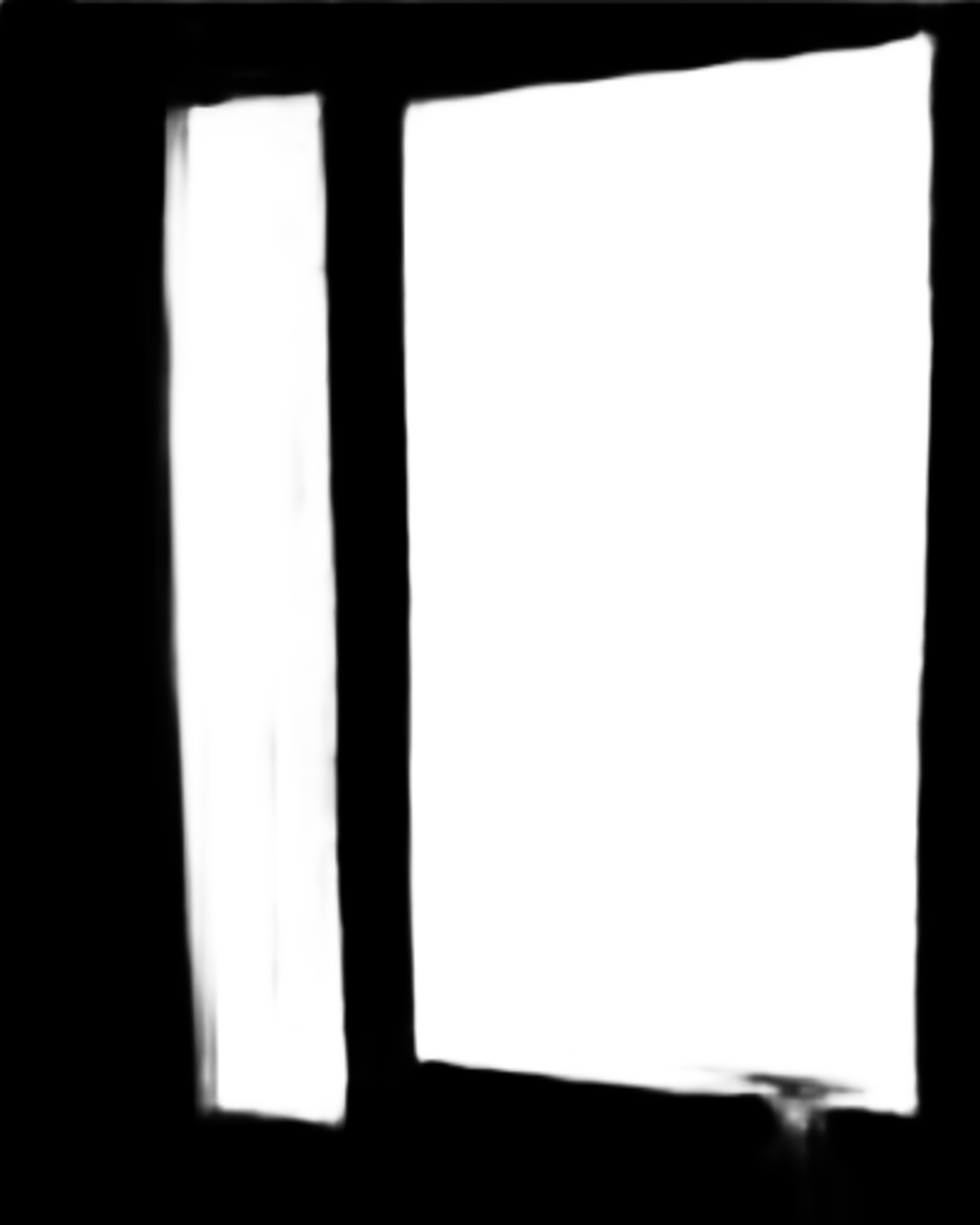}

            \includegraphics[width=1\linewidth]{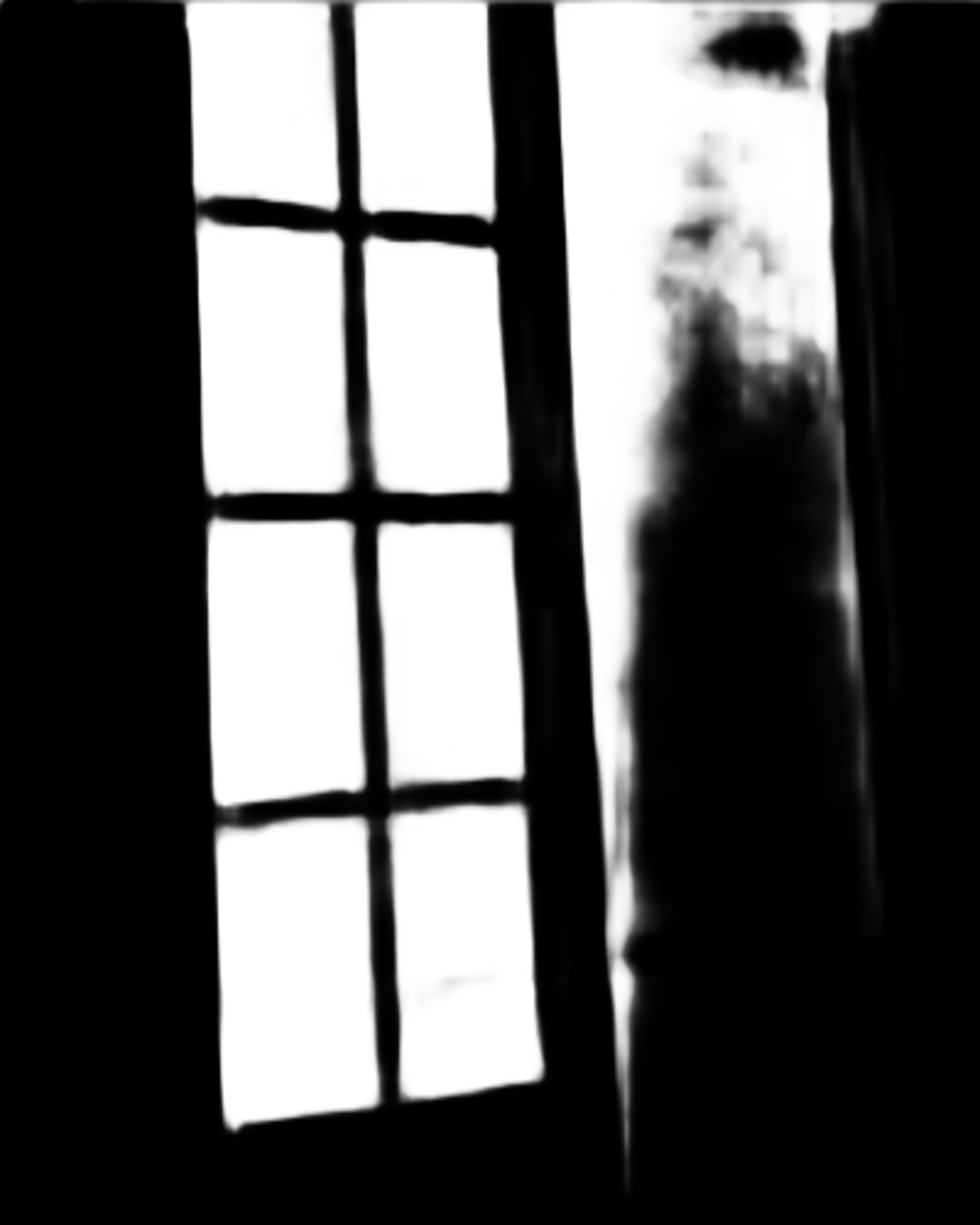}
      \end{minipage}
      }  
      \subfloat[MirrorNet]{\label{MirrorNet}
      \begin{minipage}[t]{0.07\textwidth}
            \centering
            \includegraphics[width=1\linewidth]{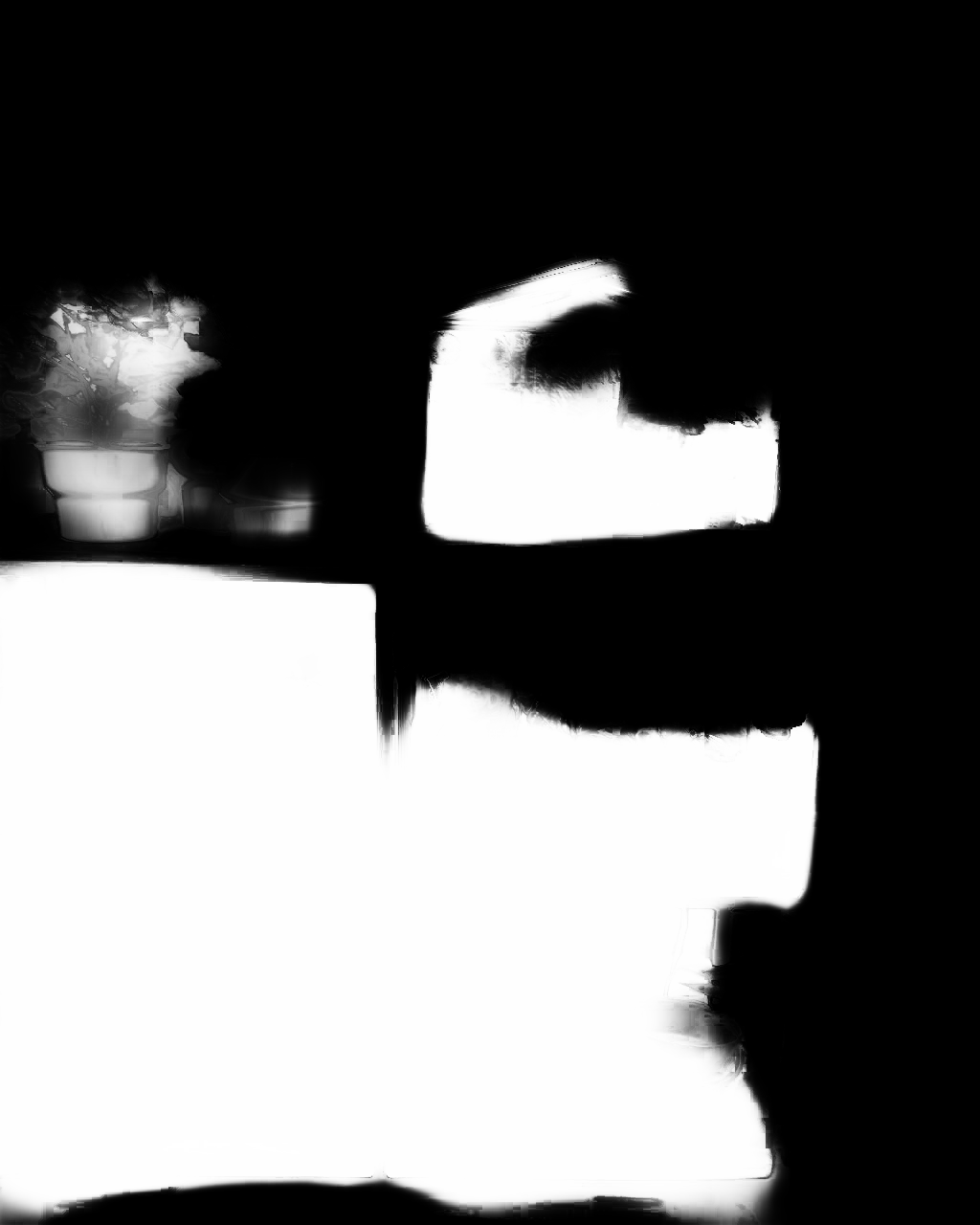}

            \includegraphics[width=1\linewidth]{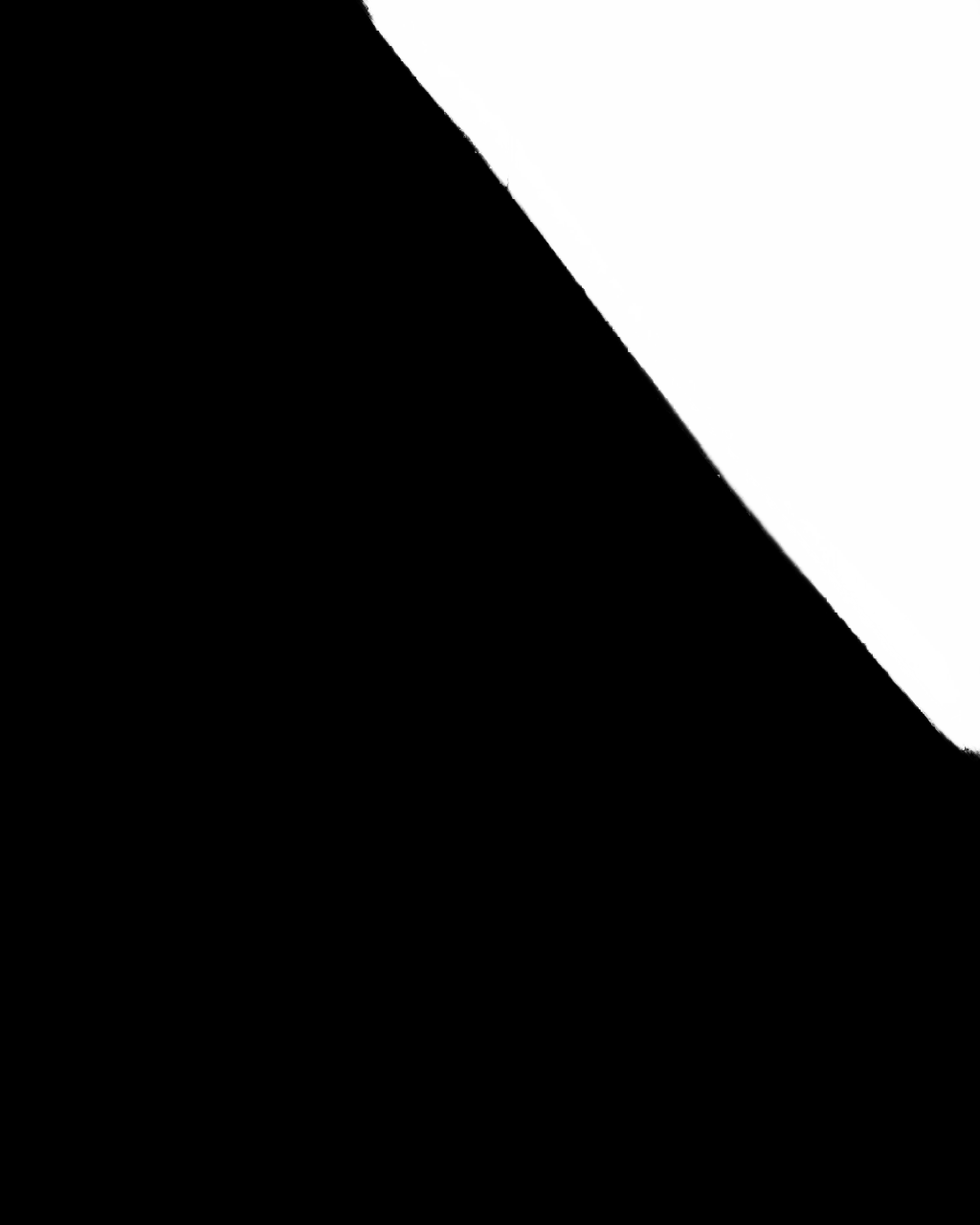}

            \includegraphics[width=1\linewidth]{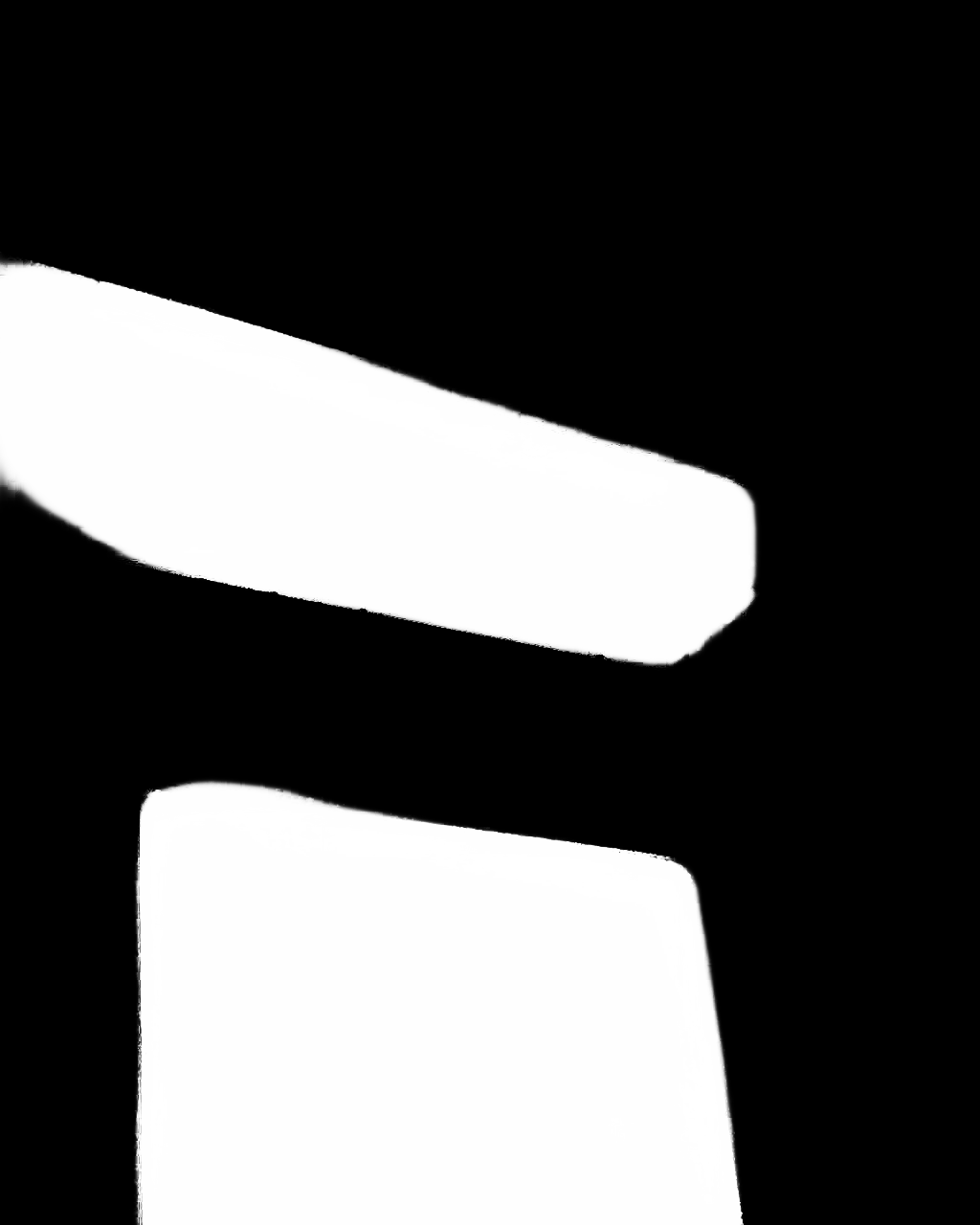}

            \includegraphics[width=1\linewidth]{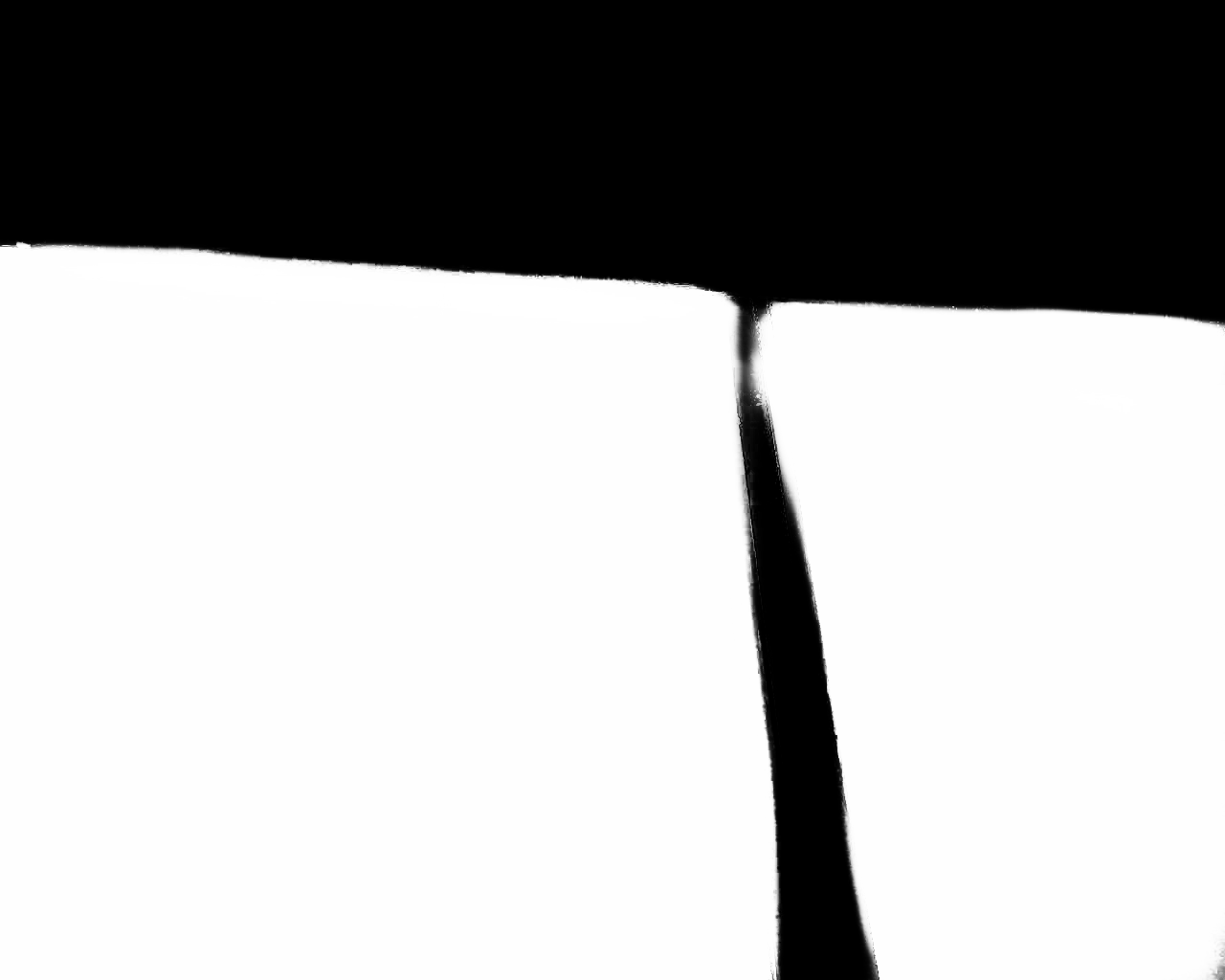}

            \includegraphics[width=1\linewidth]{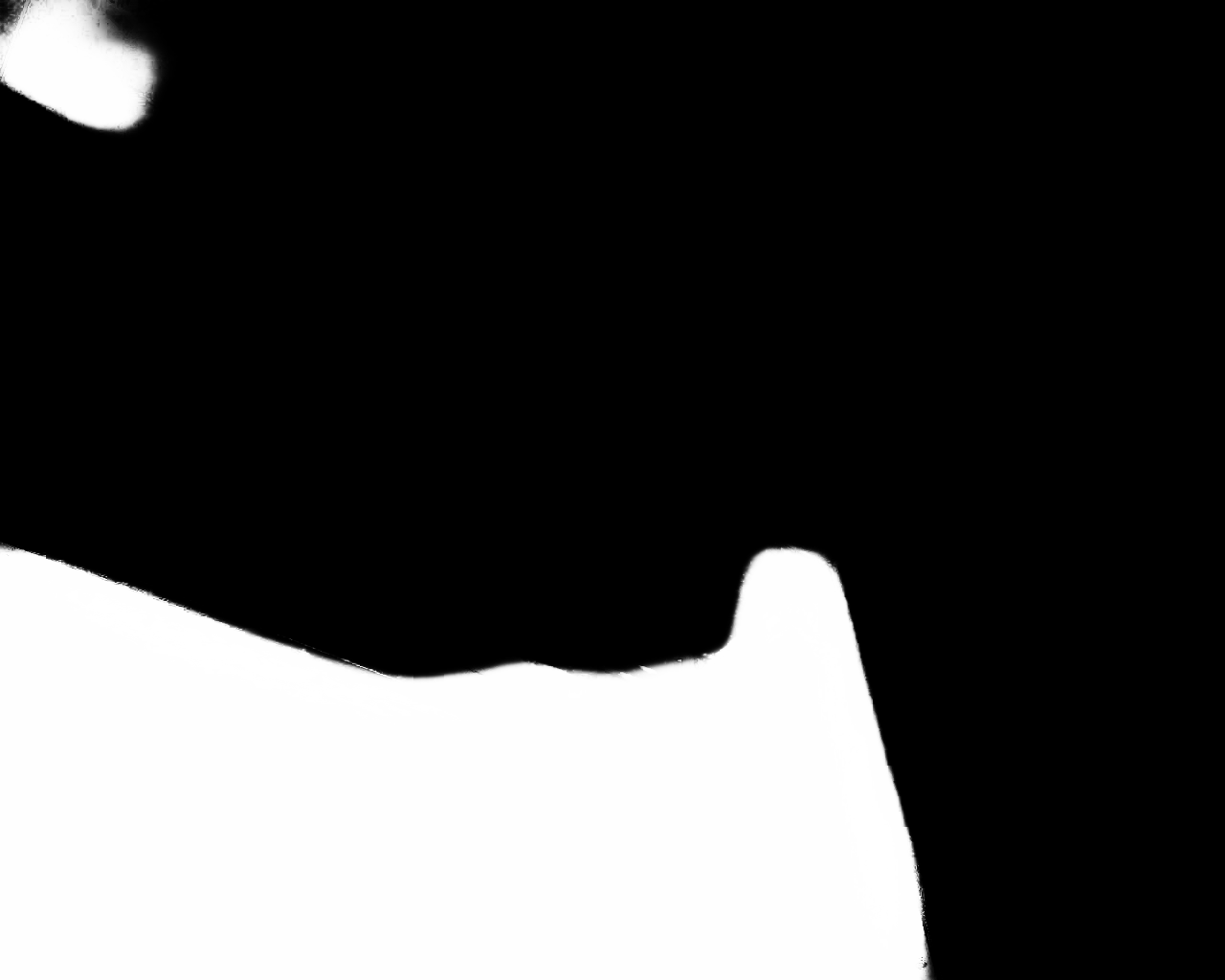}

            \includegraphics[width=1\linewidth]{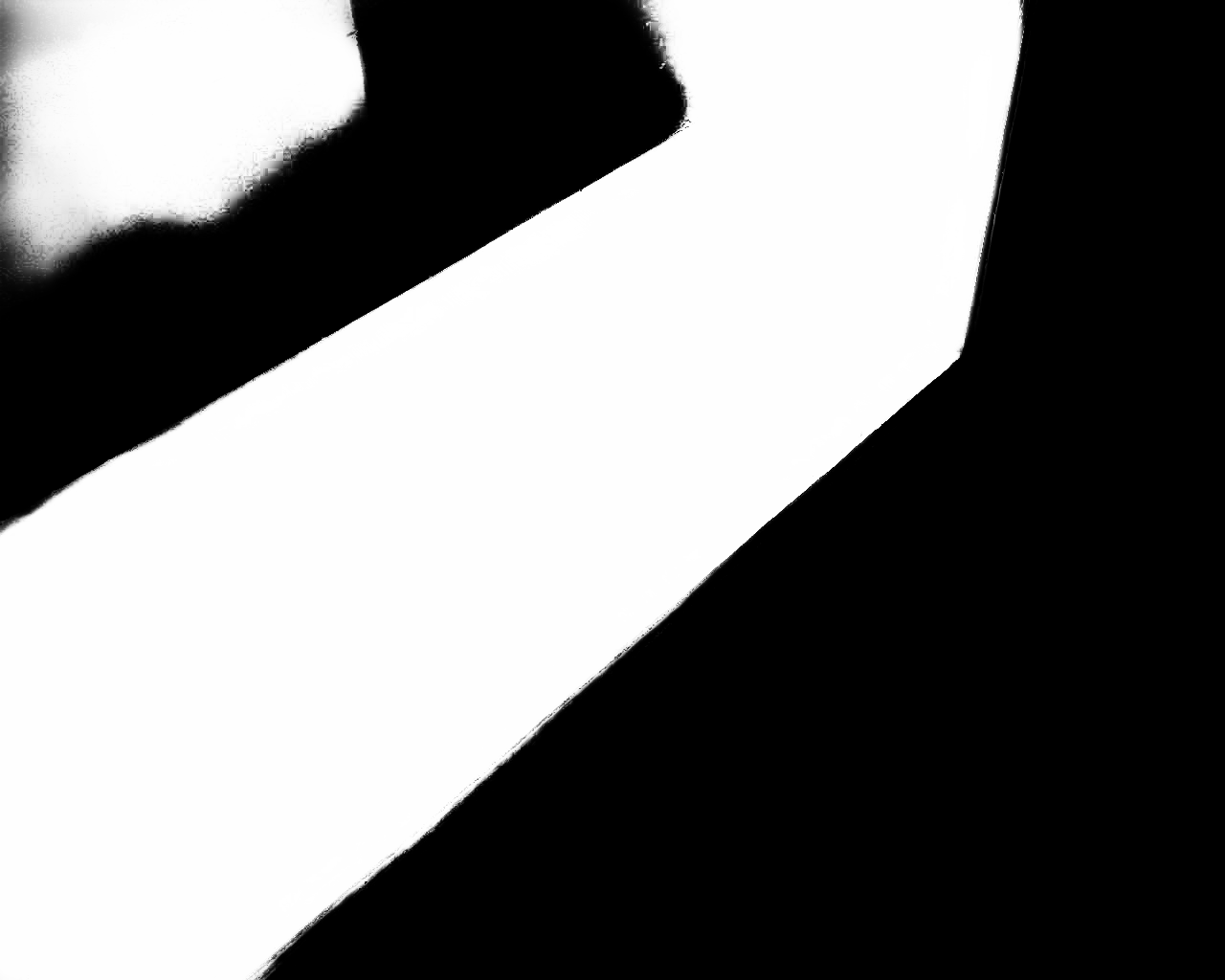}

            \includegraphics[width=1\linewidth]{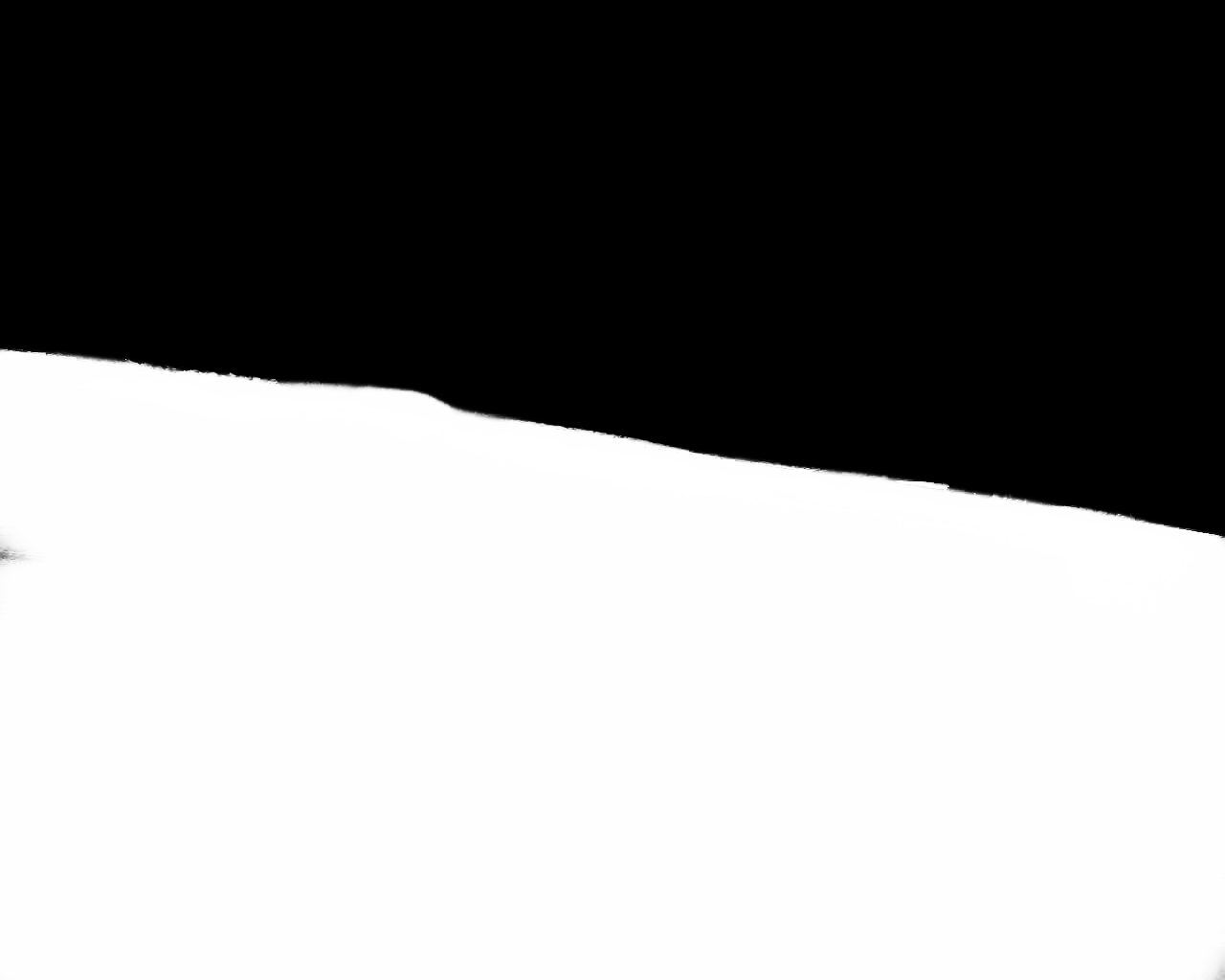}
            
            \includegraphics[width=1\linewidth]{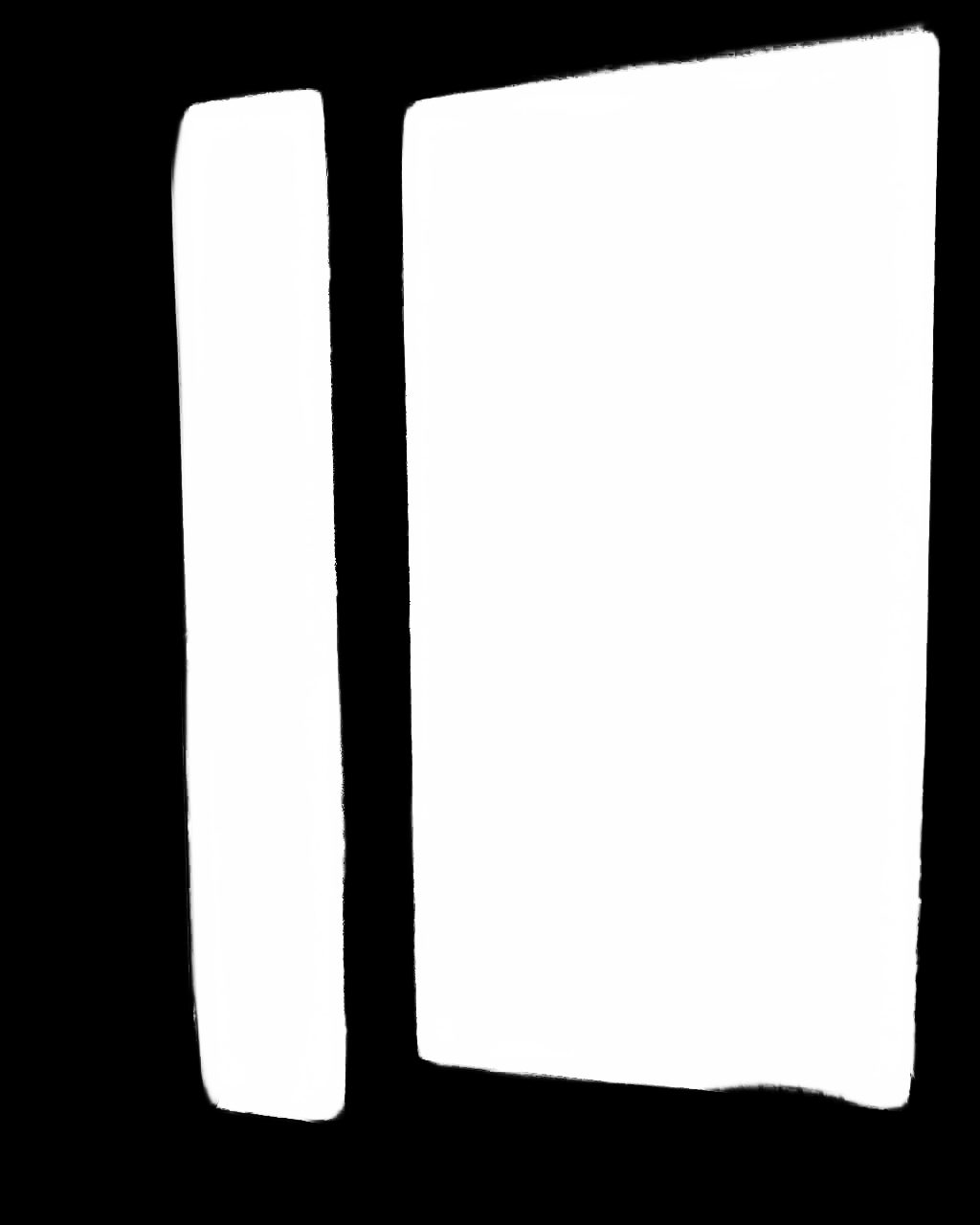}

            \includegraphics[width=1\linewidth]{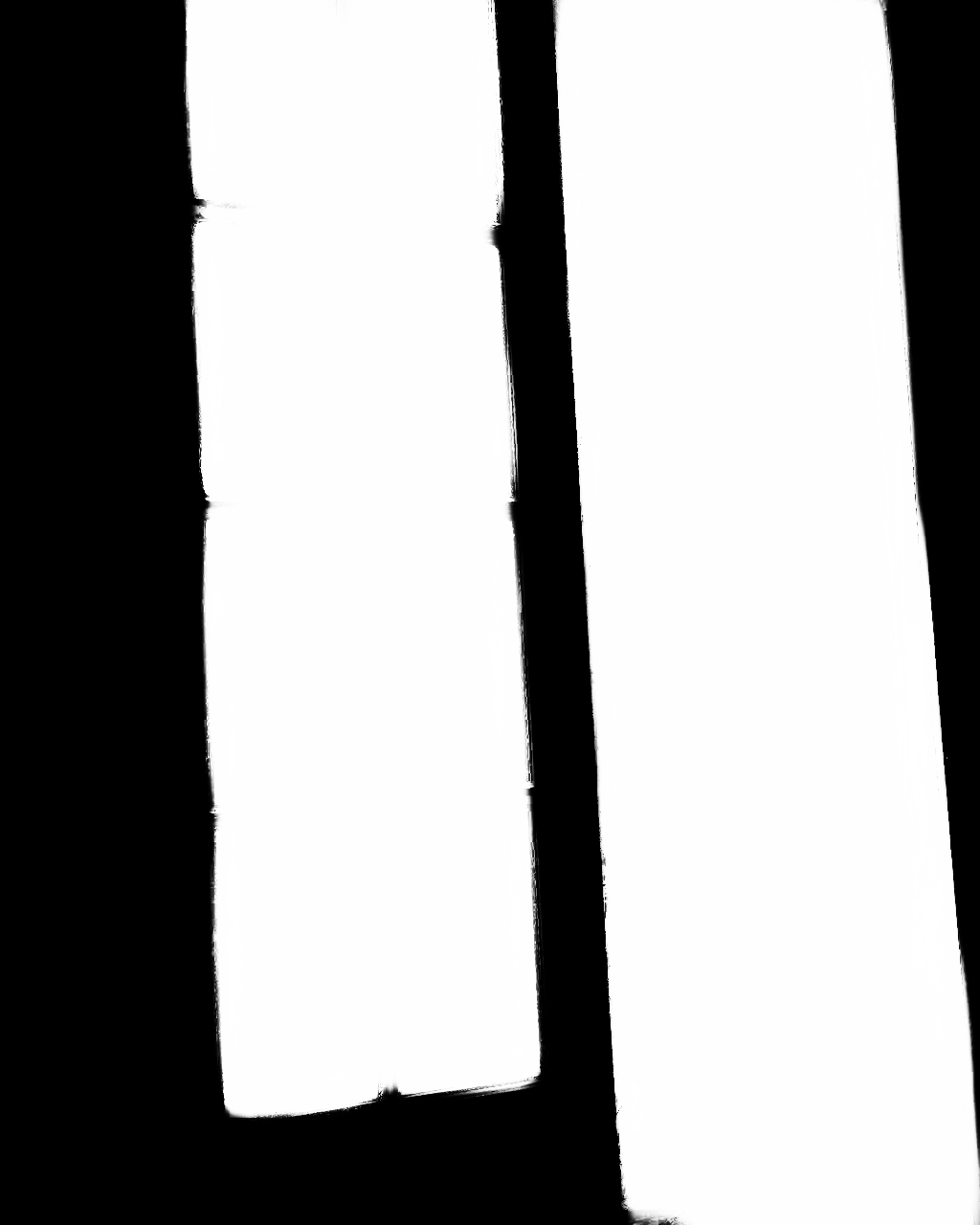}
      \end{minipage}
      }  
      \subfloat[PMD]{\label{PMD}
      \begin{minipage}[t]{0.07\textwidth}
            \centering
            \includegraphics[width=1\linewidth]{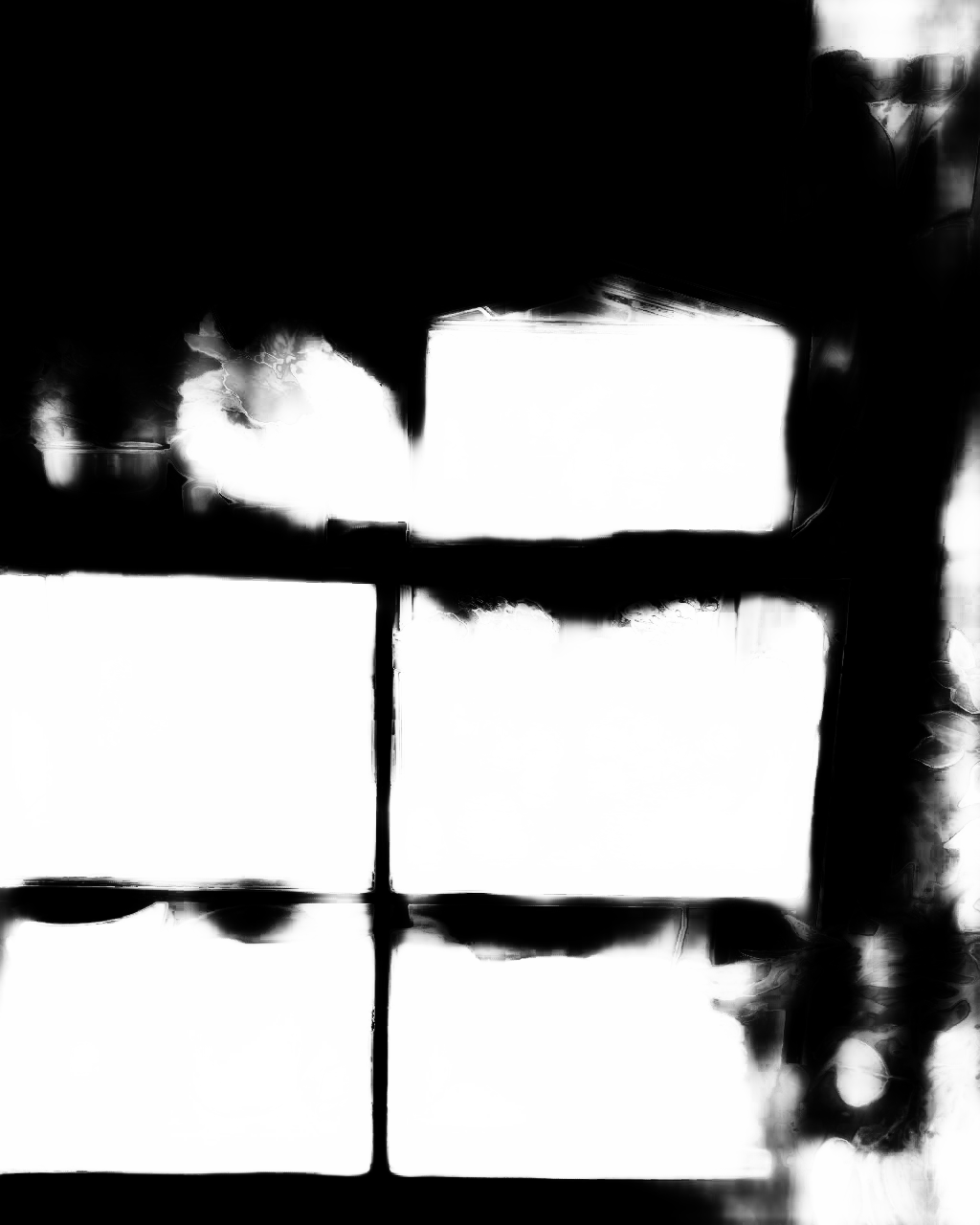}

            \includegraphics[width=1\linewidth]{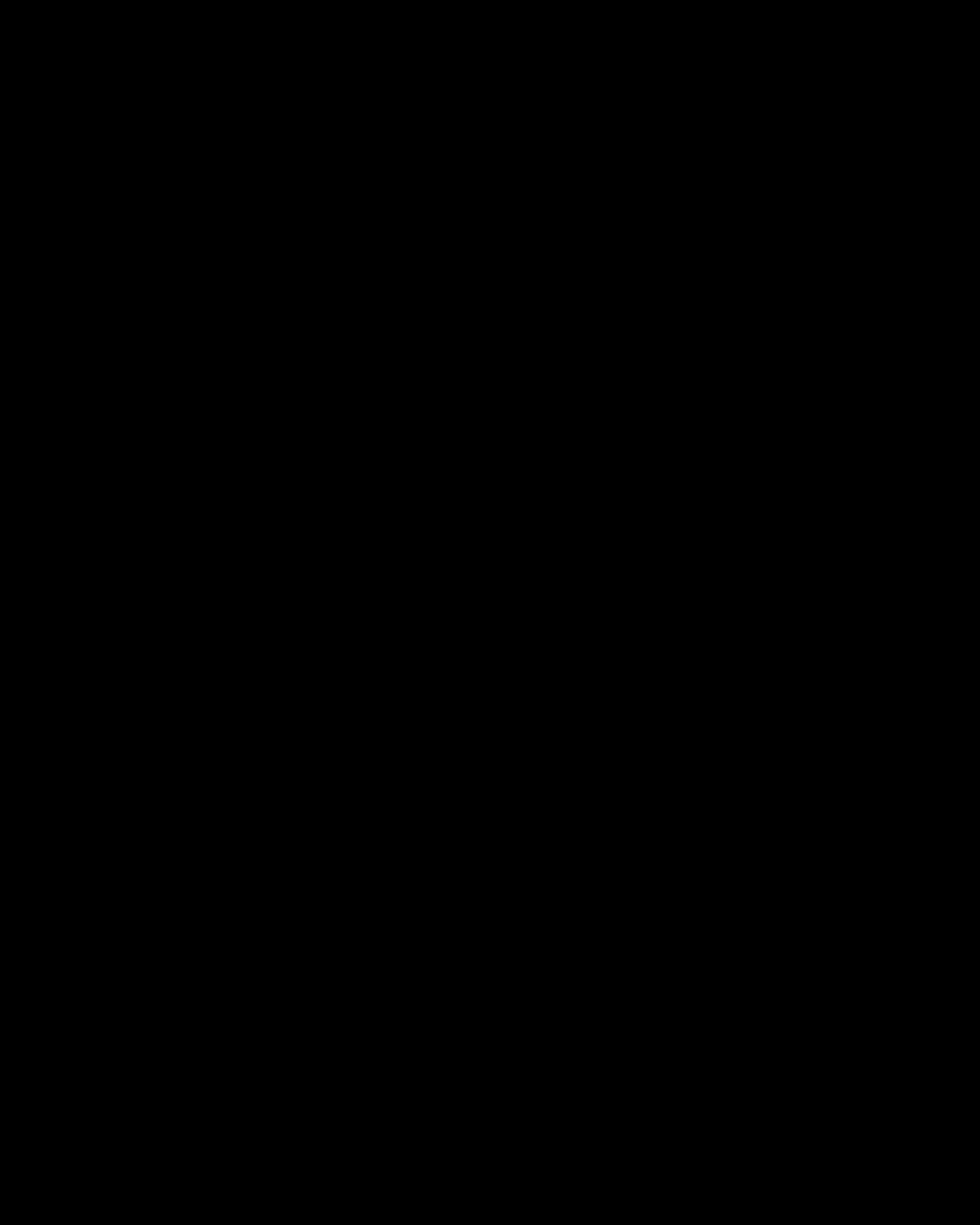}

            \includegraphics[width=1\linewidth]{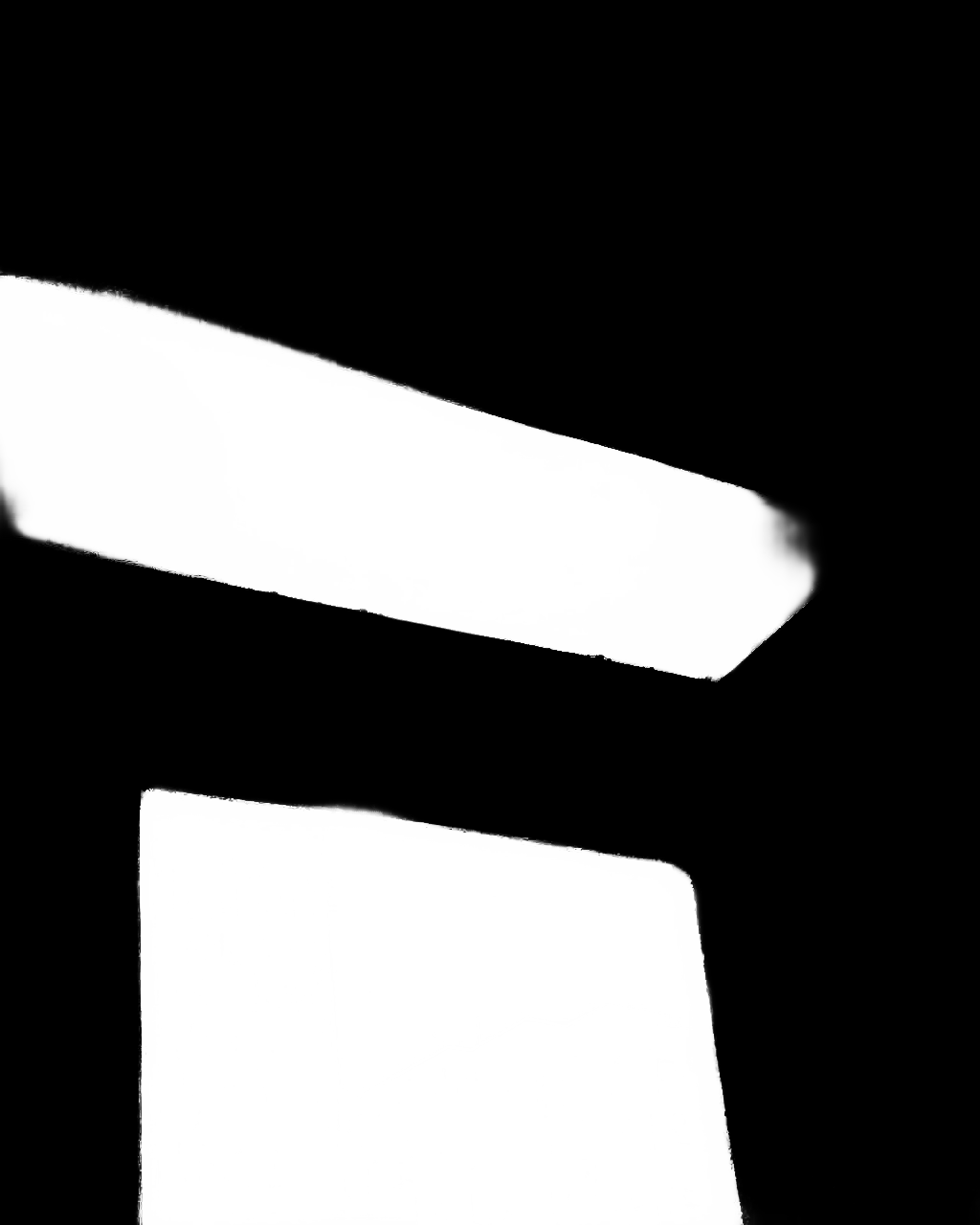}

            \includegraphics[width=1\linewidth]{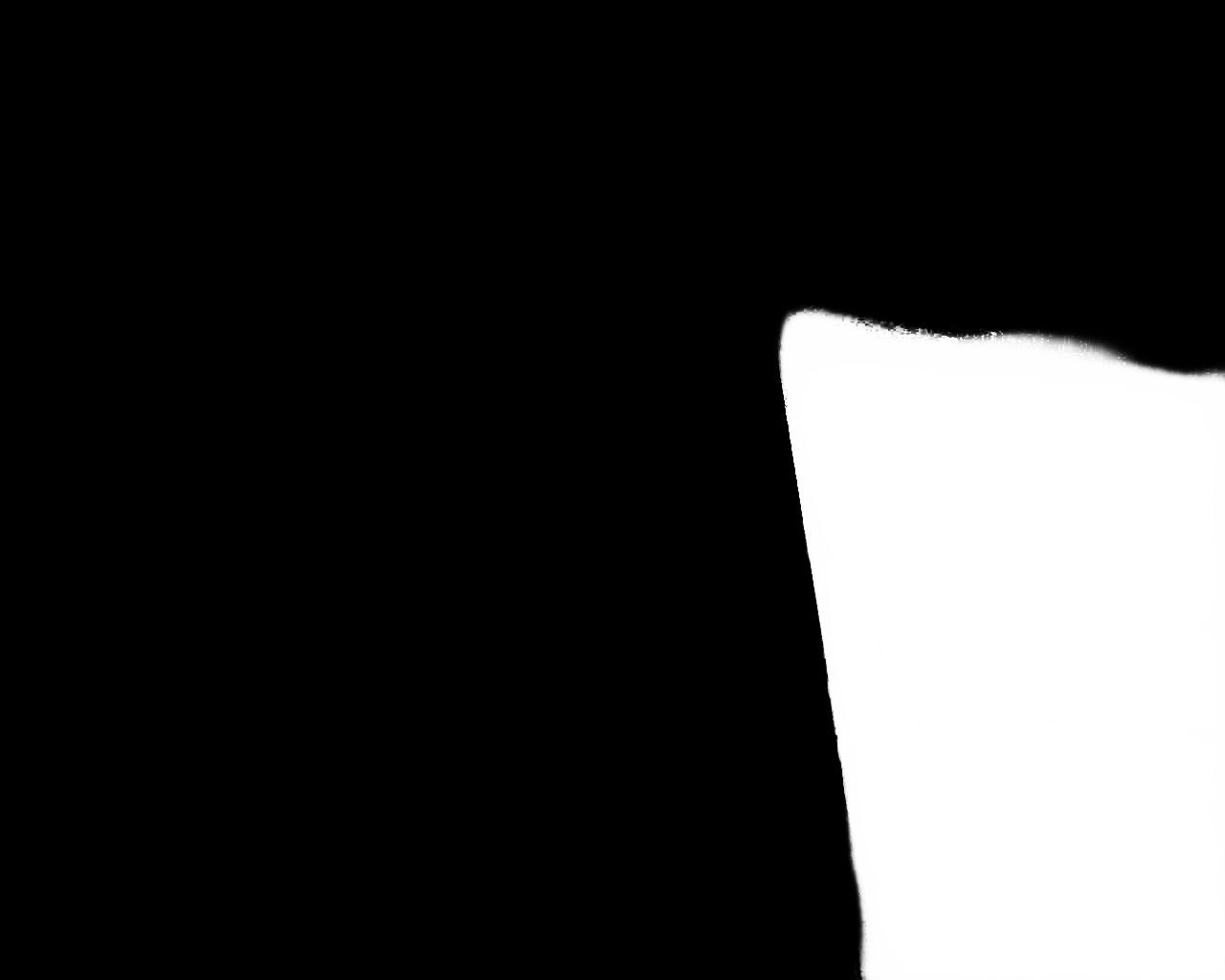}

            \includegraphics[width=1\linewidth]{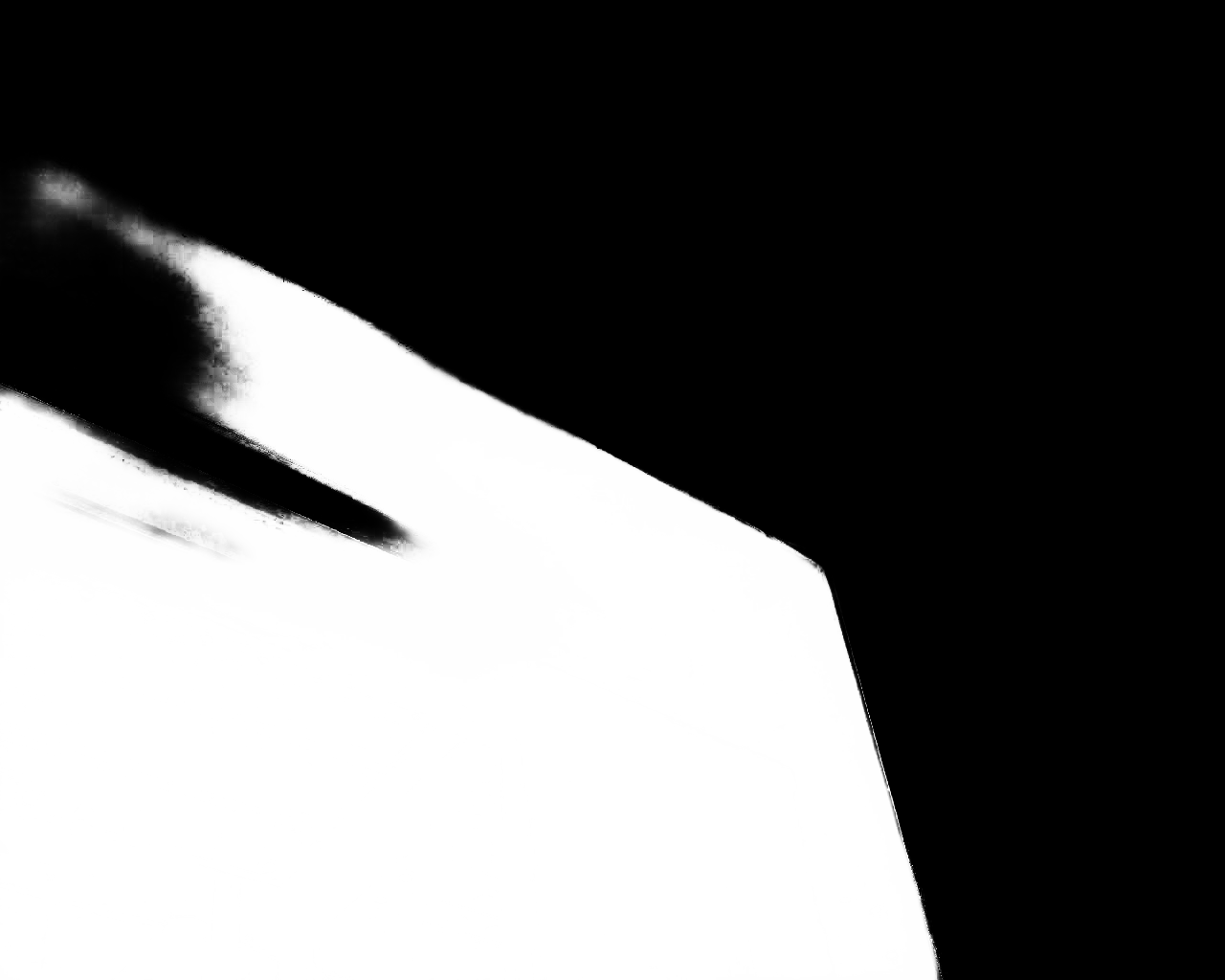}

            \includegraphics[width=1\linewidth]{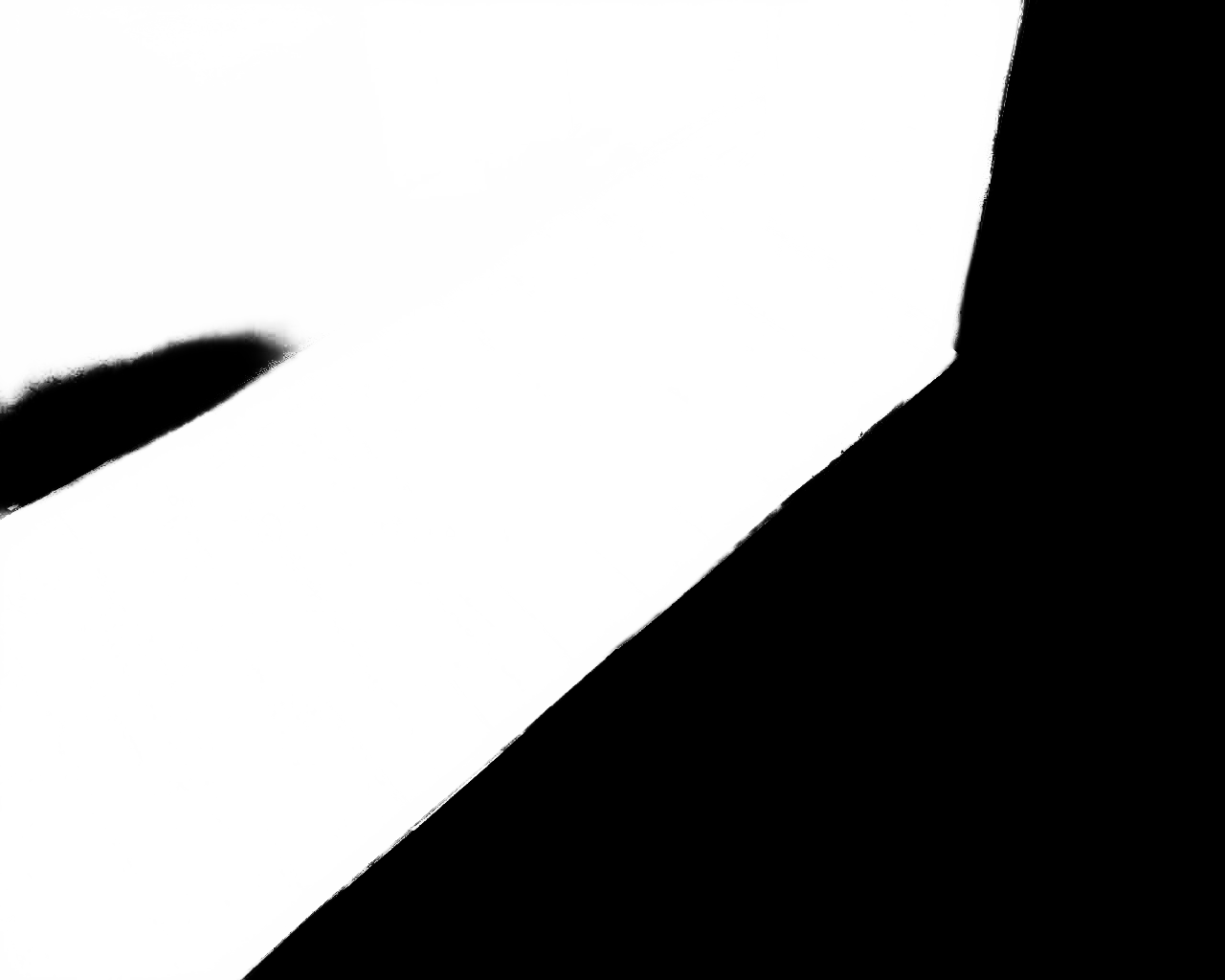}

            \includegraphics[width=1\linewidth]{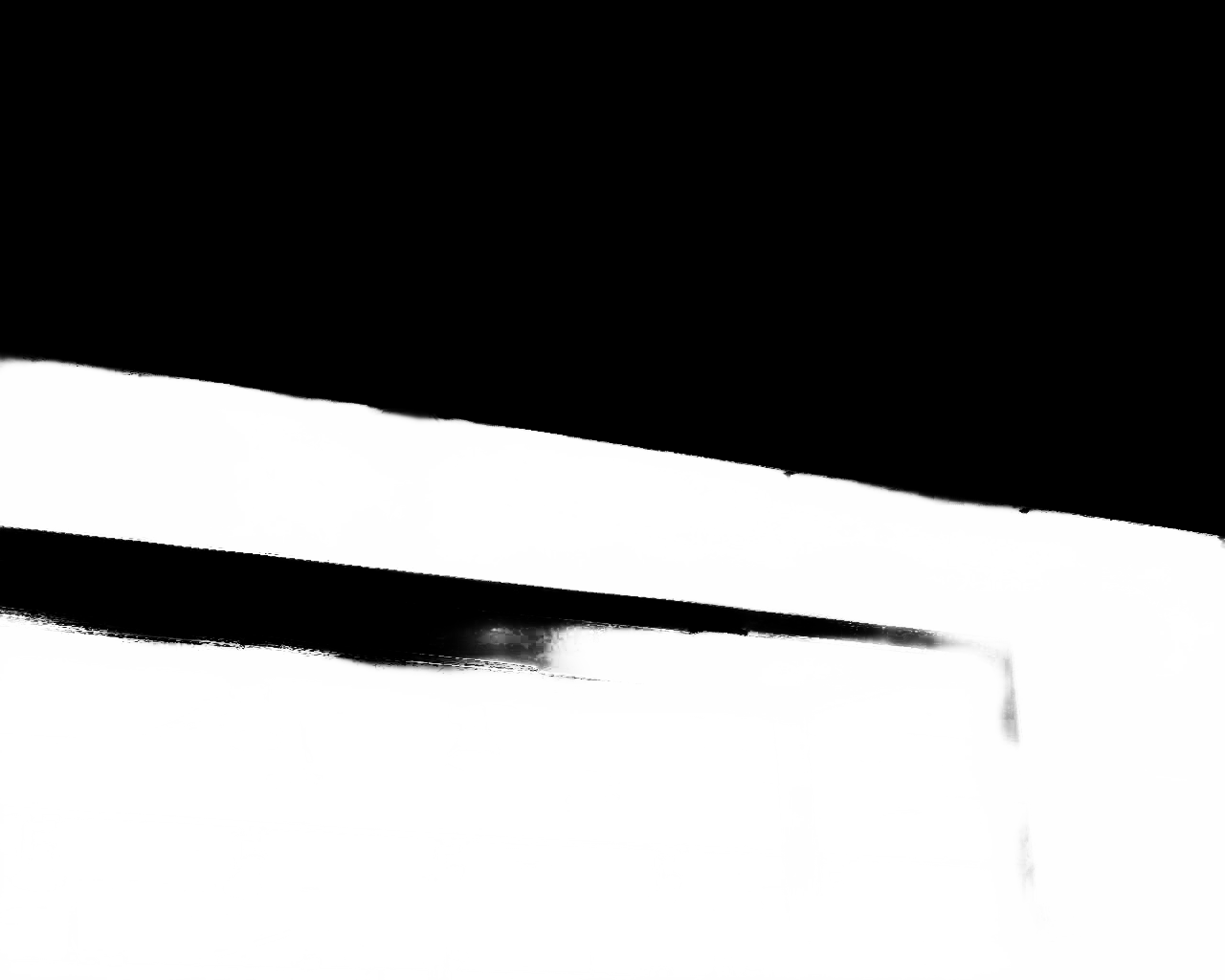}
            
            \includegraphics[width=1\linewidth]{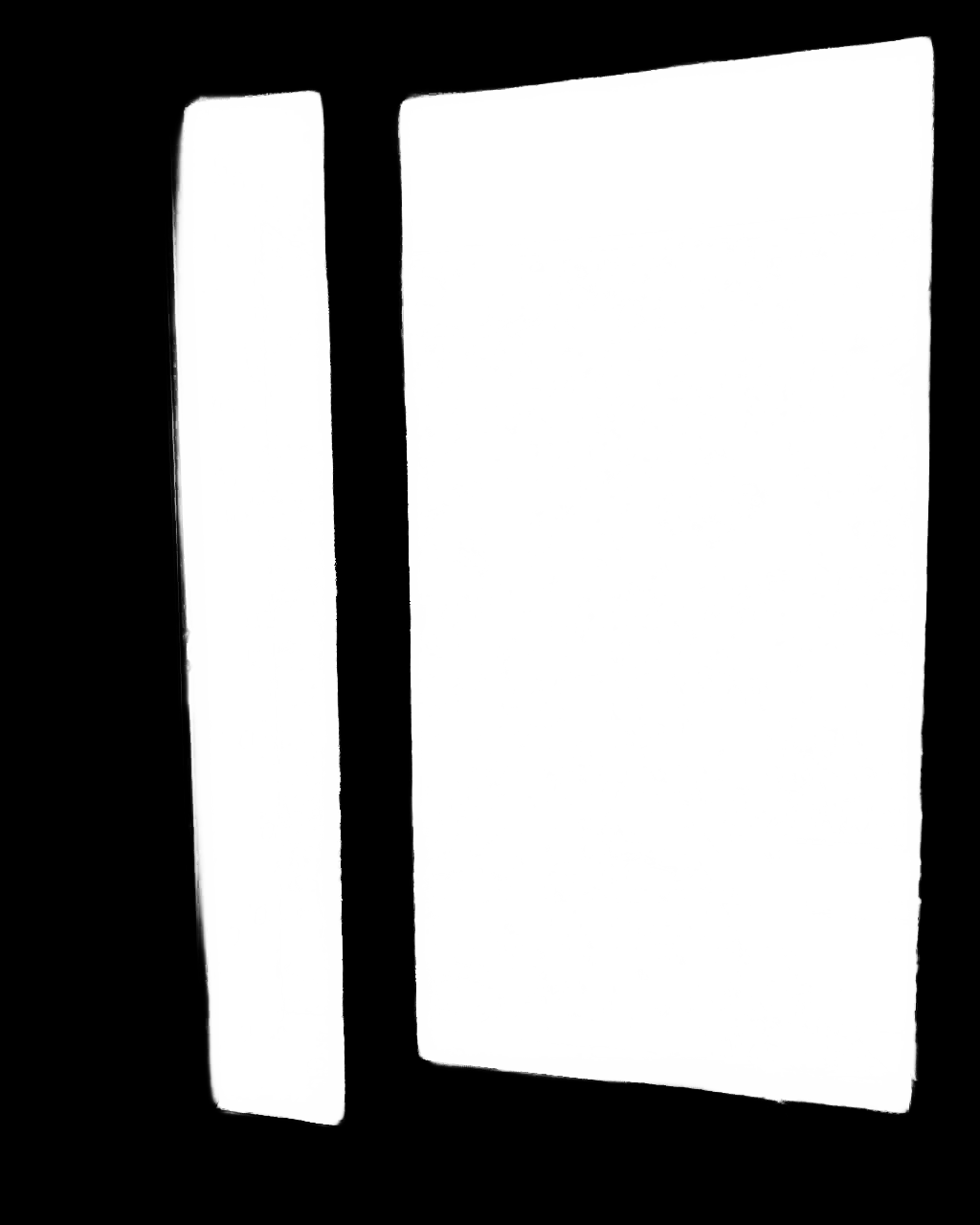}

            \includegraphics[width=1\linewidth]{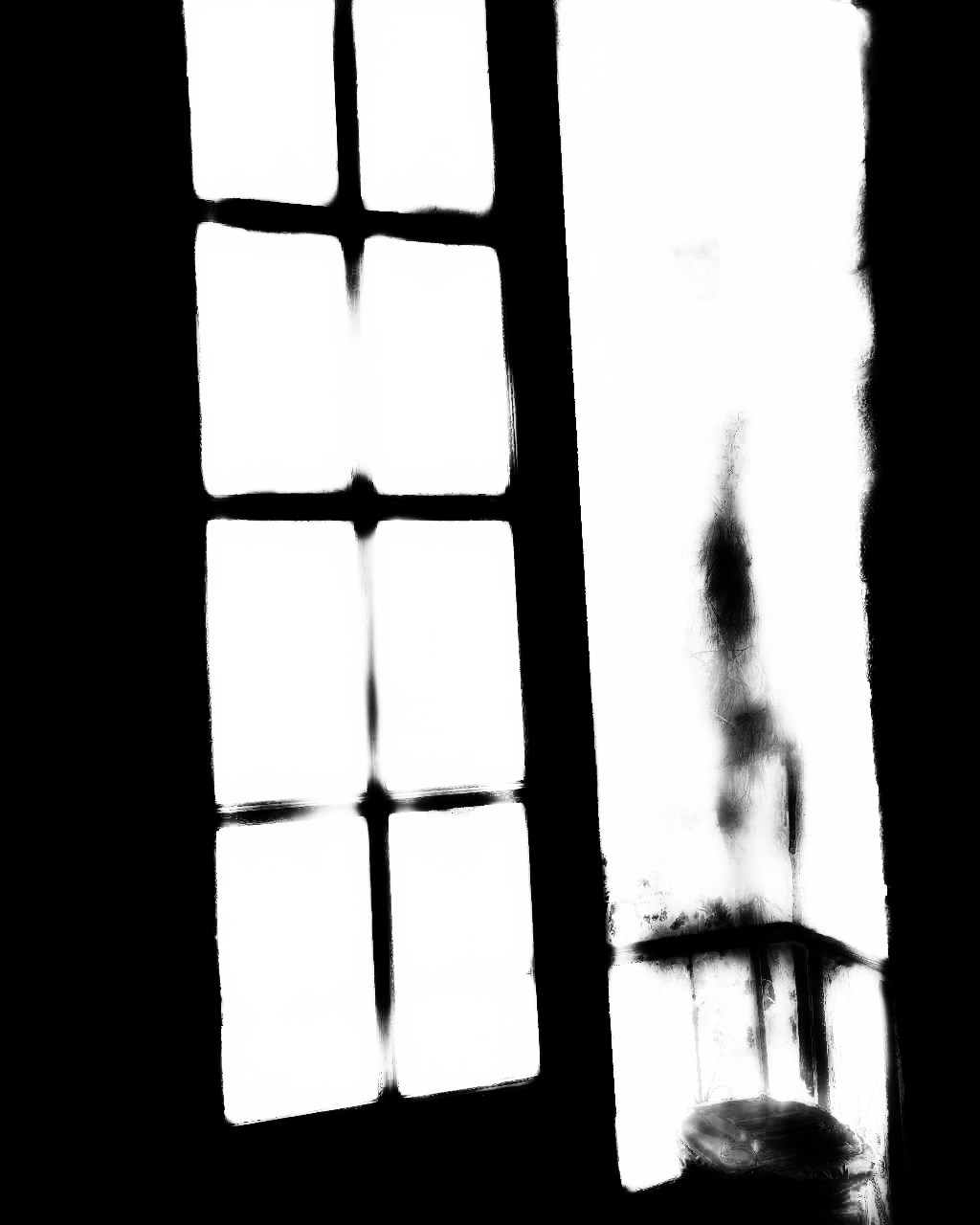}
      \end{minipage}
      }  
      \subfloat[GDNet]{\label{GD}
      \begin{minipage}[t]{0.07\textwidth}
            \centering
            \includegraphics[width=1\linewidth]{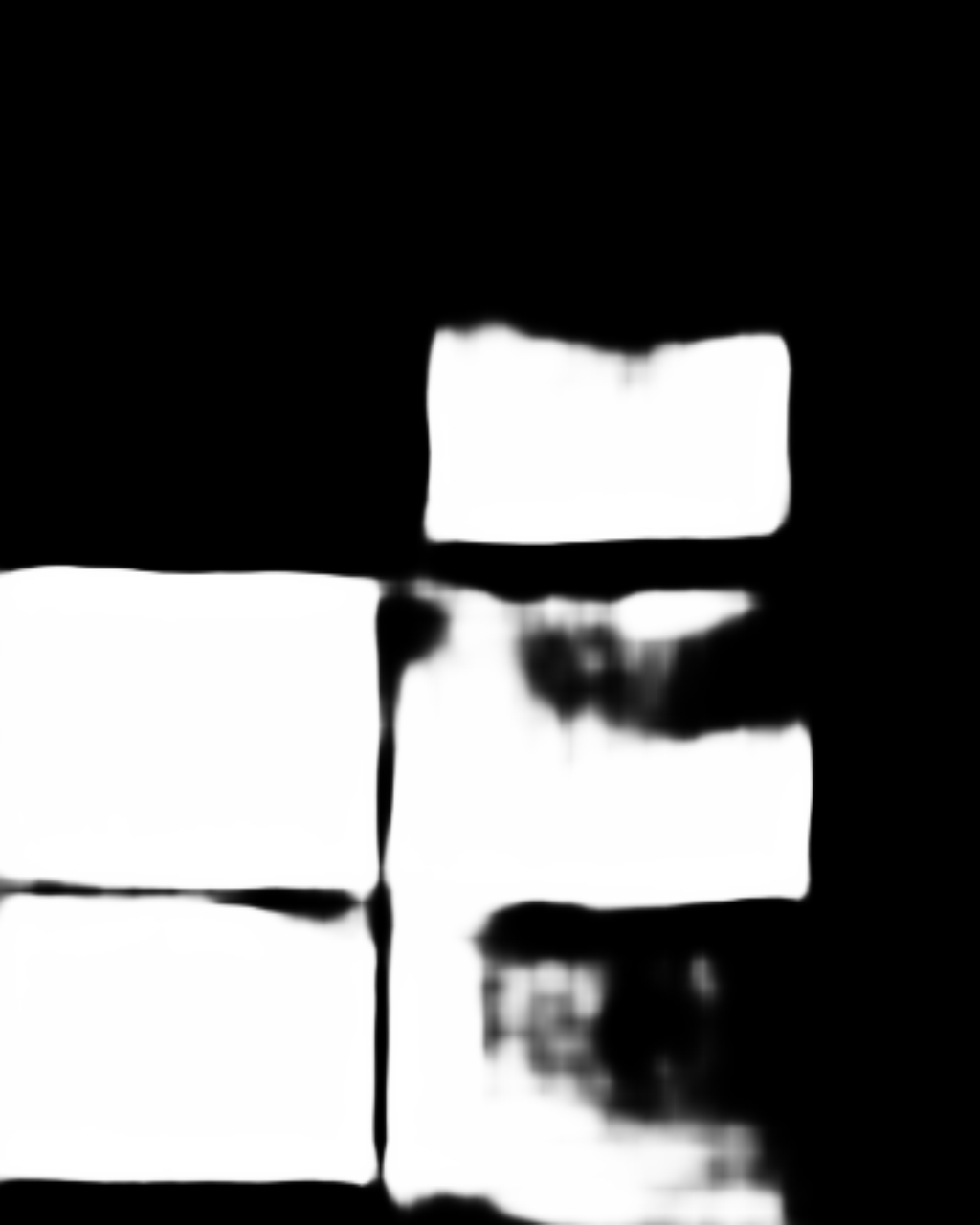}

            \includegraphics[width=1\linewidth]{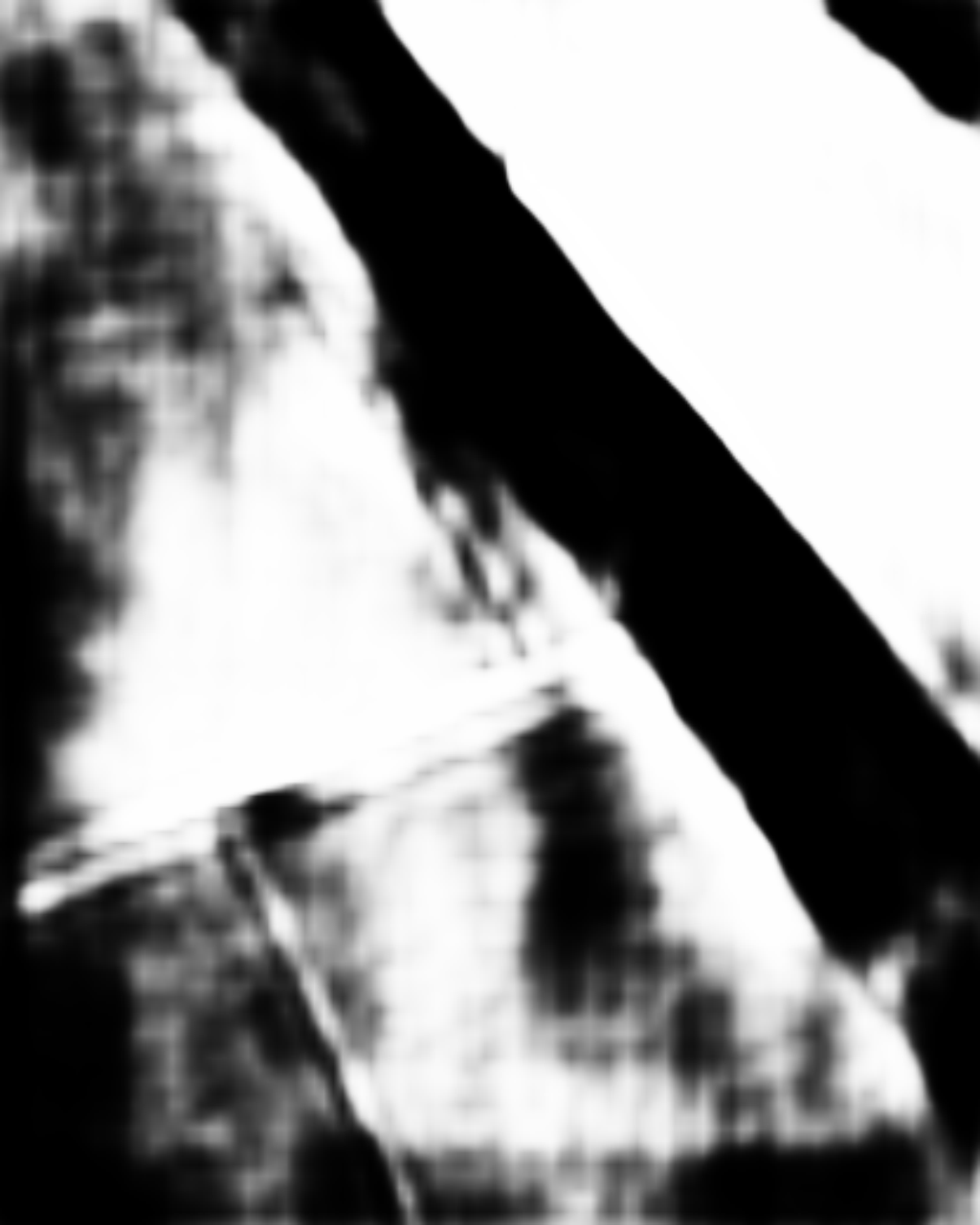}    

            \includegraphics[width=1\linewidth]{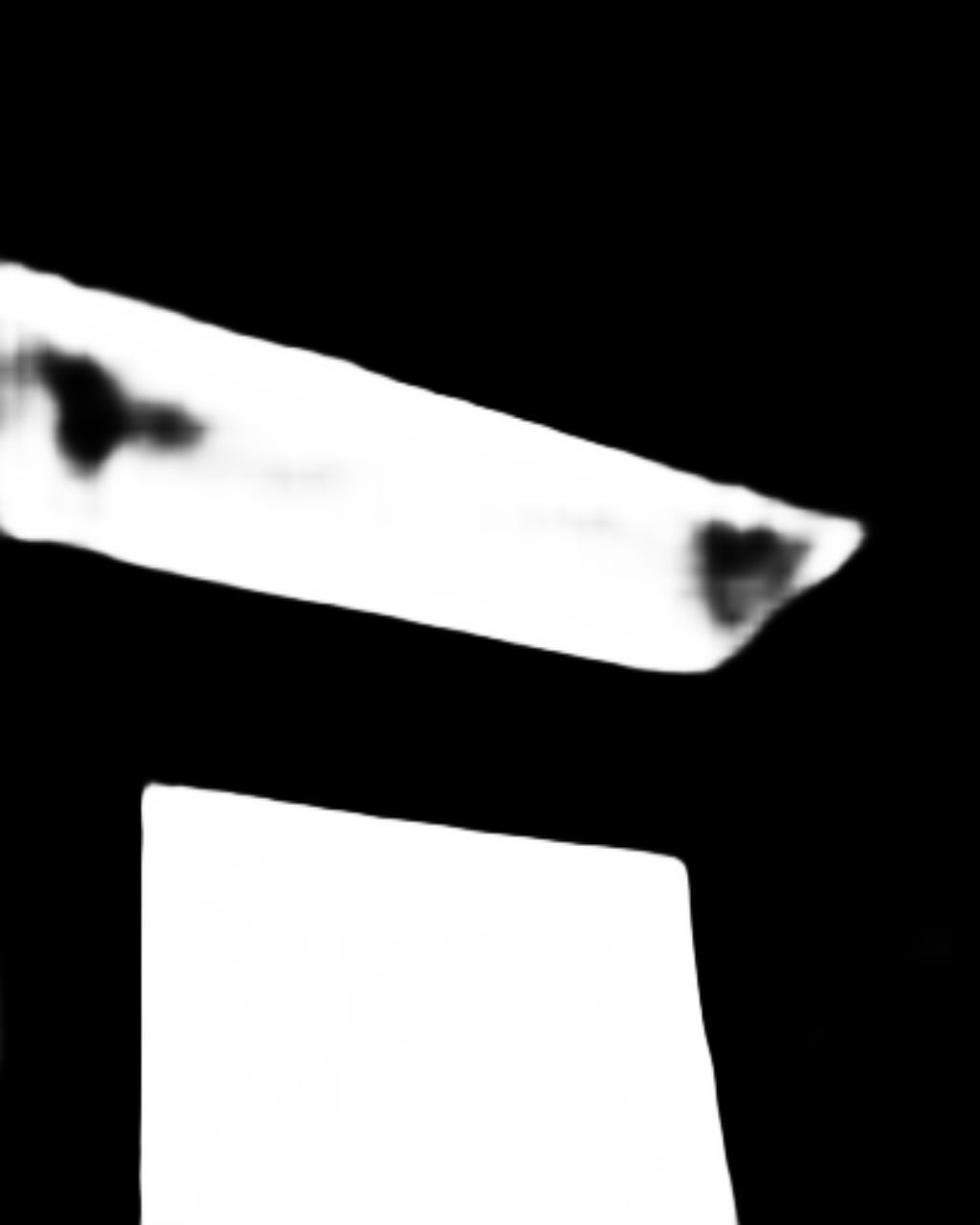}

            \includegraphics[width=1\linewidth]{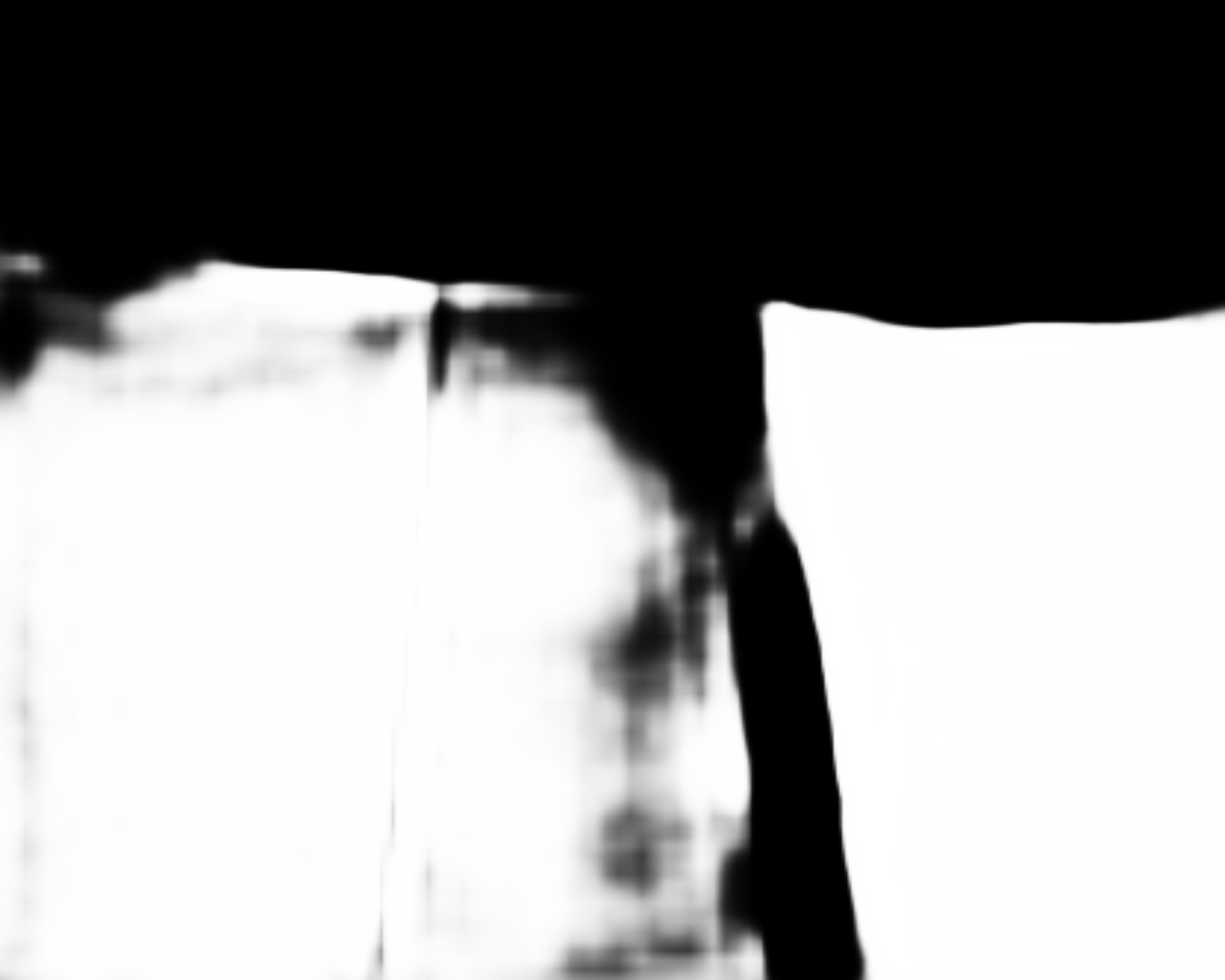}

            \includegraphics[width=1\linewidth]{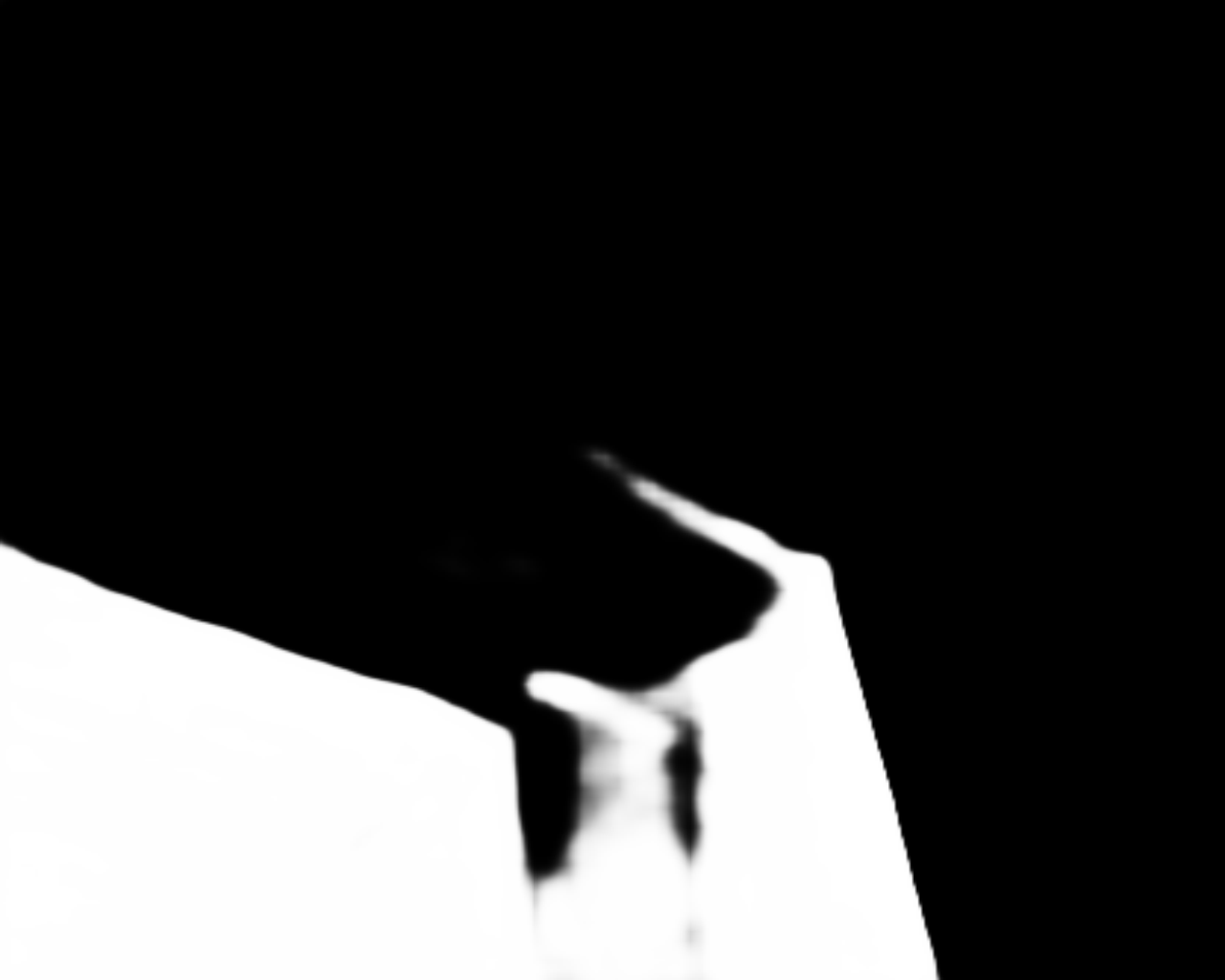}

            \includegraphics[width=1\linewidth]{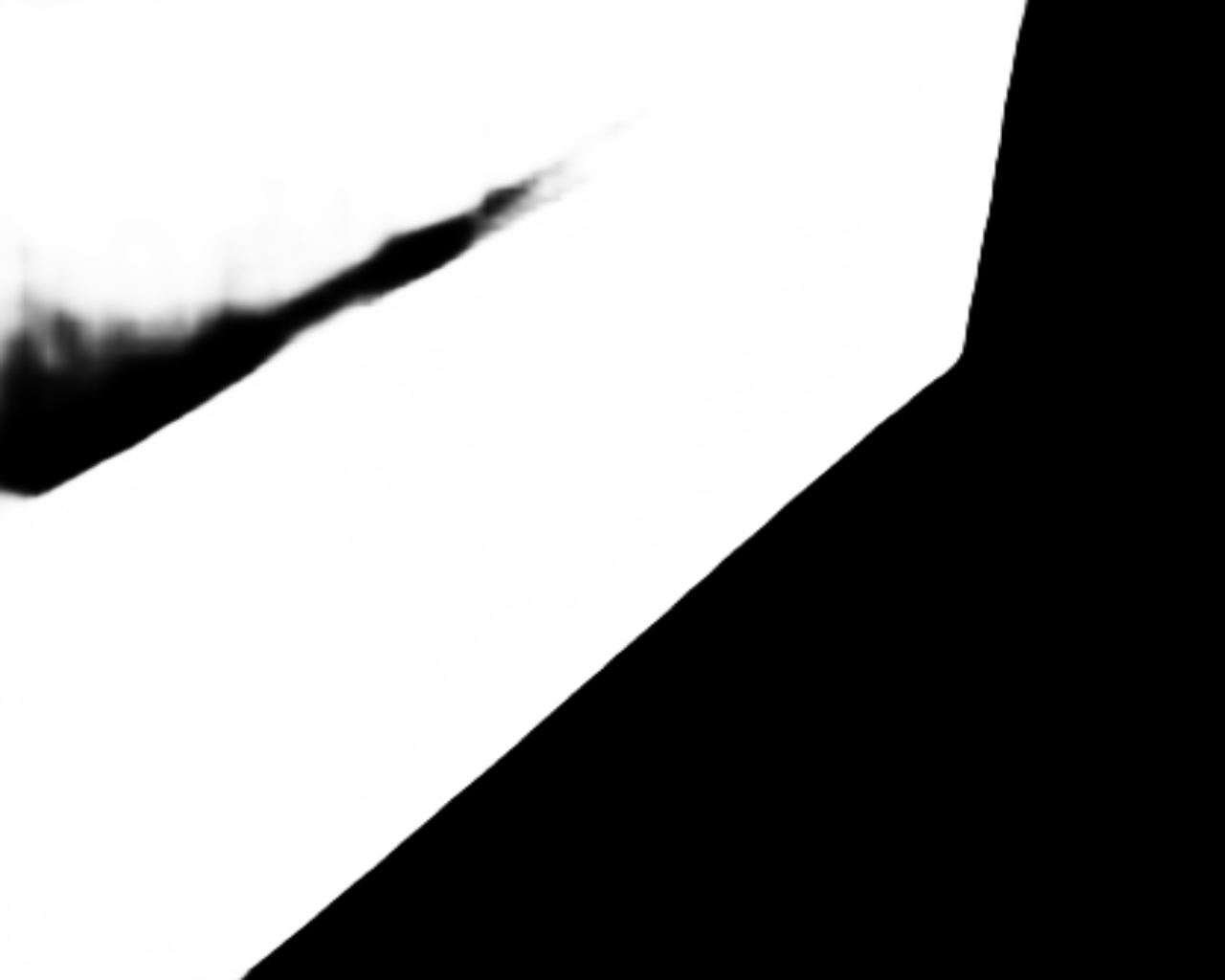}

            \includegraphics[width=1\linewidth]{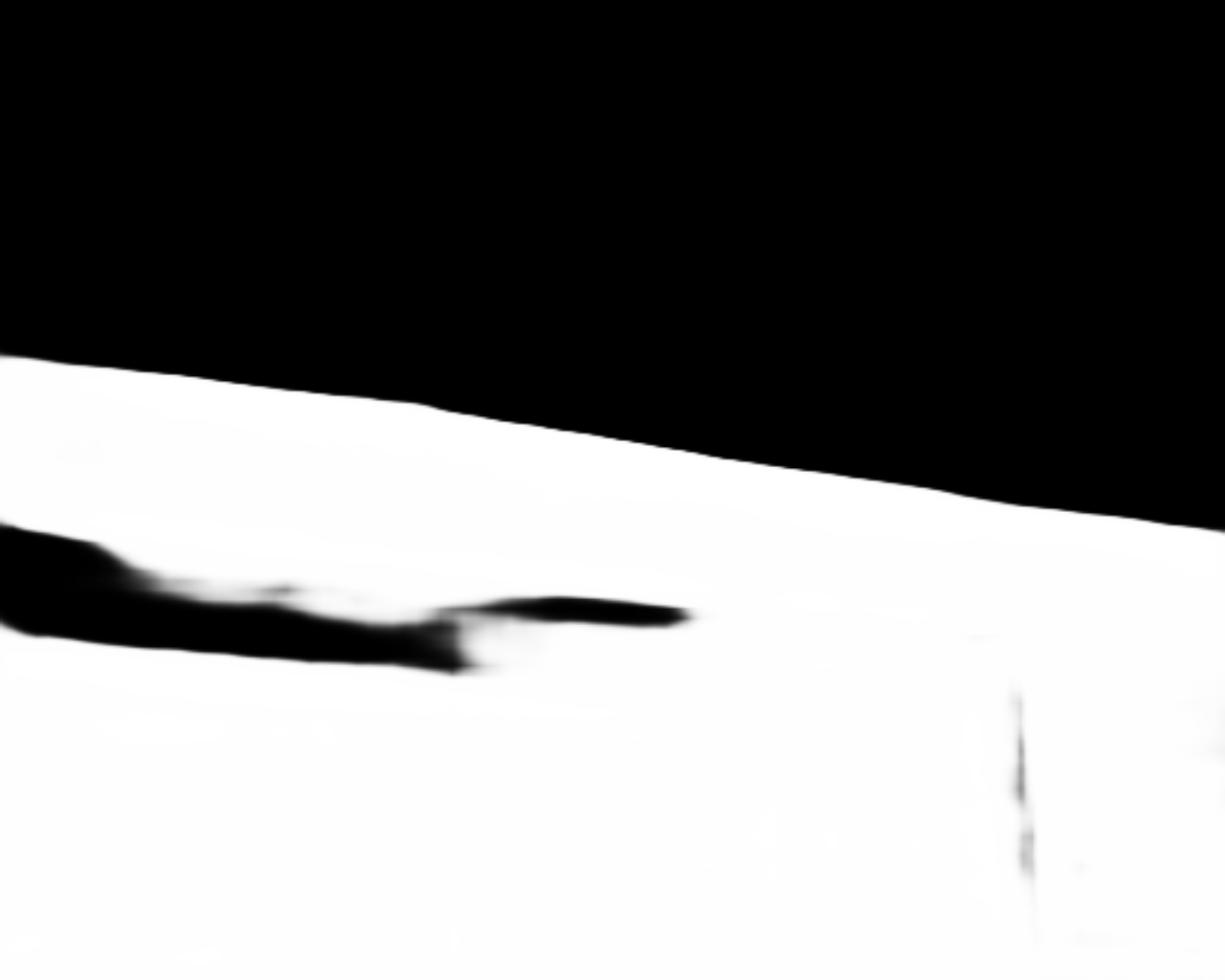}
            
            \includegraphics[width=1\linewidth]{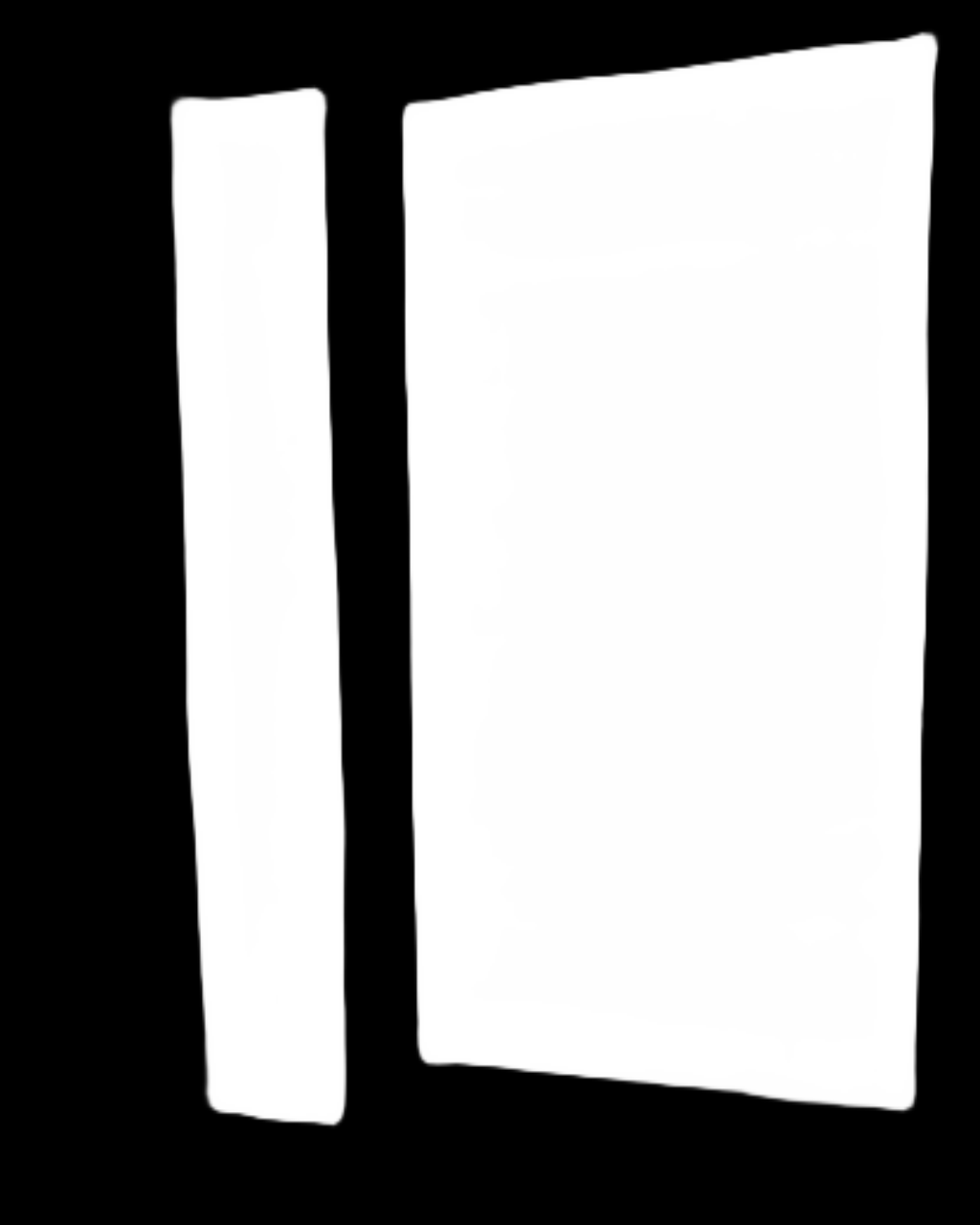}

            \includegraphics[width=1\linewidth]{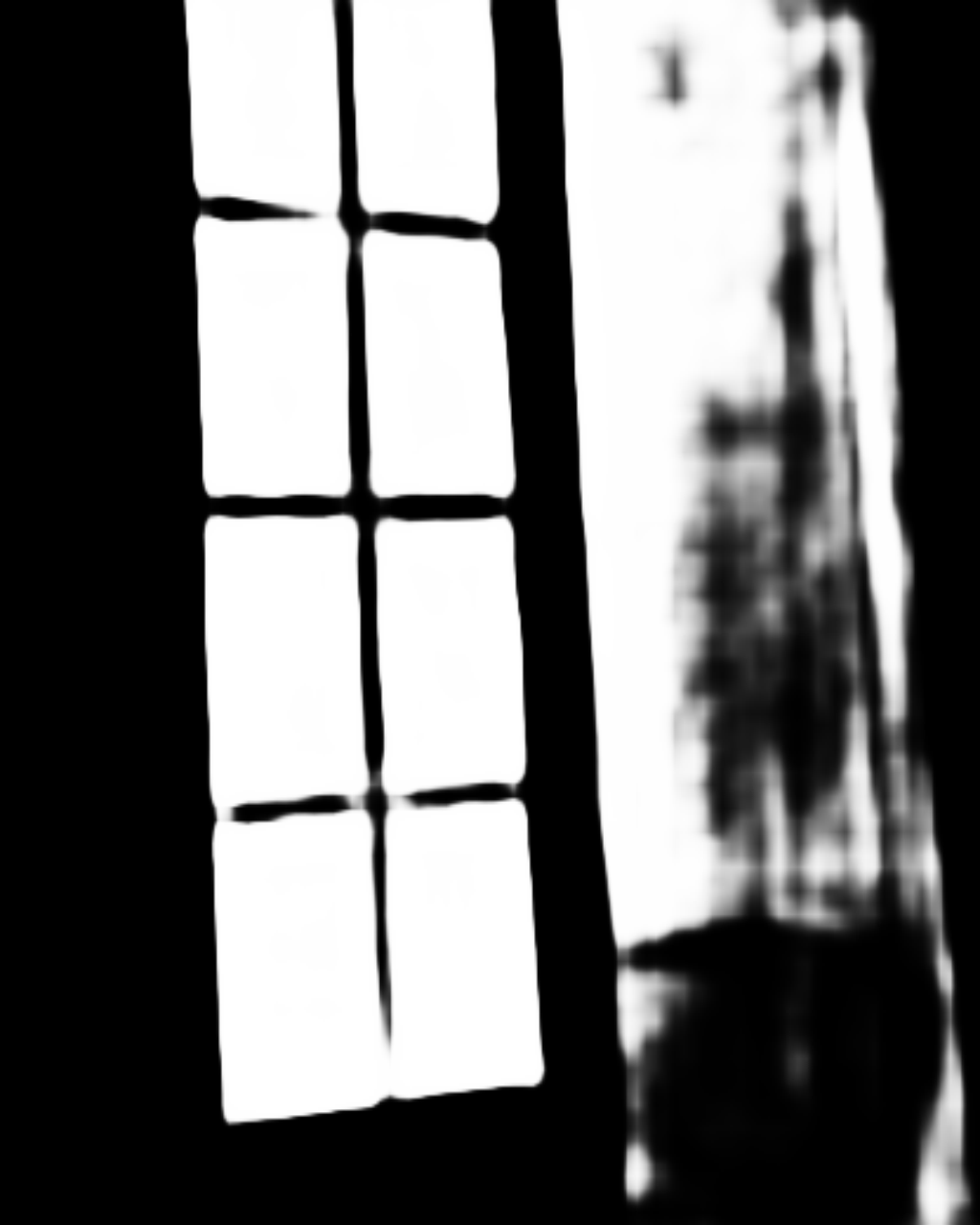}
      \end{minipage}
      }  
      \subfloat[Ours]{\label{our}
      \begin{minipage}[t]{0.07\textwidth}
            \centering
            \includegraphics[width=1\linewidth]{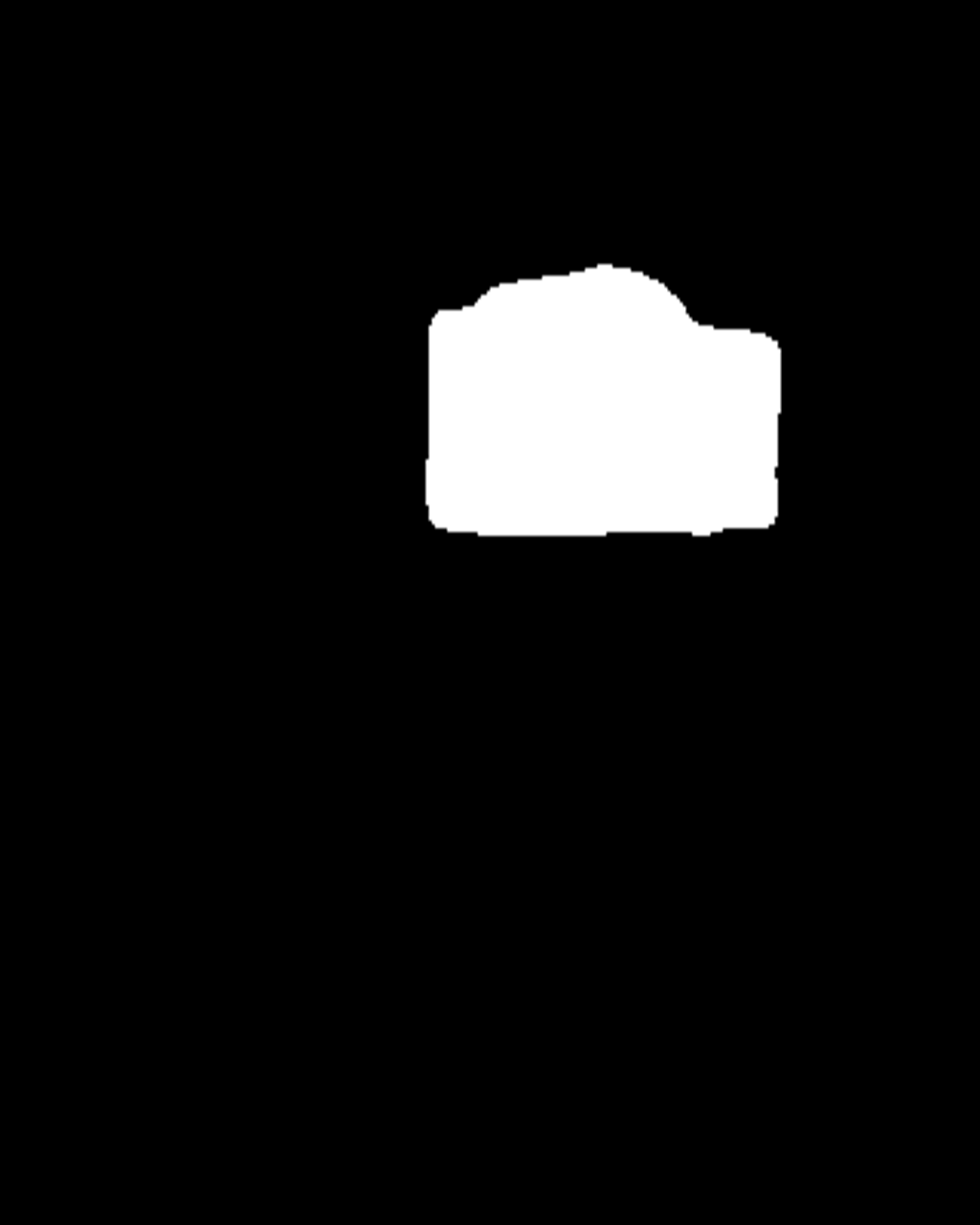}

            \includegraphics[width=1\linewidth]{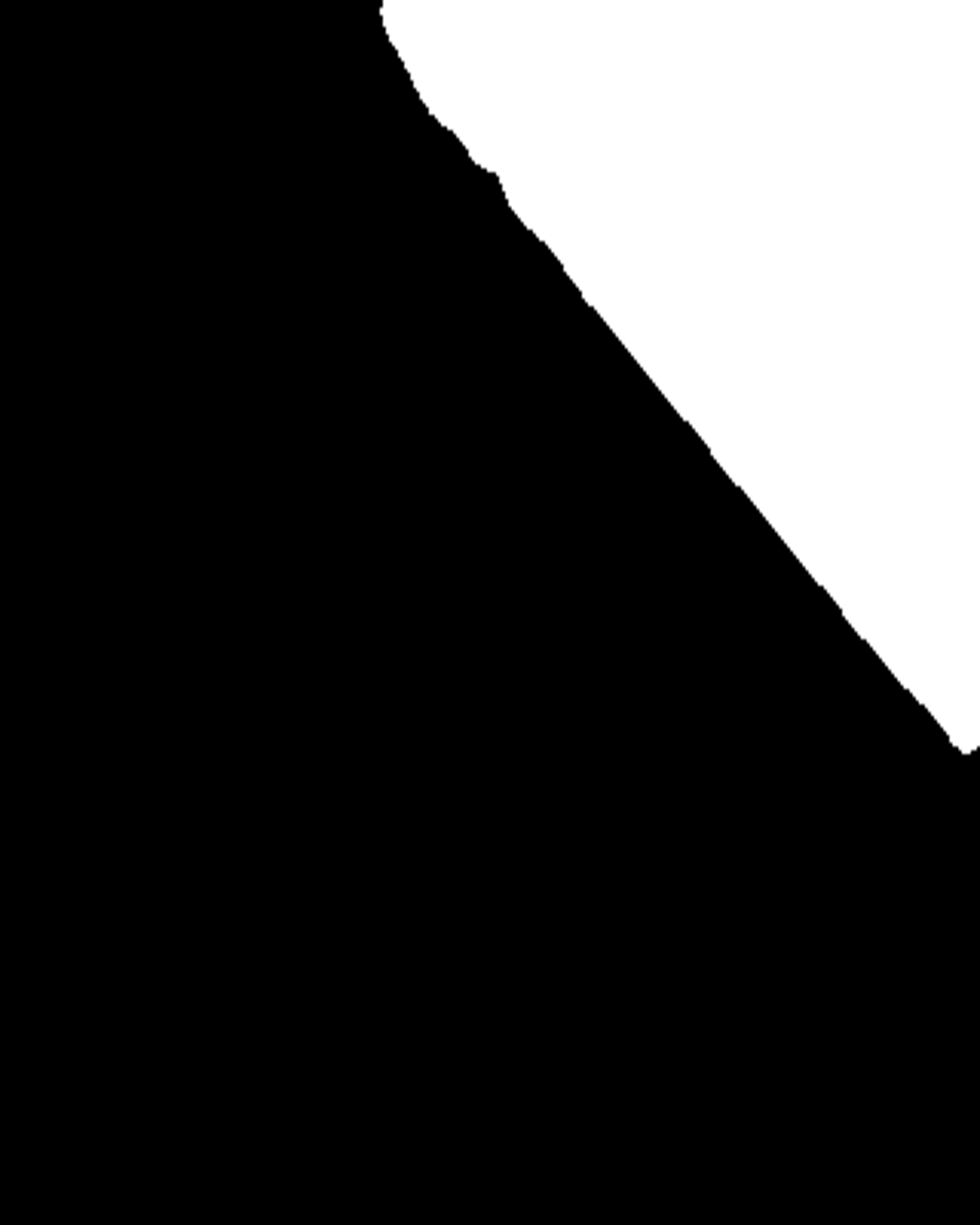}

            \includegraphics[width=1\linewidth]{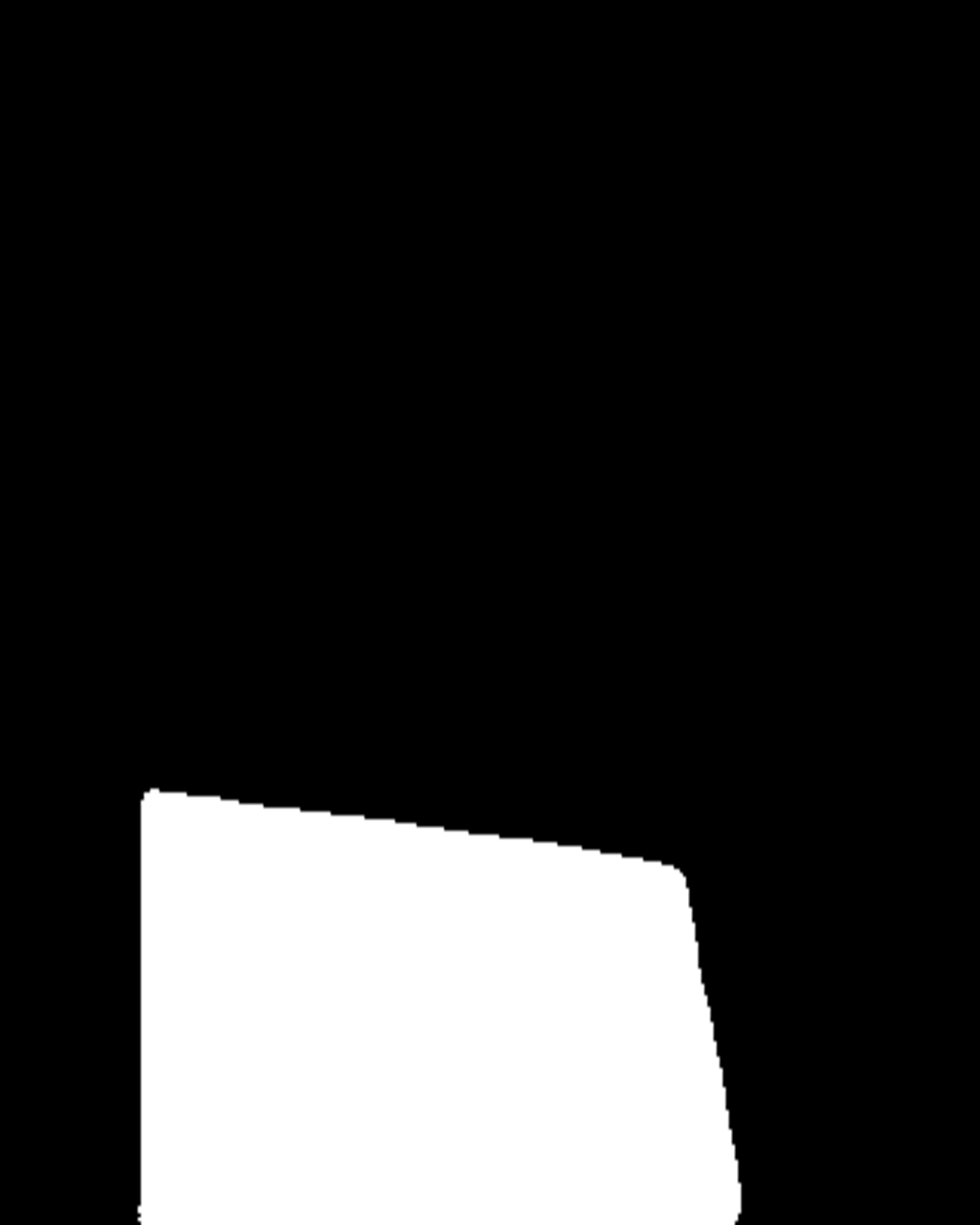}

            \includegraphics[width=1\linewidth]{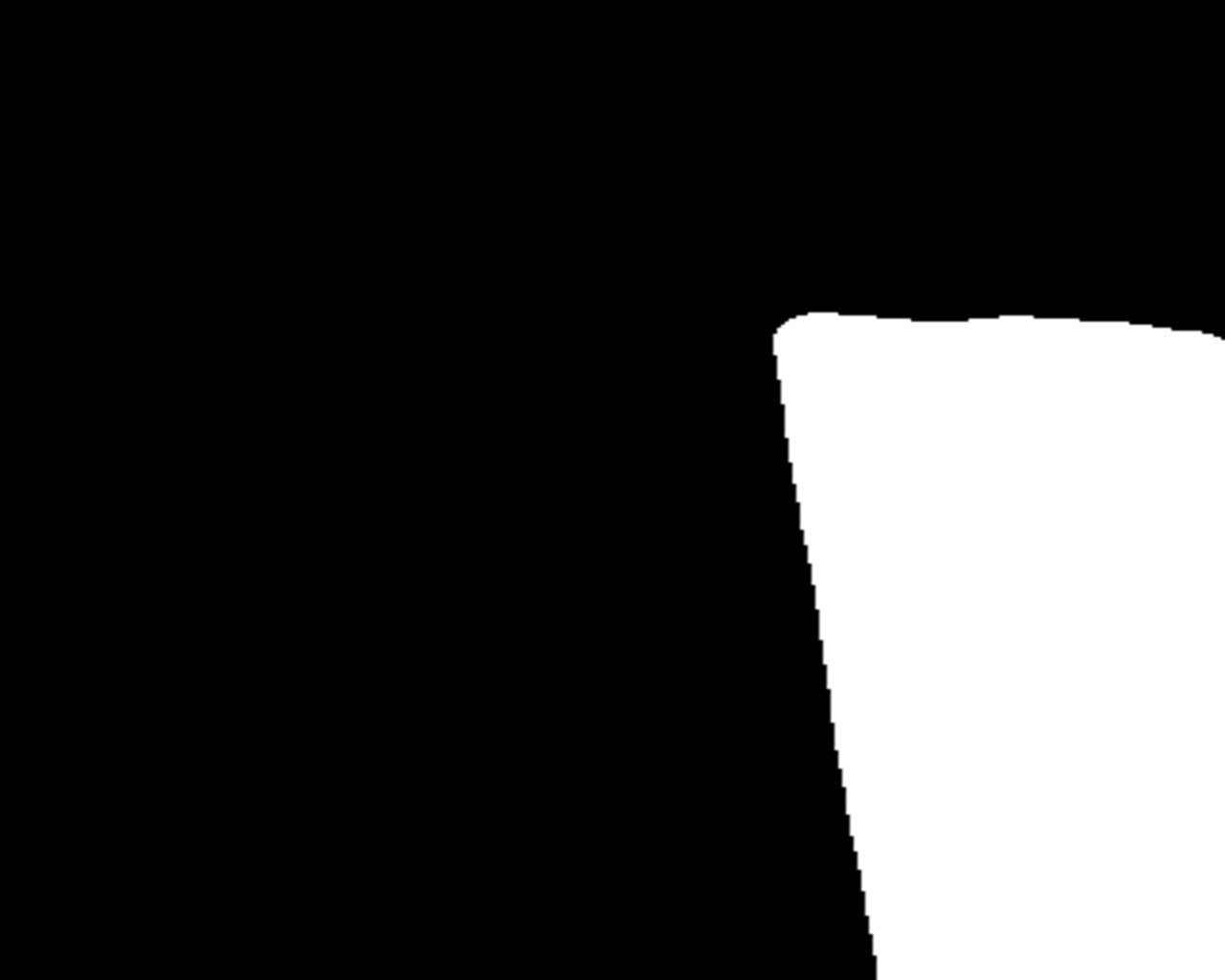}

            \includegraphics[width=1\linewidth]{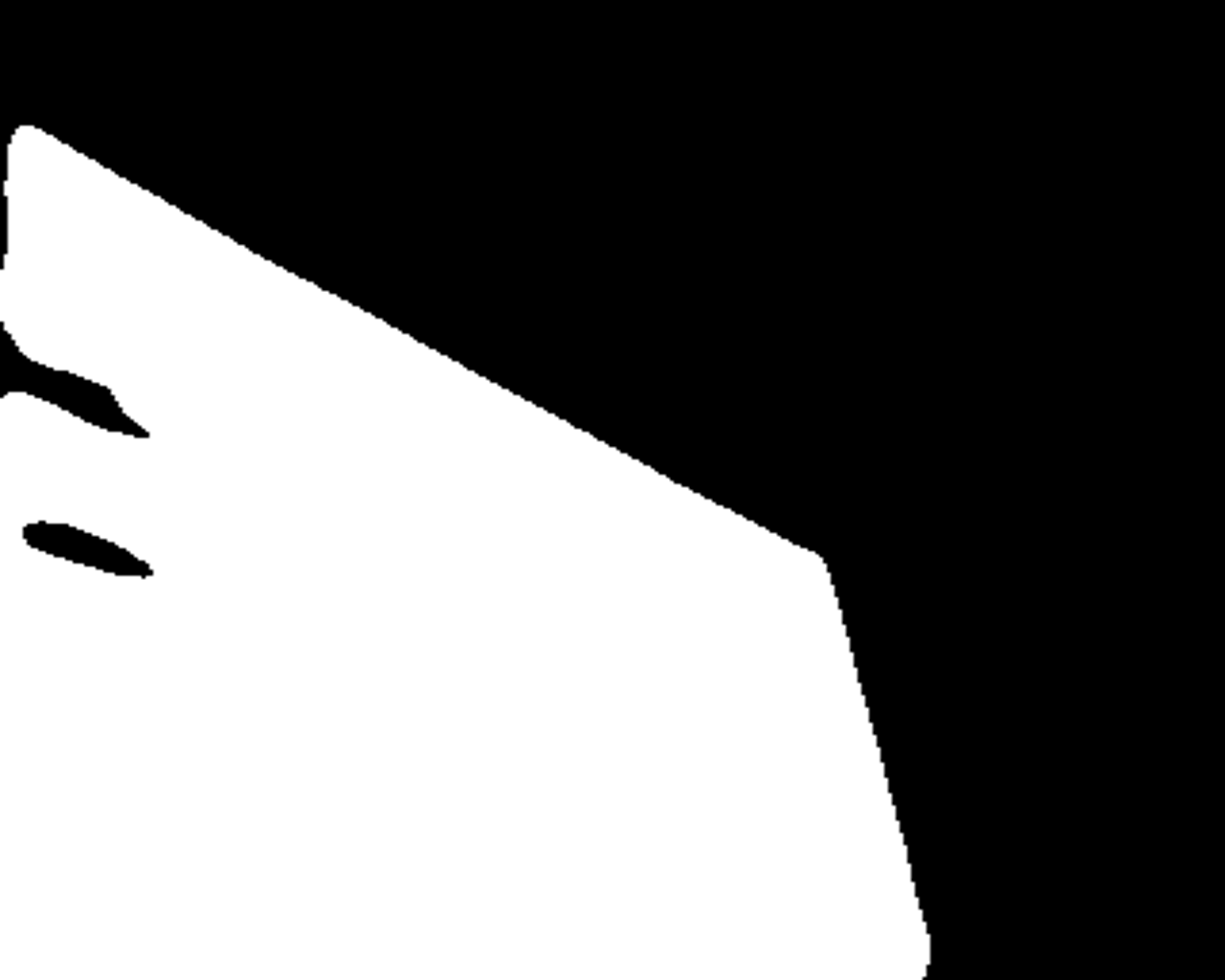}

            \includegraphics[width=1\linewidth]{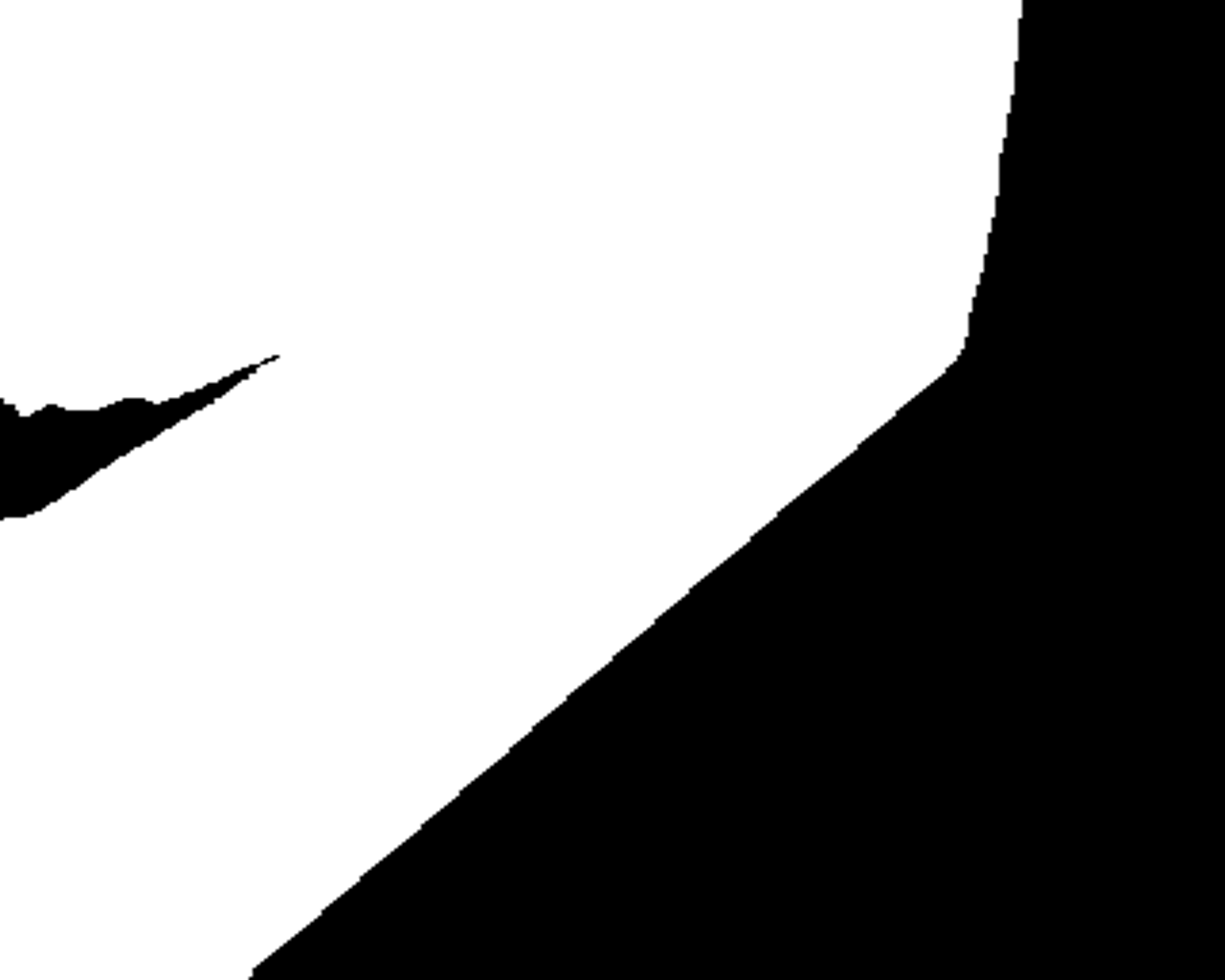}

            \includegraphics[width=1\linewidth]{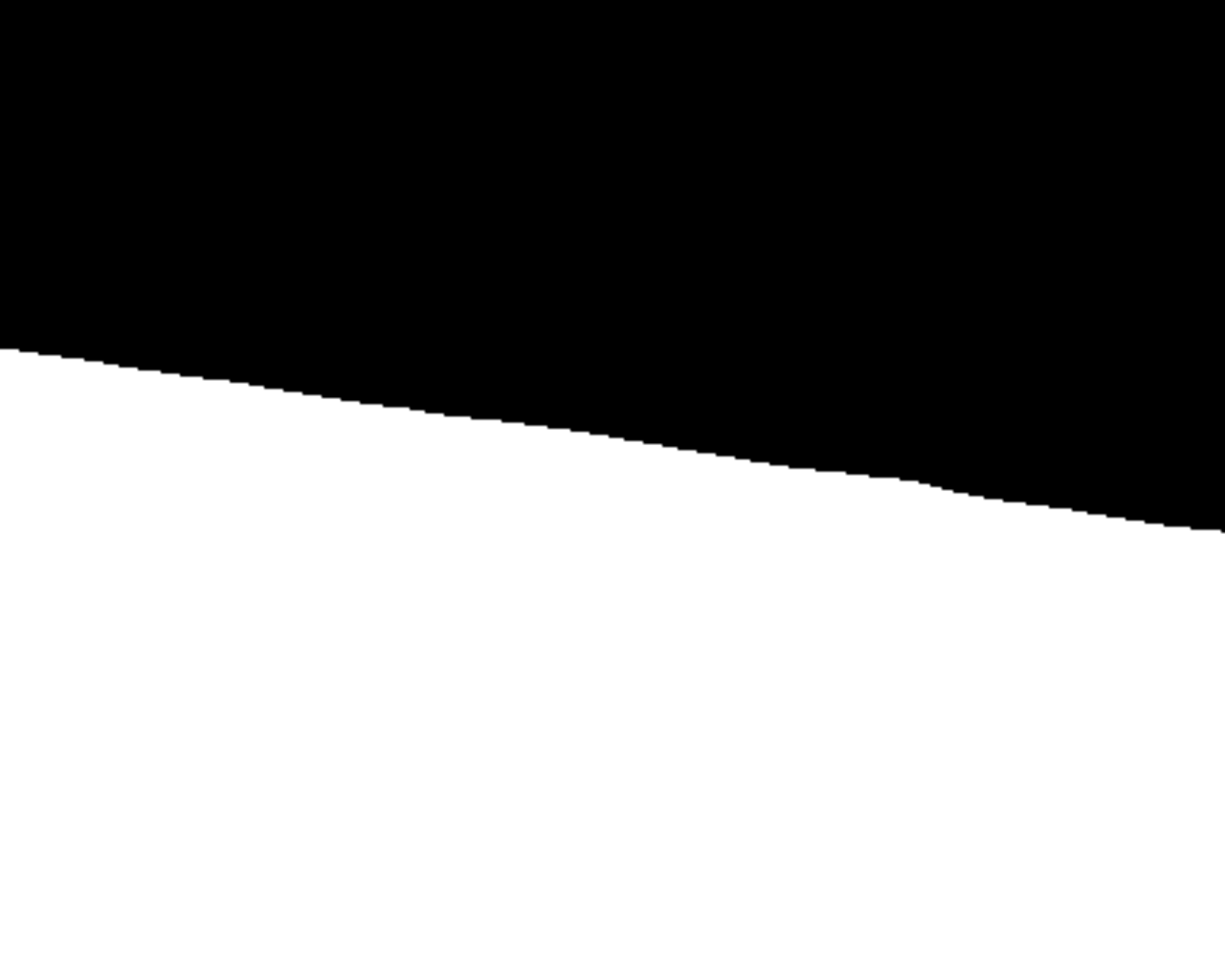}
            
            \includegraphics[width=1\linewidth]{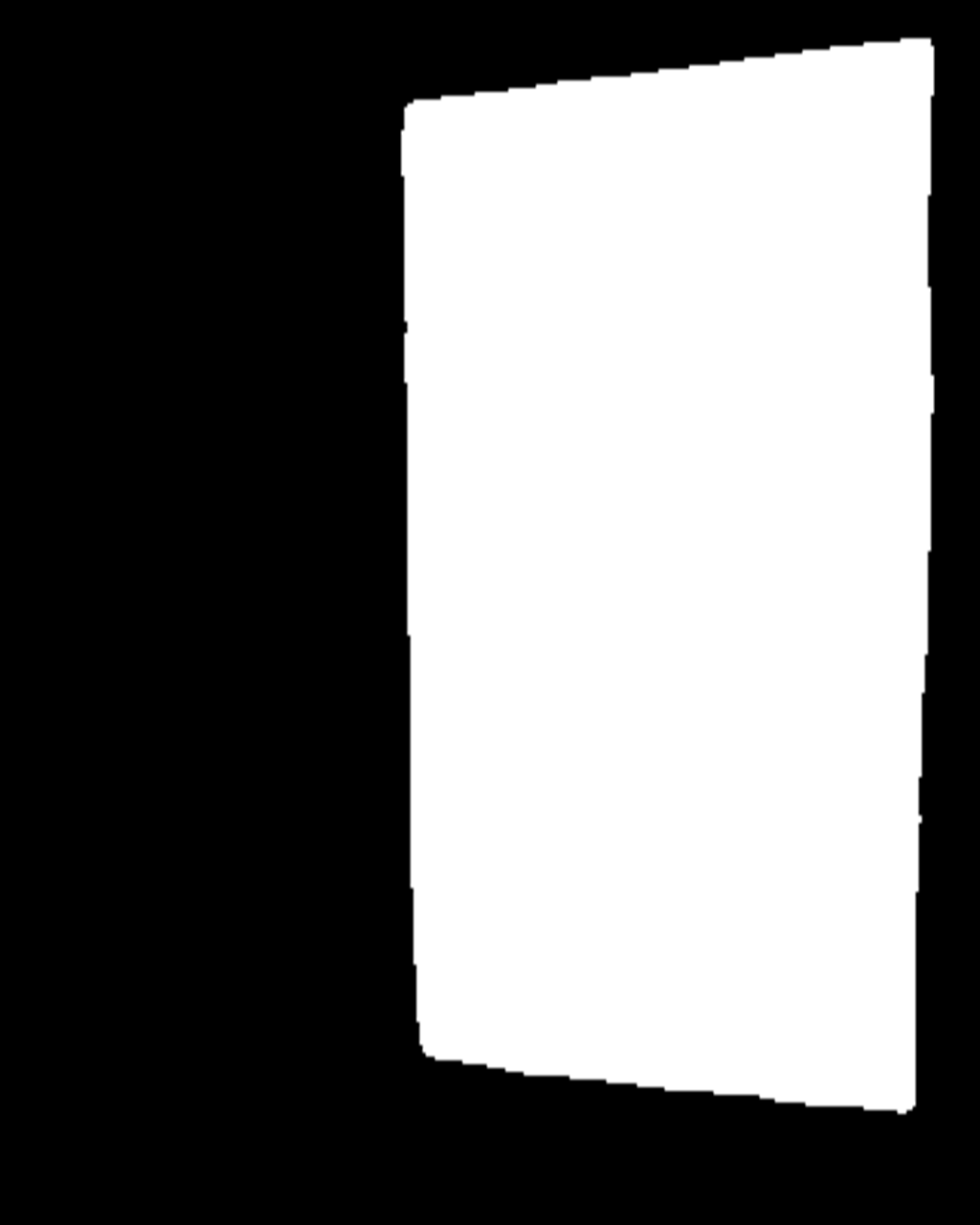}

            \includegraphics[width=1\linewidth]{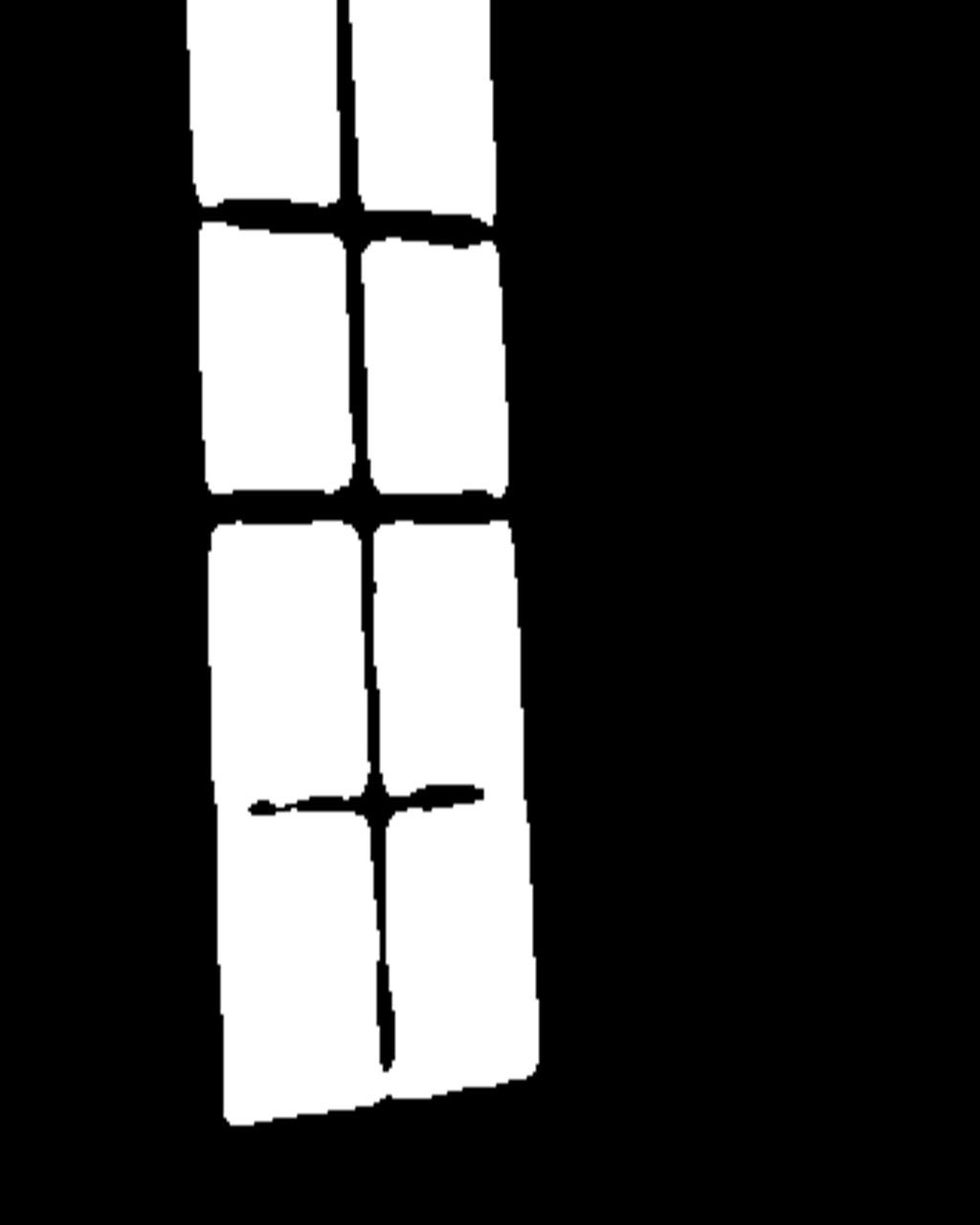}
      \end{minipage}
      }  
            \subfloat[GT]{\label{GT}
      \begin{minipage}[t]{0.07\textwidth}
            \centering
            \includegraphics[width=1\linewidth]{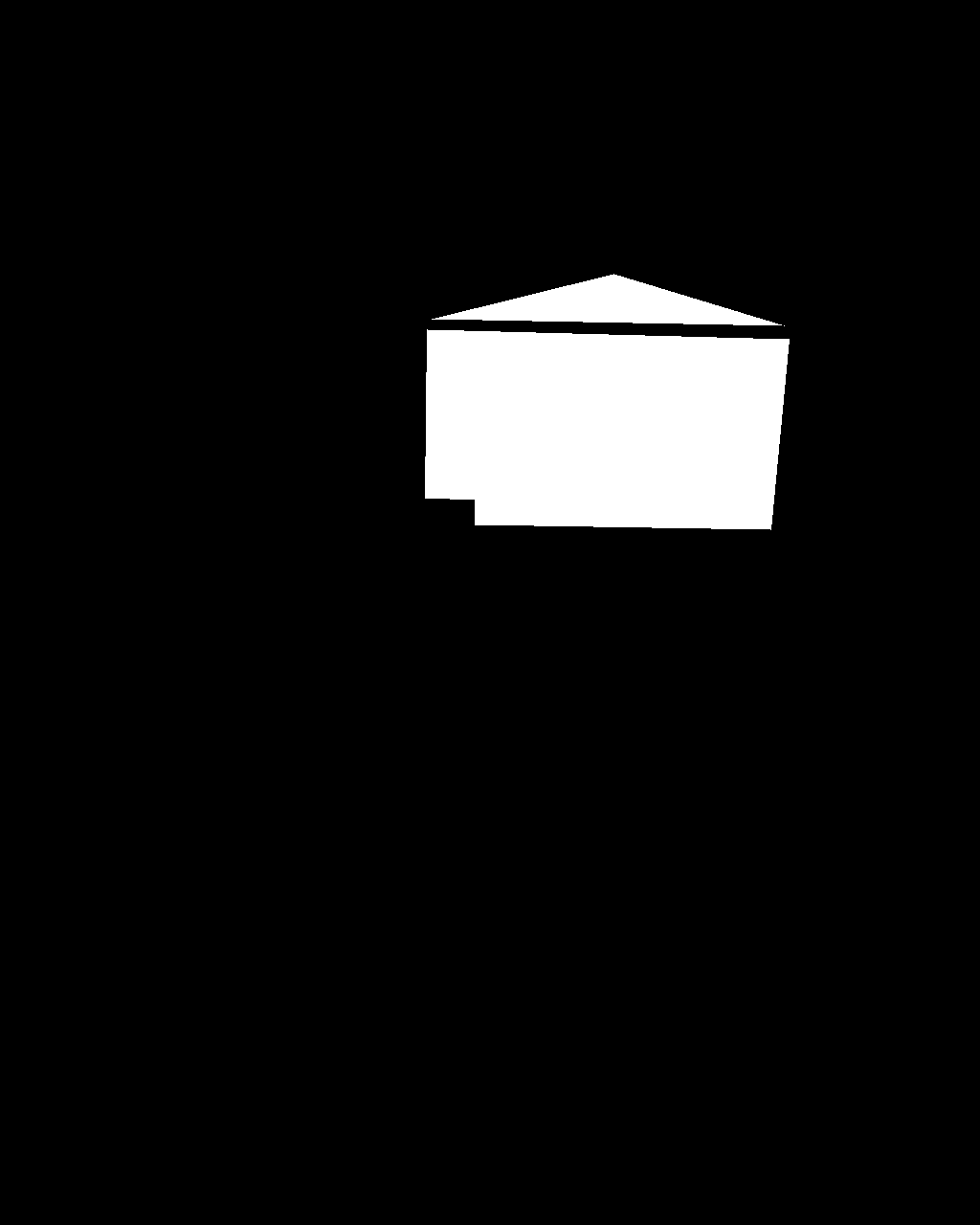}

            \includegraphics[width=1\linewidth]{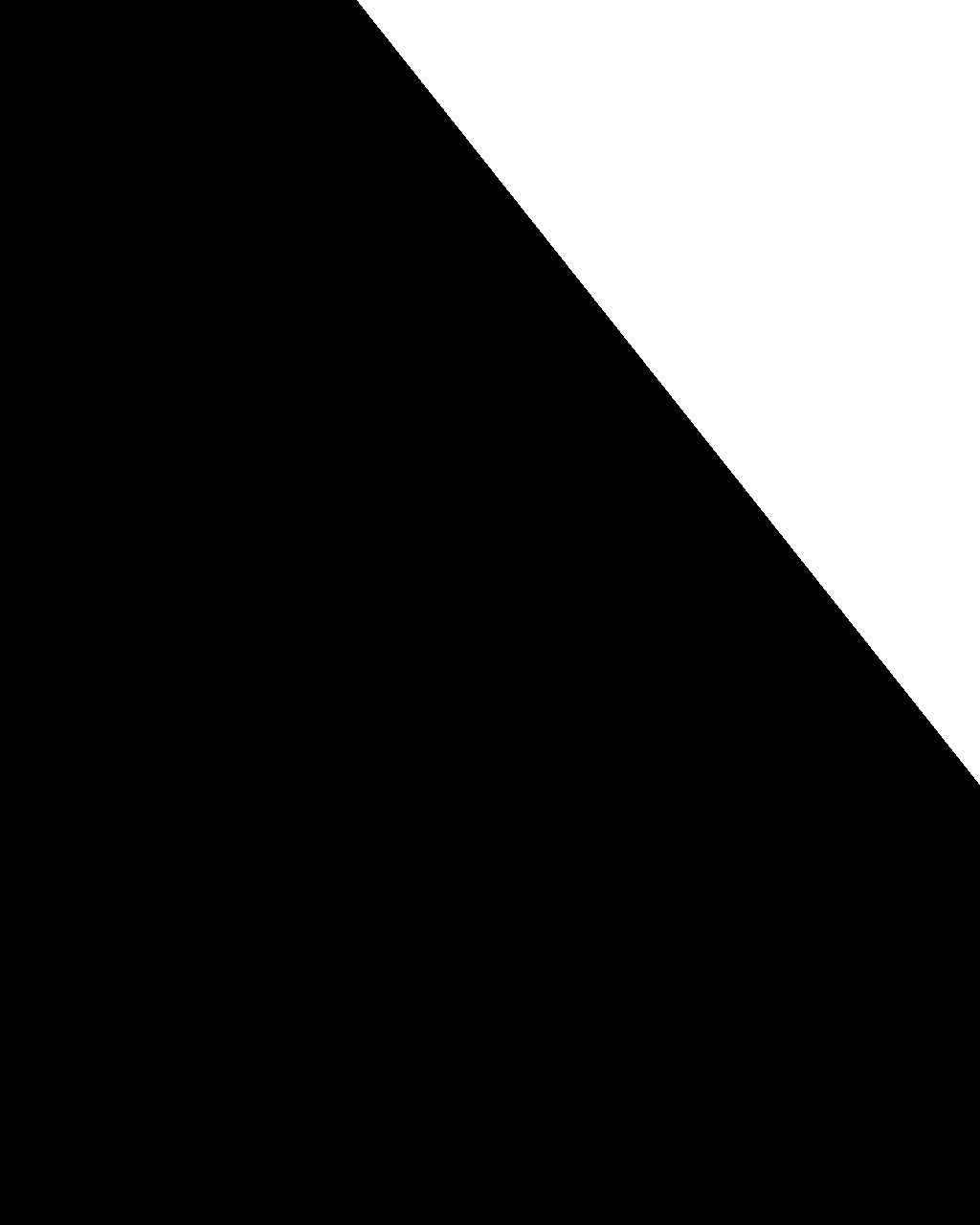}

            \includegraphics[width=1\linewidth]{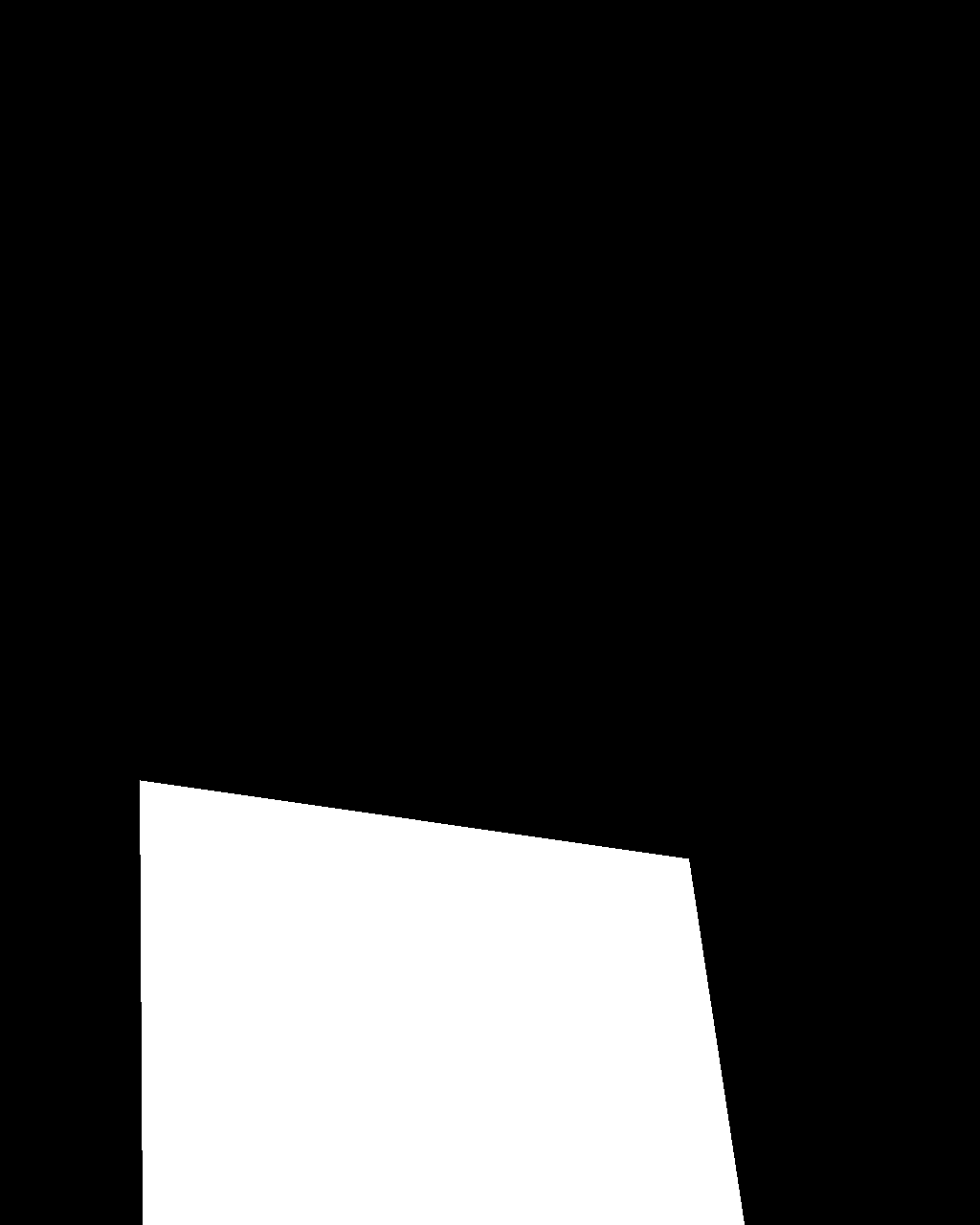}

            \includegraphics[width=1\linewidth]{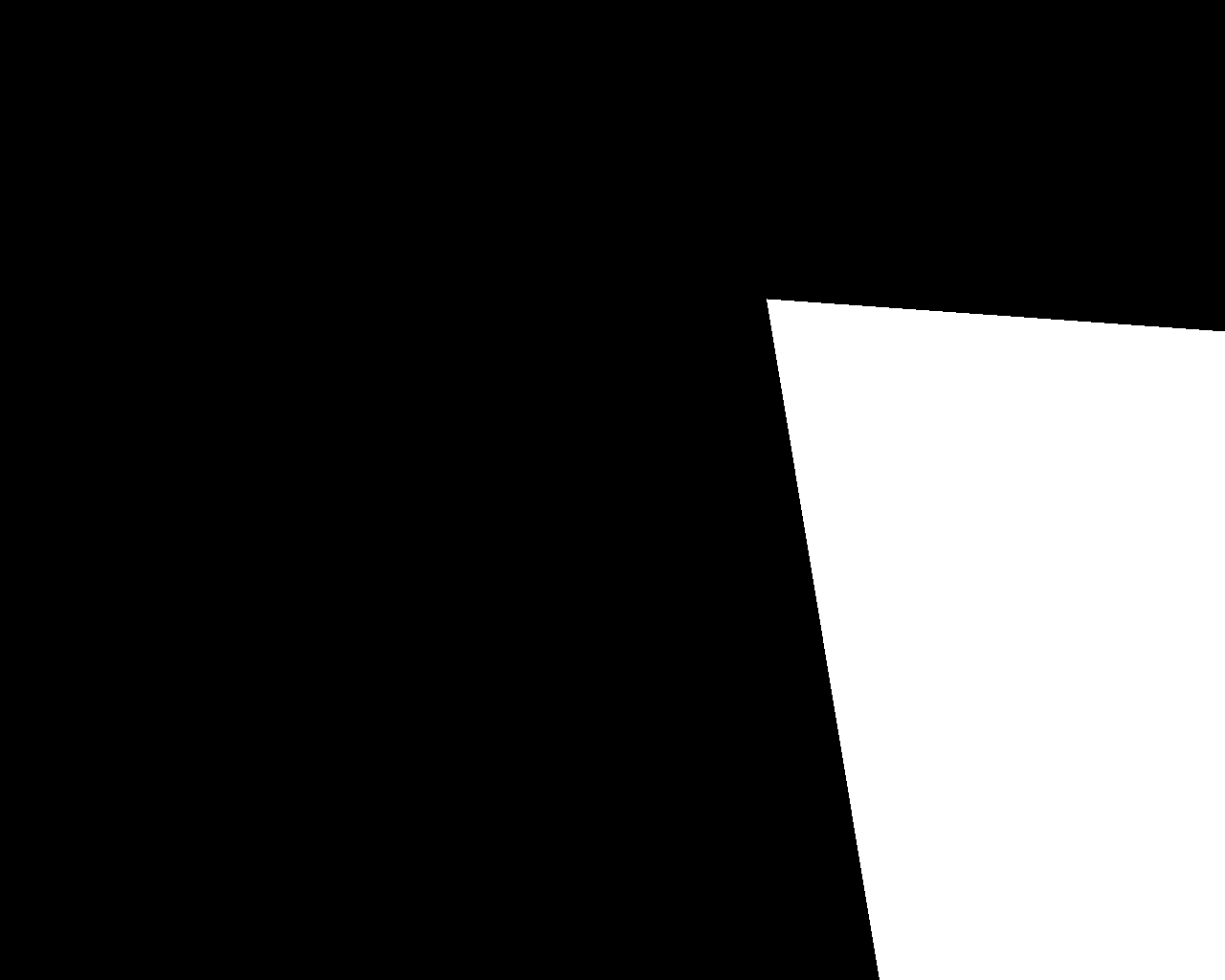}

            \includegraphics[width=1\linewidth]{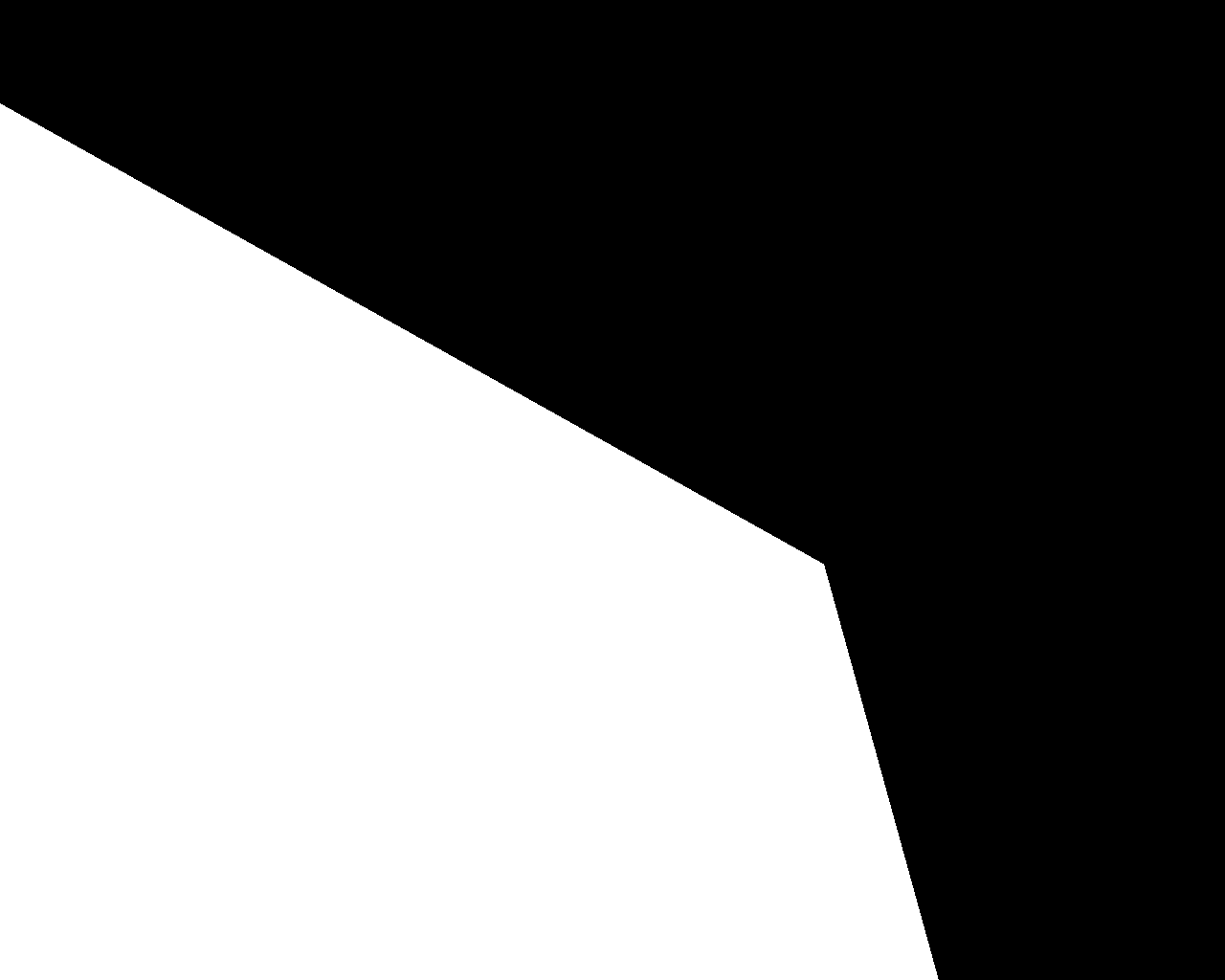}

            \includegraphics[width=1\linewidth]{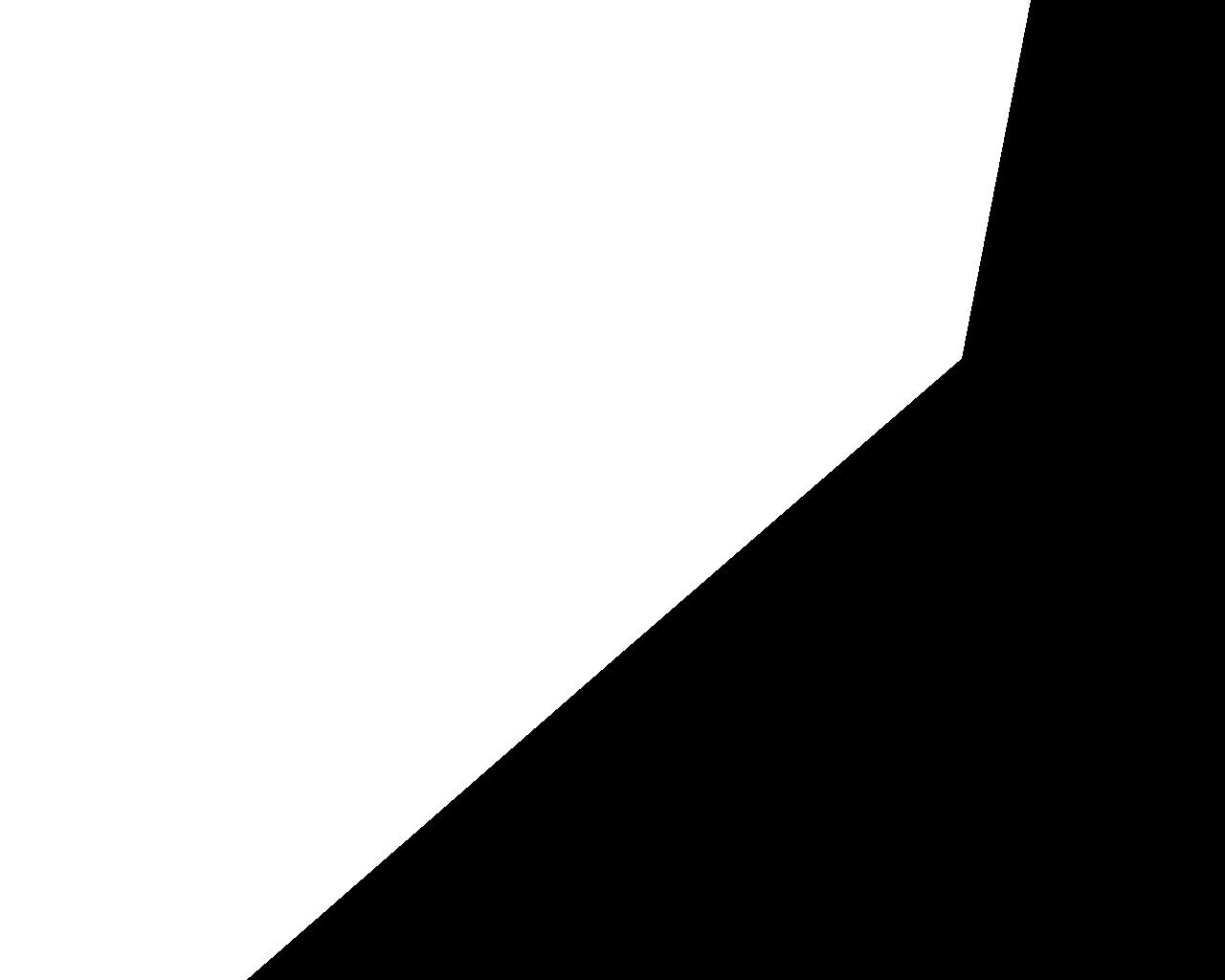}

            \includegraphics[width=1\linewidth]{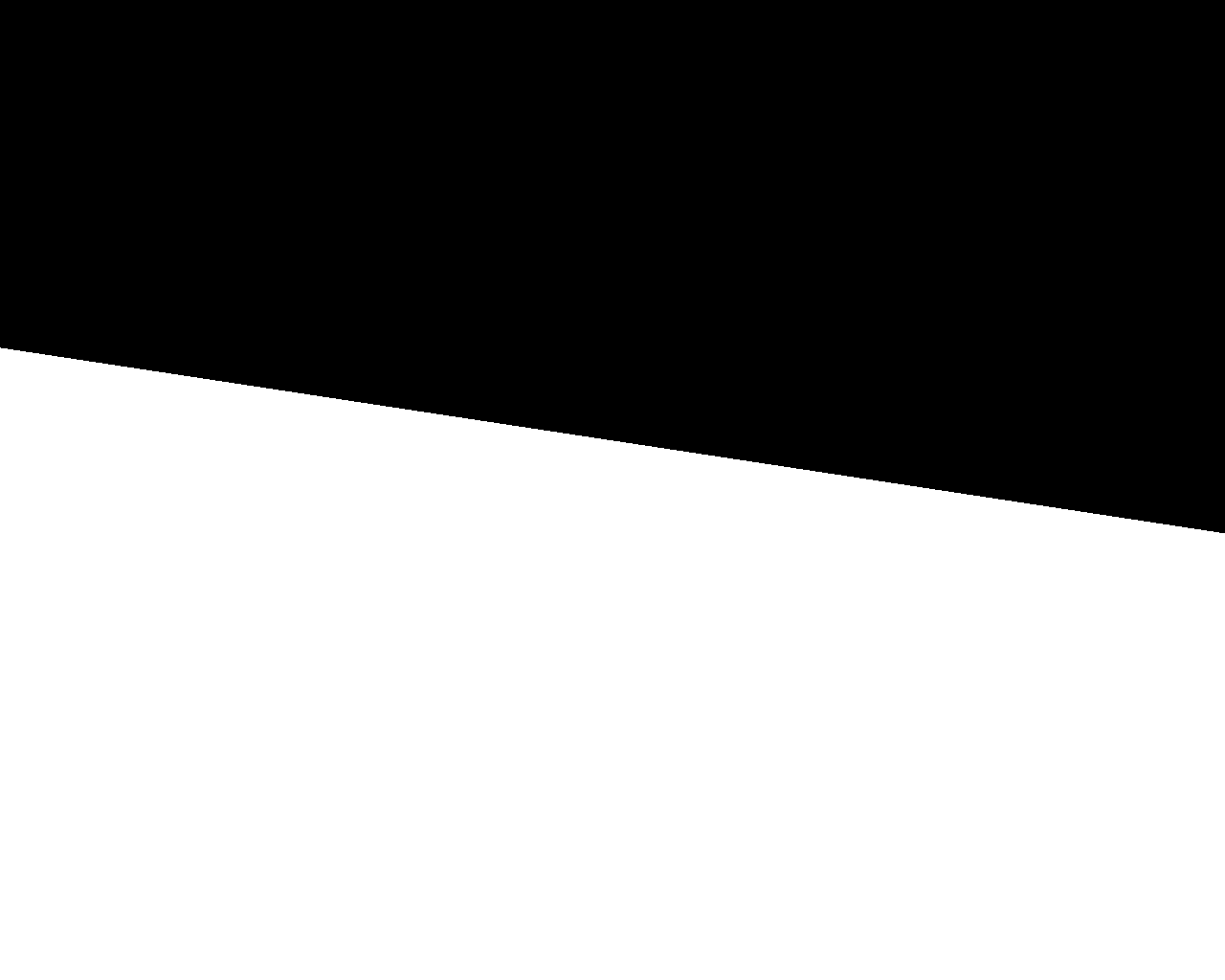}
            
            \includegraphics[width=1\linewidth]{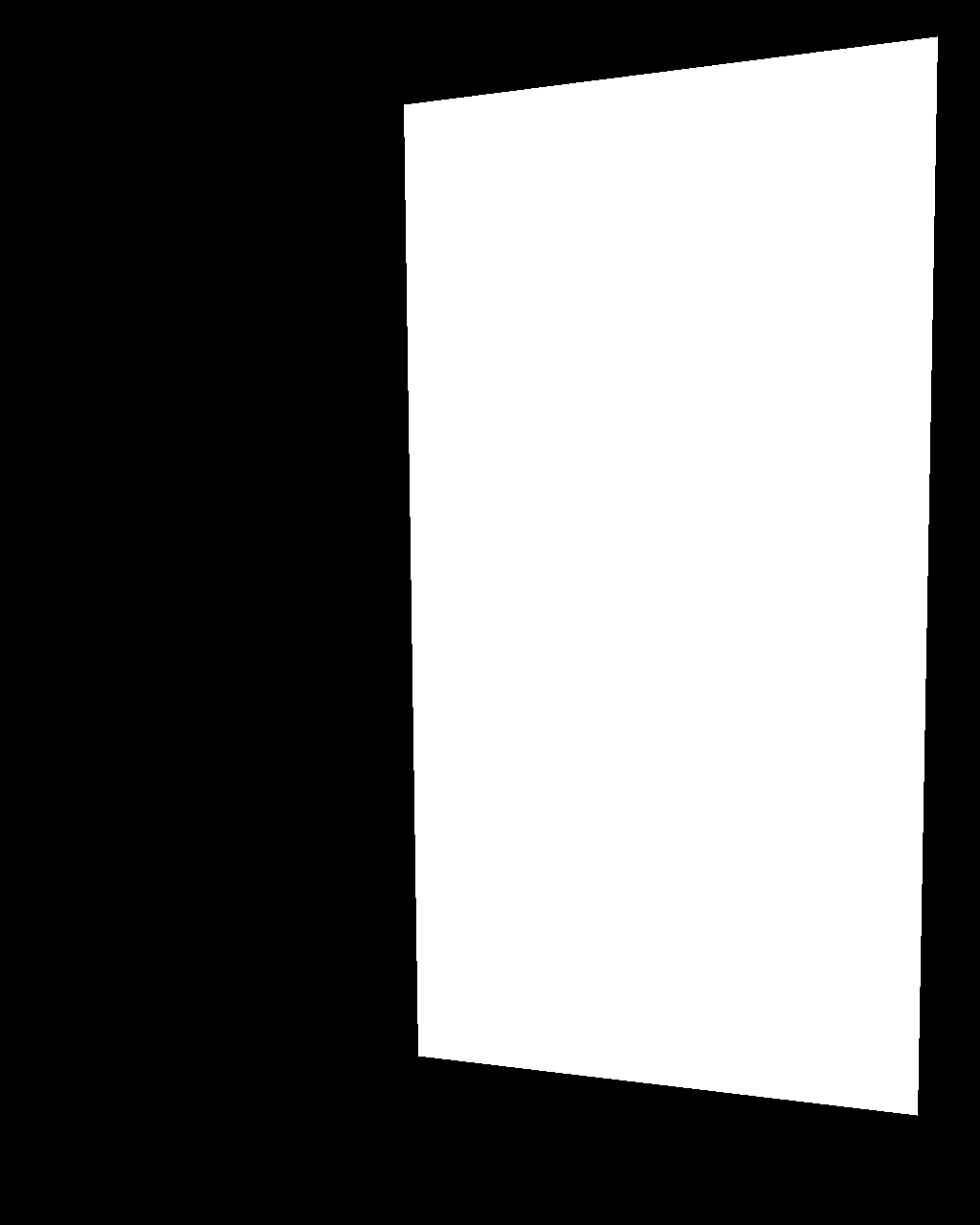}

            \includegraphics[width=1\linewidth]{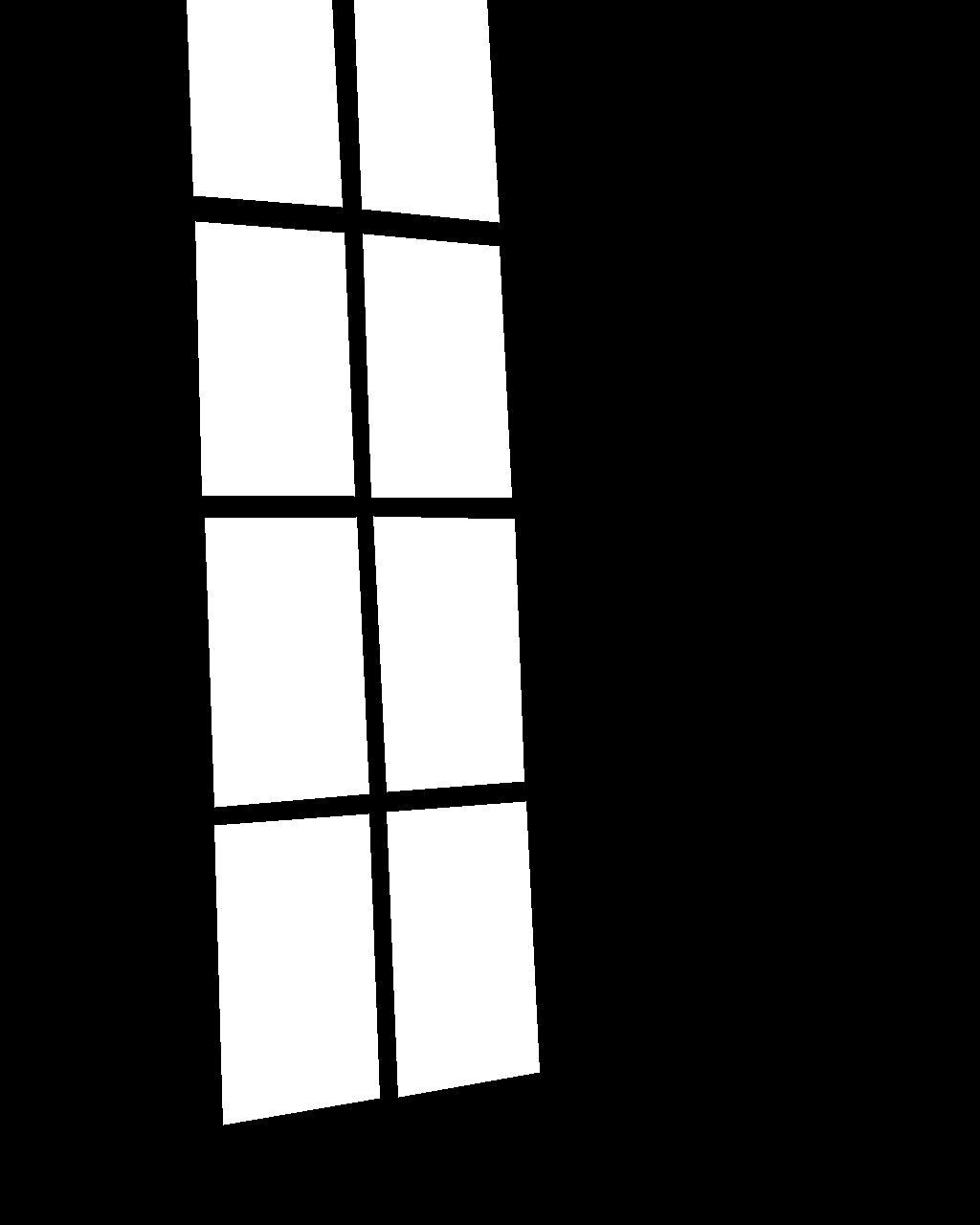}
      \end{minipage}
      } 
      \caption{\label{img7}
       Visual comparison of our GlassNet with SOTAs on the GDD testing set.}
\end{figure*}

\begin{figure*}[htb]
      \centering


      \subfloat[W/o interior and boundary]{\label{DA}
      \begin{minipage}[t]{0.12\textwidth}
            \centering
            \includegraphics[width=1\linewidth]{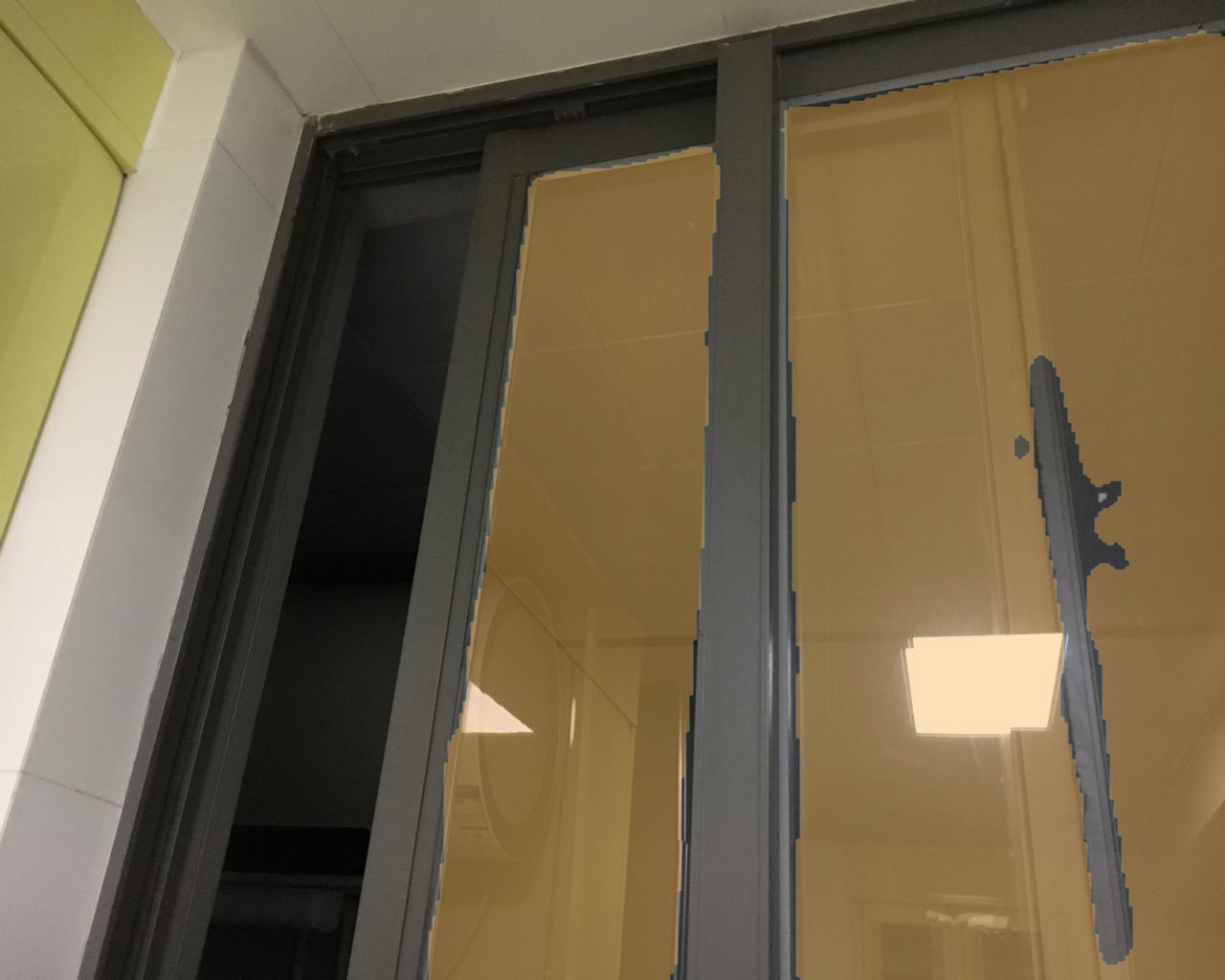}

            \includegraphics[width=1\linewidth]{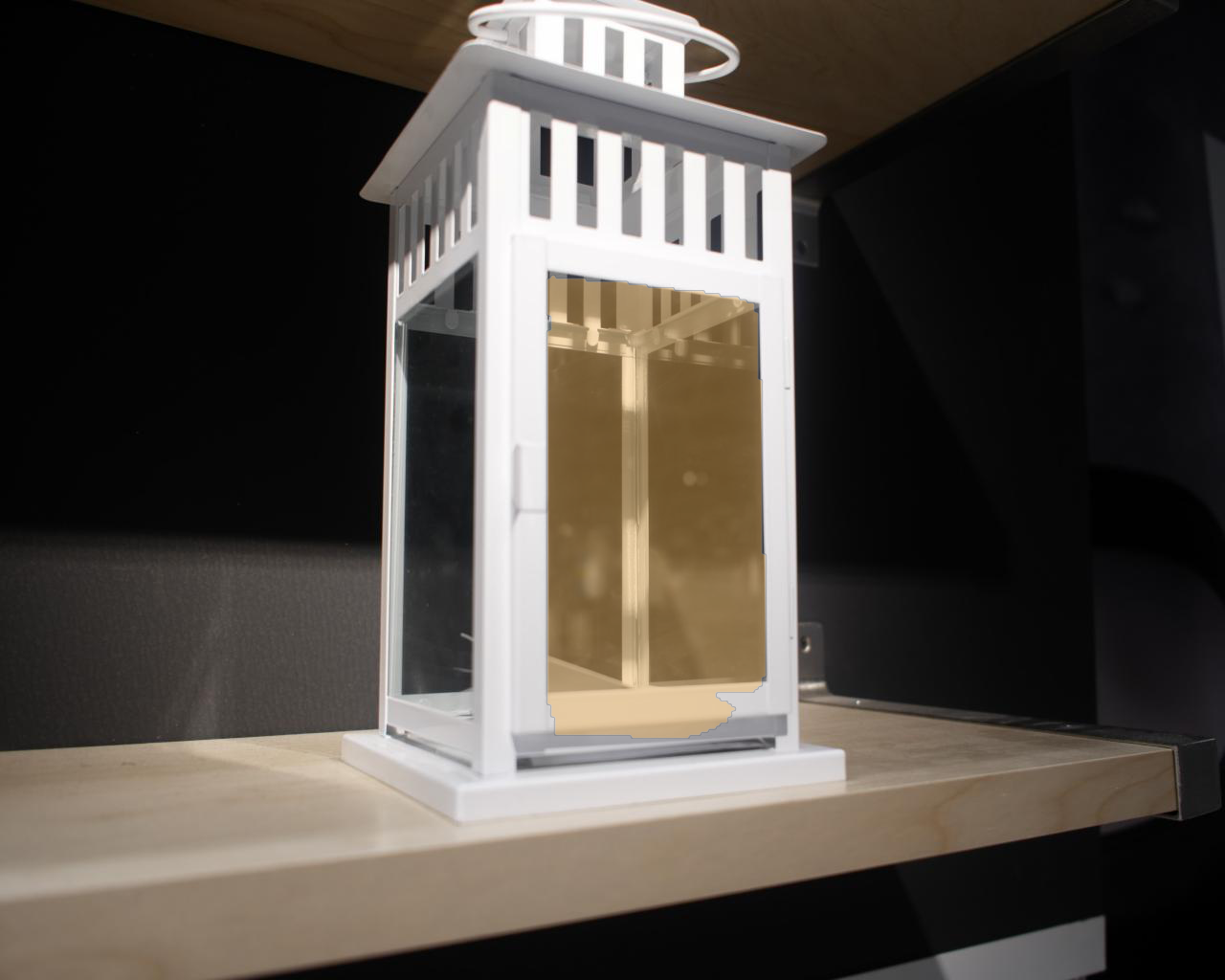}
      \end{minipage}
      }      
      \subfloat[\centering W/o boundary]{\label{CC}
      \begin{minipage}[t]{0.12\textwidth}
            \centering
            \includegraphics[width=1\linewidth]{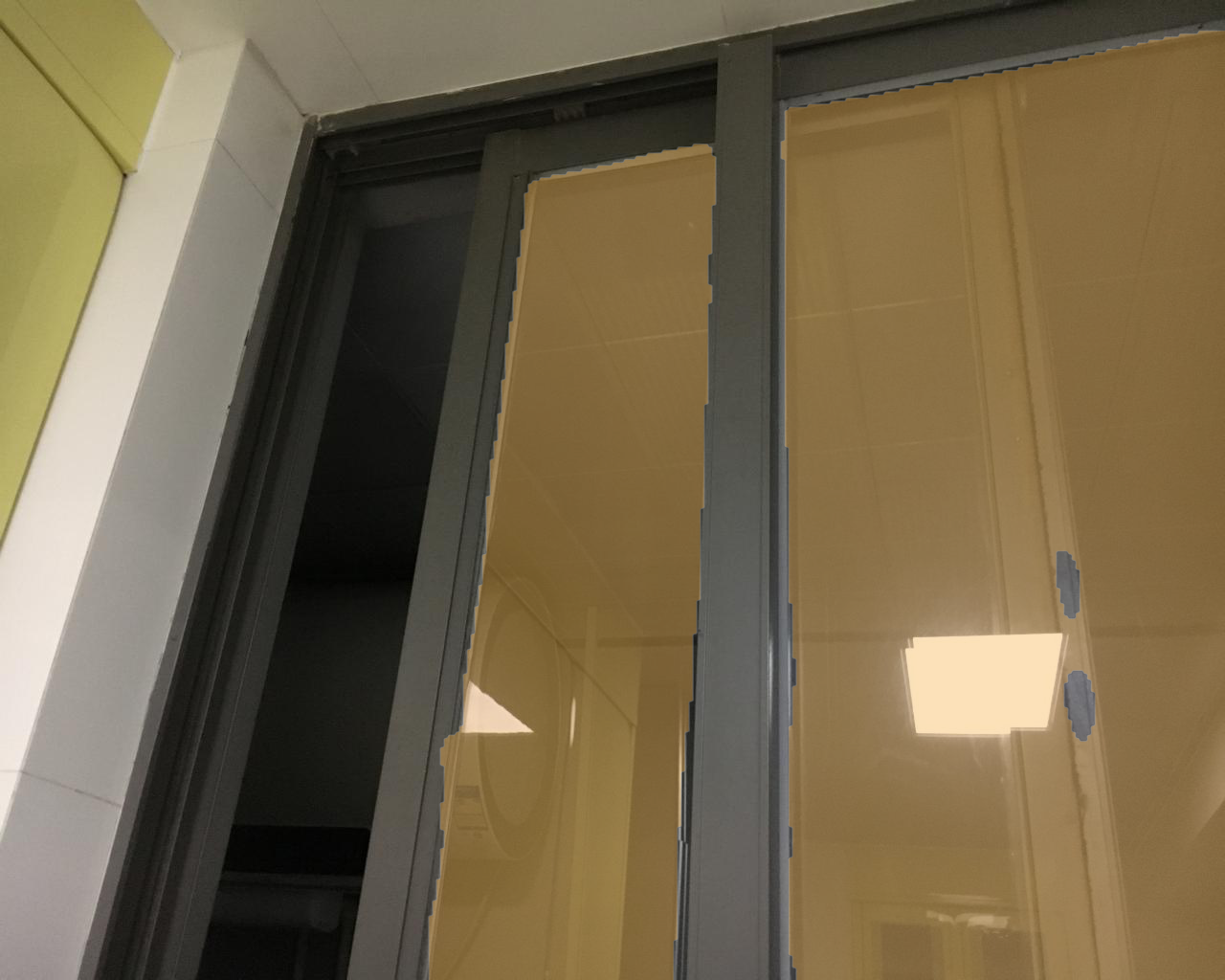}

            \includegraphics[width=1\linewidth]{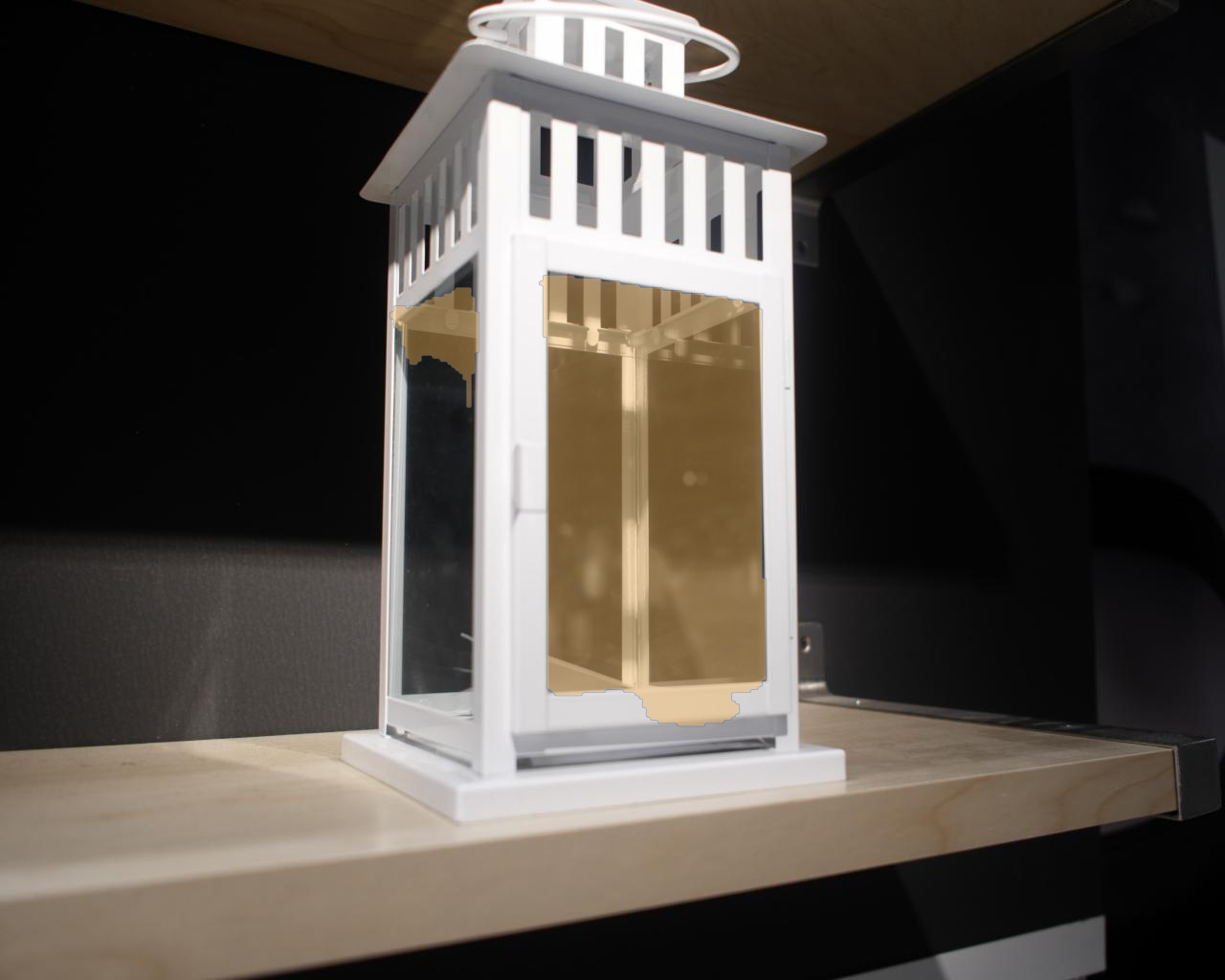}
      \end{minipage}
      }      
      \subfloat[\centering W/o interior]{\label{EG}
      \begin{minipage}[t]{0.12\textwidth}
            \centering
            \includegraphics[width=1\linewidth]{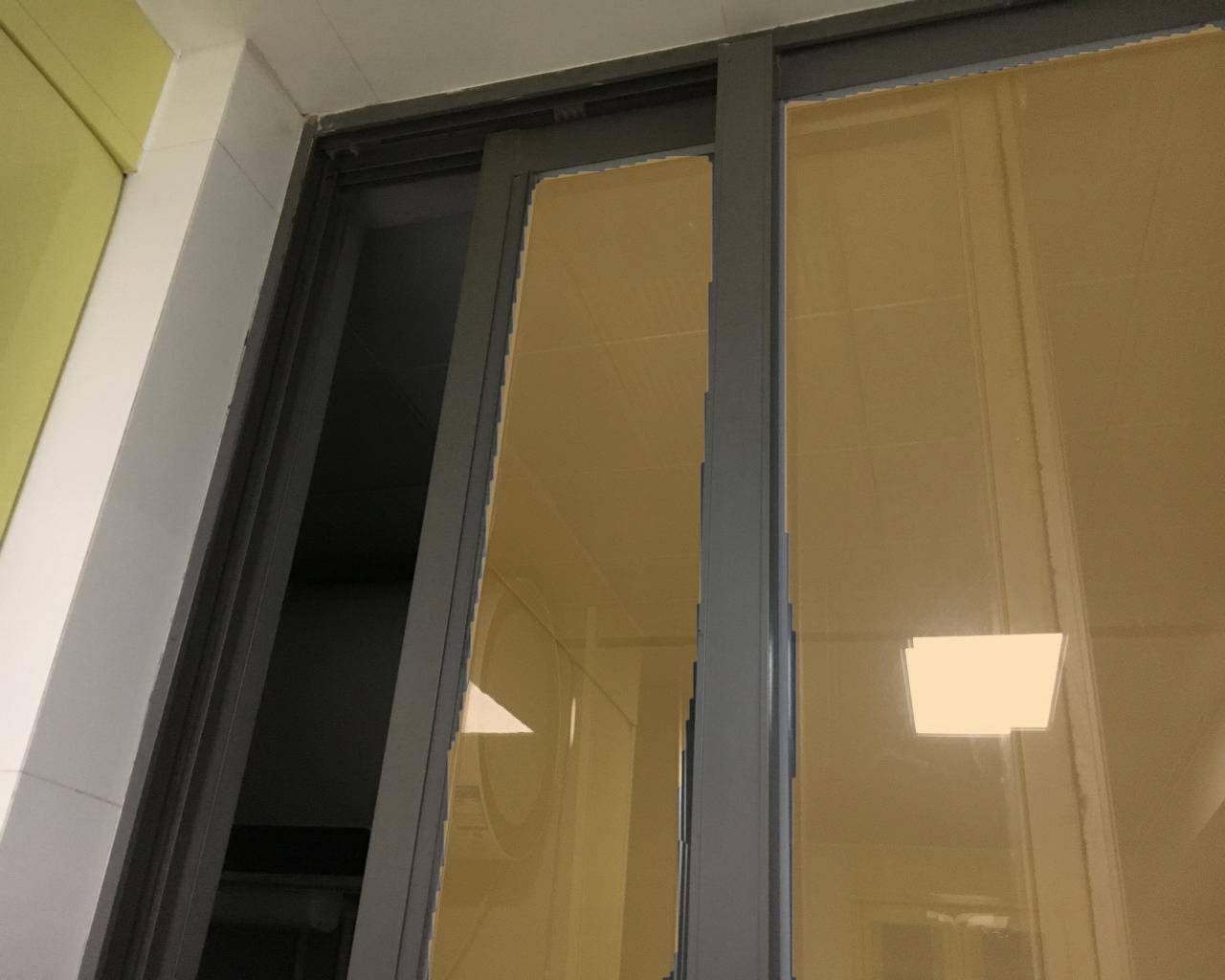}

            \includegraphics[width=1\linewidth]{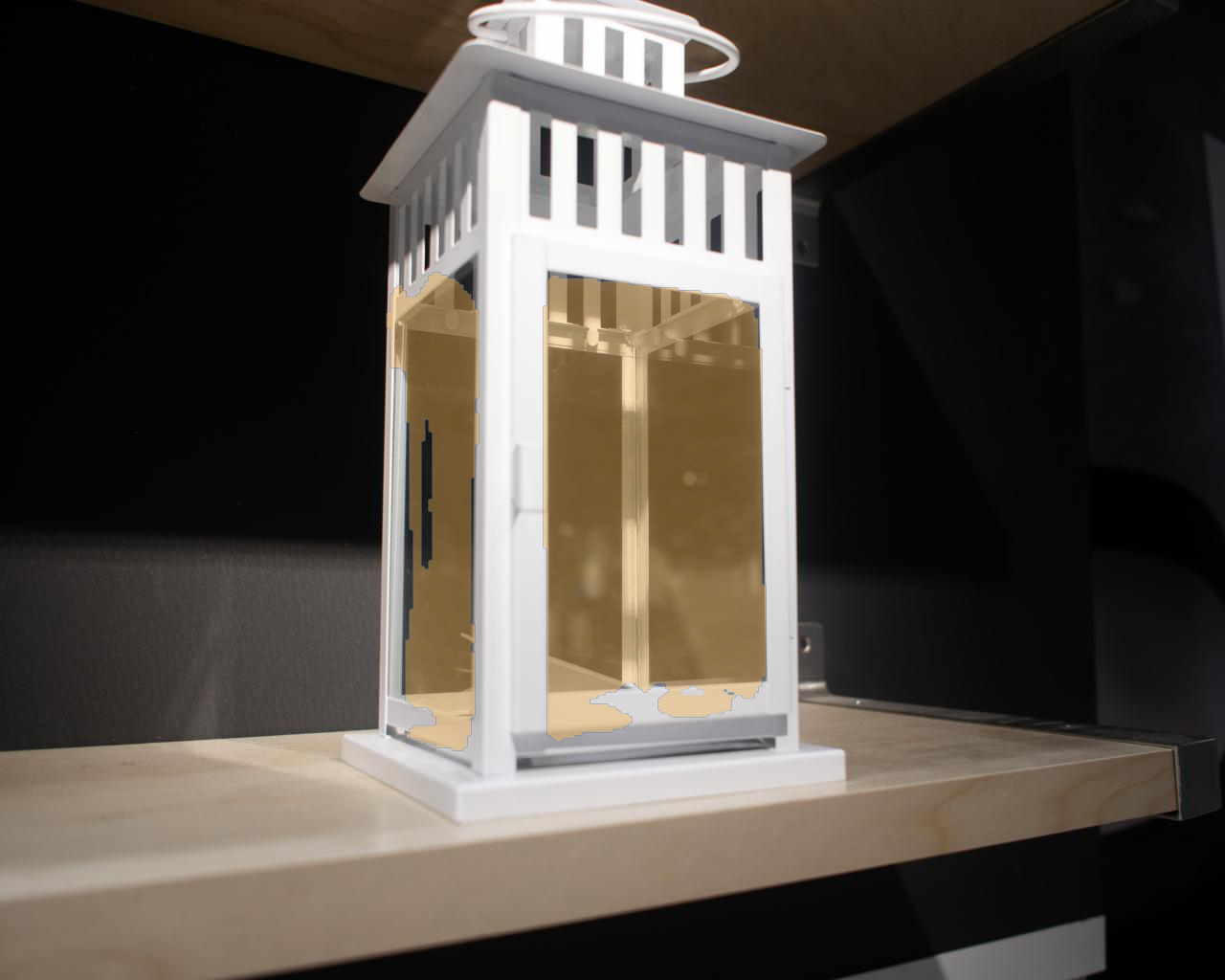}
      \end{minipage}
      }         
      \subfloat[\centering W/ only BCE loss]{\label{LDF}
      \begin{minipage}[t]{0.12\textwidth}
            \centering
            \includegraphics[width=1\linewidth]{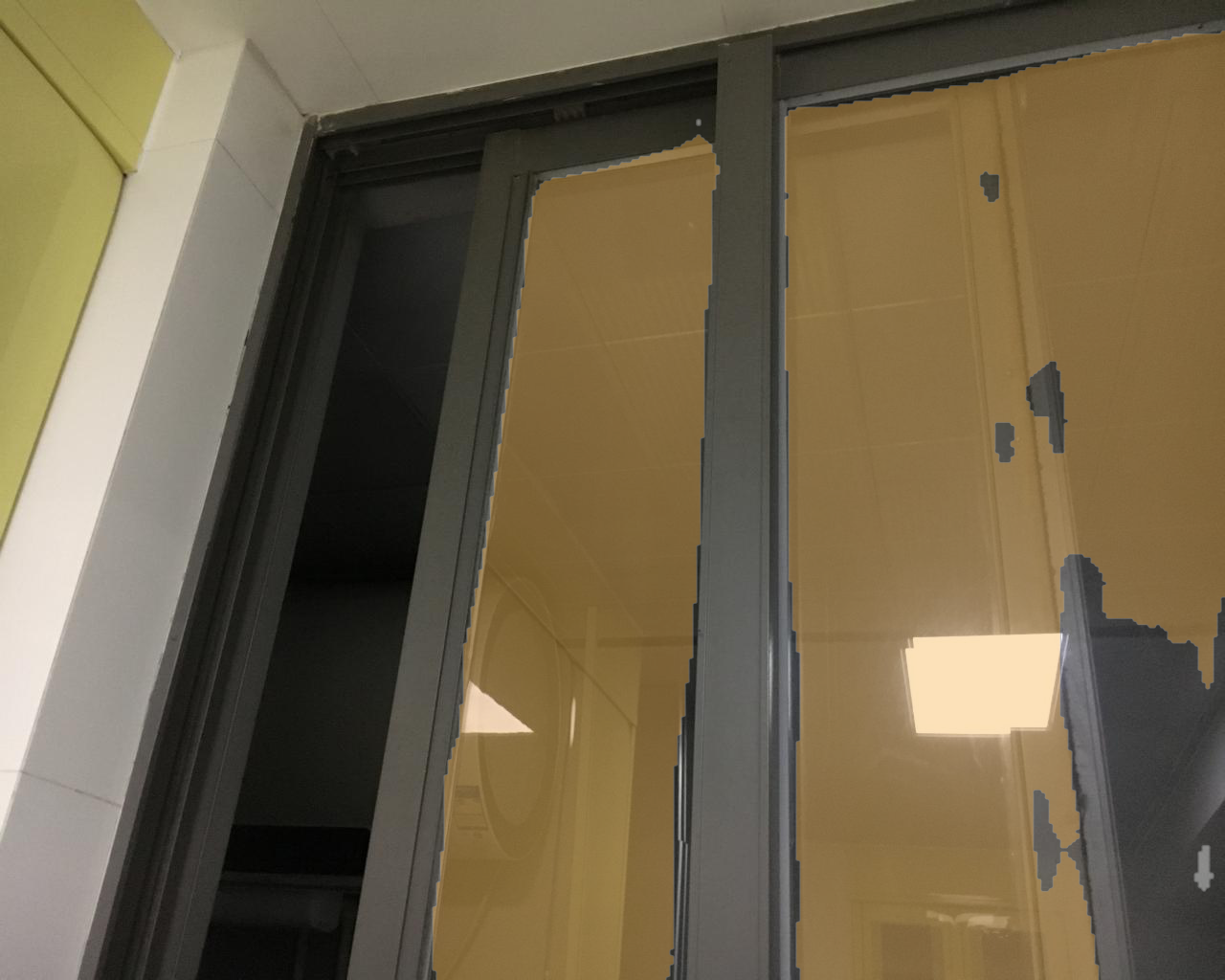}

            \includegraphics[width=1\linewidth]{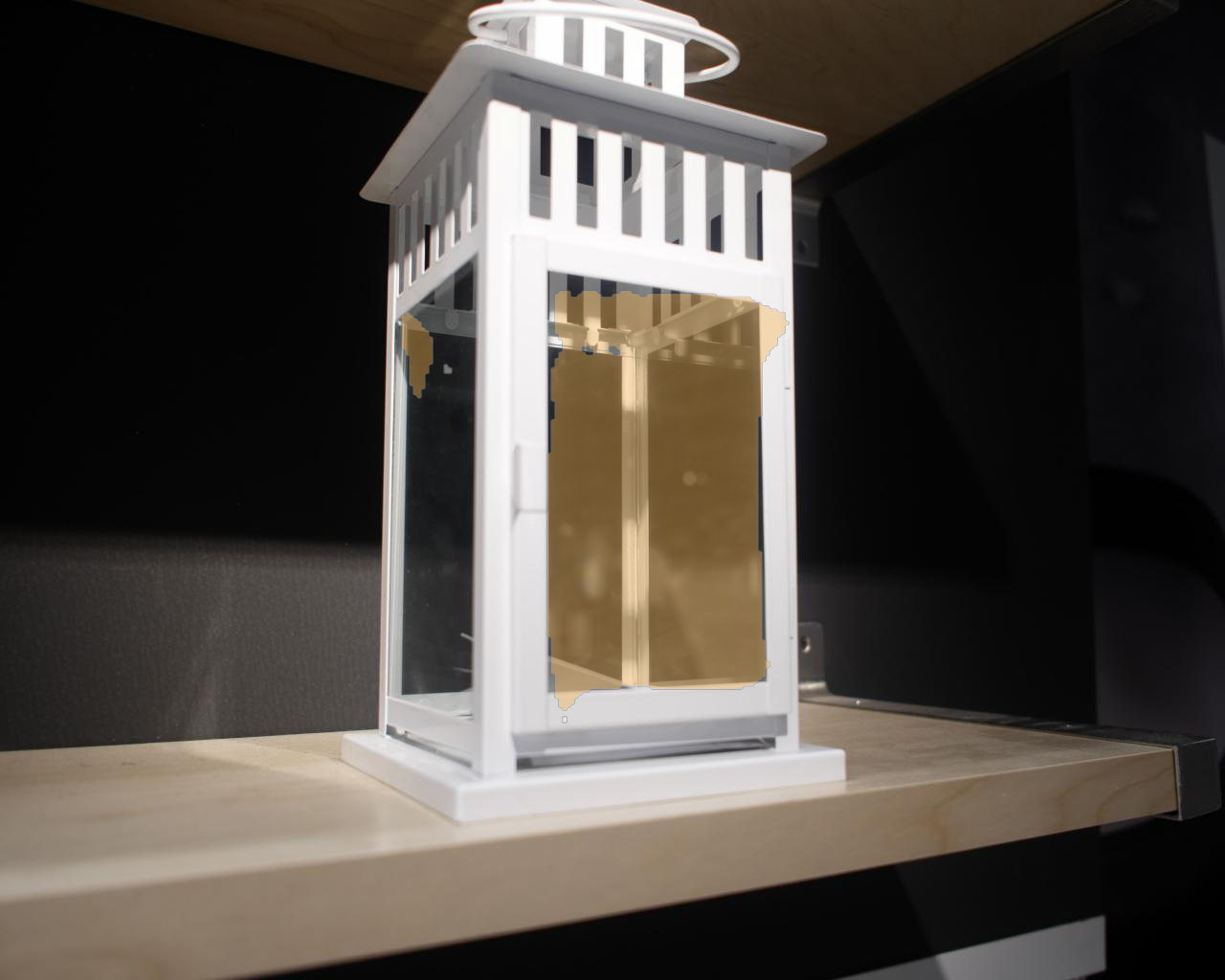}
      \end{minipage}
      }  
      \subfloat[\centering W/o IoU loss]{\label{BDRAR}
      \begin{minipage}[t]{0.12\textwidth}
            \centering
            \includegraphics[width=1\linewidth]{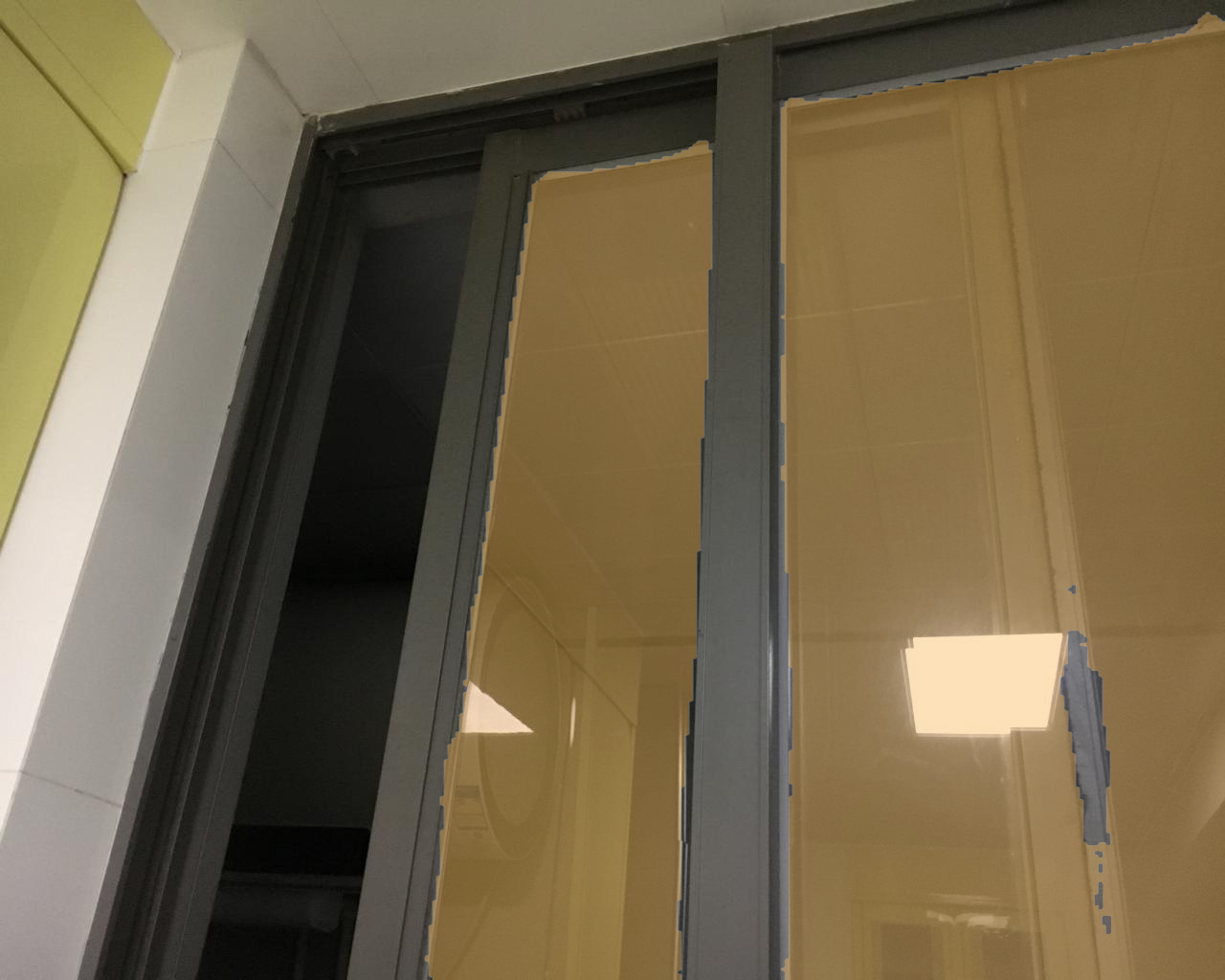}

            \includegraphics[width=1\linewidth]{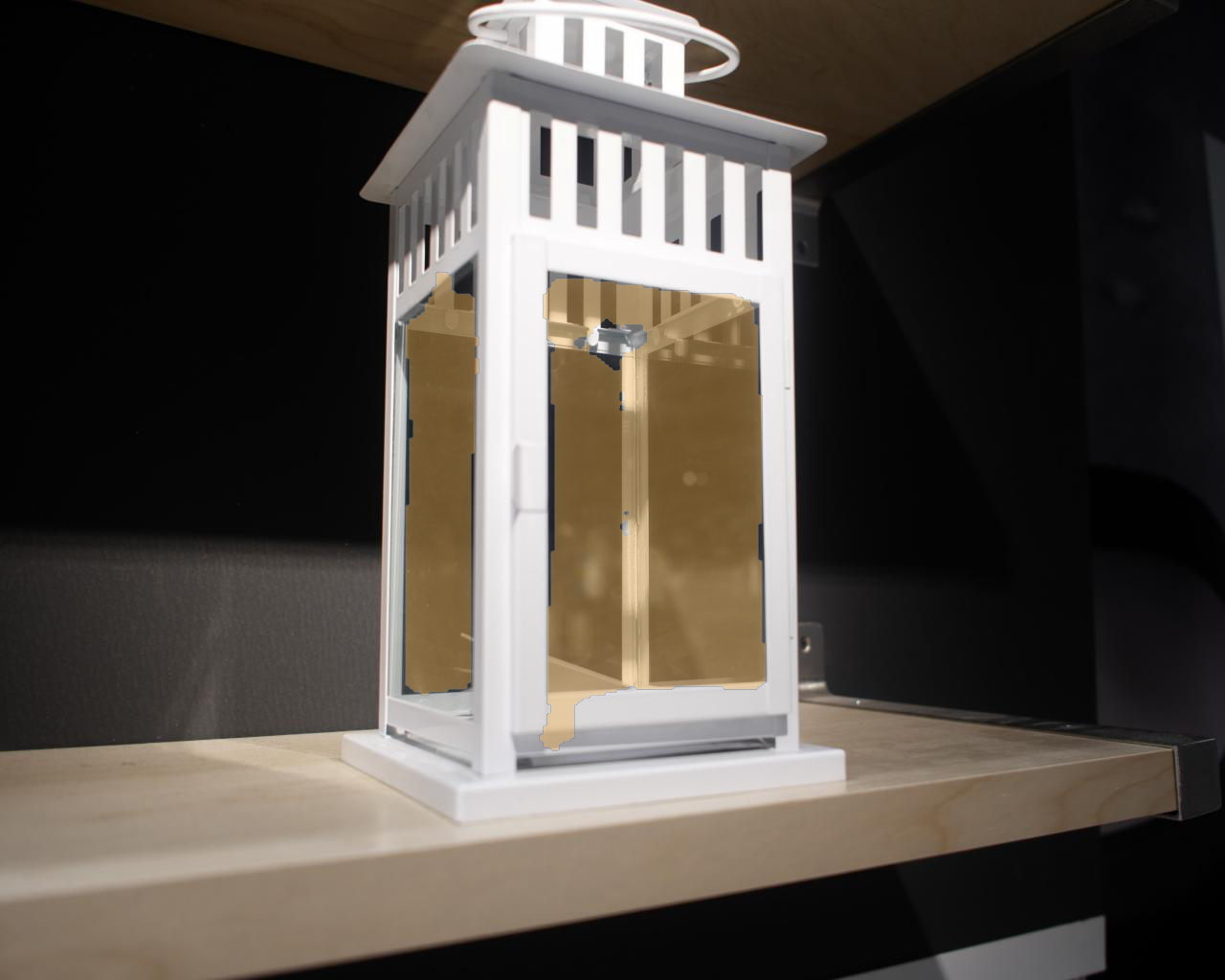}
      \end{minipage}
      }
      \subfloat[\centering W/o MID]{\label{MirrorNet}
      \begin{minipage}[t]{0.12\textwidth}
            \centering
            \includegraphics[width=1\linewidth]{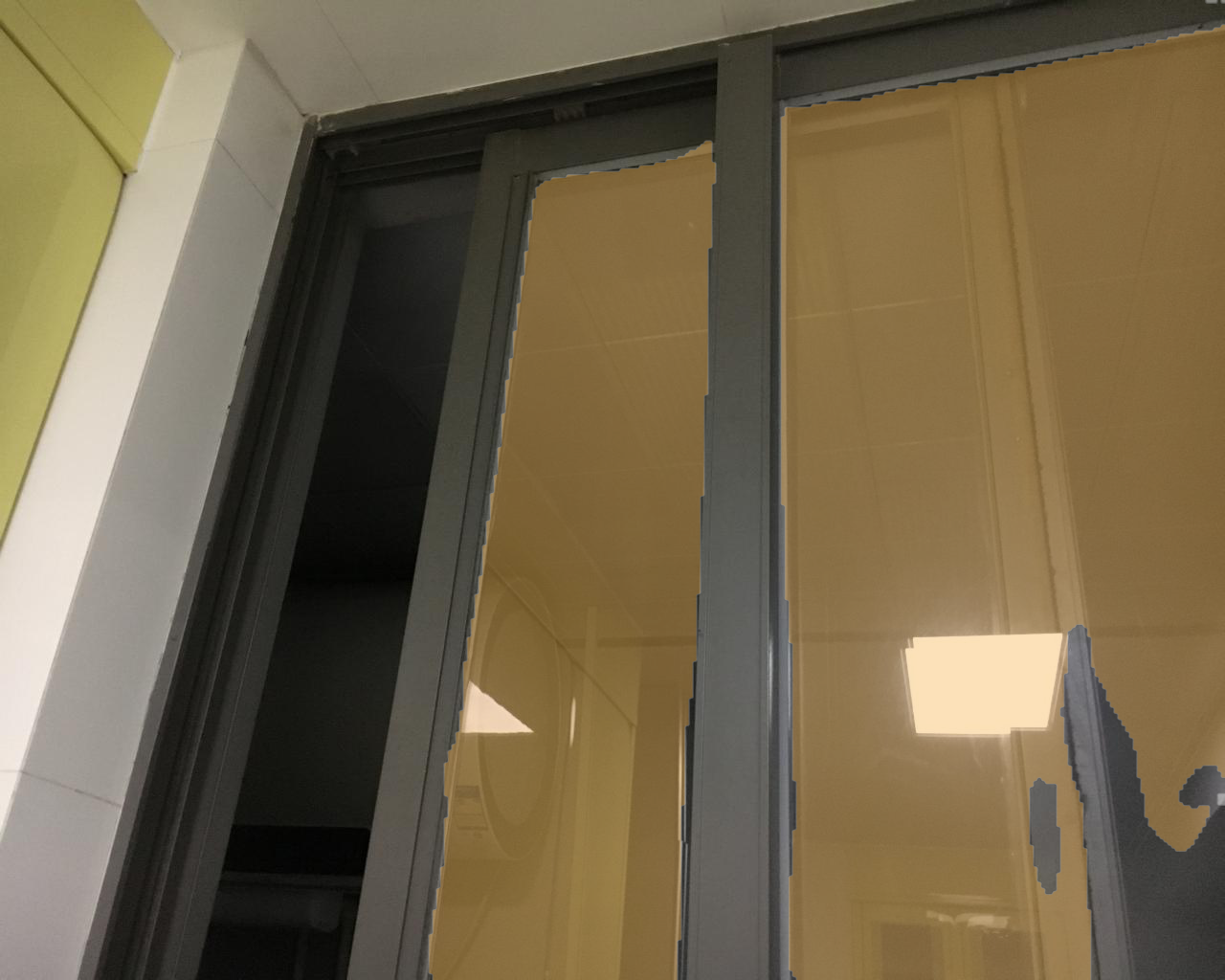}

            \includegraphics[width=1\linewidth]{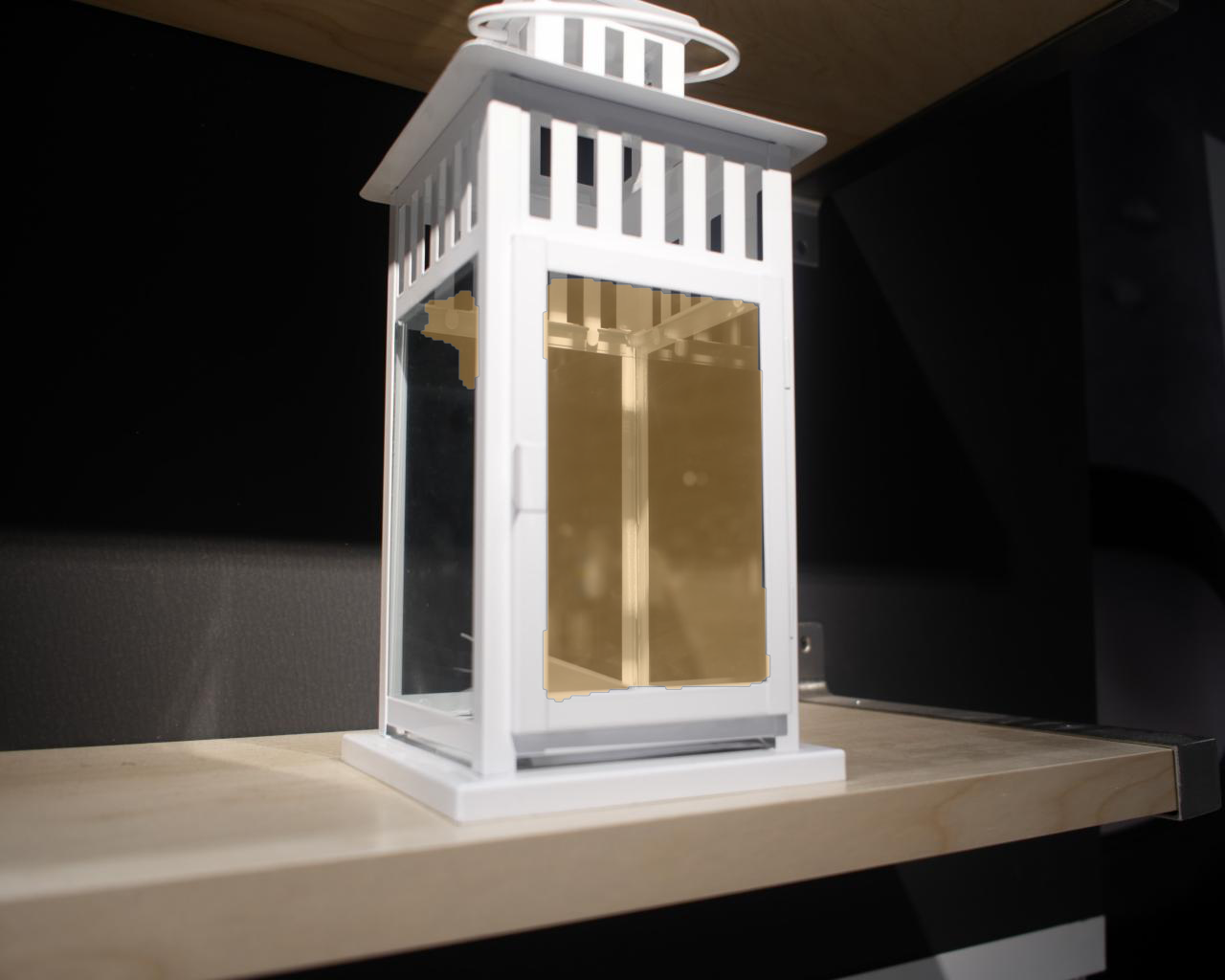}
      \end{minipage}
      }  
      \subfloat[\centering W/o BFM]{\label{PMD}
      \begin{minipage}[t]{0.12\textwidth}
            \centering
            \includegraphics[width=1\linewidth]{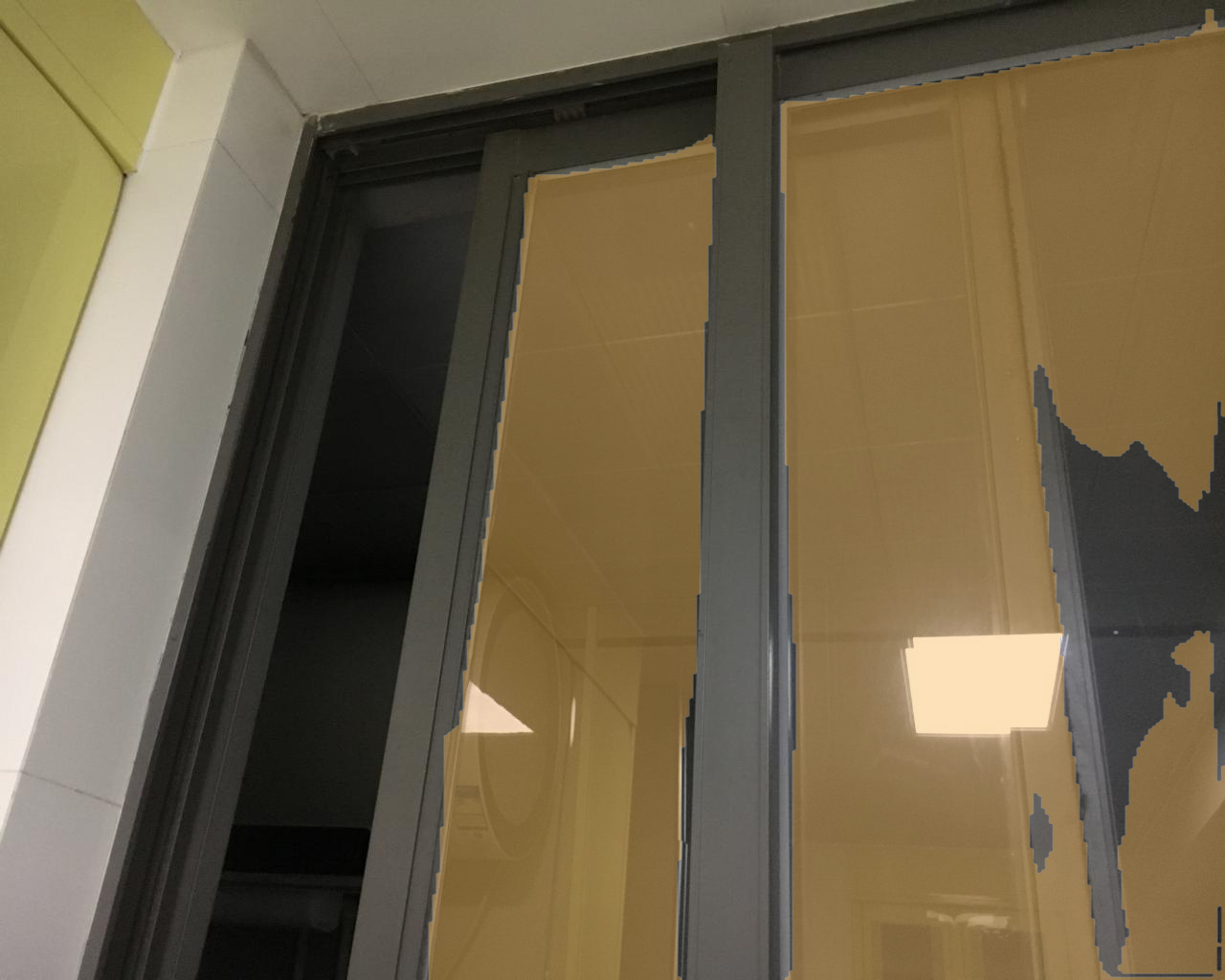}

            \includegraphics[width=1\linewidth]{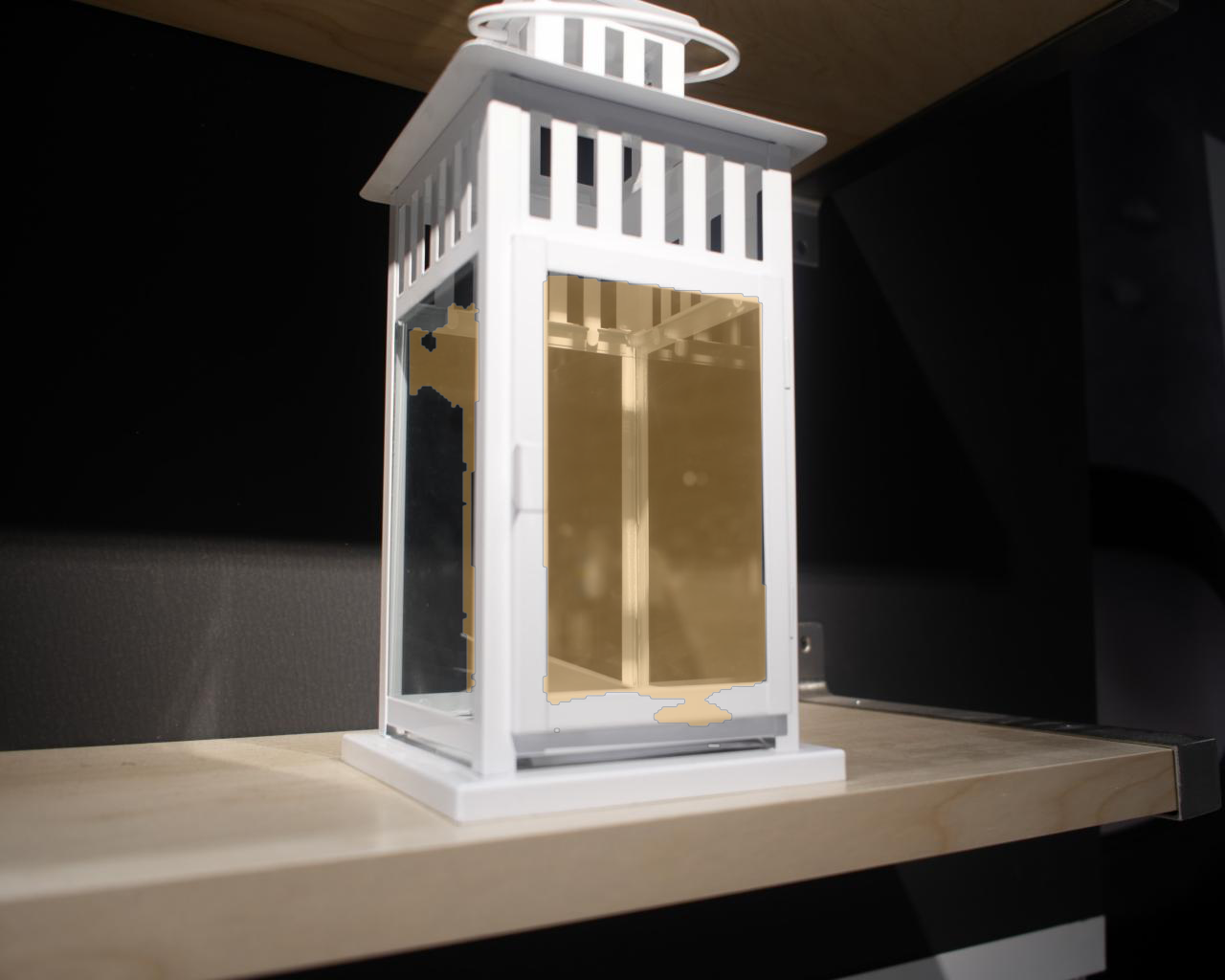}
      \end{minipage}
      }
      \subfloat[\centering Our full scheme]{\label{our}
      \begin{minipage}[t]{0.12\textwidth}
            \centering
            \includegraphics[width=1\linewidth]{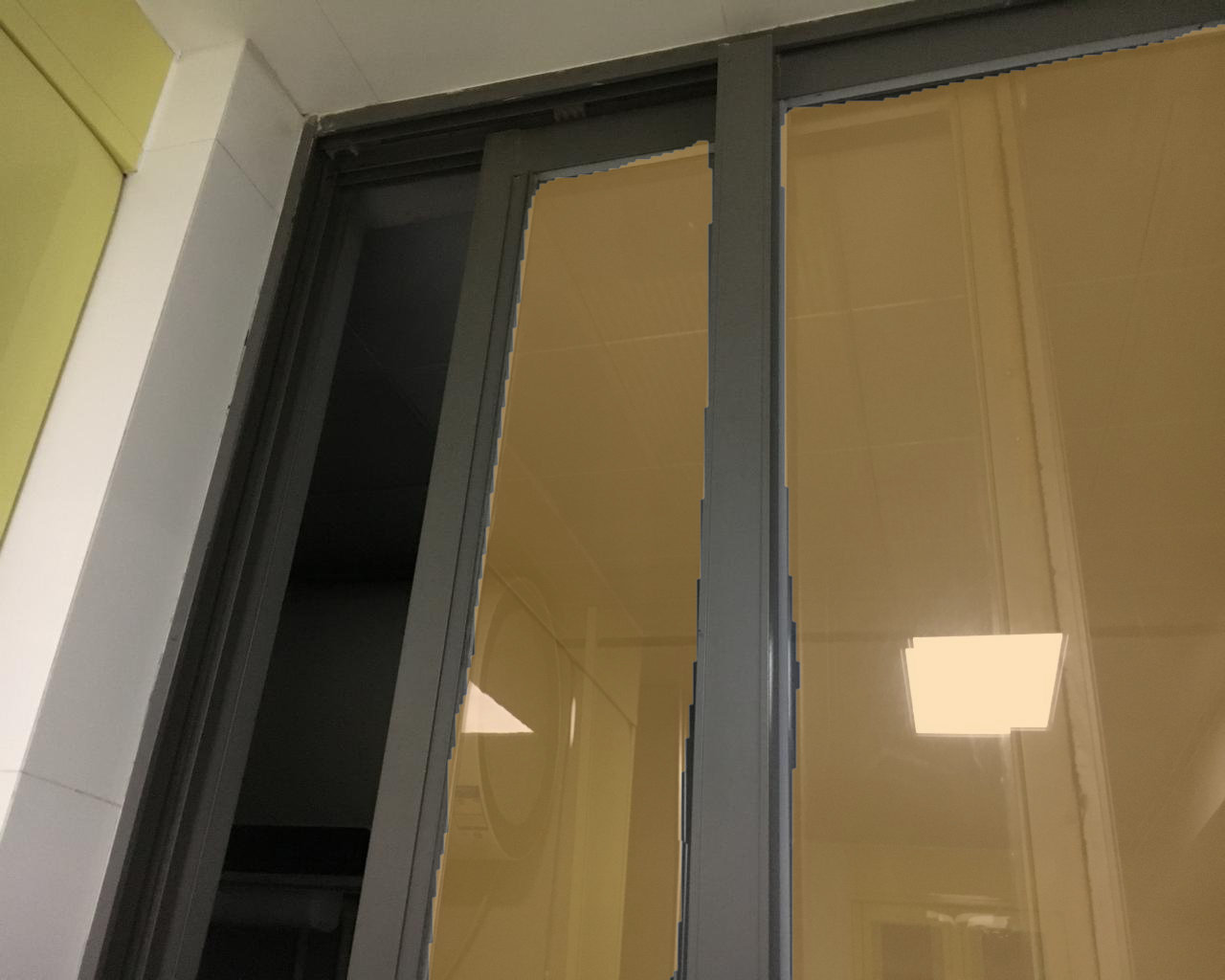}

            \includegraphics[width=1\linewidth]{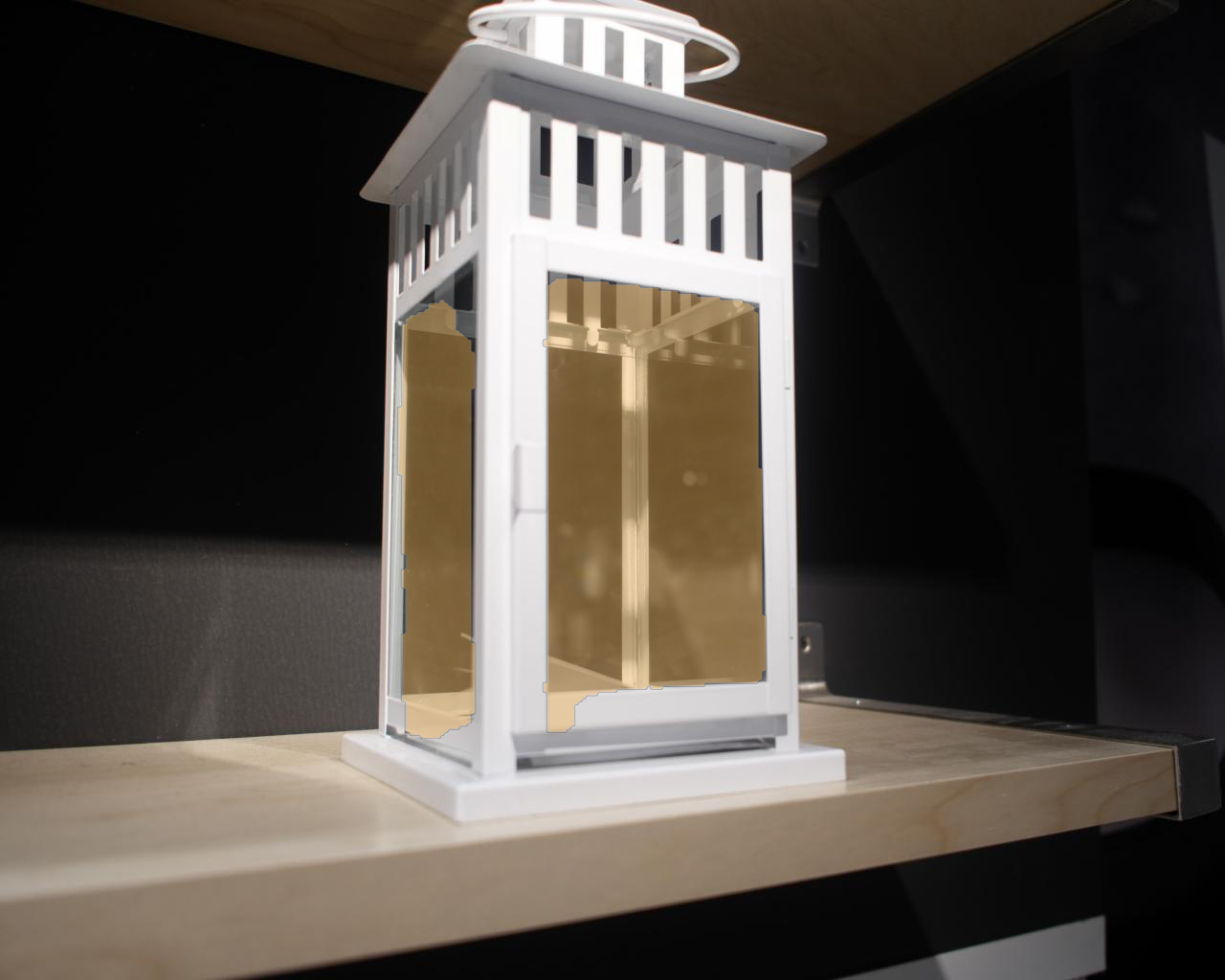}
      \end{minipage}
      }
      \caption{\label{img8}
      Visual comparison of our GlassNet with its variants.}
\end{figure*}

\subsection{Loss Function}

We use two different loss functions, i.e., the binary cross-entropy (BCE) loss $l_{bce}$, and the intersection over union (IoU) loss $l_{iou}$ \cite{basnet}, to supervise the network.
The BCE loss is a widely-used loss function in  computer vision because of its robustness:
\begin{equation}
l_{bce}=-\sum\limits_{(x,y)}{[g(x,y)log(p(x,y))+(1-g(x,y))log(1-p(x,y))]}
\end{equation}
IoU is an important metric to evaluate object detection quality by calculating the ratio of the intersection and union of the ``predicted box” and ``GT box”. Recently, it is widely used as the training loss:
\begin{equation}
l_{iou}=1-\frac{\sum\limits_{x=1}^{H}\sum\limits_{y=1}^{W}{p(x,y)g(x,y)}}{\sum\limits_{x=1}^{H}\sum\limits_{y=1}^{W}{[p(x,y)+g(x,y)-p(x,y)g(x,y)]}}  
\end{equation}

Therefore, the three different streams are supervised separately by combining different losses.
In the interior-diffusion stream, the glass stream, and the final output prediction maps, we adopt the BCE loss and the IoU loss, which can be formulated as:
\begin{equation}
L_{glass}= \sum\limits_{k=1}^{N_g}{[l_{bce}(p_k,g_{glass})+l_{iou}(p_k,g_{glass})]}
\end{equation}
\begin{equation}
L_{inner}= l_{bce}(p_k,g_{inner})+l_{iou}(p_k,g_{inner})
\end{equation}
\begin{equation}
L_{final}= l_{bce}(p_k,g_{glass})+l_{iou}(p_k,g_{glass})
\end{equation}
where $L_{glass}$ is the sum of the different levels of losses in the glass stream, and $L_{final}$ is the supervision loss of the final output of the whole network.
$p_k$ is the prediction map of the different branches in the three streams. $g_{glass}$ is the ground-truth label, and $g_{inner}$ is the interior-diffusion label decoupled by the original label. 
$N_i $ and $N_g$ are the numbers of branches in the interior-diffusion stream and the glass stream, respectively. Moreover, we only use the BCE loss in the boundary-diffusion stream:
\begin{equation}
L_{boundary}= \sum\limits_{k=1}^{N_b}{l_{bce}(p_k,g_{boundary})}
\end{equation}
where $g_{boundary}$ is the boundary-diffusion label decoupled by the original label and $N_b$ is the number of branches in the boundary-diffusion stream.

Therefore, the final loss function is formulated as:
\begin{equation} 
Loss = L_{inner}+L_{boundary}+L_{glass}+L_{final} 
\end{equation}

\section{Experiments}
\begin{table}[htb]
      \centering
      \begin{tabular}{l|c|c|c}
            \hline
            Strategy                              &   acc$\uparrow$           &   $F_\beta\uparrow$            &     BER$\downarrow$        \\
            \hline
            w/o interior and boundary   &  0.936   &  0.925  &  6.37   \\ 
            w/o boundary                  &  0.938   &  0.927  &  6.25  \\
            w/o interior                     &  0.941   &  0.932  &  5.98 \\
            \hline
            \hline
            w/ only BCE loss                      &  0.937   &  0.928 &  6.31  \\
            w/o IoU loss                          &  0.940   &  0.933 &  5.92  \\
            \hline
            \hline
            w/o MID                               &  0.943   &  0.932 &  5.53  \\
            w/o BFM                               &  0.940   &  0.923 &  5.80  \\
            \hline
            \hline
            Our GlassNet                                  &   \textbf{0.946}  &   \textbf{0.937}  &   \textbf{5.42}  \\
      \end{tabular}
      \caption{\label{table2}
      Ablation study results. Best results are highlighted in bold.}
\end{table}

\subsection{Datasets and Evaluation Metrics}
Currently, there is only a dataset available, i.e., GDD \cite{GDNet}, which is the first large-scale benchmark for glass detection and has 4,018 mirror images with their corresponding masks. 
We use five metrics widely used by other computer vision tasks to evaluate the performance of our model and existing state-of-the-art methods. 
First, we use two popular metrics, i.e., the pixel accuracy (acc) and the intersection of union (IoU). 
Besides, we apply the F-measure \cite{ftsrd} and mean absolute error (MAE) metrics from the salient object detection field, which are widely adopted in \cite{RAFnet, ckt, Picanet, dssc}. 
The F-measure is the weighted harmonic mean of precision and recall. 
We use the maximum F-measure ($F_\beta$) version as:   
\begin{equation} 
F_\beta = \frac{(1+\beta ^2)Precision\times Recall}{beta ^2Precision+Recall} 
\end{equation}
where $ \beta ^2$ is set to 0.3 as suggested in \cite{ftsrd}. MAE is the mean absolute error, i.e., the mean value of the absolute error between the prediction and the ground truth, which is defined as:
\begin{equation} 
MAE = \frac{1}{H\times W} \sum\limits_{i=1}^{H}\sum\limits_{j=1}^{W}{|p(i,j)-g(i,j) \vert} 
\end{equation}
where $ g(x,y)\in [0,1]$ is the ground-truth label of the pixel $(x, y)$ and $p(x, y) \in [0, 1]$ is the predicted probability of being glass.
In addition, we select the balance error rate (BER) \cite{lko} from the shadow detection field as our last metric, which can be obtained as:
\begin{equation}
BER = 100\times(1-\frac{1}{2}(\frac{TP}{N_p}+\frac{TN}{N_n}))
\end{equation}
where $TP$, $TN $, $N_n $ and $N_p $ is the numbers of true positives, true negatives, glass pixels and non-glass pixels, respectively.

\subsection{Implementation Details}
We implement the proposed network GlassNet based on the PyTorch framework \cite{pytorch} and train it on the benchmark dataset GDD. 
The pre-trained ResNet-50 network \cite{ResNet} on ImageNet \cite{Imagenet} is used to initialize the parameters of the backbone, and the other parameters are initialized randomly. 
We train the whole network by using the stochastic gradient descent (SGD) with a momentum of 0.9 and the weight decay of $ 5\times 10^{-4}$. 
The initial learning rate is set to 0.0001 and is adjusted by poly decay strategies \cite{Bisenet} with a power of 0.9. 
The network with a batch setting of 4 is trained on an NVIDIA GTX 1080 Ti graphics card. 
During testing, images are adjusted to the resolution of $512 \times 512$ for inference without any post-processing.

\subsection{Comparison with the SOTAs}
\textbf{Compared methods.}
It only has one deep learning-based method for glass detection from single images. Thus, we compare to this method and other 14 state-of-the-art methods, which are PSPNet \cite{PSP}, DenseASPP \cite{DenseASPP}, PSANet \cite{psanet}, DANet \cite{DANet} and CCNet \cite{CCNet} chosen from the semantic segmentation field, R$^3$Net \cite{r3net}, CPD \cite{cpd}, BASNet \cite{basnet}, EGNet \cite{egnet} and LDF \cite{LDF} chosen from the salient object detection field, DSC \cite{DSC} and BDRAR \cite{BDRAR} chosen from the shadow detection field, MirrorNet \cite{MirrorNet} and PMD \cite{PMD} from the mirror segmentation field, and GDNet \cite{GDNet} used for glass detection. For a fair comparison, we retrain all the other methods on the GDD dataset by using their publicly available codes.

\textbf{Quantitative Comparison.}
We compare the proposed network with state-of-the-art methods from the relevant fields mentioned above, which are shown in Table \ref{table1}. The first, second, and third best results are marked in bold, red, and blue, respectively. Obviously, compared with other methods in related fields, our method is better than the SOTA methods. 

\textbf{Qualitative Evaluation.}
Some prediction examples of the proposed method and state-of-the-art approaches have been shown in Figure \ref{img7}. We observe that the proposed method not only highlights the glass regions clearly but also well suppresses the background noise. It can be seen that our method can accurately detect small glass (e.g., the first four rows), large glass (e.g., the fifth to seventh rows), and others (e.g., the eighth and ninth rows). Although GDNet can locate these regions well, it has low detection accuracy for the boundary regions and cannot even detect the boundary regions correctly. In contrast, our method has higher detection accuracy in the boundary region because we use boundary information to force the network to pay more attention to the boundary region.
\begin{figure}[htb]
      \centering
      \subfloat[Input]{\label{image}
      \begin{minipage}[t]{0.22\textwidth}
            \centering
            \includegraphics[width=1\linewidth]{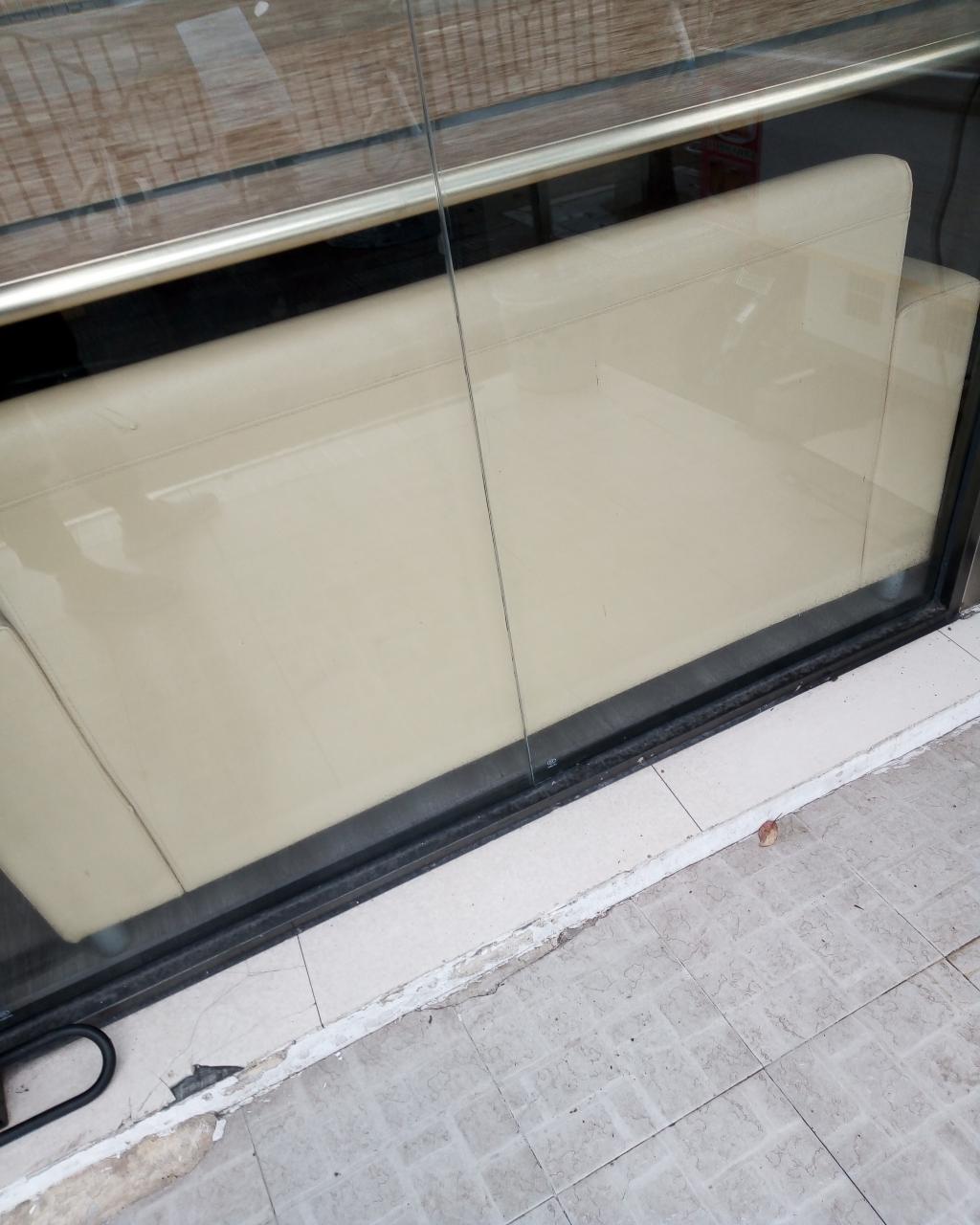}

            \includegraphics[width=1\linewidth]{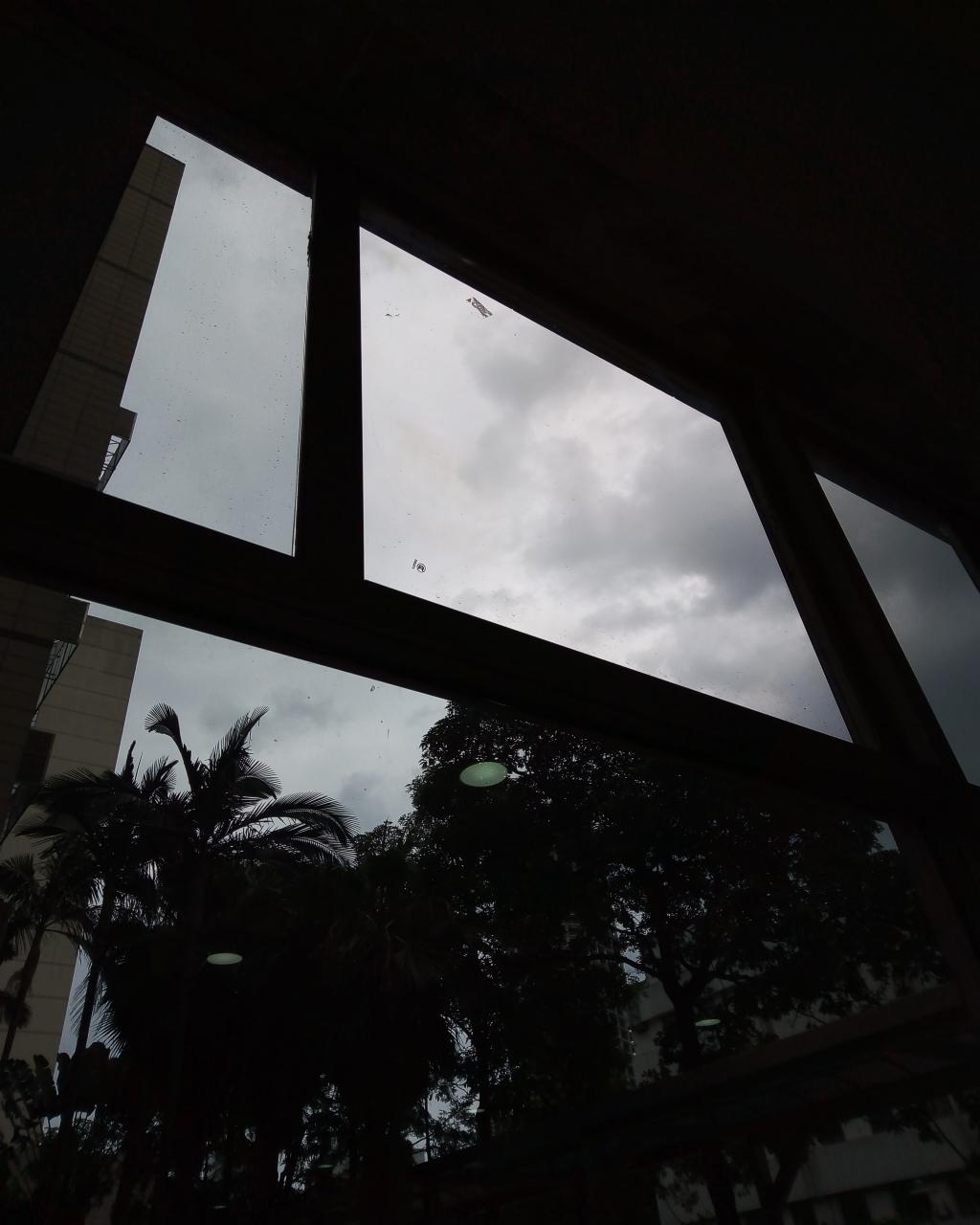}
      \end{minipage}
      }
      \subfloat[Poor detection]{\label{result}
      \begin{minipage}[t]{0.22\textwidth}
            \centering
            \includegraphics[width=1\linewidth]{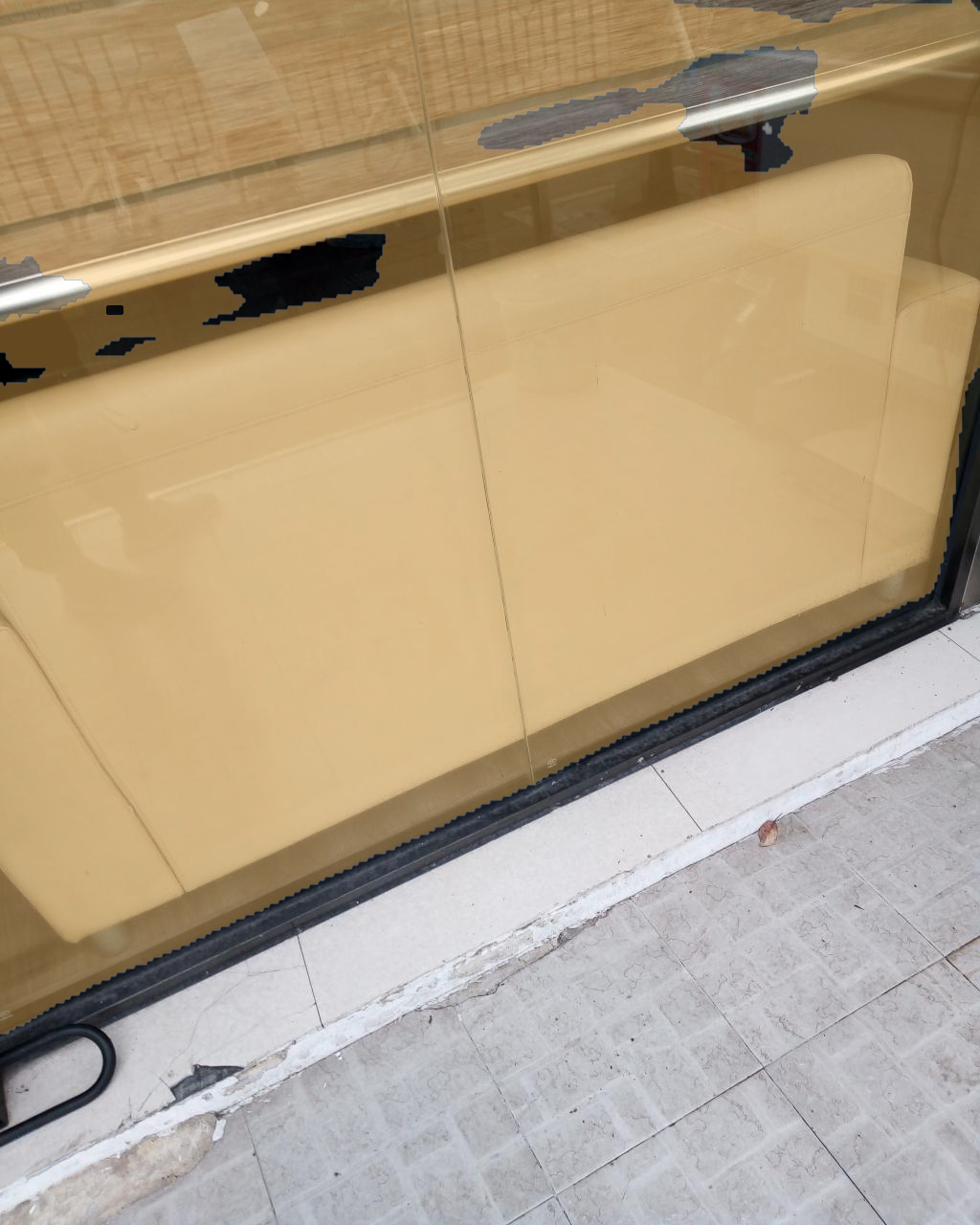}

            \includegraphics[width=1\linewidth]{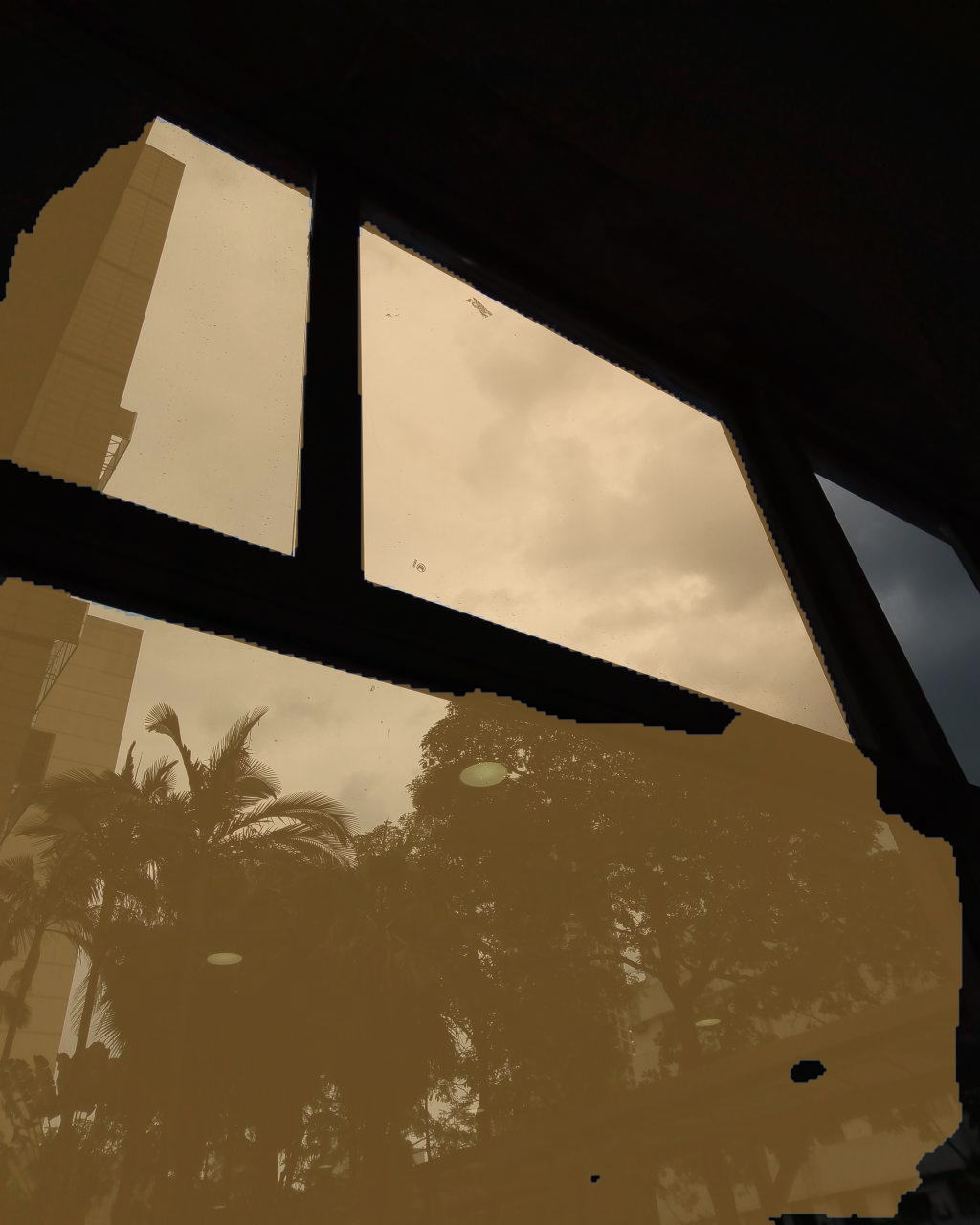}
      \end{minipage}
      }      
      \caption{\label{img9}
      Failure cases.}
\end{figure}
\subsection{Ablation Studies}
Table \ref{table2} demonstrates the effectiveness of each component in our model. From the first line to the third line, we can see that both boundary- and interior-diffusion branches can effectively improve the performance. Moreover, the effect of the network without the boundary-diffusion stream is worse than that without the interior-diffusion stream, which is consistent with our observation: Boundaries significantly improve the detection ability, which should be specially considered. 
In addition, the final detection accuracy can also be improved by the proposed multiple mixing losses, as shown in the 4th and 5th lines, where each row omits a loss, i.e., BCE and IOU, respectively.  
Finally, w/o MID and w/o BFM, respectively, indicate that we do not use any one of the two modules each time in our network, which shows their contributions on improving the glass detection quality. Figure \ref{img8} shows a visual example, proving that our method successfully addresses the glass detection problem with the help of boundaries.

\subsection{Failure cases}
GlassNet has two limitations, as shown in Figure \ref{img9}: 1) in the case of very large-scale glass, e.g., the area of the glass occupies more than 95\% or even 100\% of the whole image, it may operate poorly on such extreme cases, due to the lack of sufficient contextual information; and 2) it is nearly impossible to detect the glass in the very weak light, since under the very weak-light condition, the boundary area of the glass and the background will share very similar properties, i.e., they are all black regions with pixel values approximating to (0,0,0).

\section{Conclusion}
In this paper, we propose a three-stream network for glass detection from single images, called GlassNet. 
GlassNet consists of a label decoupling procedure that decouples the ground truth into an interior-diffusion label and a boundary-diffusion label, a multi-scale interactive dilation module for extracting and capturing contextual features, and a three-stream network integrating multi-scale and multi-modal information to generate the final prediction map. 
Besides, GlassNet utilizes an attention-based boundary-aware feature Mosaic module to integrate multi-modal information for further improving the glass detection quality. 
Experiments on the benchmark datasets demonstrate that our GlassNet outperforms the state-of-the-art methods under different evaluation metrics.

\section*{Acknowledges}
This work was supported in part by the National Natural Science Foundation of China (No. 62032011, No. 62172218), in part by the Joint Fund of National Natural Science Foundation of China and Civil Aviation Administration of China (No. U2033202), in part by the Free Exploration of Basic Research Project, Local Science and Technology Development Fund Guided by the Central Government of China (No. 2021Szvup060), and in part by the Key Program of Jiangsu Provincial Department of Culture and Tourism (No. 20ZD06).

%
\printbibliography                

@inproceedings{egnet,
  title={EGNet: Edge guidance network for salient object detection},
  author={Zhao, Jia-Xing and Liu, Jiang-Jiang and Fan, Deng-Ping and Cao, Yang and Yang, Jufeng and Cheng, Ming-Ming},
  booktitle={Proceedings of the IEEE/CVF International Conference on Computer Vision},
  pages={8779--8788},
  year={2019}
  }

@inproceedings{basnet,
  title={Basnet: Boundary-aware salient object detection},
  author={Qin, Xuebin and Zhang, Zichen and Huang, Chenyang and Gao, Chao and Dehghan, Masood and Jagersand, Martin},
  booktitle={Proceedings of the IEEE/CVF Conference on Computer Vision and Pattern Recognition},
  pages={7479--7489},
  year={2019}
  }

@inproceedings{pyramid,
  title={Pyramid feature attention network for saliency detection},
  author={Zhao, Ting and Wu, Xiangqian},
  booktitle={Proceedings of the IEEE/CVF Conference on Computer Vision and Pattern Recognition},
  pages={3085--3094},
  year={2019}
  }

@inproceedings{ITSD,
  title={Interactive two-stream decoder for accurate and fast saliency detection},
  author={Zhou, Huajun and Xie, Xiaohua and Lai, Jian-Huang and Chen, Zixuan and Yang, Lingxiao},
  booktitle={Proceedings of the IEEE/CVF Conference on Computer Vision and Pattern Recognition},
  pages={9141--9150},
  year={2020}
  }

@inproceedings{LDF,
  title={Label decoupling framework for salient object detection},
  author={Wei, Jun and Wang, Shuhui and Wu, Zhe and Su, Chi and Huang, Qingming and Tian, Qi},
  booktitle={Proceedings of the IEEE/CVF Conference on Computer Vision and Pattern Recognition},
  pages={13025--13034},
  year={2020}
}

@inproceedings{MINet,
  title={Multi-scale interactive network for salient object detection},
  author={Pang, Youwei and Zhao, Xiaoqi and Zhang, Lihe and Lu, Huchuan},
  booktitle={Proceedings of the IEEE/CVF Conference on Computer Vision and Pattern Recognition},
  pages={9413--9422},
  year={2020}
  }

@inproceedings{GDNet,
  title={Don't hit me! glass detection in real-world scenes},
  author={Mei, Haiyang and Yang, Xin and Wang, Yang and Liu, Yuanyuan and He, Shengfeng and Zhang, Qiang and Wei, Xiaopeng and Lau, Rynson WH},
  booktitle={Proceedings of the IEEE/CVF Conference on Computer Vision and Pattern Recognition},
  pages={3687--3696},
  year={2020}
  }

@inproceedings{TransNet,
   Author = {Xie, Enze and Wang, Wenjia and Wang, Wenhai and Ding, Mingyu and Shen, Chunhua and Luo, Ping},
   Title = {Segmenting Transparent Objects in the Wild},
   Address= {Glasgow, United kingdom},
   Publisher = {Springer Science and Business Media Deutschland GmbH},
   Volume = {12358 LNCS},
   Pages = {696-711},
   Year = {2020} 
   }

@inproceedings{ResNet,
  title={Deep residual learning for image recognition},
  author={He, Kaiming and Zhang, Xiangyu and Ren, Shaoqing and Sun, Jian},
  booktitle={Proceedings of the IEEE conference on computer vision and pattern recognition},
  pages={770--778},
  year={2016}
}

@inproceedings{MirrorNet,
  title={Where is my mirror?},
  author={Yang, Xin and Mei, Haiyang and Xu, Ke and Wei, Xiaopeng and Yin, Baocai and Lau, Rynson WH},
  booktitle={Proceedings of the IEEE/CVF International Conference on Computer Vision},
  pages={8809--8818},
  year={2019}
}

@inproceedings{PMD,
  title={Progressive Mirror Detection},
  author={Lin, Jiaying and Wang, Guodong and Lau, Rynson WH},
  booktitle={Proceedings of the IEEE/CVF Conference on Computer Vision and Pattern Recognition},
  pages={3697--3705},
  year={2020}
  }

@inproceedings{UNet,
  title={U-net: Convolutional networks for biomedical image segmentation},
  author={Ronneberger, Olaf and Fischer, Philipp and Brox, Thomas},
  booktitle={International Conference on Medical image computing and computer-assisted intervention},
  pages={234--241},
  year={2015},
  organization={Springer}
  }

@inproceedings{FCN,
   Author = {Long, Jonathan and Shelhamer, Evan and Darrell, Trevor},
   Title = {Fully convolutional networks for semantic segmentation},
   BookTitle = {Proceedings of the IEEE conference on computer vision and pattern recognition},
   Series= {Proceedings of the IEEE conference on computer vision and pattern recognition},
   Pages = {3431-3440},
   Year = {2015} }

@inproceedings{PSP,
  title={Pyramid scene parsing network},
  author={Zhao, Hengshuang and Shi, Jianping and Qi, Xiaojuan and Wang, Xiaogang and Jia, Jiaya},
  booktitle={Proceedings of the IEEE conference on computer vision and pattern recognition},
  pages={2881--2890},
  year={2017}
  }

@inproceedings{non-local,
  title={Non-local neural networks},
  author={Wang, Xiaolong and Girshick, Ross and Gupta, Abhinav and He, Kaiming},
  booktitle={Proceedings of the IEEE conference on computer vision and pattern recognition},
  pages={7794--7803},
  year={2018}
  }

@article{DeepLab,
   Author = {Chen, Liang-Chieh and Papandreou, George and Kokkinos, Iasonas and Murphy, Kevin and Yuille, Alan L.},
   Title = {DeepLab: Semantic Image Segmentation with Deep Convolutional Nets, Atrous Convolution, and Fully Connected CRFs},
   Journal = {IEEE Transactions on Pattern Analysis and Machine Intelligence},
   Volume = {40},
   Number = {4},
   Pages = {834-848},
   DOI = {10.1109/TPAMI.2017.2699184},
   Year = {2018} 
   }

@inproceedings{DenseASPP,
  title={Denseaspp for semantic segmentation in street scenes},
  author={Yang, Maoke and Yu, Kun and Zhang, Chi and Li, Zhiwei and Yang, Kuiyuan},
  booktitle={Proceedings of the IEEE conference on computer vision and pattern recognition},
  pages={3684--3692},
  year={2018}
}

@inproceedings{DANet,
  title={Dual attention network for scene segmentation},
  author={Fu, Jun and Liu, Jing and Tian, Haijie and Li, Yong and Bao, Yongjun and Fang, Zhiwei and Lu, Hanqing},
  booktitle={Proceedings of the IEEE/CVF Conference on Computer Vision and Pattern Recognition},
  pages={3146--3154},
  year={2019}
  }

@inproceedings{CCNet,
  title={Ccnet: Criss-cross attention for semantic segmentation},
  author={Huang, Zilong and Wang, Xinggang and Huang, Lichao and Huang, Chang and Wei, Yunchao and Liu, Wenyu},
  booktitle={Proceedings of the IEEE/CVF International Conference on Computer Vision},
  pages={603--612},
  year={2019}
  }

@inproceedings{BDRAR,
  title={Bidirectional feature pyramid network with recurrent attention residual modules for shadow detection},
  author={Zhu, Lei and Deng, Zijun and Hu, Xiaowei and Fu, Chi-Wing and Xu, Xuemiao and Qin, Jing and Heng, Pheng-Ann},
  booktitle={Proceedings of the European Conference on Computer Vision (ECCV)},
  pages={121--136},
  year={2018}
  }

@inproceedings{DSC,
  title={Direction-aware spatial context features for shadow detection},
  author={Hu, Xiaowei and Zhu, Lei and Fu, Chi-Wing and Qin, Jing and Heng, Pheng-Ann},
  booktitle={Proceedings of the IEEE Conference on Computer Vision and Pattern Recognition},
  pages={7454--7462},
  year={2018}
  }

@article{RAC,
  title={Rethinking atrous convolution for semantic image segmentation},
  author={Chen, Liang-Chieh and Papandreou, George and Schroff, Florian and Adam, Hartwig},
  journal={arXiv preprint arXiv:1706.05587},
  year={2017}
  }

@inproceedings{CCF,
  title={Context contrasted feature and gated multi-scale aggregation for scene segmentation},
  author={Ding, Henghui and Jiang, Xudong and Shuai, Bing and Liu, Ai Qun and Wang, Gang},
  booktitle={Proceedings of the IEEE Conference on Computer Vision and Pattern Recognition},
  pages={2393--2402},
  year={2018}
  }

@inproceedings{CE,
  title={Context encoding for semantic segmentation},
  author={Zhang, Hang and Dana, Kristin and Shi, Jianping and Zhang, Zhongyue and Wang, Xiaogang and Tyagi, Ambrish and Agrawal, Amit},
  booktitle={Proceedings of the IEEE conference on Computer Vision and Pattern Recognition},
  pages={7151--7160},
  year={2018}
  }

@inproceedings{SAC,
  title={Scale-adaptive convolutions for scene parsing},
  author={Zhang, Rui and Tang, Sheng and Zhang, Yongdong and Li, Jintao and Yan, Shuicheng},
  booktitle={Proceedings of the IEEE International Conference on Computer Vision},
  pages={2031--2039},
  year={2017}
  }

@article{attention,
  title={Attention is all you need},
  author={Vaswani, Ashish and Shazeer, Noam and Parmar, Niki and Uszkoreit, Jakob and Jones, Llion and Gomez, Aidan N and Kaiser, Lukasz and Polosukhin, Illia},
  journal={arXiv preprint arXiv:1706.03762},
  year={2017}
  }

@article{LSM,
  title={Long short-term memory-networks for machine reading},
  author={Cheng, Jianpeng and Dong, Li and Lapata, Mirella},
  journal={arXiv preprint arXiv:1601.06733},
  year={2016}
  }

@inproceedings{Transformer,
  title={End-to-end object detection with transformers},
  author={Carion, Nicolas and Massa, Francisco and Synnaeve, Gabriel and Usunier, Nicolas and Kirillov, Alexander and Zagoruyko, Sergey},
  booktitle={European Conference on Computer Vision},
  pages={213--229},
  year={2020},
  organization={Springer}
  }

@article{ids,
  title={Salient object detection in the deep learning era: An in-depth survey},
  author={Wang, Wenguan and Lai, Qiuxia and Fu, Huazhu and Shen, Jianbing and Ling, Haibin and Yang, Ruigang},
  journal={IEEE Transactions on Pattern Analysis and Machine Intelligence},
  year={2021},
  publisher={IEEE}
  }

@inproceedings{rdcfp,
  title={Reverse densely connected feature pyramid network for object detection},
  author={Xin, Yongjian and Wang, Shuhui and Li, Liang and Zhang, Weigang and Huang, Qingming},
  booktitle={Asian Conference on Computer Vision},
  pages={530--545},
  year={2018},
  organization={Springer}
  }

@inproceedings{gmr,
  title={Saliency detection via graph-based manifold ranking},
  author={Yang, Chuan and Zhang, Lihe and Lu, Huchuan and Ruan, Xiang and Yang, Ming-Hsuan},
  booktitle={Proceedings of the IEEE conference on computer vision and pattern recognition},
  pages={3166--3173},
  year={2013}
  }

@inproceedings{rbd,
  title={Saliency optimization from robust background detection},
  author={Zhu, Wangjiang and Liang, Shuang and Wei, Yichen and Sun, Jian},
  booktitle={Proceedings of the IEEE conference on computer vision and pattern recognition},
  pages={2814--2821},
  year={2014}
  }

@inproceedings{om,
  title={Salient object detection via objectness measure},
  author={Srivatsa, R Sai and Babu, R Venkatesh},
  booktitle={2015 IEEE International Conference on Image Processing (ICIP)},
  pages={4481--4485},
  year={2015},
  organization={IEEE}
  }

@inproceedings{mb,
  title={Minimum barrier salient object detection at 80 fps},
  author={Zhang, Jianming and Sclaroff, Stan and Lin, Zhe and Shen, Xiaohui and Price, Brian and Mech, Radomir},
  booktitle={Proceedings of the IEEE international conference on computer vision},
  pages={1404--1412},
  year={2015}
  }

@article{imgnet,
  title={Imagenet classification with deep convolutional neural networks},
  author={Krizhevsky, Alex and Sutskever, Ilya and Hinton, Geoffrey E},
  journal={Advances in neural information processing systems},
  volume={25},
  pages={1097--1105},
  year={2012}
  }

@article{vdcn,
  title={Very deep convolutional networks for large-scale image recognition},
  author={Simonyan, Karen and Zisserman, Andrew},
  journal={arXiv preprint arXiv:1409.1556},
  year={2014}
  }

@inproceedings{Amulet,
   Author = {Zhang, Pingping and Wang, Dong and Lu, Huchuan and Wang, Hongyu and Ruan, Xiang},
   Title = {Amulet: Aggregating multi-level convolutional features for salient object detection},
   BookTitle = {Proceedings of the IEEE International Conference on Computer Vision},
   Series= {Proceedings of the IEEE International Conference on Computer Vision},
   Pages = {202-211},
   Year = {2017} 
   }

@inproceedings{PAGR,
  title={Progressive attention guided recurrent network for salient object detection},
  author={Zhang, Xiaoning and Wang, Tiantian and Qi, Jinqing and Lu, Huchuan and Wang, Gang},
  booktitle={Proceedings of the IEEE Conference on Computer Vision and Pattern Recognition},
  pages={714--722},
  year={2018}
  }

@inproceedings{dgrl,
  title={Detect globally, refine locally: A novel approach to saliency detection},
  author={Wang, Tiantian and Zhang, Lihe and Wang, Shuo and Lu, Huchuan and Yang, Gang and Ruan, Xiang and Borji, Ali},
  booktitle={Proceedings of the IEEE conference on computer vision and pattern recognition},
  pages={3127--3135},
  year={2018}
  }

@inproceedings{stagerefine,
  title={A stagewise refinement model for detecting salient objects in images},
  author={Wang, Tiantian and Borji, Ali and Zhang, Lihe and Zhang, Pingping and Lu, Huchuan},
  booktitle={Proceedings of the IEEE International Conference on Computer Vision},
  pages={4019--4028},
  year={2017}
  }

@inproceedings{Imagenet,
  title={Imagenet: A large-scale hierarchical image database},
  author={Deng, Jia and Dong, Wei and Socher, Richard and Li, Li-Jia and Li, Kai and Fei-Fei, Li},
  booktitle={2009 IEEE conference on computer vision and pattern recognition},
  pages={248--255},
  year={2009},
  organization={Ieee}
  }

@inproceedings{dssc,
  title={Deeply supervised salient object detection with short connections},
  author={Hou, Qibin and Cheng, Ming-Ming and Hu, Xiaowei and Borji, Ali and Tu, Zhuowen and Torr, Philip HS},
  booktitle={Proceedings of the IEEE conference on computer vision and pattern recognition},
  pages={3203--3212},
  year={2017}
  }

@inproceedings{ftsrd,
  title={Frequency-tuned salient region detection},
  author={Achanta, Radhakrishna and Hemami, Sheila and Estrada, Francisco and Susstrunk, Sabine},
  booktitle={2009 IEEE conference on computer vision and pattern recognition},
  pages={1597--1604},
  year={2009},
  organization={IEEE}
  }

@inproceedings{RAFnet,
  title={Reverse attention for salient object detection},
  author={Chen, Shuhan and Tan, Xiuli and Wang, Ben and Hu, Xuelong},
  booktitle={Proceedings of the European Conference on Computer Vision (ECCV)},
  pages={234--250},
  year={2018}
  }

@inproceedings{ckt,
  title={Contour knowledge transfer for salient object detection},
  author={Li, Xin and Yang, Fan and Cheng, Hong and Liu, Wei and Shen, Dinggang},
  booktitle={Proceedings of the European Conference on Computer Vision (ECCV)},
  pages={355--370},
  year={2018}
  }

@inproceedings{Picanet,
  title={Picanet: Learning pixel-wise contextual attention for saliency detection},
  author={Liu, Nian and Han, Junwei and Yang, Ming-Hsuan},
  booktitle={Proceedings of the IEEE Conference on Computer Vision and Pattern Recognition},
  pages={3089--3098},
  year={2018}
  }

@inproceedings{lko,
  title={Leave-one-out kernel optimization for shadow detection},
  author={Vicente, Tom{\'a}s F Yago and Hoai, Minh and Samaras, Dimitris},
  booktitle={Proceedings of the IEEE International Conference on Computer Vision},
  pages={3388--3396},
  year={2015}
  }

@article{pytorch,
  title={Pytorch: An imperative style, high-performance deep learning library},
  author={Paszke, Adam and Gross, Sam and Massa, Francisco and Lerer, Adam and Bradbury, James and Chanan, Gregory and Killeen, Trevor and Lin, Zeming and Gimelshein, Natalia and Antiga, Luca and others},
  journal={arXiv preprint arXiv:1912.01703},
  year={2019}
  }

@inproceedings{Bisenet,
  title={Bisenet: Bilateral segmentation network for real-time semantic segmentation},
  author={Yu, Changqian and Wang, Jingbo and Peng, Chao and Gao, Changxin and Yu, Gang and Sang, Nong},
  booktitle={Proceedings of the European conference on computer vision (ECCV)},
  pages={325--341},
  year={2018}
  }

@inproceedings{psanet,
  title={Psanet: Point-wise spatial attention network for scene parsing},
  author={Zhao, Hengshuang and Zhang, Yi and Liu, Shu and Shi, Jianping and Loy, Chen Change and Lin, Dahua and Jia, Jiaya},
  booktitle={Proceedings of the European Conference on Computer Vision (ECCV)},
  pages={267--283},
  year={2018}
  }

@inproceedings{r3net,
  title={R3net: Recurrent residual refinement network for saliency detection},
  author={Deng, Zijun and Hu, Xiaowei and Zhu, Lei and Xu, Xuemiao and Qin, Jing and Han, Guoqiang and Heng, Pheng-Ann},
  booktitle={Proceedings of the 27th International Joint Conference on Artificial Intelligence},
  pages={684--690},
  year={2018},
  organization={AAAI Press}
  }

@inproceedings{cpd,
  title={Cascaded partial decoder for fast and accurate salient object detection},
  author={Wu, Zhe and Su, Li and Huang, Qingming},
  booktitle={Proceedings of the IEEE/CVF Conference on Computer Vision and Pattern Recognition},
  pages={3907--3916},
  year={2019}
  }

@article{crf,
  title={Efficient inference in fully connected crfs with gaussian edge potentials},
  author={Kr{\"a}henb{\"u}hl, Philipp and Koltun, Vladlen},
  journal={Advances in neural information processing systems},
  volume={24},
  pages={109--117},
  year={2011}
  }

@inproceedings{richNet,
  title={Rich Context Aggregation With Reflection Prior for Glass Surface Detection},
  author={Lin, Jiaying and He, Zebang and Lau, Rynson WH},
  booktitle={Proceedings of the IEEE/CVF Conference on Computer Vision and Pattern Recognition},
  pages={13415--13424},
  year={2021}
}

@inproceedings{hu2018squeeze,
  title={Squeeze-and-excitation networks},
  author={Hu, Jie and Shen, Li and Sun, Gang},
  booktitle={Proceedings of the IEEE conference on computer vision and pattern recognition},
  pages={7132--7141},
  year={2018}
}

@article{hard,
  title={Contour loss: Boundary-aware learning for salient object segmentation},
  author={Chen, Zixuan and Zhou, Huajun and Xie, Xiaohua and Lai, Jianhuang},
  journal={arXiv preprint arXiv:1908.01975},
  year={2019}
  }

\newpage

\end{document}